\documentclass[a4paper,fleqn]{cas-sc}

 \usepackage[numbers]{natbib}
\usepackage{subcaption} 
\usepackage{bm}
\usepackage{amsmath}
\usepackage{hyperref}

\DeclareCaptionLabelFormat{empty}{}
\def\tsc#1{\csdef{#1}{\textsc{\lowercase{#1}}\xspace}}
\tsc{WGM}
\tsc{QE}
\tsc{EP}
\tsc{PMS}
\tsc{BEC}
\tsc{DE}

\begin{document}
\let\WriteBookmarks\relax
\def\floatpagepagefraction{1}
\def\textpagefraction{.001}

\shorttitle{Kolmogorov-Arnold PointNet}


\title [mode = title]{Kolmogorov-Arnold PointNet: Deep learning for prediction of fluid fields on irregular geometries}  



%
\author[1]{Ali Kashefi}[
                        orcid=0000-0003-0014-9051]

\cormark[1]


\ead{kashefi@stanford.edu}



\affiliation[1]{organization={Department of Civil and Environmental Engineering, Stanford University},
    city={Stanford},
    postcode={94305, CA}, 
    country={USA}}







\cortext[cor1]{Corresponding author}



\begin{abstract}
Kolmogorov-Arnold Networks (KANs) have emerged as a promising alternative to traditional Multilayer Perceptrons (MLPs) in deep learning. KANs have already been integrated into various architectures, such as convolutional neural networks, graph neural networks, and transformers, and their potential has been assessed for predicting physical quantities. However, the combination of KANs with point-cloud-based neural networks (e.g., PointNet) for computational physics has not yet been explored. To address this, we present Kolmogorov-Arnold PointNet (KA-PointNet) as a novel supervised deep learning framework for the prediction of incompressible steady-state fluid flow fields in irregular domains, where the predicted fields are a function of the geometry of the domains. In KA-PointNet, we implement shared KANs in the segmentation branch of the PointNet architecture. We utilize Jacobi polynomials to construct shared KANs. As a benchmark test case, we consider incompressible laminar steady-state flow over a cylinder, where the geometry of its cross-section varies over the data set. We investigate the performance of Jacobi polynomials with different degrees as well as special cases of Jacobi polynomials such as Legendre polynomials, Chebyshev polynomials of the first and second kinds, and Gegenbauer polynomials, in terms of the computational cost of training and accuracy of prediction of the test set. Furthermore, we examine the robustness of KA-PointNet in the presence of noisy training data and missing points in the point clouds of the test set. Additionally, we compare the performance of PointNet with shared KANs (i.e., KA-PointNet) and PointNet with shared MLPs. It is observed that when the number of trainable parameters is approximately equal, PointNet with shared KANs (i.e., KA-PointNet) outperforms PointNet with shared MLPs. Moreover, KA-PointNet predicts the pressure and velocity distributions along the surface of cylinders more accurately, resulting in more precise computations of lift and drag.
\end{abstract}





\maketitle

\section{Introduction and motivation}
\label{Sect1}

Kolmogorov-Arnold Networks (KANs) \citep{liu2024kan} were recently proposed as an alternative modeling approach to Multilayer Perceptrons (MLPs) \citep{cybenko1989approximation,hornik1989multilayer,Goodfellow2016}. KANs are primarily based on the principles of the Kolmogorov-Arnold representation theorem \citep{arnold2009representation,arnold2009functions,kolmogorovSuperposition,hecht1987kolmogorov,girosi1989representation,braun2009constructive,ismayilova2024kolmogorov,borri2024one}. The core idea of KANs is to propose a neural network where the objective is to train activation functions rather than training weights and biases with fixed activation functions, as in MLPs \citep{liu2024kan}. The efficiency and capability of KANs have been demonstrated in various areas and applications, such as physics-informed machine learning \citep{wang2024KANinformed,shukla2024comprehensive,howard2024finite,toscano2024inferring,wang2024kolmogorov,rigas2024adaptive,patra2024physics}, deep operator networks \citep{abueidda2024deepokan,shukla2024comprehensive}, neural ordinary differential equations \citep{koenig2024kan}, image classification \citep{azam2024suitability,cheon2024kolmogorovRemote,seydi2024exploringPolynomial,seydi2024KANwavelets,ta2024bsrbf,cheon2024Vision,lobanov2024hyperkan,yu2024kan,ExploreClassification,altarabichi2024rethinking}, image segmentation \citep{li2024UKAN,tang20243d}, image detection \citep{wang2024spectralkan}, audio classification \citep{yu2024kan}, and many other scientific and industrial fields \citep{bozorgasl2024wav,vaca2024kolmogorov,genet2024tkan,samadi2024smooth,liu2024ikan,li2024KANradial,aghaei2024Fractional,xu2024kolmogorovPower,xu2024fourierkan,peng2024predictivePump,genet2024temporalKAN,nehma2024leveragingOperator,herbozo2024kan,liu2024initialHumanActivity,poeta2024Table,kundu2024KANquantum,aghaei2024RationalKAN,li2024coeff,pratyush2024calmphoskan,liu2024complexity}. Additionally, KANs have been integrated into Convolutional Neural Networks (CNNs) \citep{azam2024suitability,bodner2024CNNkan} and graph neural networks \citep{kiamari2024gkan,bresson2024kagnns,zhang2024graphKAN,de2024kolmogorovGraph}. In the current study, our focus is on the field of computational physics and mechanics.

One of the most important applications of supervised deep learning in computational mechanics is to accelerate the investigation of geometric parameters for device optimization. Supervised deep learning models are first trained on labeled data obtained from numerical simulations or lab experiments, and then used to predict quantities of interest for unseen geometries. For this specific application, geometric deep learning models such as CNNs and their extensions (e.g., see Refs. \citep{ronneberger2015u,guo2016convolutional,gao2021phygeonet,tompson2017accelerating,thuerey2020deep,bhatnagar2019prediction}), graph neural networks (e.g., see Refs. \citep{xu2018powerful,pfaff2020learning,belbute2020combining,maurizi2022predicting,gao2022physicsGraph}), and point cloud-based neural networks such as PointNet and its derived versions (e.g., see Refs. \citep{qi2017pointnet,qi2017pointnet++,kashefi2021PointNet,kashefi2022physics,kashefi2023PIPNelasticity,CompressiblePointNet,DeepONetPointNetElasticity}) are particularly useful. Focusing on this aspect, the goal of the current study is to investigate the capability of KANs, embedded in a geometric deep learning framework, for rapid predictions of quantities of interest in computational mechanics when the geometry of the domains varies across the data set. In these types of networks, the geometry of a domain is learned in a latent space, making the network's prediction a function of the geometry.

Graph neural networks \citep{Graph2020Review} and PointNet \citep{qi2017pointnet} have advantages for handling complicated geometries, which are common in industrial design. These benefits have been extensively discussed in the literature (e.g., see Ref. \cite{kashefi2022physics}), and a summary is presented here. Using graph neural networks or PointNet, the geometry of the domain of interest or the object under investigation for shape optimization can be represented without pixelation, unlike CNN-based models. This approach avoids artifacts in the representation of object or domain boundaries (e.g., see Fig. 1 in Ref. \citep{kashefi2021PointNet}). Moreover, because the geometry of the target object is precisely illustrated in graph neural networks or PointNet, the network's predictions are highly responsive to small variations in the geometry from one labeled data to another, such as minor adjustments in the angle of attack of an airfoil. Additionally, graph neural networks or PointNet allow for adaptive variation in the spatial distribution of points from fine to coarse scales (e.g., see Figs. 7--10 in Ref. \citep{kashefi2021PointNet}), a feature similar to unstructured grids. This optimizes the computational cost of training, rather than imposing high computational costs due to uniform pixels across the entire domain, as in CNN-based models. Finally, the spatial dimensions of the domain under study can vary from one labeled data to another, and are not constrained to a fixed size, unlike in CNN-based networks (e.g., see Figs. 7--11 in Ref. \citep{kashefi2021PointNet}). Due to these advantages, we embed KANs (instead of MLPs) into PointNet to create a new geometric deep learning framework. For the rest of the article, we will refer to this framework as KA-PointNet.

PointNet \citep{qi2017pointnet} was initially introduced in 2017 for the classification and segmentation of three-dimensional objects modeled as three-dimensional points. PointNet \citep{qi2017pointnet} and its advanced versions \cite{qi2017pointnet++,thomas2019kpconv} have received remarkable attention in the areas of computer graphics and computer vision (e.g., see Refs. \citep{3DPointSurvay,Qi_2018_CVPR,zeng2022lion,zhao2021point}). Historically, Kashefi et al. \citep{kashefi2021PointNet} were the first to use PointNet, with necessary adaptations and adjustments, in the area of computational physics, specifically for steady-state incompressible flow in two dimensions. Subsequently, other researchers employed PointNet and its advanced versions, such as PointNet++ \citep{qi2017pointnet++}, in other areas such as compressible flow \citep{CompressiblePointNet}, elasticity \citep{DeepONetPointNetElasticity}, and others \citep{PointNetMelting,singleholePointNet,PointNetPorousMedia}. Kashefi et al. \citep{kashefi2022physics} also introduced physics-informed PointNet (PIPN) in 2022 to solve forward and inverse problems involving incompressible flows \citep{kashefi2022physics}, thermal fields \citep{kashefi2022physics}, linear elasticity \citep{kashefi2023PIPNelasticity}, and flow transport in porous media \citep{kashefi2023PIPNporous}. In simple terms, PIPN combines PointNet with the concept of physics-informed deep learning introduced by Raissi et al. \citep{RAISSI2019PIPN}, integrating the residual of governing equations associated with the physics of a problem into the loss function of PIPN.

Although PointNet adapted for computational fluid dynamics \citep{kashefi2021PointNet} has proven to be a promising machine learning framework, several concerns remain. The first one is the use of a fixed activation function, such as the sigmoid or hyperbolic tangent functions, in the last layer of the network. This design choice implies that all variables of interest, regardless of the physics or geometry of the problem, must ultimately be predicted by a predefined function with a fixed range of output, which is inherent to the nature of MLPs embedded in PointNet. Part of the motivation for the current study is to address this concern by implementing KANs in PointNet, particularly in its last layer. In a supervised learning framework, this approach enables labeled data (i.e., pairs of geometry and physical variables defined on the geometry) to influence the formation of the activation function in the last layer of PointNet during the training process. In other words, since the labeled data depends on both the physics and the geometry of the problem, activation functions in KANs (e.g., polynomials) learn to represent both the physics and the geometries of the training dataset. The importance of this feature becomes critical when the network is expected to provide accurate predictions on the boundaries of geometries, such as pressure on the surface of an airfoil for calculating drag and lift.

Beyond the limitation of the activation function in the last layer of PointNet with MLPs, a second concern relates to the simplicity of its architecture. The original PointNet \cite{qi2017pointnet} (implemented with MLPs) has a straightforward architecture, and its efficiency has been demonstrated for computer graphics applications \cite{qi2017pointnet,qi2017pointnet++}. According to this architecture (see, for example, Fig. 2 of Ref. \cite{qi2017pointnet} and Fig. 5 of Ref. \cite{kashefi2021PointNet}), PointNet's size is mainly controlled by two parameters: the number of layers, which determines its depth, and the number of neurons per layer. However, this simplicity might potentially result in a high degree of bias during training, particularly in computational physics applications, as discussed by \citet{kashefi2021PointNet} (e.g., see Table II in Ref. \cite{kashefi2021PointNet}) and as will be demonstrated later in this article. Embedding KAN layers into PointNet (i.e., KA-PointNet) introduces the degree of a polynomial as an additional key parameter that controls the network's size and allows its configuration to vary for each layer. One of the motivations for proposing KA-PointNet is that its added complexity enhances PointNet's robustness against high bias during training in computational physics.

To build KA-PointNet, we fundamentally use the segmentation branch of PointNet \citep{qi2017pointnet}. We implement shared KANs in all layers of PointNet instead of shared MLPs. Furthermore, we utilize Jacobi polynomials in shared KANs. Jacobi polynomials have already been used for KAN implementations (e.g., see Refs. \citep{shukla2024comprehensive,seydi2024exploringPolynomial,aghaei2024Fractional,zhang2024rpnreconciledpolynomialnetwork,KANwithTANH}). Moreover, we investigate the capability of special cases of Jacobi polynomials, such as Legendre polynomials, Chebyshev polynomials of the first and second kinds, and Gegenbauer polynomials. Additionally, we assess the performance of KA-PointNet for different degrees of Jacobi polynomials. Flow past a cylinder with different cross-sectional shapes is considered as a representative benchmark problem. To construct point cloud data readable for feeding into PointNet \citep{qi2017pointnet}, grid vertices of finite volume meshes of computational domains are viewed as points. A comparison between the outcomes of PointNet with shared KANs (i.e., KA-PointNet) and PointNet with shared MLPs is made in terms of prediction accuracy, training time, and the number of trainable parameters. The Adam optimizer \citep{kingma2014adam}, along with the mean squared error loss function, is used to perform the gradient descent optimization. It is noteworthy that, to the best of the author's knowledge, this is the first time a geometric deep learning model combined with KANs is used to predict solutions for computational fields over domains with distinct geometries. Although KANs have already been integrated into CNNs and graph neural networks, they have been utilized for other applications in computer vision \citep{azam2024suitability,bodner2024CNNkan,kiamari2024gkan,bresson2024kagnns,zhang2024graphKAN,de2024kolmogorovGraph}. 


The rest of this article is organized as follows. Governing equations and data generation are explained in Sect. \ref{Sect2}. The architecture of KA-PointNet with shared KANs and its training on the desired data set are elaborated in Sect. \ref{Sect3}. Analysis of the performance of KA-PointNet and the effect of different parameters on the accuracy of its predictions are given in Sect. \ref{Sect4}. Additionally, a comparison between the performance of KA-PointNet and PointNet with shared MLPs is provided in Sect. \ref{Sect4} as well. Finally, a summary of the research article and potential future directions are explained in Sect. \ref{Sect5}.

\begin{figure}[!htbp]
  \centering 
      \begin{subfigure}[b]{0.49\textwidth}
        \centering
        \includegraphics[width=\textwidth]{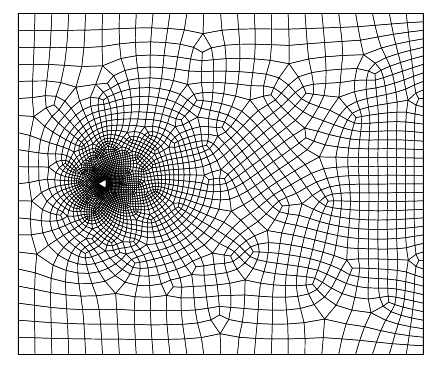}
    \end{subfigure}
    \begin{subfigure}[b]{0.49\textwidth}
        \centering
        \includegraphics[width=\textwidth]{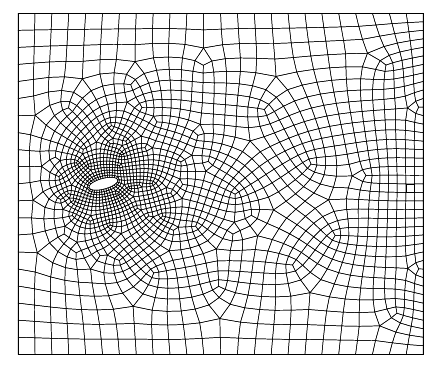}
    \end{subfigure}

    \caption{Finite volume meshes used for the numerical simulation of flow over a cylinder with a triangular cross section, with 2775 vertices on the left panel, and an elliptical cross section, with 2672 vertices on the right panel.}
  
  \label{Fig1}
\end{figure}

\section{Problem formulation}
\label{Sect2}

In the current article, we propose a supervised version of KA-PointNet for predicting the velocity and pressure fields of steady-state incompressible flow. To train the deep learning framework and perform error analysis on the predictions of KA-PointNet for unseen data, we require labeled data. We use the classical problem of steady-state flow past a cylinder as a benchmark test case. Accordingly, the input of KA-PointNet is the geometry of the domain, and the output is the prediction of velocity and pressure fields. We describe the governing equations of the problem and the procedure of data generation in the subsections of \ref{Sect21} and \ref{Sect22}, respectively.

\subsection{Governing equations of interest}
\label{Sect21}

The behavior of incompressible viscous Newtonian fluid flow is described by the conservation of mass and conservation of momentum, along with the associated boundary conditions, as follows:

\begin{equation}
    \nabla \cdot \boldsymbol{u} = 0 \quad \text{in } V,
    \label{Eq1}
\end{equation}

\begin{equation}
\rho \left(\boldsymbol{u} \cdot \nabla \right)\boldsymbol{u} - \mu \Delta \boldsymbol{u} + \nabla p = \boldsymbol{0} \quad \text{in } V,
\label{Eq2}
\end{equation}

\begin{equation}
    \boldsymbol{u} = \boldsymbol{u}_{\Gamma_\text{D}} \quad \text{on } \Gamma_\text{D},
    \label{Eq3}
\end{equation}

\begin{equation}
    -p\boldsymbol{n} + \mu \nabla \boldsymbol{u} \cdot \boldsymbol{n} = \boldsymbol{t}_{\Gamma_\text{Ne}} \quad \text{on } \Gamma_\text{Ne},
    \label{Eq4}
\end{equation}
where $\boldsymbol{u}$ and $p$ are the velocity vector and absolute pressure of the fluid in the domain $V$, respectively. Fluid density and dynamic viscosity are denoted by $\rho$ and $\mu$, respectively. The Dirichlet and Neumann boundaries of domain $V$ are represented by $\Gamma_\text{D}$ and $\Gamma_\text{Ne}$, respectively, with no overlap between $\Gamma_\text{D}$ and $\Gamma_\text{Ne}$. The stress vector acting on $\Gamma_\text{Ne}$ is denoted by $\boldsymbol{t}_{\Gamma_\text{Ne}}$, where $\boldsymbol{n}$ indicates the outward unit normal vector to $\Gamma_\text{Ne}$. The components of the velocity vector $\boldsymbol{u}$ are demonstrated by $u$ and $v$ in the $x$ and $y$ directions, respectively.

We are interested in solving Eqs. (\ref{Eq1})--(\ref{Eq2}) for flow over an infinite cylinder with different cross-sectional shapes. To achieve this goal, we consider a rectangular domain $V =$ [0, 38 m] $\times$ [0, 32 m]. The cross-section of the cylinder is represented by a two-dimensional object with its center of mass at the point (8 m, 16 m) in the domain $V$. The cylinder is rigid, and we impose no-slip conditions on its surfaces. The free stream velocity, with a magnitude of $u_\infty$ and parallel to the $x$ axis, is applied at the inflow, bottom, and top boundaries of the domain $V$. To satisfy far-field assumptions \citep{DING2004FDscheme,behr1995incompressible,kashefiCoarse2,kashefiCoarse4,kashefiCoarse5}, the outflow velocity is described by the Neumann stress-free conditions (i.e., $\boldsymbol{t}_{\Gamma_{Ne}}=\boldsymbol{0}$) as follows:

\begin{equation}
    -p\boldsymbol{n} + \mu \nabla \boldsymbol{u} \cdot \boldsymbol{n} = \boldsymbol{0}.
    \label{Eq5}
\end{equation}
Working in the International System of Units, the fluid density ($\rho$), free stream velocity ($u_\infty$), and dynamic viscosity ($\mu$) are set to 1.00 kg/m$^3$, 1.00 m/s, and 0.05 Pa$\cdot$s, respectively.


\begin{table}[width=.9\linewidth,cols=6, pos=!htbp]
\caption{Details of the generated data}\label{Table1}
\begin{tabular*}{\tblwidth}{@{} LLLLLLL@{}}
\toprule
 Shape & 
\vtop{\hbox{\strut Schematic}\hbox{\strut figure}}
 & \vtop{\hbox{\strut Variation in}\hbox{\strut orientation}} & \vtop{\hbox{\strut Variation in}\hbox{\strut length scale}} & \vtop{\hbox{\strut Number}\hbox{\strut of data}}\\
\hline
Circle & \includegraphics[width=0.08\linewidth]{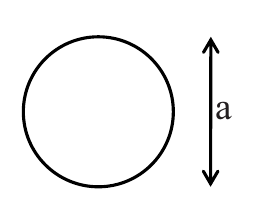} & - & $a=1$ m & 1 \\
Equilateral hexagon & \includegraphics[width=0.08\linewidth]{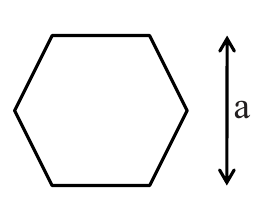} & $3^\circ$, $6^\circ$, \ldots, $60^\circ$ & $a=1$ m & 20 \\
Equilateral pentagon & \includegraphics[width=0.08\linewidth]{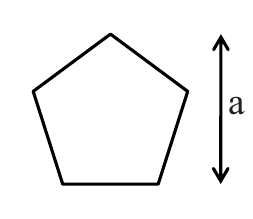} & $3^\circ$, $6^\circ$, \ldots, $72^\circ$ & $a=1$ m & 24 \\
Square & \includegraphics[width=0.08\linewidth]{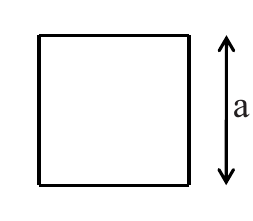} &  $3^\circ$, $6^\circ$, \ldots, $90^\circ$ & $a=1$ m & 30 \\
Equilateral triangle & \includegraphics[width=0.08\linewidth]{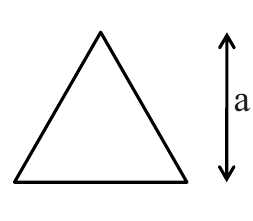} & $3^\circ$, $6^\circ$, \ldots, $180^\circ$ & $a=1$ m & 60\\
Rectangle & \includegraphics[width=0.08\linewidth]{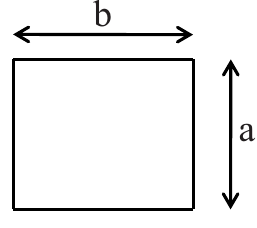} &  $3^\circ$, $6^\circ$, \ldots, $180^\circ$ & $a=1$ m; $b/a=$ 1.2, 1.4, \ldots, 3.6 & 780 \\
Ellipse & \includegraphics[width=0.08\linewidth]{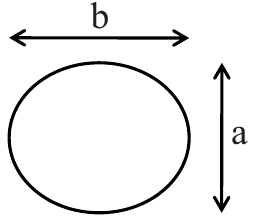} &  $3^\circ$, $6^\circ$, \ldots, $180^\circ$ & $a=1$ m; $b/a=$ 1.2, 1.4, \ldots, 3.8 & 840\\
Triangle & \includegraphics[width=0.08\linewidth]{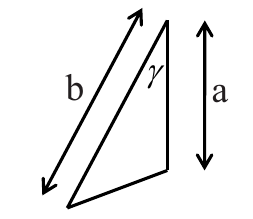} &  $3^\circ$, $6^\circ$, \ldots, $360^\circ$ & \vtop{\hbox{\strut $a=1$ m; $b/a=$ 1.5, 1.75}\hbox{\strut $\gamma=$  60$^\circ$, 80$^\circ$}} & 480 \\
\bottomrule
\end{tabular*}
\end{table}


\begin{figure}[!htbp]
  \centering 
      \begin{subfigure}[b]{0.24\textwidth}
      \caption{Input, spatial coordinates}
        \centering
        \includegraphics[width=\textwidth]{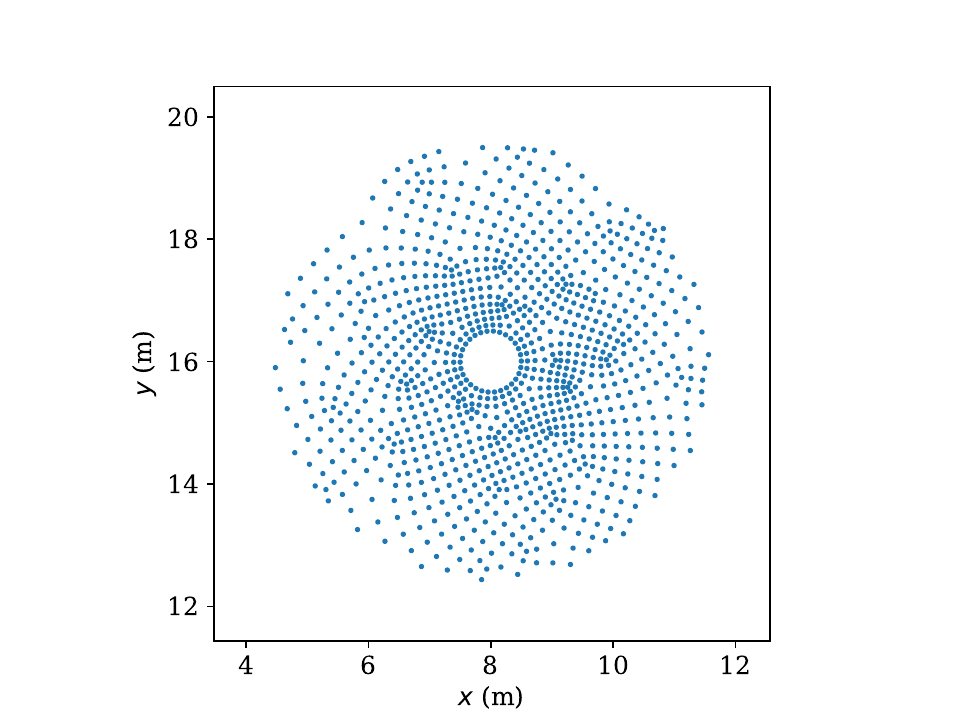}
    \end{subfigure}
    \begin{subfigure}[b]{0.24\textwidth}
    \caption{Output, $u$ (m/s)}
        \centering
        \includegraphics[width=\textwidth]{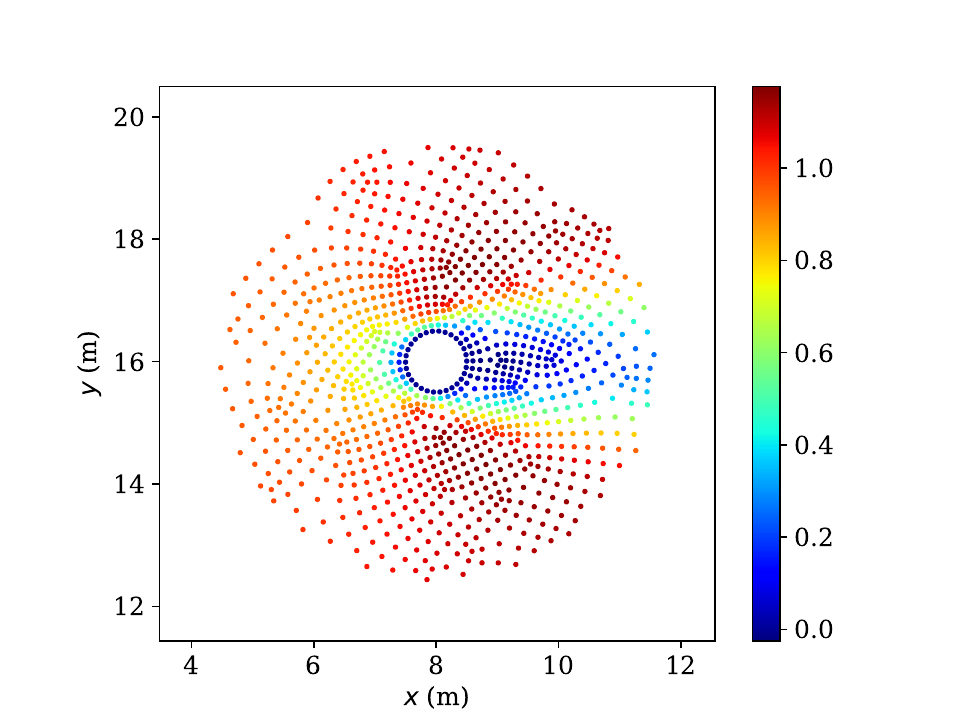}
    \end{subfigure}
    \begin{subfigure}[b]{0.24\textwidth}
    \caption{Output, $v$ (m/s)}
        \centering
        \includegraphics[width=\textwidth]{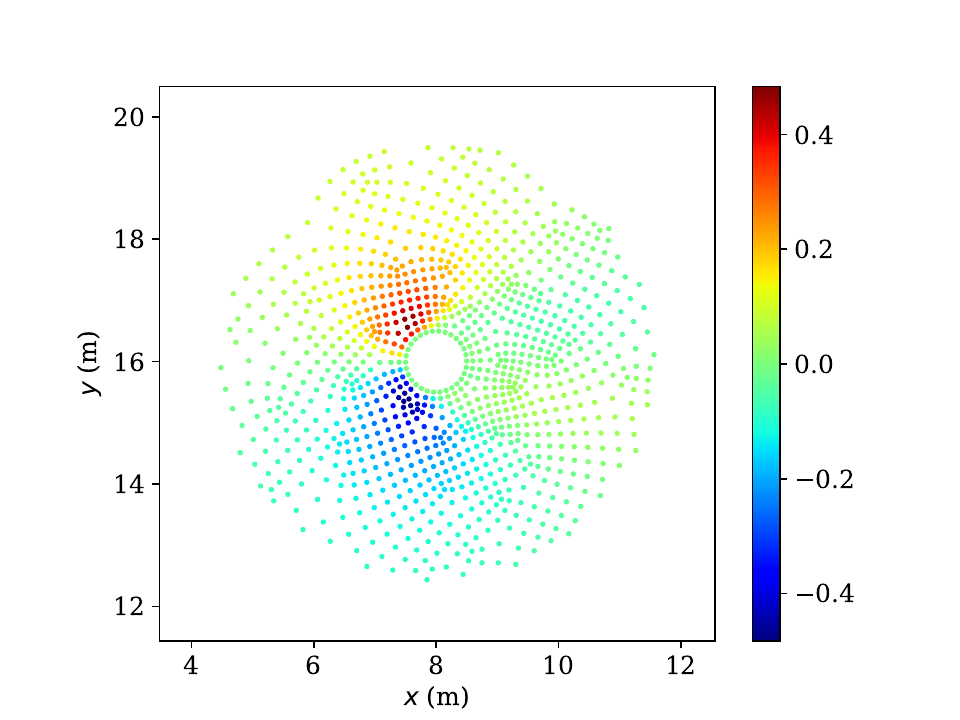}
    \end{subfigure}
     \begin{subfigure}[b]{0.24\textwidth}
     \caption{Output, gauge pressure (Pa)}
        \centering
        \includegraphics[width=\textwidth]{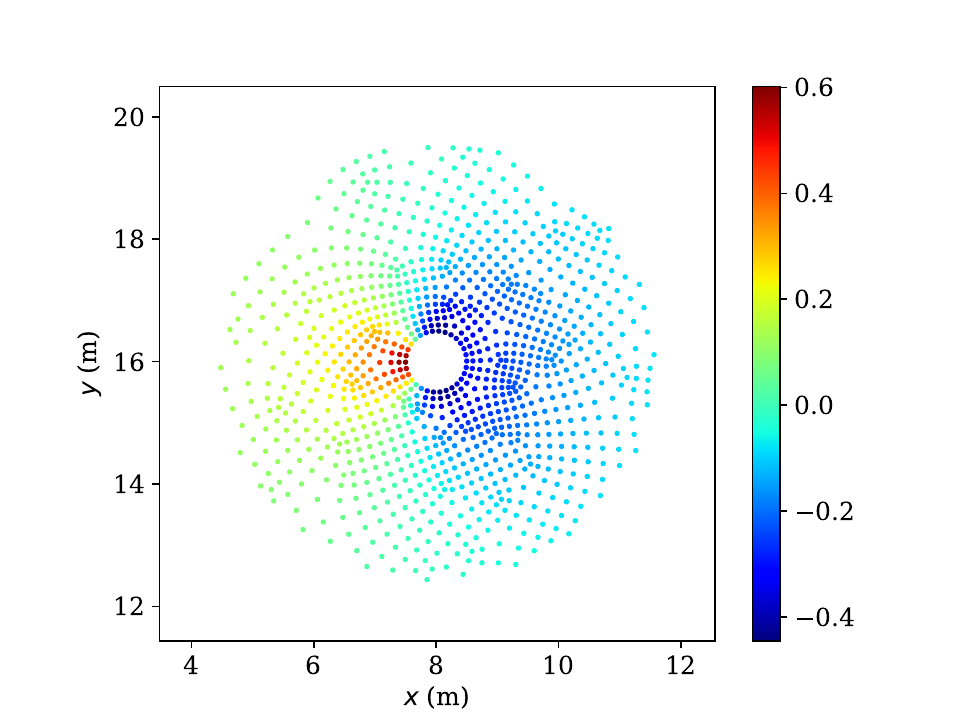}
    \end{subfigure}

       \begin{subfigure}[b]{0.24\textwidth}
        \centering
        \includegraphics[width=\textwidth]{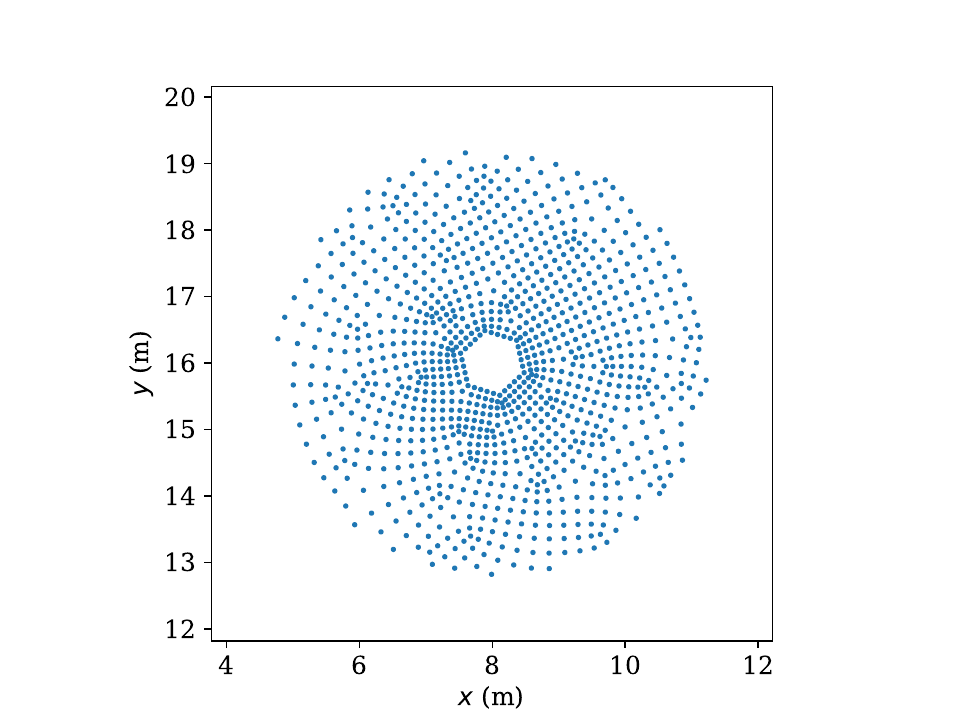}
    \end{subfigure}
    \begin{subfigure}[b]{0.24\textwidth}
        \centering
        \includegraphics[width=\textwidth]{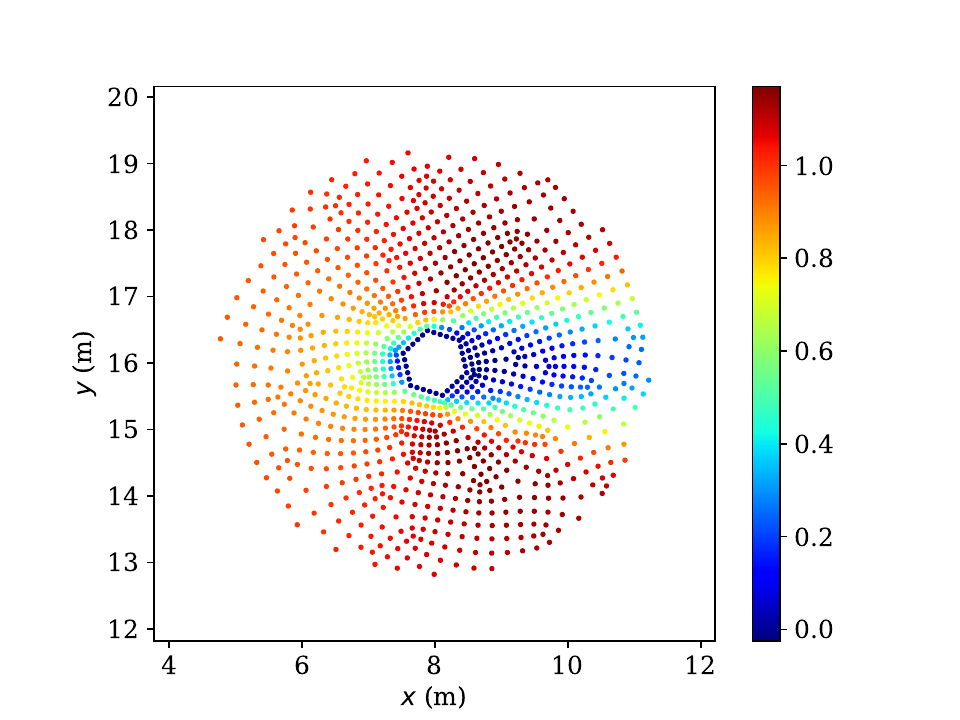}
    \end{subfigure}
    \begin{subfigure}[b]{0.24\textwidth}
        \centering
        \includegraphics[width=\textwidth]{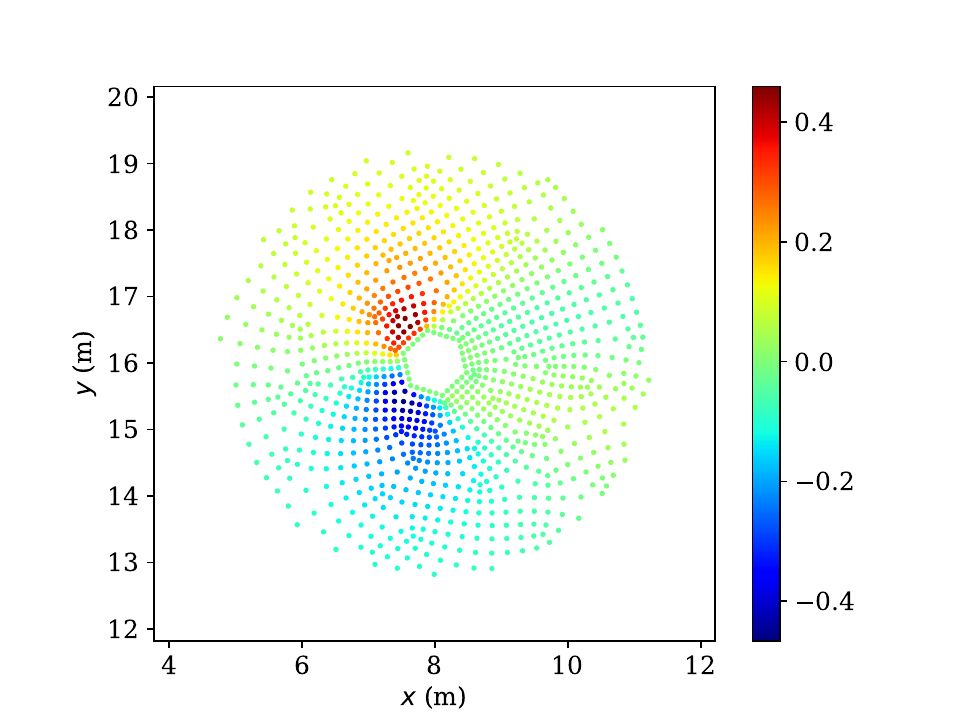}
    \end{subfigure}
     \begin{subfigure}[b]{0.24\textwidth}
        \centering
        \includegraphics[width=\textwidth]{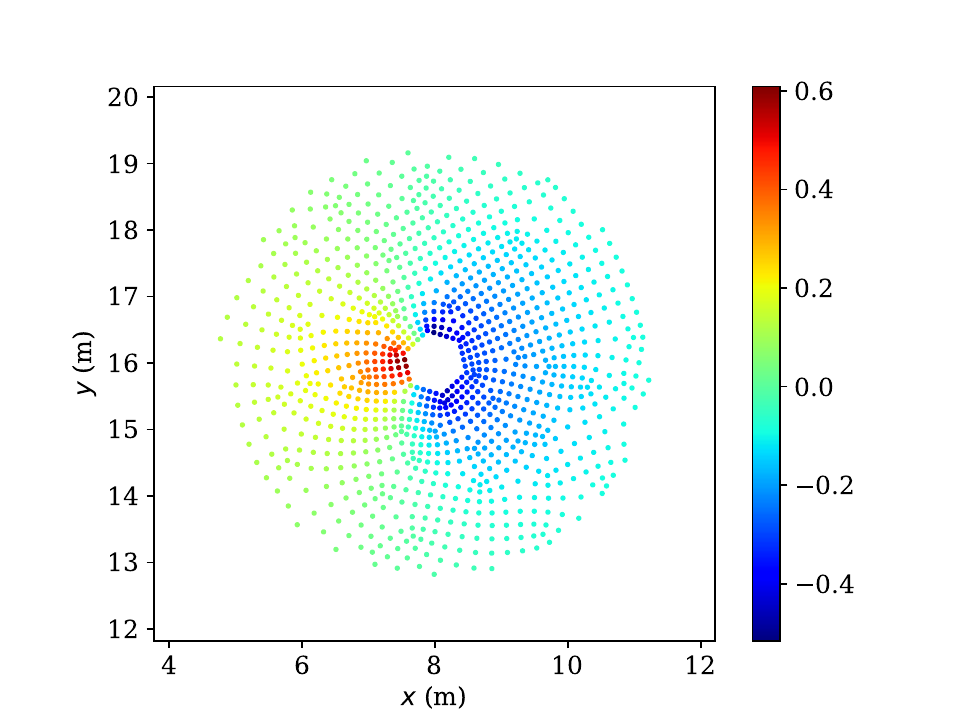}
    \end{subfigure}

    \begin{subfigure}[b]{0.24\textwidth}
        \centering
        \includegraphics[width=\textwidth]{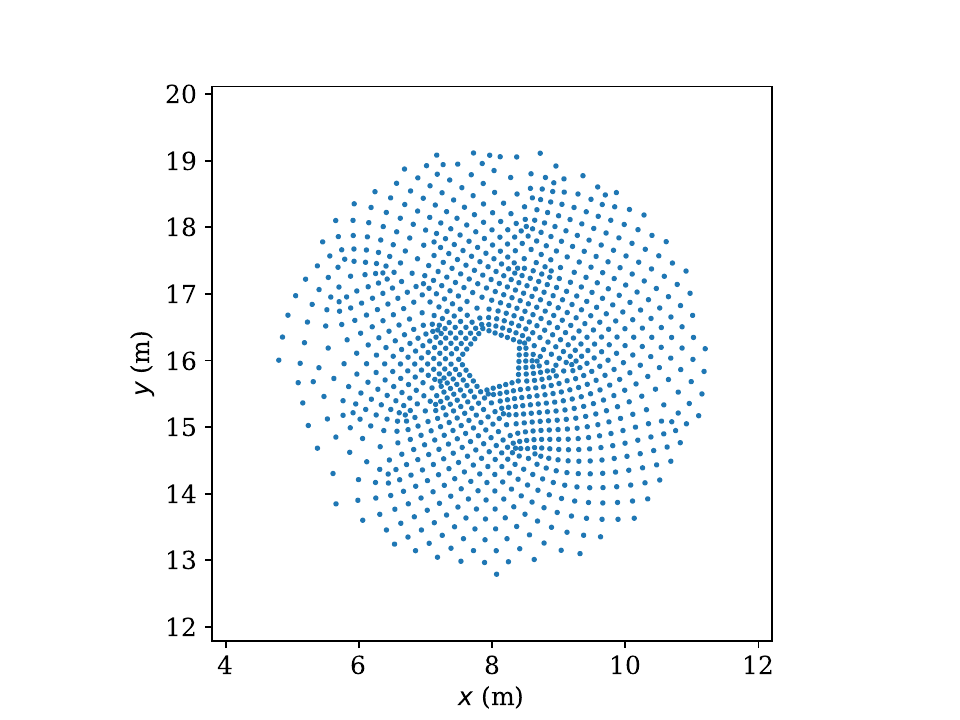}
    \end{subfigure}
    \begin{subfigure}[b]{0.24\textwidth}
        \centering
        \includegraphics[width=\textwidth]{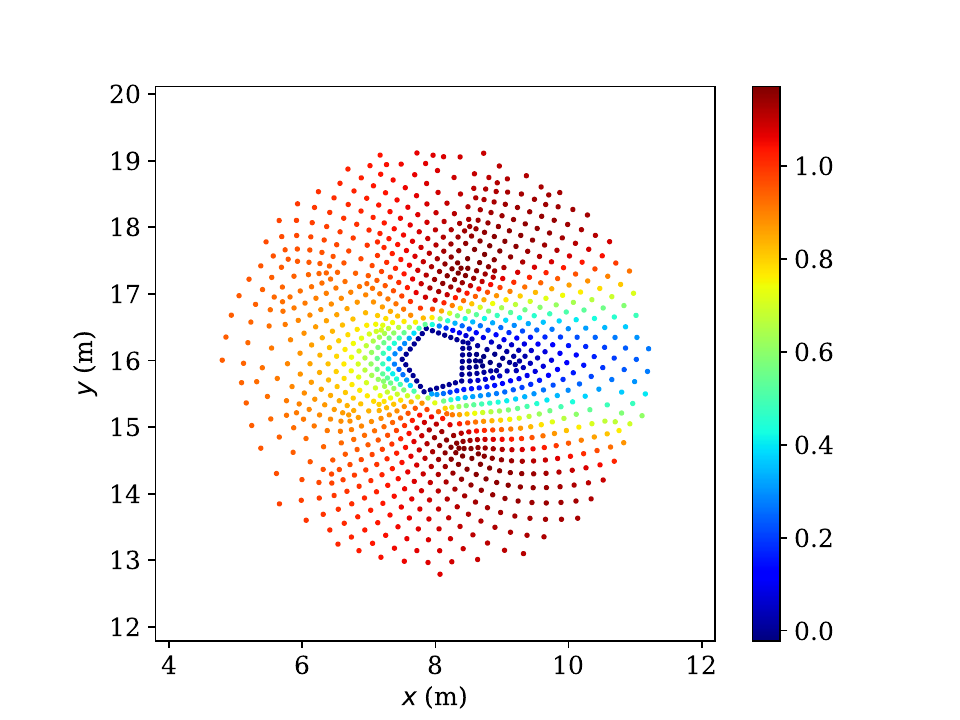}
    \end{subfigure}
    \begin{subfigure}[b]{0.24\textwidth}
        \centering
        \includegraphics[width=\textwidth]{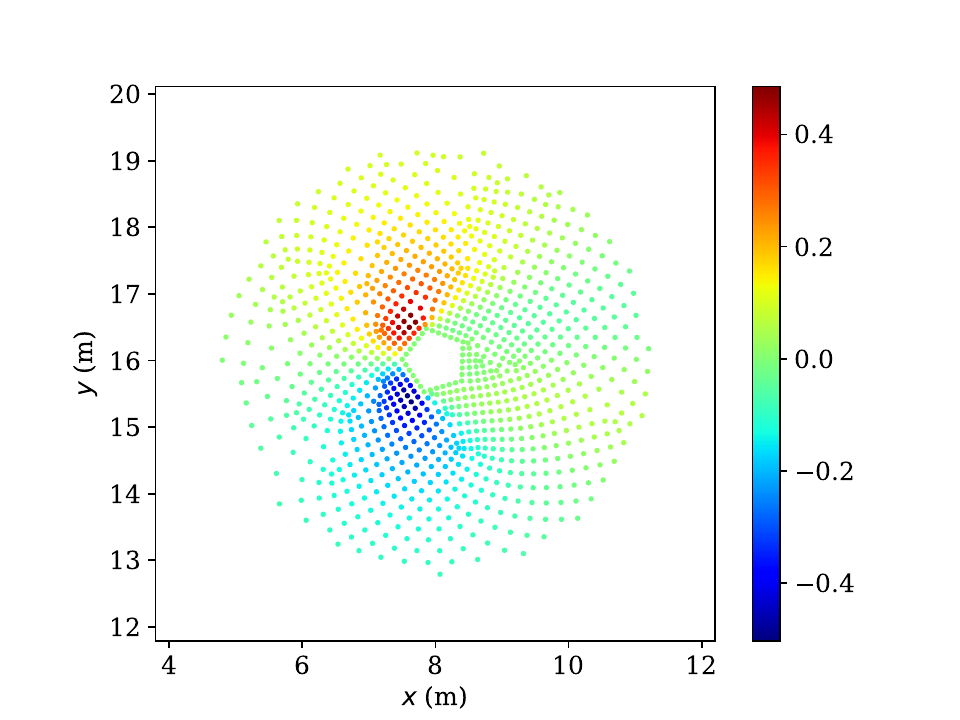}
    \end{subfigure}
     \begin{subfigure}[b]{0.24\textwidth}
        \centering
        \includegraphics[width=\textwidth]{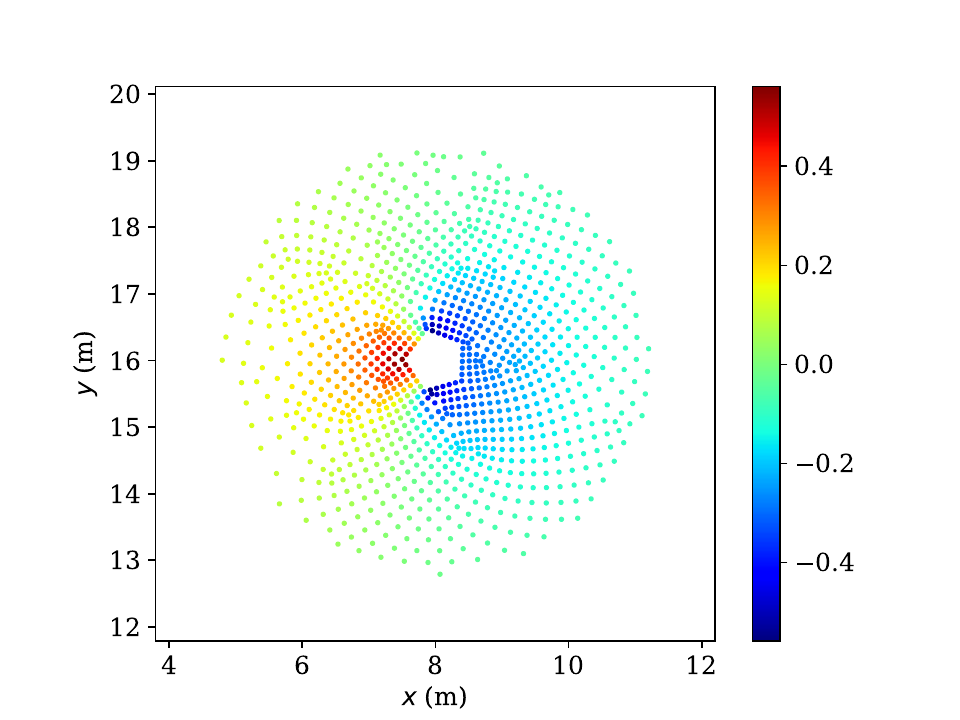}
    \end{subfigure}

    \begin{subfigure}[b]{0.24\textwidth}
        \centering
        \includegraphics[width=\textwidth]{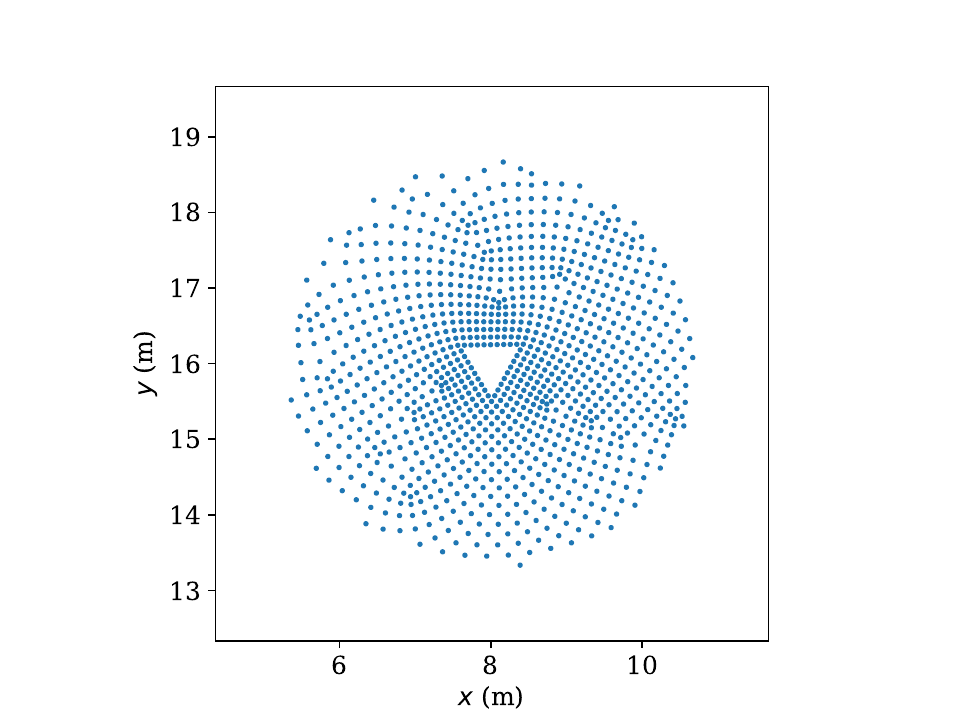}
    \end{subfigure}
    \begin{subfigure}[b]{0.24\textwidth}
        \centering
        \includegraphics[width=\textwidth]{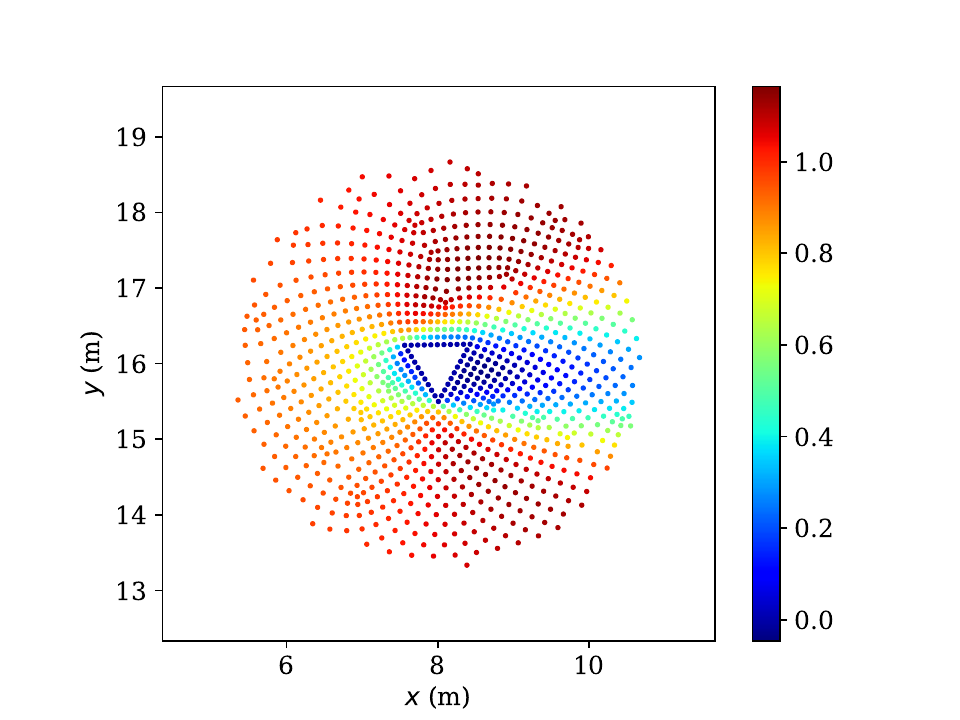}
    \end{subfigure}
    \begin{subfigure}[b]{0.24\textwidth}
        \centering
        \includegraphics[width=\textwidth]{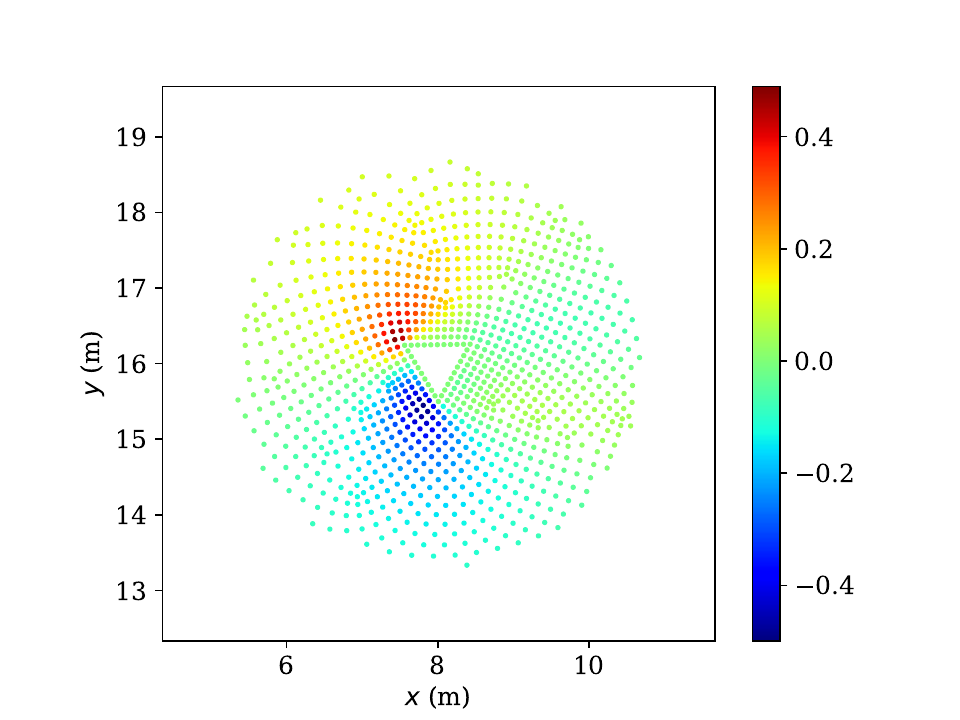}
    \end{subfigure}
     \begin{subfigure}[b]{0.24\textwidth}
        \centering
        \includegraphics[width=\textwidth]{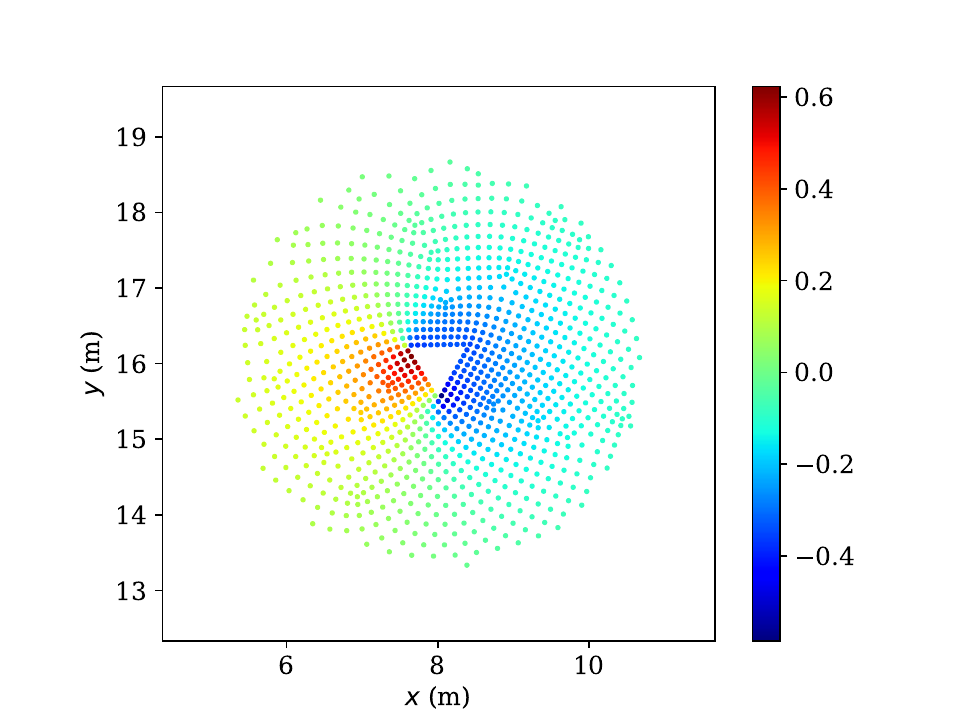}
    \end{subfigure}


        \begin{subfigure}[b]{0.24\textwidth}
        \centering
        \includegraphics[width=\textwidth]{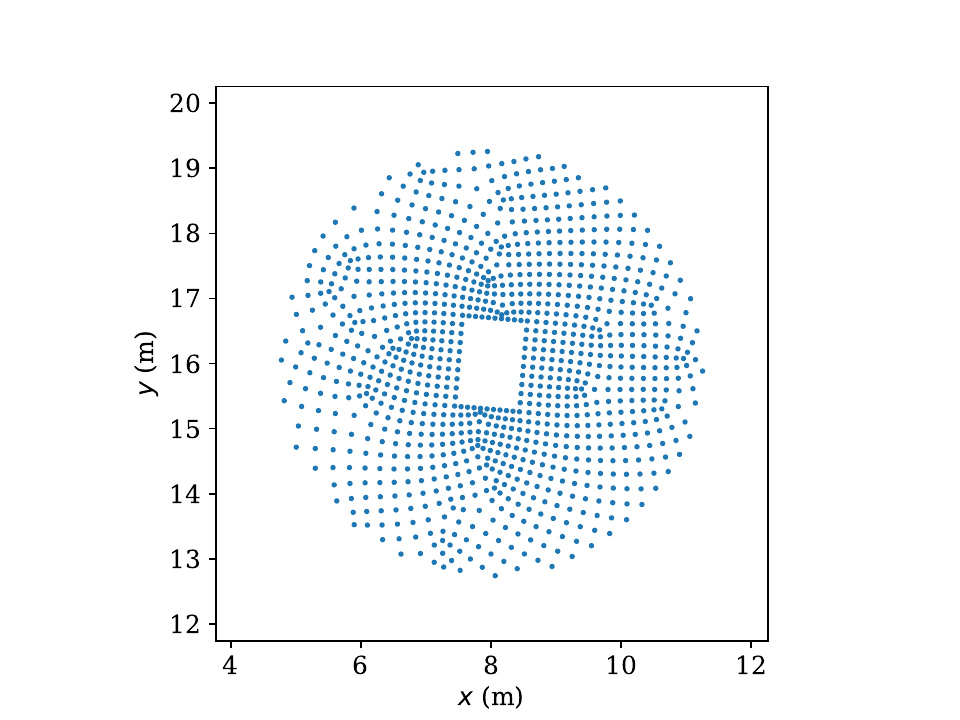}
    \end{subfigure}
    \begin{subfigure}[b]{0.24\textwidth}
        \centering
        \includegraphics[width=\textwidth]{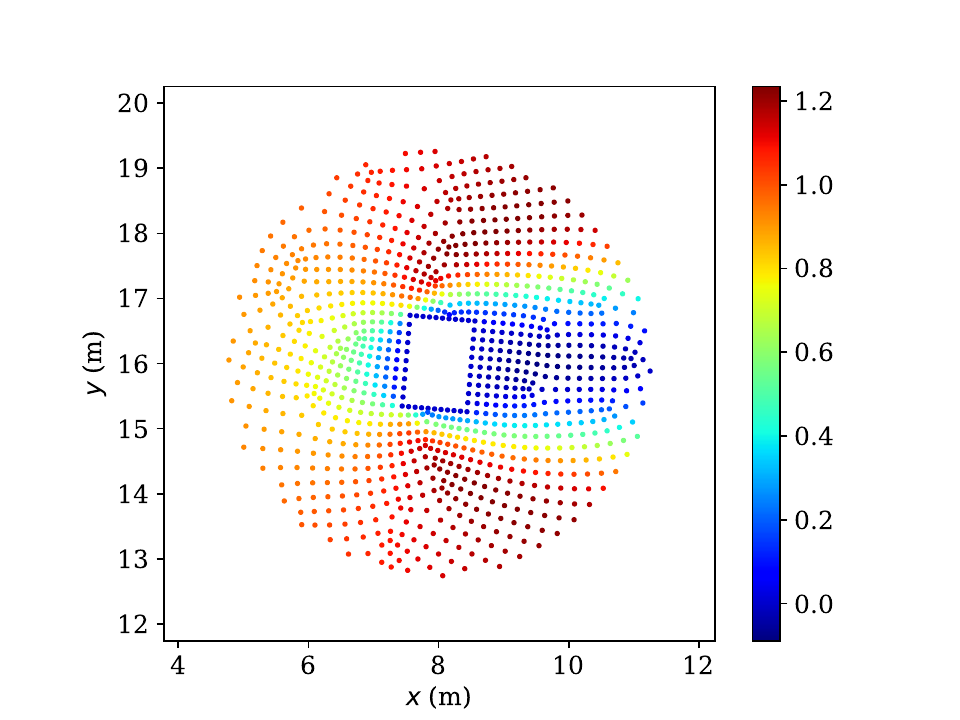}
    \end{subfigure}
    \begin{subfigure}[b]{0.24\textwidth}
        \centering
        \includegraphics[width=\textwidth]{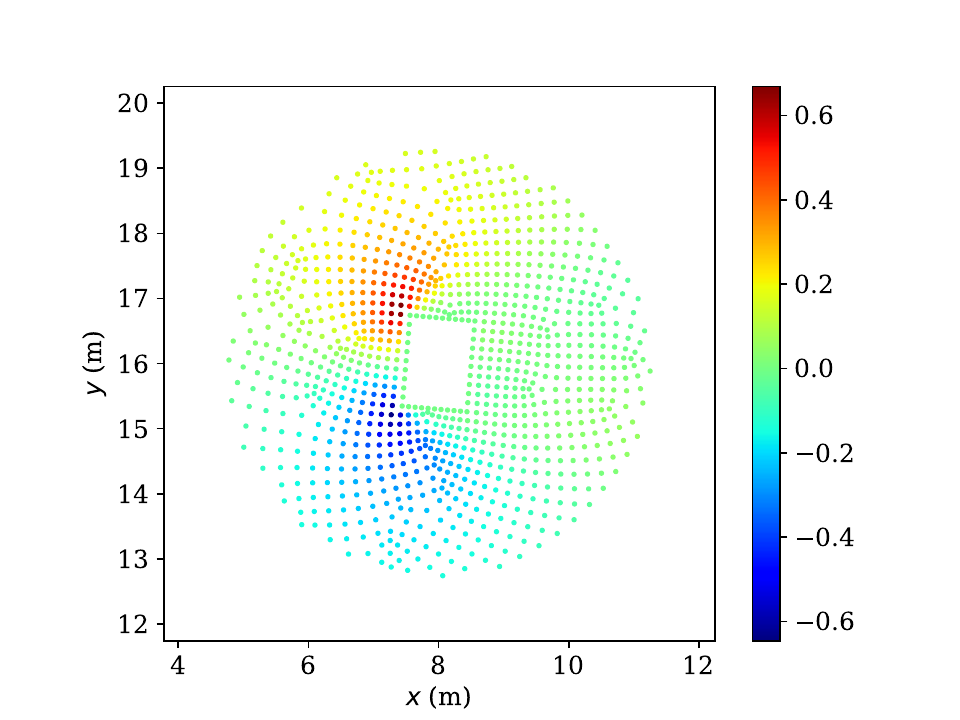}
    \end{subfigure}
     \begin{subfigure}[b]{0.24\textwidth}
        \centering
        \includegraphics[width=\textwidth]{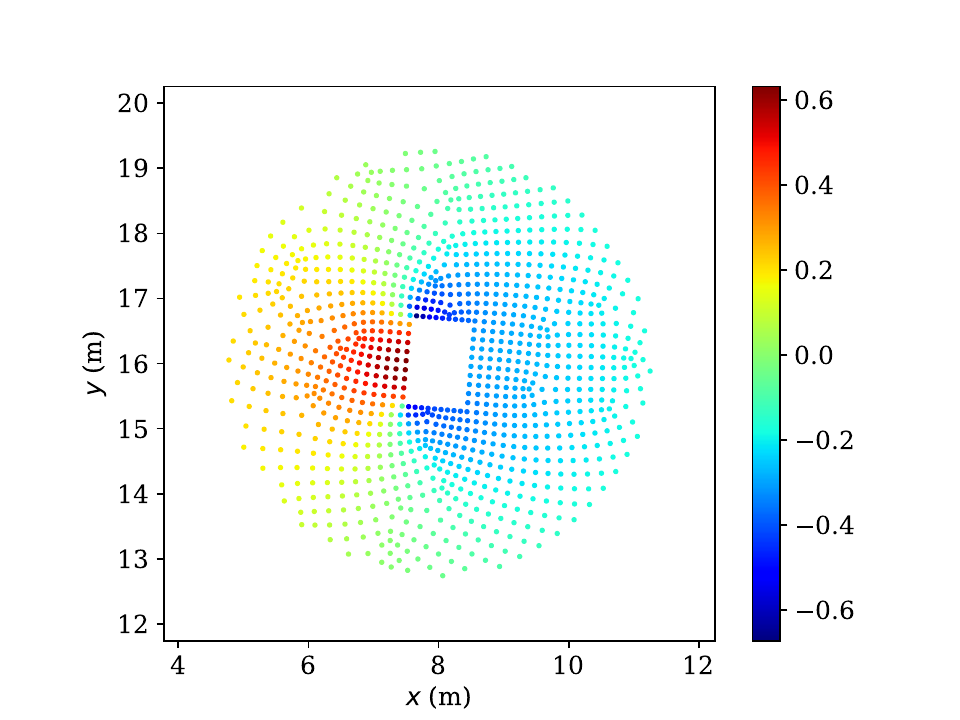}
    \end{subfigure}
    
    
    \begin{subfigure}[b]{0.24\textwidth}
        \centering
        \includegraphics[width=\textwidth]{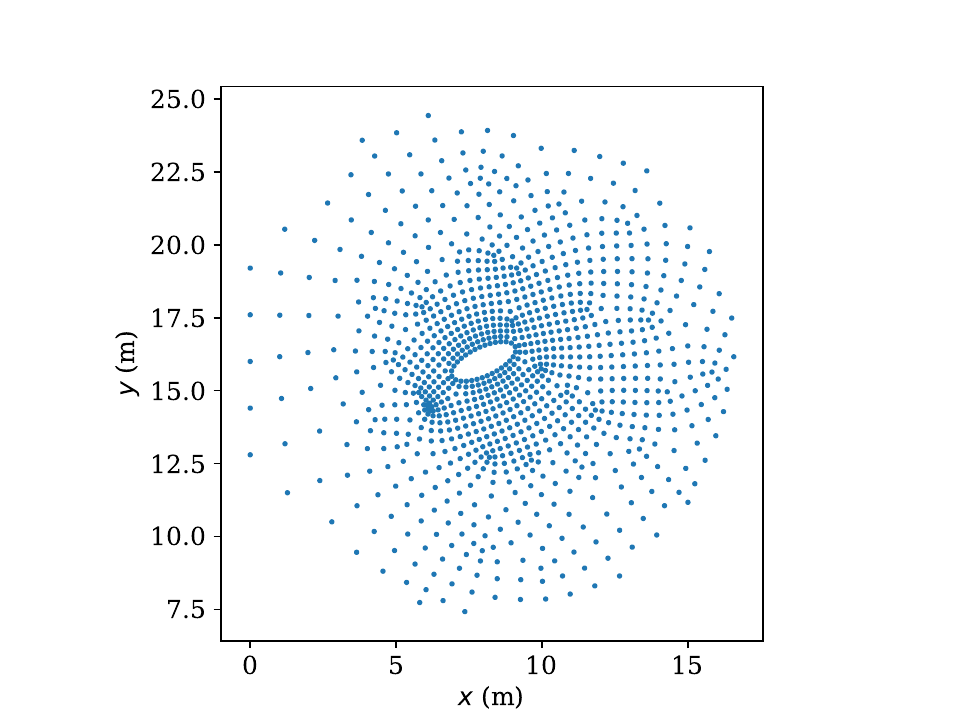}
    \end{subfigure}
    \begin{subfigure}[b]{0.24\textwidth}
        \centering
        \includegraphics[width=\textwidth]{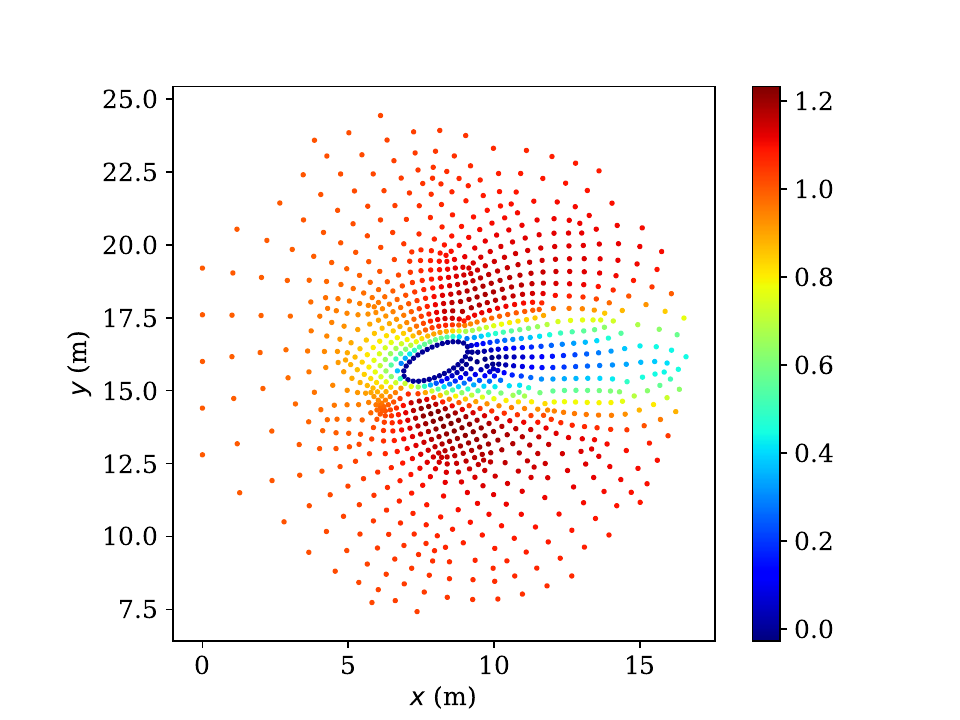}
    \end{subfigure}
    \begin{subfigure}[b]{0.24\textwidth}
        \centering
        \includegraphics[width=\textwidth]{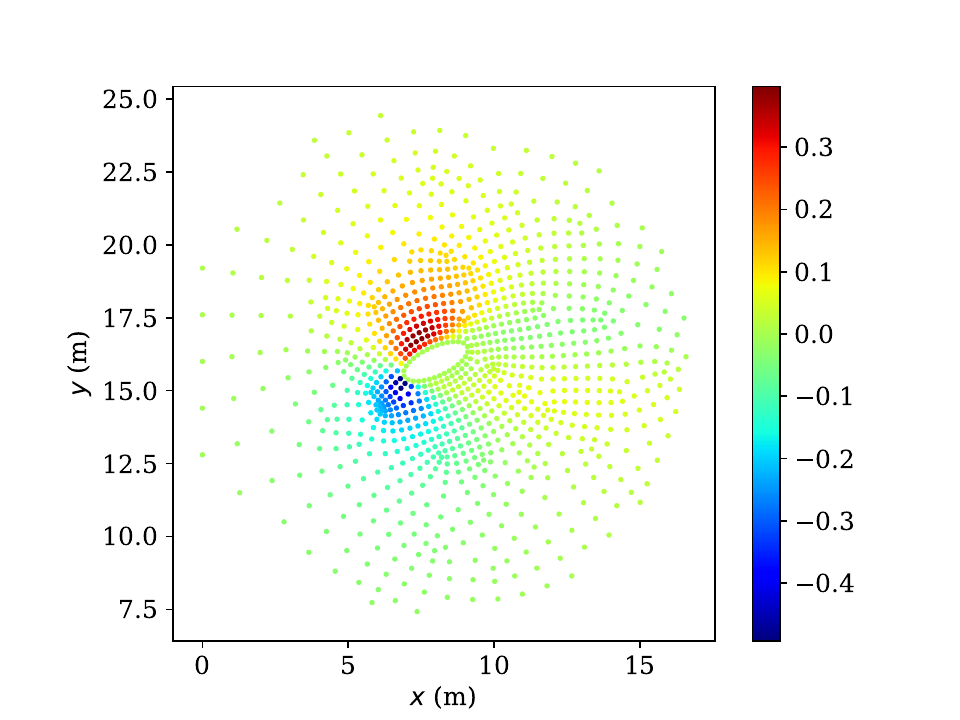}
    \end{subfigure}
     \begin{subfigure}[b]{0.24\textwidth}
        \centering
        \includegraphics[width=\textwidth]{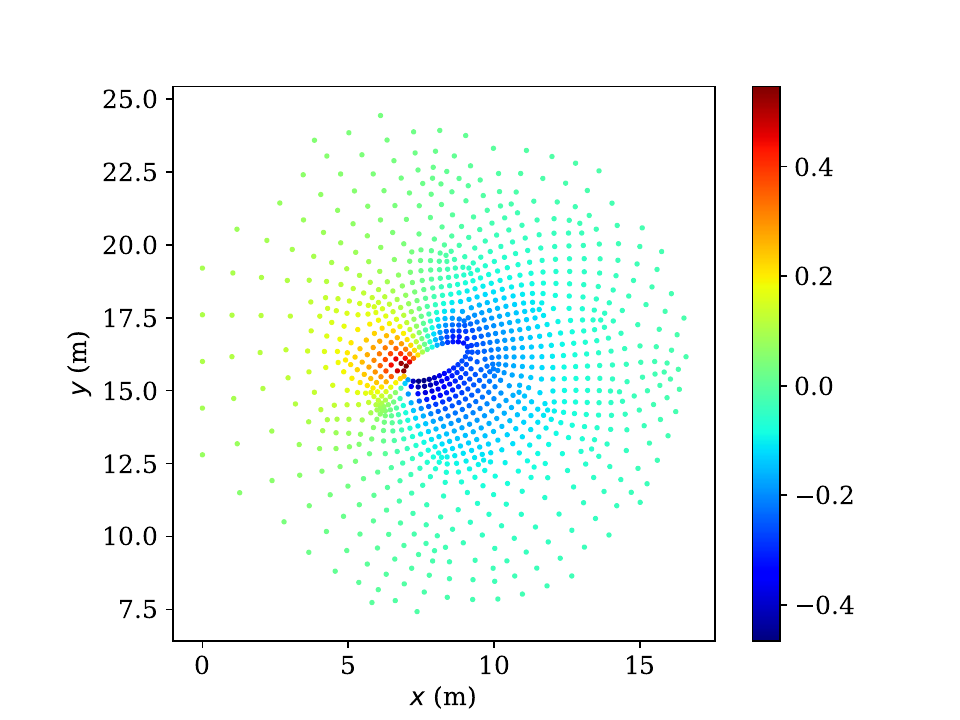}
    \end{subfigure}

  \caption{Examples of input and output of the generated dataset in the form of point clouds}
  \label{Fig2}
\end{figure}


\subsection{Data generation}
\label{Sect22}

We employ Gmsh \cite{geuzaine2009gmsh} to discretize the domain $V$ using unstructured finite volume meshes. Figure \ref{Fig1} exhibits two examples of generated meshes for cylinders with triangular and elliptical cross sections. We utilize the OpenFOAM solver \citep{weller1998tensorial} with the Semi-Implicit Method for Pressure Linked Equations (SIMPLE) \citep{caretto1973two} to numerically solve Eqs. (\ref{Eq1})--(\ref{Eq2}) under the conditions and assumptions described above. Numerical simulations are conducted until the $L^2$ norm of residuals of the discretized Eqs. (\ref{Eq1})--(\ref{Eq2}) reaches $10^{-3}$. A solution is considered a steady state once this criterion is met. The solution of the velocity and pressure fields in Eqs. (\ref{Eq1})--(\ref{Eq2}) is a function of the geometry of the cylinder's cross-section. Hence, to generate the labeled data, we consider seven different classical geometries for the cross-section, with varying sizes and rotations, as depicted and described in Table \ref{Table1}. In our computational setting, the cross-sectional geometry of the cylinder has a characteristic length $L$. Note that the Reynolds number can be expressed as

\begin{equation}
    \text{Re} = \frac{\rho L u_\infty}{\mu},
    \label{Eq6}
\end{equation}
and thus, it is a function of $L$. As illustrated in Table \ref{Table1}, the data set includes cross-sections with the shapes of circle \citep{behr1995incompressible,kashefiCoarse1,kashefiCoarse3}, square \citep{sen2011flow}, triangle \citep{kumar2006numerical}, rectangle \citep{zhong2019flow}, ellipse \citep{mittal1996direct}, pentagon \citep{abedin2017simulation}, and hexagon \citep{abedin2017simulation}, each with different length scales and orientations. Referring to Table \ref{Table1}, the characteristic length ($L$) is $a$ for cross-sections shaped as a circle, equilateral hexagon, equilateral pentagon, square, and equilateral triangle, whereas for the rectangle, ellipse, and triangle, the characteristic length ($L$) is $b$. The range of the Reynolds number in the data set varies from 20.0 to 76.0. The total number of labeled data is 2235. A similar data set was generated and used by Kashefi et al. \citep{kashefi2021PointNet}. It is noteworthy that the process of data generation, starting from defining the domain geometry, mesh generation, executing the numerical solver, and storing the velocity and pressure fields, is fully automated using C++ codes, Python scripts, journal files, and batch files.

In the next step, we introduce $u^{*}$, $v^{*}$, and $p^{*}$ as dimensionless variables associated respectively with $u$, $v$, and $p$. They are defined as follows:

\begin{equation}
    u^{*} = \frac{u}{u_\infty},
    \label{Eq7}
\end{equation}

\begin{equation}
    v^{*} = \frac{v}{u_\infty},
    \label{Eq8}
\end{equation}

\begin{equation}
    p^{*} = \frac{p-p_0}{\rho u_\infty^2},
    \label{Eq9}
\end{equation}
where the atmospheric pressure is indicated by $p_0$. As the next stage, we scale the spatial coordinates as well as the velocity and pressure fields of the training set to a range of [$-1$, 1] using the following formulation:

\begin{equation}
    \left\{\phi'\right\} = 2\left(\frac{\left\{\phi\right\}  - \min(\left\{\phi\right\} )}{\max(\left\{\phi\right\} ) - \min(\left\{\phi\right\} )}\right) - 1,
    \label{Eq10}
\end{equation}
where the set $\left\{\phi\right\}$ includes $\left\{x^*\right\}$, $\left\{y^*\right\}$, $\left\{u^*\right\}$, $\left\{v^*\right\}$, and $\left\{p^*\right\}$. According to Eq. (\ref{Eq10}), the scaled input and output data are represented by $\left\{x'\right\}$, $\left\{y'\right\}$, $\left\{u'\right\}$, $\left\{v'\right\}$, and $\left\{p'\right\}$. Note that scaling the input spatial coordinates to the range of [$-1$, 1] is critical for KA-PointNet; otherwise, the training loss diverges. The prediction of KA-PointNet will be scaled back to the physical domain using a similar process.

The input of KA-PointNet is a set of spatial points, and the output is the corresponding velocity and pressure values at those points. To represent the generated data set as point clouds suitable for feeding into KA-PointNet, we select the grid vertices of finite volume meshes as points. Note that a similar practice has been conducted by Kashefi et al. \citep{kashefi2021PointNet}. Additionally, since we are more interested in the solution of the velocity and pressure fields around the cylinder and in the wake region compared to the far field, we select the first $N$ closest grid vertices to the center of mass of the cylinder to establish the point clouds. In this study, we set $N=1024$, so each point cloud contains 1024 points. Because the distribution of grid vertices in the unstructured finite volume meshes is nonuniform, the range of spatial coordinates of points in each point cloud varies. In the described data set, $x_\text{min} \in$ [0, 5.46 m], $x_\text{max} \in$ [10.55 m, 22.14 m], $y_\text{min} \in$ [0, 13.41 m], and $y_\text{max} \in$ [18.57 m, 32 m]. A few examples of input and output of the labeled data set are shown in Fig. \ref{Fig2}. Note that these types of flexibility in input domain definition are the advantage of point-cloud and graph-based neural networks compared to traditional CNNs (e.g., see Ref \citep{thuerey2020deep}). We randomly split the data into three categories of the training set (1772 data), validation set (241 data), and test set (222 data).


\begin{figure}[!htbp]
  \centering 
        \includegraphics[width=\textwidth]{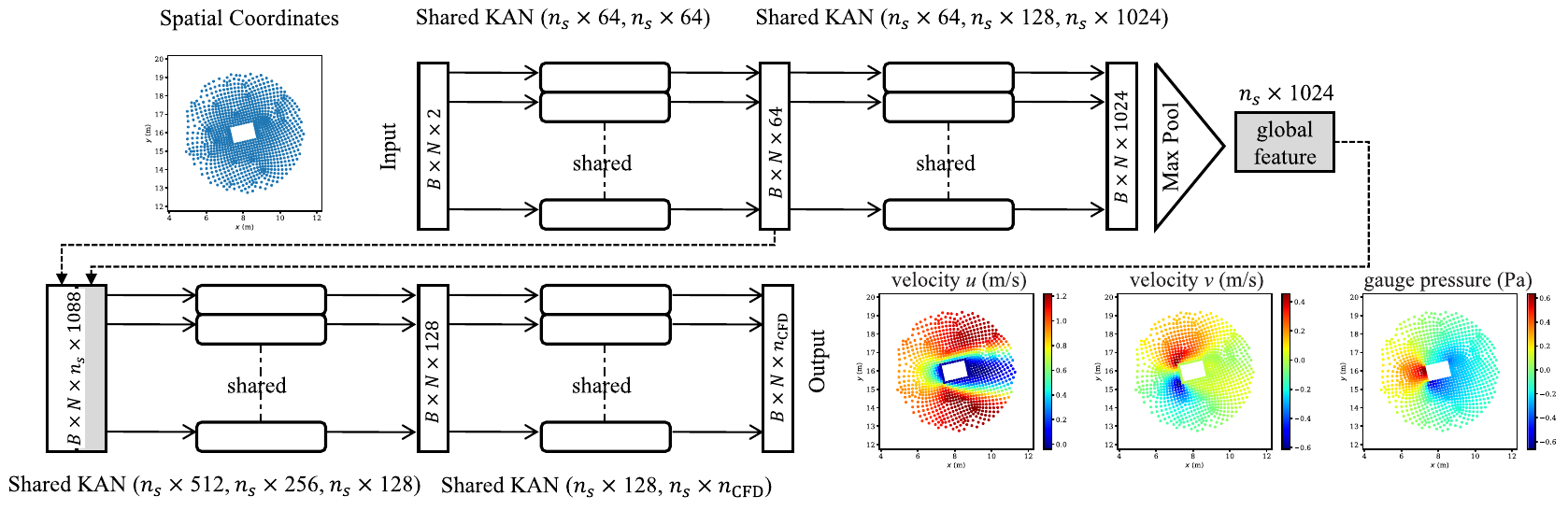}
  \caption{Architecture of the Kolmogorov-Arnold PointNet. Shared KANs with the labels $(\mathcal{B}_1, \mathcal{B}_2)$ and $(\mathcal{B}_1, \mathcal{B}_2, \mathcal{B}_3)$ are explained in the text. $n_{\text{CFD}}$ denotes the number of CFD variables. $N$ is the number of points in the point clouds. $B$ represents the batch size. $n_s$ is the global scaling parameter used to control the network size. Note that the velocity and pressure fields shown are schematic.}
  \label{Fig3}
\end{figure}


\section{Kolmogorov-Arnold PointNet}
\label{Sect3}

\subsection{Kolmogorov-Arnold network layers with Jacobi polynomials}
\label{Sect31}

In this subsection, we explain the concept of Kolmogorov-Arnold network layers. For simplicity, let us consider a Kolmogorov-Arnold network with one hidden layer, where the input of the network is a vector $\mathbf{r}$ of size $d_\text{input}$ and the output of the network is a vector $\mathbf{s}$ of size $d_\text{output}$. In this setup, the one-layer Kolmogorov-Arnold network relates the input to the output as

\begin{equation}
    \mathbf{s}_{d_\text{output}\times 1} =  \mathbf{\Phi}_{d_\text{output}\times d_\text{input}} \mathbf{r}_{d_\text{input}\times 1},
    \label{Eq13}
\end{equation}
where the matrix $\mathbf{\Phi}_{d_\text{output}\times d_\text{input}}$ is expressed as

\begin{equation}
    \mathbf{\Phi}_{d_\text{output}\times d_\text{input}} = 
    \left[
\begin{array}{cccc}
\psi_{1,1}(\cdot) & \psi_{1,2}(\cdot) & \cdots & \psi_{1,d_\text{input}}(\cdot) \\
\psi_{2,1}(\cdot) & \psi_{2,2}(\cdot) & \cdots & \psi_{2,d_\text{input}}(\cdot) \\
\vdots & \vdots & \ddots & \vdots \\
\psi_{d_\text{output},1}(\cdot) & \psi_{d_\text{output},2}(\cdot) & \cdots & \psi_{d_\text{output},d_\text{input}}(\cdot) \\
\end{array}
\right],
\label{Eq14}
\end{equation}
where $\psi(z)$ is defined as

\begin{equation}
    \psi(z) = \sum_{i=0}^n \Lambda_i  P_i^{(\alpha,\beta)}(z),
    \label{Eq15}
\end{equation}
where $P_i^{(\alpha,\beta)}(z)$ is the Jacobi polynomial of order $i$, $n$ is the order of the polynomial $\psi$, and $\Lambda_i$ are trainable parameters. Hence, the number of trainable parameters (i.e., $\Lambda_i$) of this KAN layer is equal to $(n+1)\times d_\text{output} \times d_\text{input}$. We implement $P_n^{(\alpha,\beta)}(z)$ recursively, using the following relationship \citep{Szego1939Orthogonal}

\begin{equation}
    P_n^{(\alpha,\beta)}(z) = (A_n z + B_n)P_{n-1}^{(\alpha,\beta)}(z) + C_n P_{n-2}^{(\alpha,\beta)}(z), 
     \label{Eq16}
\end{equation}
where $A_n$, $B_n$, and $C_n$ coefficients are defined as follow

\begin{equation}
    A_n = \frac{(2n+\alpha+\beta-1)(2n+\alpha+\beta)}{2n(n+\alpha+\beta)},
     \label{Eq17}
\end{equation}

\begin{equation}
    B_n = \frac{(2n+\alpha+\beta-1)(\alpha^2 - \beta^2)}{2n(n+\alpha+\beta)(2n+\alpha+\beta-2)},
     \label{Eq18}
\end{equation}

\begin{equation}
    C_n = \frac{-2(n+\alpha-1)(n+\beta-1)(2n+\alpha+\beta)}{2n(n+\alpha+\beta)(2n+\alpha+\beta-2)},
     \label{Eq19}
\end{equation}
along with the following two initial conditions:

\begin{equation}
    P_0^{(\alpha,\beta)}(z) = 1,
     \label{Eq20}
\end{equation}

\begin{equation}
    P_1^{(\alpha,\beta)}(z) = \frac{1}{2}(\alpha+\beta+2)z + \frac{1}{2}(\alpha-\beta).
     \label{Eq21}
\end{equation}
Since $P_n^{(\alpha,\beta)}(z)$ is constructed recursively, $P_i^{(\alpha,\beta)}(z)$, for ($0 \leq i \leq n$), are constructed along the way as well. Moreover, because the input of Jacobi polynomials must fall in the interval [$-$1, 1], the input vector $\mathbf{r}$ should be scaled in this interval before being fed into the KAN layer. To handle this situation, we use the hyperbolic tangent function defined as follows

\begin{equation}
 \tanh(z) = \frac{e^{2z} - 1}{e^{2z} + 1}.
  \label{Eq22}
\end{equation}
Other researchers have also used this strategy (e.g., see Refs. \citep{shukla2024comprehensive,KANwithTANH,aghaei2024Fractional}). Other reasonable strategies might be used for this scaling. For example, one option is to use the following formula:

\begin{equation}
    \left\{z\right\} = 2\left(\frac{\left\{z\right\} - \min(\left\{z\right\})}{\max(\left\{z\right\}) - \min(\left\{z\right\})}\right) - 1.
    \label{EqAA}
\end{equation}
The motivation for this choice is that since we scale both input and output data using the formula given in Eq. (\ref{Eq10}), a similar scaling scheme for the intermediate layers of KA-PointNet might be beneficial. However, our machine learning experiments show that using the hyperbolic tangent function is more efficient in terms of prediction accuracy. Therefore, we use it in this study.

Several sequential KAN layers establish a KAN component. Specifically, we implement shared KAN components in PointNet. We denote a shared KAN component with two layers of sizes $\mathcal{B}_1$ and $\mathcal{B}_2$ as ($\mathcal{B}_1$, $\mathcal{B}_2$). Similarly, we define a shared KAN component with three layers as ($\mathcal{B}_1$, $\mathcal{B}_2$, $\mathcal{B}_3$). After each KAN layer, we implement batch normalization \citep{ioffe2015batch}. This implementation is mandatory; otherwise, the training loss will diverge.

At the end of this subsection, we note that with the choice of $\alpha = \beta = 0$, the Legendre polynomial is obtained \citep{MiltonHandbook,Szego1939Orthogonal}. The Chebyshev polynomial of the first and second kinds are respectively covered with the choice of $\alpha = \beta = -0.5$ and $\alpha = \beta = 0.5$ \citep{MiltonHandbook,Szego1939Orthogonal}. Additionally, the Gegenbauer polynomial (or ultraspherical polynomials) is obtained if $\alpha = \beta$ \citep{Szego1939Orthogonal}.

\subsection{Architecture}
\label{Sect32}

The fundamental concept of KA-PointNet is to use shared KANs instead of shared MLPs. To explain the architecture of KA-PointNet, we first discuss the general aspects and then delve into the details and formulation of shared KANs. More specifically, we will explain why it is essential to implement KANs as ``shared'' layers.

Figure \ref{Fig3} depicts the architecture of KA-PointNet. To explain the architecture, let us first denote the number of labeled data in a desired set by $m$. The desired set can be the training set, validation set, or test set. In a data set with $m$ labeled data, each domain $V_i$ ($1 \leq i \leq m$) in the set is represented by a point cloud $\mathcal{X}_i$ with $N$ points, where $\mathcal{X}_i = \left\{ \boldsymbol{x}_j \in \mathbb{R}^d \right\}_{j=1}^{N}$. The spatial dimension of $V_i$ and consequently $\mathcal{X}_i$ is denoted by $d$. Because our focus is on two-dimensional problems in this study, we set $d=2$. Accordingly, $\boldsymbol{x}_j$ ($1 \leq j \leq N$) is the vector representing the spatial coordinates of each point in the point cloud $\mathcal{X}_i$. We indicate the $x$ and $y$ components of each $\boldsymbol{x}_j$, respectively, by $x'_j$ and $y'_j$. The task of KA-PointNet is to provide an end-to-end mapping from $\mathcal{X}_i$ to $\mathcal{Y}_i$, where $\mathcal{Y}_i = \left\{ \boldsymbol{y}_j \in \mathbb{R}^{n_{\text{CFD}}} \right\}_{j=1}^{N}$. We demonstrate the number of desired predicted fields by $n_\text{CFD}$. In this study, we are particularly interested in the prediction of the velocity vector (in two dimensions) and the pressure field. Hence, $n_\text{CFD}=3$. Based on the given definition, $\boldsymbol{y}_j$ is the vector corresponding to the predicted fields at the spatial point $\boldsymbol{x}_j$, and thus, has three components of $u'_j$, $v'_j$, and $p'_j$. This procedure can be mathematically formulated as: 

\begin{equation}
\left(u'_j, v'_j, p'_j\right) = f \left(\left(x'_j, y'_j\right), g\left(\mathcal{X}_i\right)\right); \quad \forall \left(x'_j, y'_j\right) \in \mathcal{X}_i \text{ and } \forall \left(u'_j, v'_j, p'_j\right) \in \mathcal{Y}_i \text{ with } 1 \leq i \leq m \text{ and } 1 \leq j \leq N,
 \label{Eq11}
\end{equation}
where $f$ is the mapping function representing KA-PointNet. 
KA-PointNet is invariant to any of the $N!$ permutations of the input vector. In other words, if the input vector ($\mathcal{X}_i$) is randomly permuted, the geometry of the domain remains unchanged, and thus the solution ($\mathcal{Y}_i$) should also remain unchanged. KA-PointNet achieves this permutation invariance through a symmetric function and the use of shared KANs. In this sense, $g$ is a symmetric function encoding the geometric features of the point cloud $\mathcal{X}_i$. Among the possible options for the symmetric function $g$, we follow the original function introduced in Ref. \cite{qi2017pointnet}, which is the maximum function, such that

\begin{equation}
g\left(\mathcal{X}_i\right) = \max \left(h\left(x'_1, y'_1\right), \ldots, h\left(x'_N, y'_N\right)\right); \quad \forall \left(x'_j, y'_j\right) \in \mathcal{X}_i \text{ with } 1 \leq i \leq m \text{ and } 1 \leq j \leq N, 
 \label{Eq12}
\end{equation}
where $h$ is a function representing two shared KAN layers in the first branch of KA-PointNet (see Fig. \ref{Fig3}). From a computer science perspective, $g\left(\mathcal{X}_i\right)$ is regarded as the global feature in the KA-PointNet architecture, as shown in Fig. \ref{Fig3}. The key idea behind using KA-PointNet, as a geometric deep learning model, is that the predicted fields at each spatial point depend on both the spatial coordinates of that specific point and the overall geometry of the domain constructed by all the points, including the specific point. This fact can be seen in Eqs. (\ref{Eq11})--(\ref{Eq12}).

We introduce $n_s$ as a global scaling variable that controls the size of KA-PointNet. Furthermore, We denote the batch size by $B$, referring to the number of point clouds (i.e., $\mathcal{X}_i$ and $\mathcal{Y}_i$ pairs) fed into KA-PointNet at each epoch. In Fig. \ref{Fig3} and similarly in Fig. \ref{Fig17}, the notation ($\mathcal{B}_1$,$\mathcal{B}_2$) represents two sequential hidden layers, where the first layer contains $\mathcal{B}_1$ neurons and the second layer contains $\mathcal{B}_2$ neurons. The notation of ($\mathcal{B}_1$,$\mathcal{B}_2$,$\mathcal{B}_3$) is similarly defined. As can be seen from Fig. \ref{Fig3}, the input of KA-PointNet is a three-dimensional tensor of size $B \times N \times 2$. Following this, two sequential shared KANs are applied, with sizes ($n_s \times 64$, $n_s \times 64$) and ($n_s \times 64$, $n_s \times 128$, $n_s \times 1024$), as shown in Fig. \ref{Fig3}. The maximum function then generates the global feature with a size of $n_s \times 1024$. As illustrated in Fig. \ref{Fig3}, this global feature is concatenated with an intermediate feature tensor of size $B \times N \times (n_s \times 64)$, resulting in a new tensor with dimension $B \times N \times (n_s \times 1088)$. Next, two additional sequential shared KANs are applied within the KA-PointNet architecture, with sizes ($n_s \times 512$, $n_s \times 256$, $n_s \times 128$) and ($n_s \times 128$, $n_\text{CFD}$), respectively. The result of the previous step is a tensor with dimensions $B \times N \times n_\text{CFD}$, as depicted in Fig. \ref{Fig3}. It is important to note that suitable values for $n_s$ should lead to positive integers for the size of shared KANs.

\subsection{Shared Kolmogorov-Arnold networks}
\label{Sect33}

The idea of shared KANs is distinct from regular fully connected layers. Here, we illustrate the concept of shared KANs with a straightforward example. Consider the initial layer in the shared KAN given in the first branch of KA-PointNet, which has a size of ($n_s \times 64$, $n_s \times 64$), as shown in Fig. \ref{Fig3}. Assume $n_s = 1$ for simplicity. Particularly, we focus on the first shared KAN layer with a size of 64. The transposed input vector $\mathcal{X}_i$ can be expressed as

\begin{equation}
    \mathcal{X}_i^{\text{tr}} =
\begin{bmatrix}
x'_1 & x'_2 & \cdots & x'_N \\
\\
y'_1 & y'_2 & \cdots & y'_N
\end{bmatrix}.
 \label{Eq23}
\end{equation}
After processing $\mathcal{X}_i$ through the first shared KAN layer, the result is a matrix of size $64 \times N$ and can be described as:

\begin{equation}
\begin{bmatrix}
\mathbf{s}_{64\times1}^{(1)} & \mathbf{s}_{64\times1}^{(2)} & \cdots & \mathbf{s}_{64\times1}^{(N)}
\end{bmatrix},
 \label{Eq24}
\end{equation}
where $\mathbf{s}_{64\times1}^{(1)}$, $\mathbf{s}_{64\times1}^{(2)}$, $\cdots$, $\mathbf{s}_{64\times1}^{(N)}$ are vectors, computed as follows:

\begin{equation}
\begin{aligned}
\mathbf{s}^{(1)}_{64 \times 1} &=   \mathbf{\Phi}_{64 \times 2} \begin{bmatrix} x'_1 \\ \\ y'_1 \end{bmatrix}, \\
\mathbf{s}^{(2)}_{64 \times 1} &=   \mathbf{\Phi}_{64 \times 2} \begin{bmatrix} x'_2 \\ \\ y'_2 \end{bmatrix}, \\
&\vdots \\
\mathbf{s}^{(N)}_{64 \times 1} &= \mathbf{\Phi}_{64 \times 2} \begin{bmatrix} x'_N \\ \\ y'_N \end{bmatrix},
\end{aligned}
 \label{Eq25}
\end{equation}
where $\mathbf{\Phi}_{64 \times 2}$ is the shared KAN layer. As can be observed in Eq. (\ref{Eq25}), the same (shared) $\mathbf{\Phi}_{64 \times 2}$ used for each spatial point in the domain, corresponding to the vector [$x'_j$ $y'_j$]$^\text{tr}$, where $1 \leq j \leq N$. This is the reason it is called shared KANs. This procedure is similarly applied to the remaining layers. With this strategy, it becomes evident that each point is independently processed within the KA-PointNet framework. The only instance when the points interact is when the global feature is determined (see Eq. (\ref{Eq12})).

\subsection{Training}
\label{Sect34}

Mean squared error is used as the loss function, expressed as
\begin{equation}
\mathcal{L} = \frac{1}{3\times N} \left(\sum_{i=1}^{N} \left[ (u'_i - \tilde{u}'_i)^2 + (v'_i - \tilde{v}'_i)^2 + (p'_i - \tilde{p}'_i)^2 \right]\right),
\label{Eq26}
\end{equation}
where $\tilde{u}'$, $\tilde{v}'$, and $\tilde{p}'$ are the predicted velocity and pressure fields, respectively. Note that for error analysis and visualization, predicted variables are scaled back into the physical domain. The Adam optimizer \citep{kingma2014adam} is used with parameters $\beta_1=0.9$, $\beta_2=0.999$, and $\hat{\epsilon}=10^{-8}$. We refer the audience to Ref. \citep{kingma2014adam} for the mathematical definitions of $\beta_1$, $\beta_2$, and $\hat{\epsilon}$ in the Adam optimizer \citep{kingma2014adam}. For a fair comparison between different setups for KA-PointNet, we perform all the training procedures on an NVIDIA A100 Tensor Core GPU with 80 gigabytes of RAM. We use mini-batch gradient descent with a batch size of 128 and a constant learning rate of $5 \times 10^{-4}$. To avoid overfitting, the evolution of the loss function for both the training and validation sets is monitored during the training procedure and an early stopping scheme is used. We discuss the number of trainable parameters, the history of the loss function, and the computational cost of training in detail in Sect. \ref{Sect4}.


\begin{table}[width=.9\linewidth,cols=6,pos=!htbp]
\caption{Computational cost and error analysis of the velocity and pressure fields predicted by Kolmogorov-Arnold PointNet (i.e., KA-PointNet) for the test set containing 222 unseen geometries for different degrees of the Jacobi polynomial. Here, $n_s=1$ is set. In the Jacobi polynomial, $\alpha=\beta=1$ is set. $||\cdots||$ indicates the $L^2$ norm.}\label{Table2}
\begin{tabular*}{\tblwidth}{@{} LLLLLL@{} }
\toprule
Polynomial degree &  2 & 3  & 4 & 5 & 6 \\
\midrule
Average $||\Tilde{u}-u||/||u||$ & 1.43377E$-$2 & 1.73537E$-$2 & 1.67633E$-$2 & 1.17467E$-$2 & 1.07525E$-$2\\
Maximum $||\Tilde{u}-u||/||u||$ & 1.66709E$-$1 & 1.40088E$-$1 & 1.37067E$-$1 & 1.40614E$-$1 & 1.68387E$-$1\\
Minimum $||\Tilde{u}-u||/||u||$ & 5.40694E$-$3 & 7.22247E$-$3 & 6.70142E$-$3 & 5.01482E$-$3 & 4.82974E$-$3\\
\midrule
Average $||\Tilde{v}-v||/||v||$ & 5.28306E$-$2 & 5.75906E$-$2 & 5.38728E$-$2 & 4.81333E$-$2 & 5.08094E$-$2\\
Maximum $||\Tilde{v}-v||/||v||$ & 5.25991E$-$1 & 4.42245E$-$1 & 4.48167E$-$1 & 4.61926E$-$1 & 5.44426E$-$1\\
Minimum $||\Tilde{v}-v||/||v||$ & 2.79826E$-$2 & 2.05299E$-$2 & 2.51025E$-$2 & 2.11034E$-$2 & 2.36179E$-$2\\
\midrule
Average $||\Tilde{p}-p||/||p||$ & 3.49754E$-$2 & 4.75586E$-$2 & 4.67659E$-$2 & 3.30555E$-$2 & 3.47443E$-$2\\
Maximum $||\Tilde{p}-p||/||p||$ & 1.73144E$-$1 & 1.55694E$-$1 & 1.69091E$-$1 & 1.64282E$-$1 & 1.90192E$-$1\\
Minimum $||\Tilde{p}-p||/||p||$ & 1.71494E$-$2 & 1.75673E$-$2 & 1.93152E$-$2 & 1.50220E$-$2 & 1.44066E$-$2\\
\midrule
Training time & 3.71432 & 6.52910 & 10.12893 & 14.28542 & 19.26667\\
per epoch (s) &  &  &  &  & \\
\midrule
Number of trainable & 2660480 & 3545728 & 4430976 & 5316224 & 6201472 \\
parameters &  &  &  &  & \\
\bottomrule
\end{tabular*}
\end{table}


\begin{figure}[!htbp]
  \centering 
      \begin{subfigure}[b]{0.32\textwidth}
        \centering
        \includegraphics[width=\textwidth]{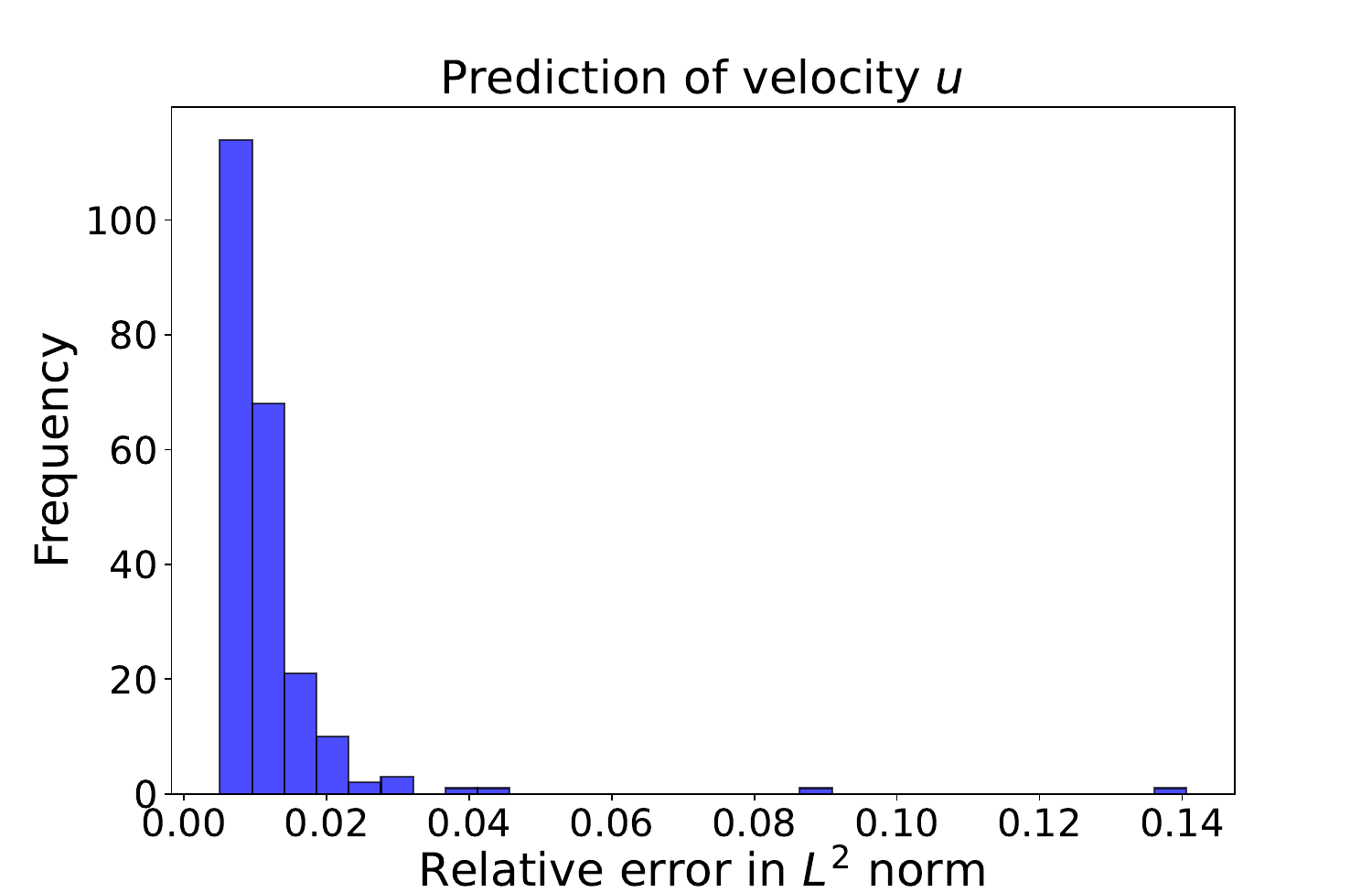}
    \end{subfigure}
    \begin{subfigure}[b]{0.32\textwidth}
        \centering
        \includegraphics[width=\textwidth]{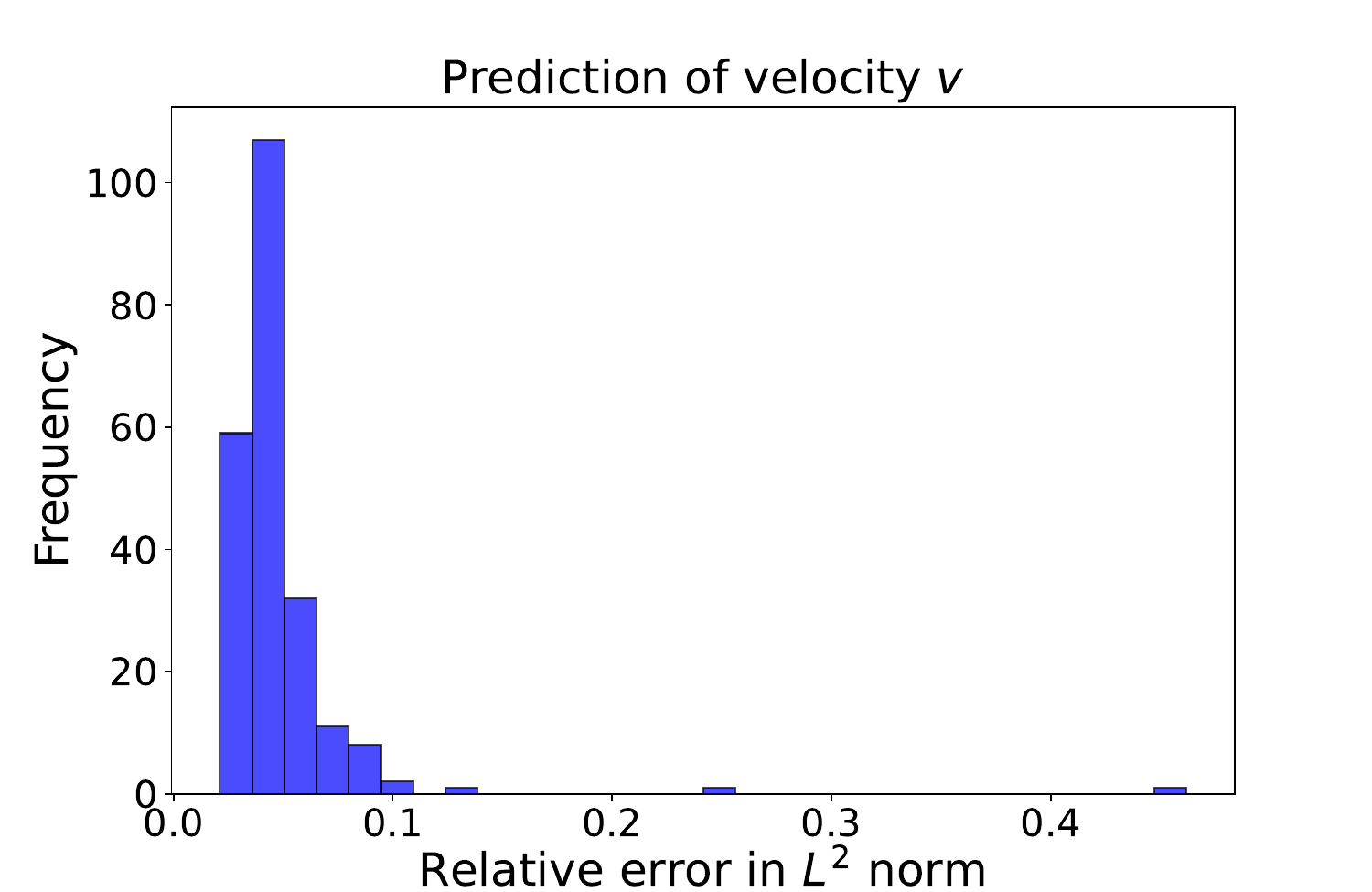}
    \end{subfigure}
    \begin{subfigure}[b]{0.32\textwidth}
        \centering
        \includegraphics[width=\textwidth]{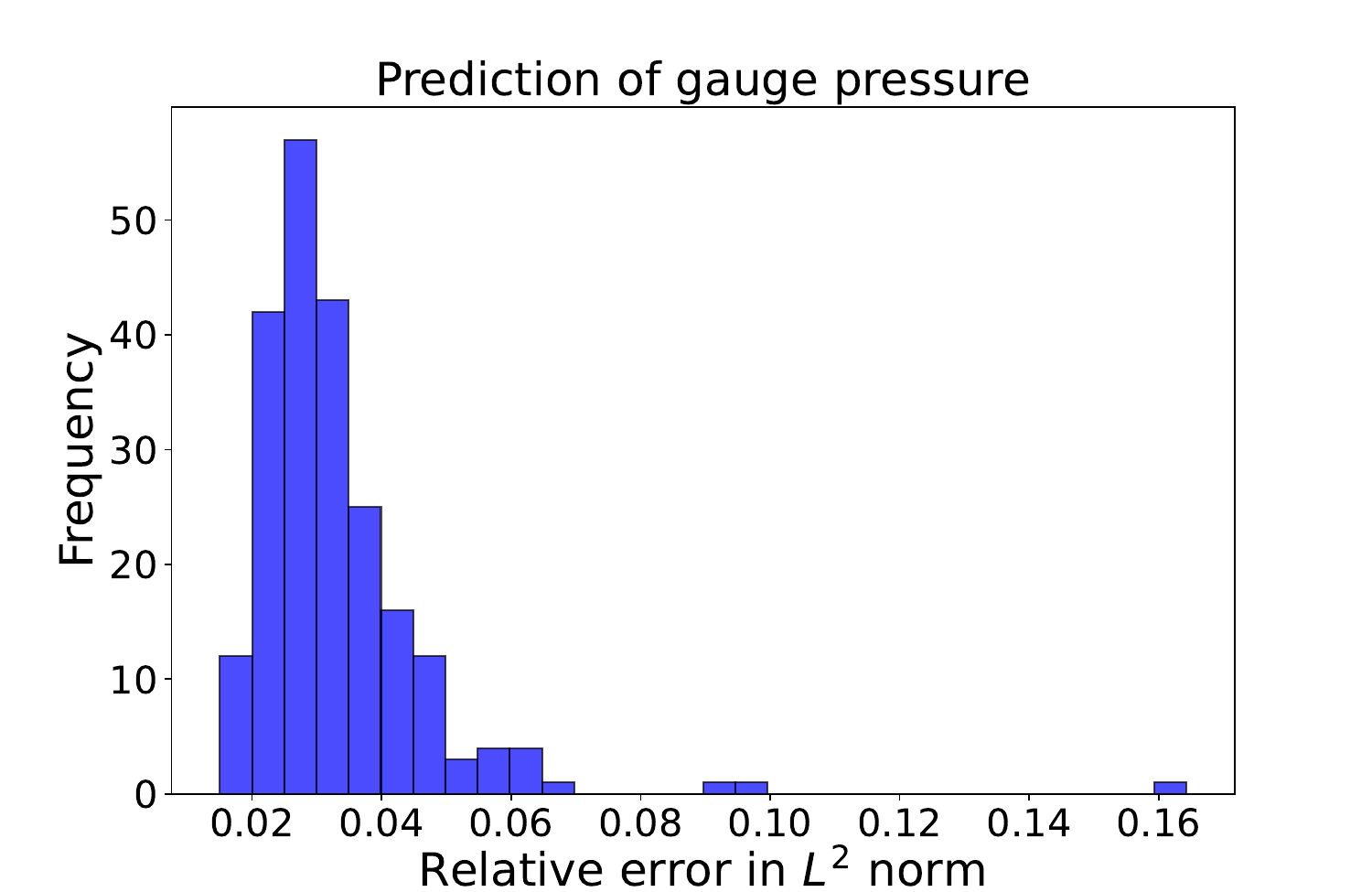}
    \end{subfigure}

  \caption{Histograms of the relative pointwise error in $L^2$ norm for the velocity and pressure fields predicted by Kolmogorov-Arnold PointNet (i.e., KA-PointNet). The Jacobi polynomial used has a degree of 5, with $\alpha = \beta = 1$. Here, $n_s=1$ is set.}
  \label{Fig15}
\end{figure}


\begin{figure}[!htbp]
  \centering 
      \begin{subfigure}[b]{0.24\textwidth}
      \caption{Ground truth}
        \centering
        \includegraphics[width=\textwidth]{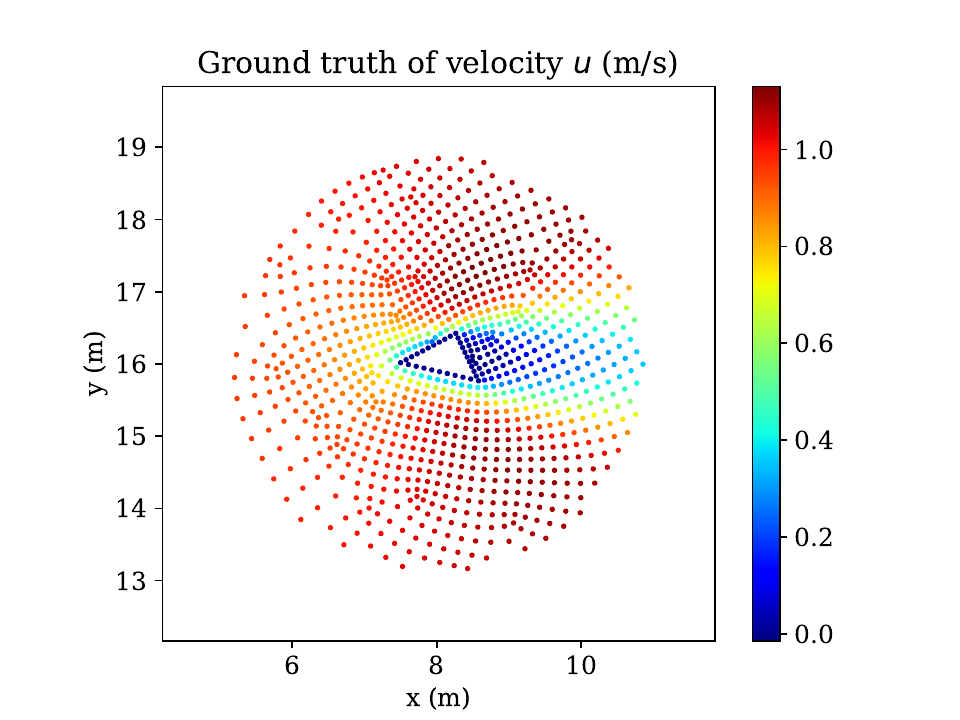}
    \end{subfigure}
    \begin{subfigure}[b]{0.24\textwidth}
    \caption{Prediction after 10 epochs}
        \centering
        \includegraphics[width=\textwidth]{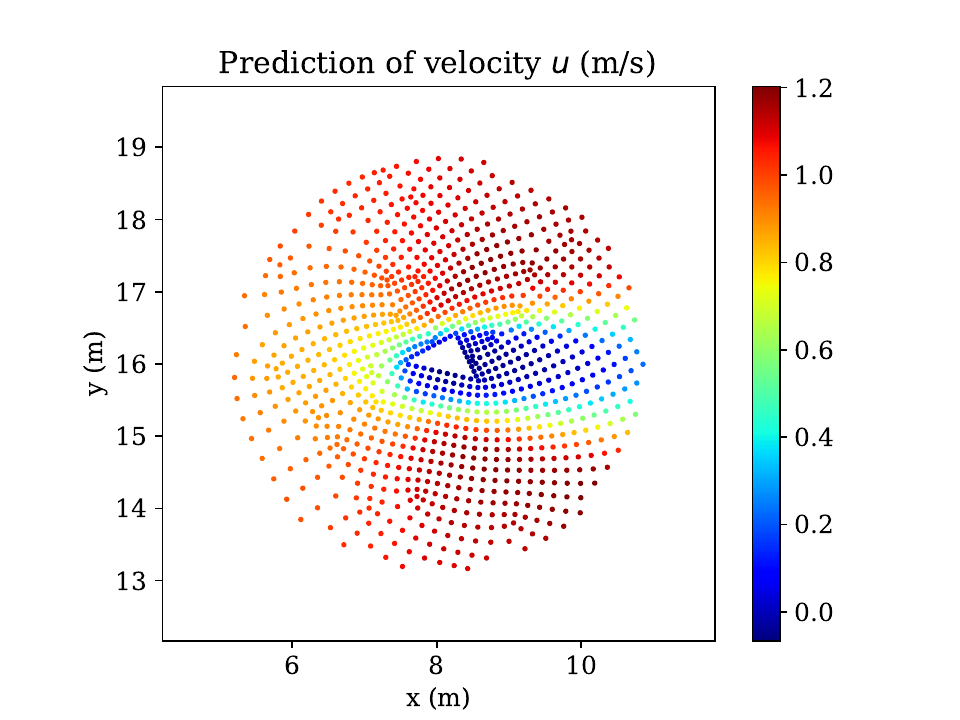}
    \end{subfigure}
    \begin{subfigure}[b]{0.24\textwidth}
    \caption{Prediction after 100 epochs}
        \centering
        \includegraphics[width=\textwidth]{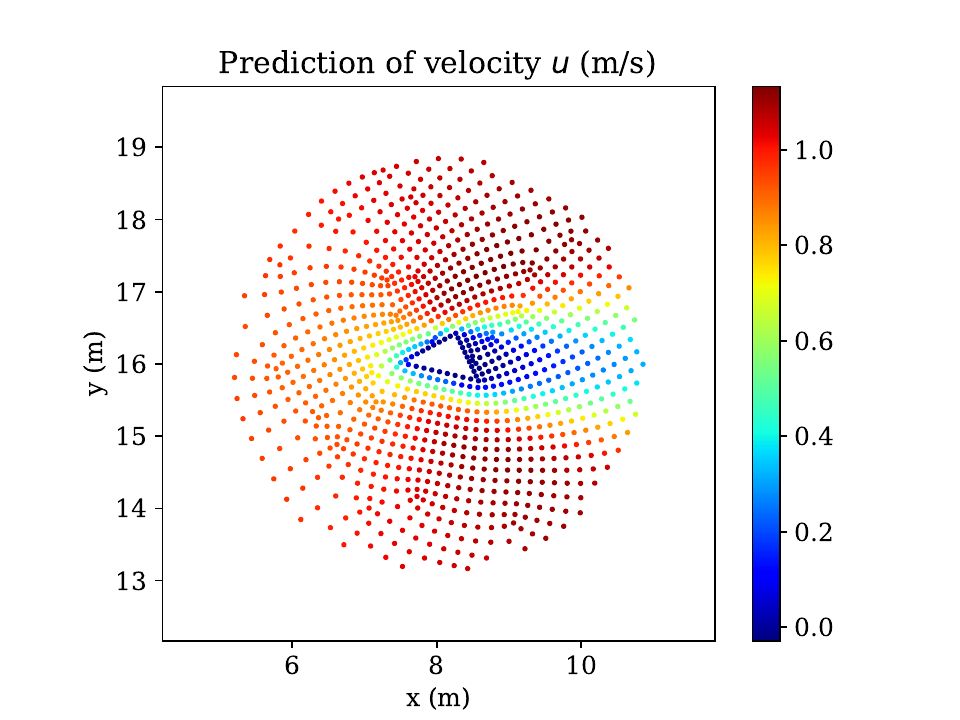}
    \end{subfigure}
     \begin{subfigure}[b]{0.24\textwidth}
     \caption{Prediction after 1000 epochs}
        \centering
        \includegraphics[width=\textwidth]{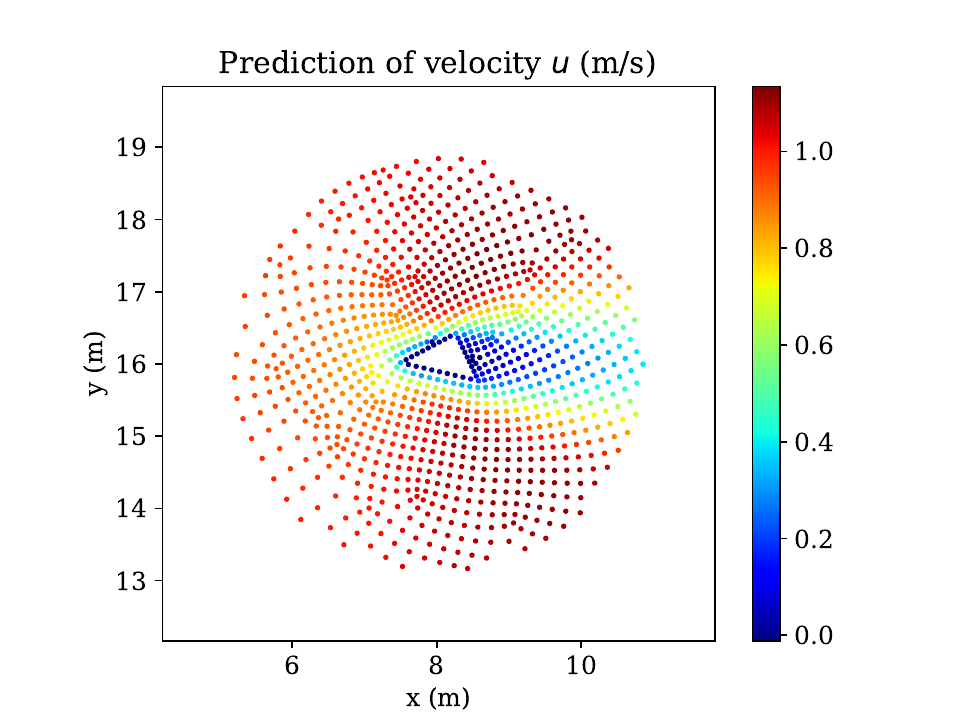}
    \end{subfigure}

    
    \begin{subfigure}[b]{0.24\textwidth}
        \centering
        \includegraphics[width=\textwidth]{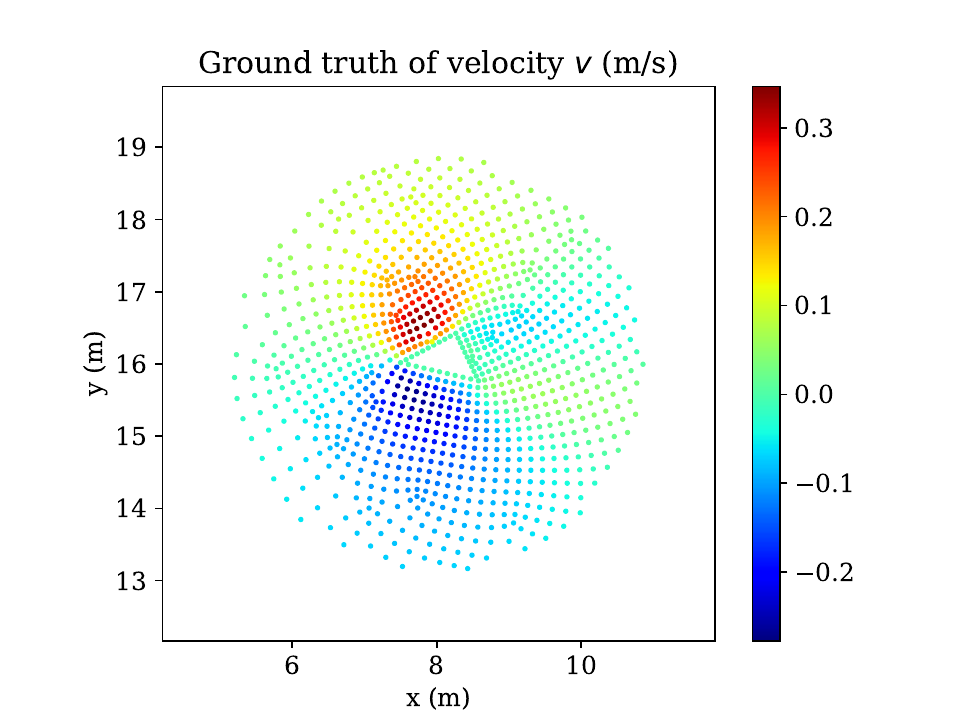}
    \end{subfigure}
    \begin{subfigure}[b]{0.24\textwidth}
        \centering
        \includegraphics[width=\textwidth]{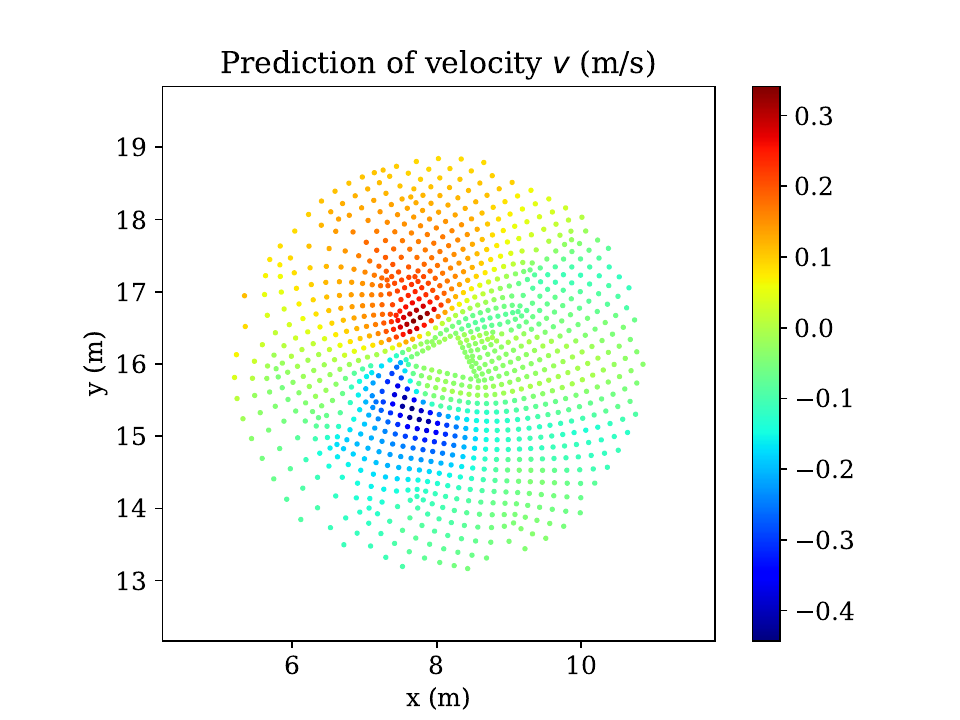}
    \end{subfigure}
    \begin{subfigure}[b]{0.24\textwidth}
        \centering
        \includegraphics[width=\textwidth]{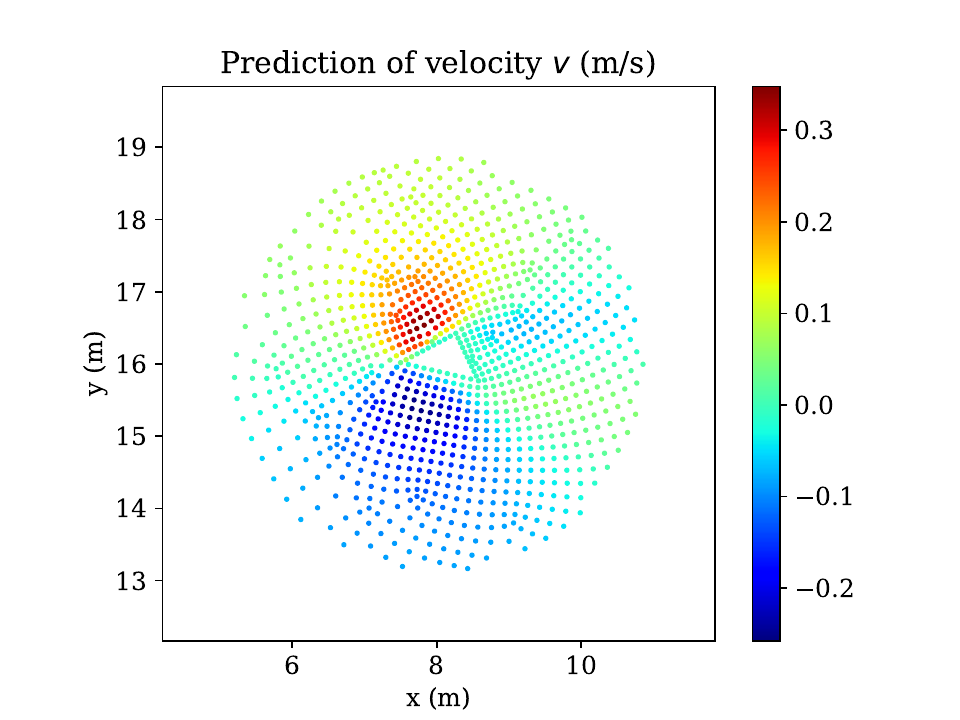}
    \end{subfigure}
     \begin{subfigure}[b]{0.24\textwidth}
        \centering
        \includegraphics[width=\textwidth]{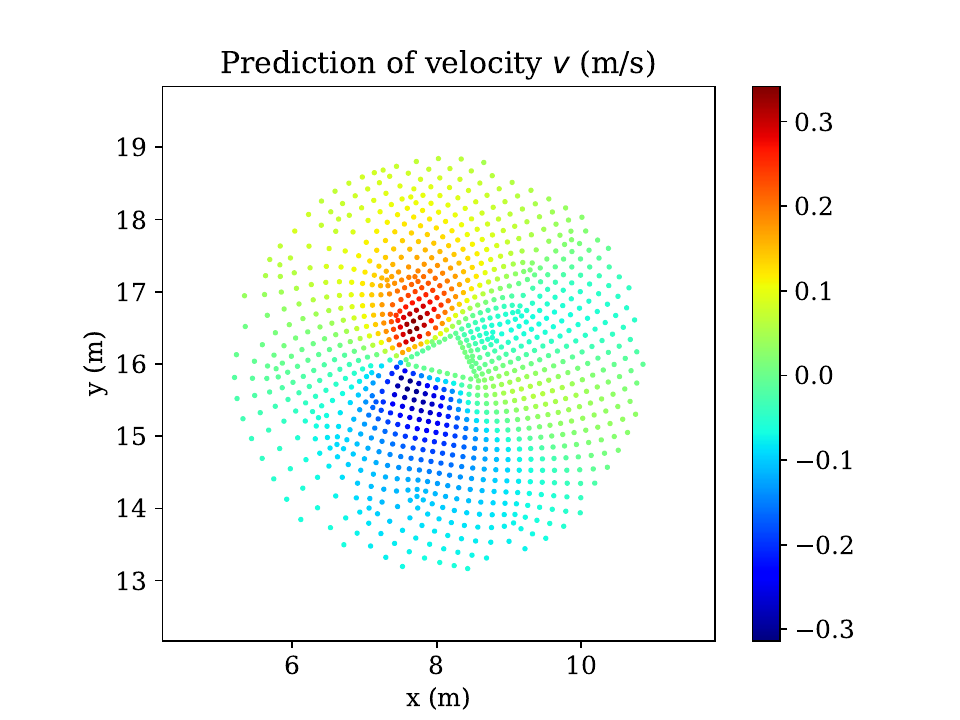}
    \end{subfigure}

    
    \begin{subfigure}[b]{0.24\textwidth}
        \centering
        \includegraphics[width=\textwidth]{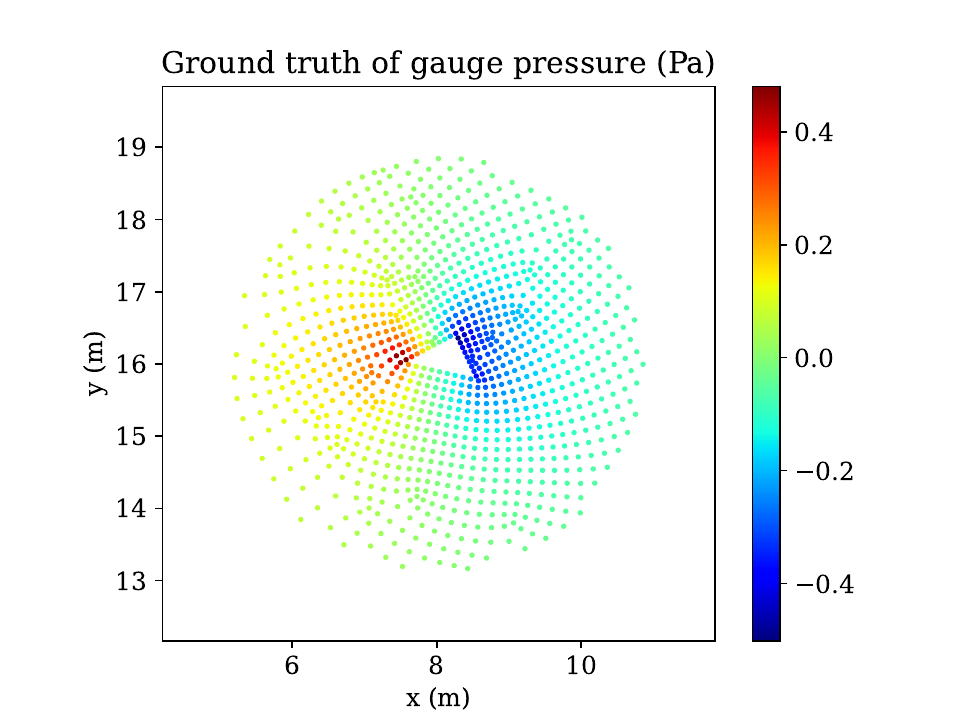}
    \end{subfigure}
    \begin{subfigure}[b]{0.24\textwidth}
        \centering
        \includegraphics[width=\textwidth]{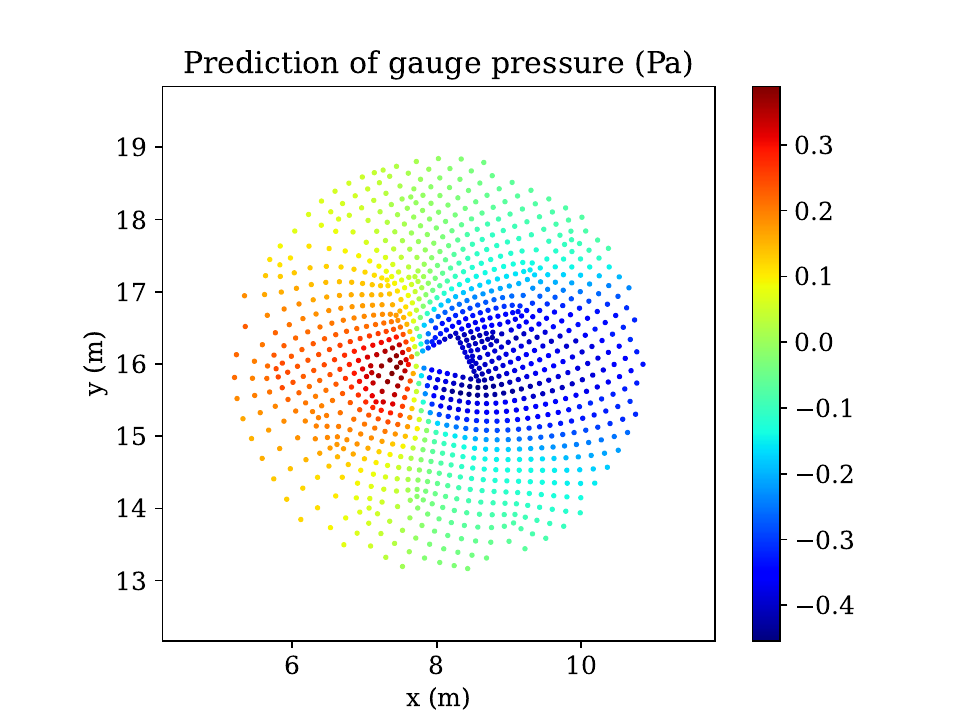}
    \end{subfigure}
    \begin{subfigure}[b]{0.24\textwidth}
        \centering
        \includegraphics[width=\textwidth]{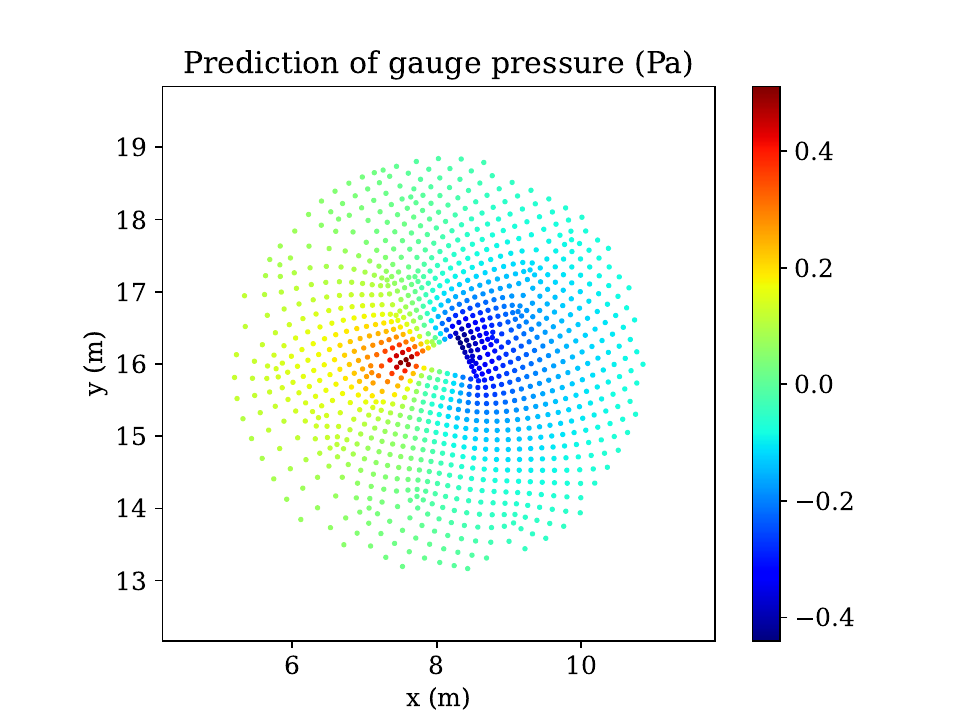}
    \end{subfigure}
     \begin{subfigure}[b]{0.24\textwidth}
        \centering
        \includegraphics[width=\textwidth]{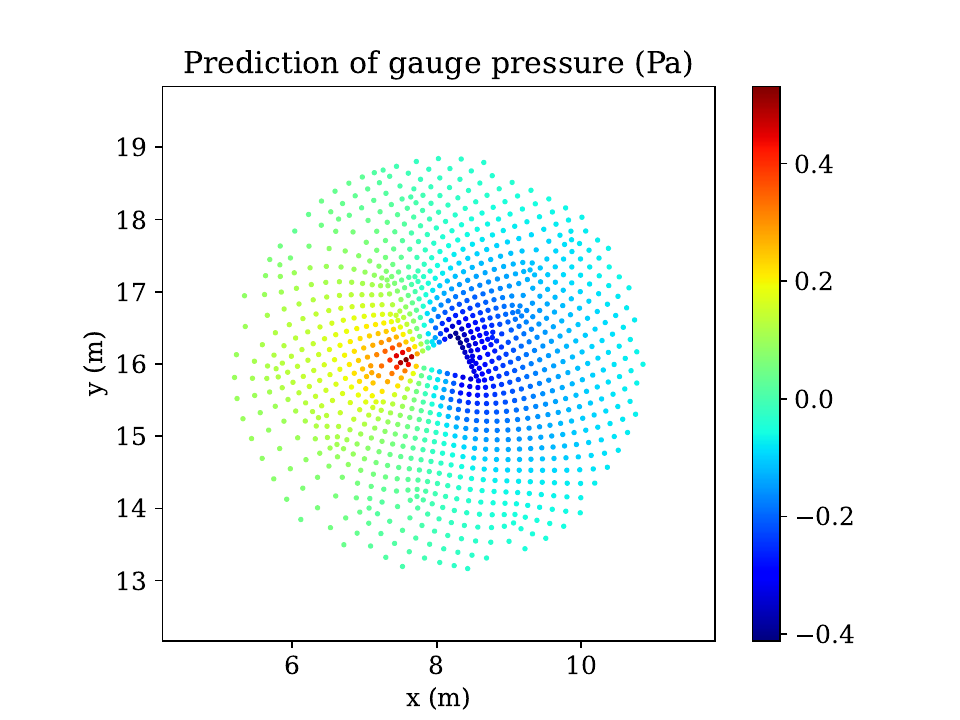}
    \end{subfigure}

  \caption{A comparison between the ground truth and prediction of the Kolmogorov-Arnold PointNet for the velocity and pressure fields after 10, 100, and 1000 epochs. The Jacobian polynomial used has a degree of 5, with $\alpha=\beta=1$. Here, $n_s=1$ is set.}
  \label{Fig4}
\end{figure}

\begin{table}[width=.9\linewidth,cols=7,pos=!htbp]
\caption{Error analysis of the velocity and pressure fields predicted by Kolmogorov-Arnold PointNet (i.e., KA-PointNet) for the test set containing 222 unseen geometries for different values of $\alpha$ and $\beta$ in Jacobi polynomials. The degree of the Jacobi polynomials used is 3. Here, $n_s=1$ is set. $||\cdots||$ indicates the $L^2$ norm.}\label{Table3}
\begin{tabular*}{\tblwidth}{@{} LLLLLLL@{} }
\toprule
 & $\alpha = \beta = 0$ & $\alpha = \beta = -0.5$ & $\alpha = \beta = 0.5$ & $\alpha = \beta = 1$ & $2 \alpha = \beta = 2$ & $ \alpha = 2\beta = 2$ \\
\midrule
Average $||\Tilde{u}-u||/||u||$ & 1.36404E$-$2 & 1.21807E$-$2 & 1.24601E$-$2 & 1.73537E$-$2 & 1.65403E$-$2 & 2.98503E$-$2 \\
Maximum $||\Tilde{u}-u||/||u||$ & 1.37743E$-$1 & 1.43526E$-$1 & 1.42853E$-$1 & 1.40088E$-$1 & 1.51214E$-$1 & 1.54898E$-$1 \\
Minimum $||\Tilde{u}-u||/||u||$ & 5.46195E$-$3 & 6.51784E$-$3 & 6.28078E$-$3 & 7.22247E$-$3 & 7.76342E$-$3 & 1.00267E$-$2 \\
\midrule
Average $||\Tilde{v}-v||/||v||$ & 6.38935E$-$2 & 4.70539E$-$2 & 4.32044E$-$2 & 5.75906E$-$2 & 5.26547E$-$2 & 9.47980E$-$2 \\
Maximum $||\Tilde{v}-v||/||v||$ & 4.34468E$-$1 & 4.40415E$-$1 & 4.76479E$-$1 & 4.42245E$-$1 & 5.06639E$-$1 & 4.42623E$-$1 \\
Minimum $||\Tilde{v}-v||/||v||$ & 2.76012E$-$2 & 1.94763E$-$2 & 1.92582E$-$2 & 2.05299E$-$2 & 2.09325E$-$2 & 3.83024E$-$2 \\
\midrule
Average $||\Tilde{p}-p||/||p||$ & 4.13612E$-$2 & 2.49465E$-$2 & 3.55946E$-$2 & 4.75586E$-$2 & 3.70187E$-$2 & 6.39906E$-$2 \\
Maximum $||\Tilde{p}-p||/||p||$ & 1.56200E$-$1 & 1.66064E$-$1 & 1.62299E$-$1 & 1.55694E$-$1 & 1.68458E$-$1 & 1.61651E$-$1 \\
Minimum $||\Tilde{p}-p||/||p||$ & 1.53982E$-$2 & 9.89027E$-$3 & 1.36161E$-$2 & 1.75673E$-$2 & 1.13910E$-$2 & 2.14740E$-$2 \\
\bottomrule
\end{tabular*}
\end{table}


\begin{table}[width=.9\linewidth,cols=7,pos=!htbp]
\caption{Computational cost and error analysis of the velocity and pressure fields predicted by Kolmogorov-Arnold PointNet (i.e., KA-PointNet) for the test set containing 222 unseen geometries for different choices of $n_s$. The Jacobi polynomial used has a degree of 3, with $\alpha=\beta=1$. $||\cdots||$ indicates the $L^2$ norm.}\label{Table4}
\begin{tabular*}{\tblwidth}{@{} LLLLLLL@{}}
\toprule
$n_s$ & 0.5 & 0.75 & 1 & 1.25 & 1.5 & 2 \\
\midrule
Average $||\Tilde{u}-u||/||u||$ & 2.63597E$-$2 & 2.57360E$-$2 & 1.73537E$-$2 & 1.82822E$-$2 & 1.68094E$-$2 & 1.69208E$-$2\\
Maximum $||\Tilde{u}-u||/||u||$ & 1.30750E$-$1 & 1.56881E$-$1 & 1.40088E$-$1 & 1.49949E$-$1 & 1.47257E$-$1 & 1.50143E$-$1\\
Minimum $||\Tilde{u}-u||/||u||$ & 1.30994E$-$2 & 1.04247E$-$2 & 7.22247E$-$3 & 7.63074E$-$3 & 6.98780E$-$3 & 1.08372E$-$2\\
\midrule
Average $||\Tilde{v}-v||/||v||$ & 1.00292E$-$1 & 8.05285E$-$2 & 5.75906E$-$2 & 5.34776E$-$2 & 5.66996E$-$2 & 5.48352E$-$2\\
Maximum $||\Tilde{v}-v||/||v||$ & 4.19706E$-$1 & 4.83207E$-$1 & 4.42245E$-$1 & 5.15754E$-$1 & 4.59807E$-$1 & 4.85161E$-$1\\
Minimum $||\Tilde{v}-v||/||v||$ & 5.58101E$-$2 & 3.13250E$-$2 & 2.05299E$-$2 & 1.95743E$-$2 & 2.54641E$-$2 & 2.49093E$-$2\\
\midrule
Average $||\Tilde{p}-p||/||p||$ & 9.86696E$-$2 & 6.35050E$-$2 & 4.75586E$-$2 & 4.21600E$-$2 & 3.63245E$-$2 & 4.25412E$-$2\\
Maximum $||\Tilde{p}-p||/||p||$ & 2.88919E$-$1 & 1.82941E$-$1 & 1.55694E$-$1 & 1.60423E$-$1 & 1.76526E$-$1 & 1.63125E$-$1\\
Minimum $||\Tilde{p}-p||/||p||$ & 4.67306E$-$2 & 3.69896E$-$2 & 1.75673E$-$2 & 1.75982E$-$2 & 1.53689E$-$2 & 1.74415E$-$2\\
\midrule
Training time & 2.83741 & 4.53013 & 6.52910 & 8.85618 & 11.15280 & 16.69120\\
per epoch (s) &  &  &  &  &  & \\
\midrule
Number of trainable & 888128 & 1995744 & 3545728 & 5538080 & 7972800 & 14169344\\
parameters &  &  &  &  &  & \\
\bottomrule
\end{tabular*}
\end{table}


\begin{figure}[!htbp]
  \centering 
      \begin{subfigure}[b]{0.32\textwidth}
        \centering
        \includegraphics[width=\textwidth]{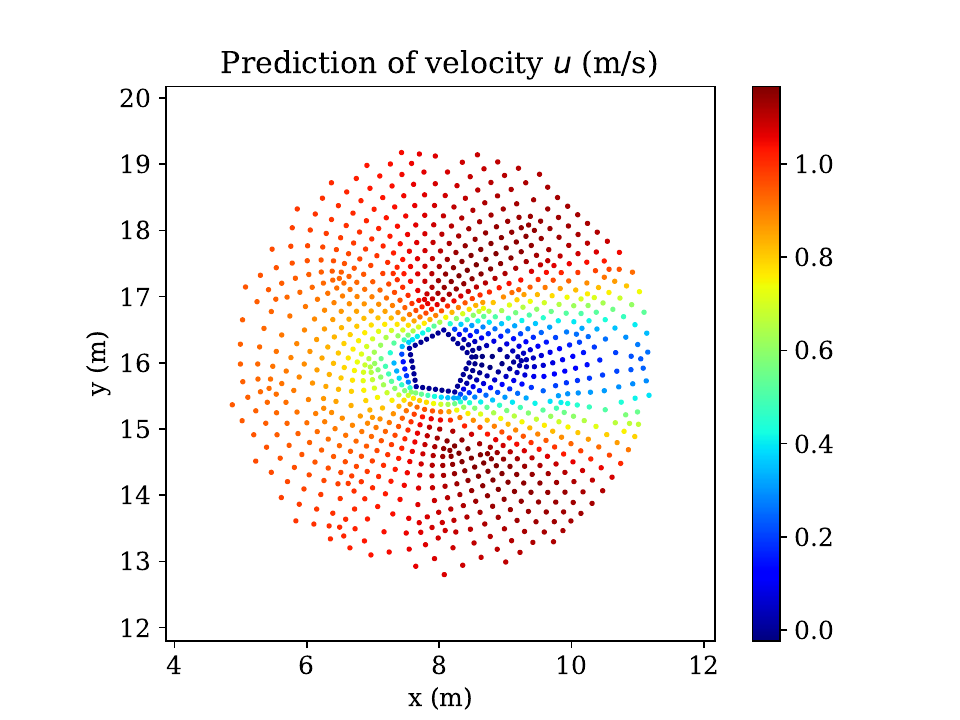}
    \end{subfigure}
    \begin{subfigure}[b]{0.32\textwidth}
        \centering
        \includegraphics[width=\textwidth]{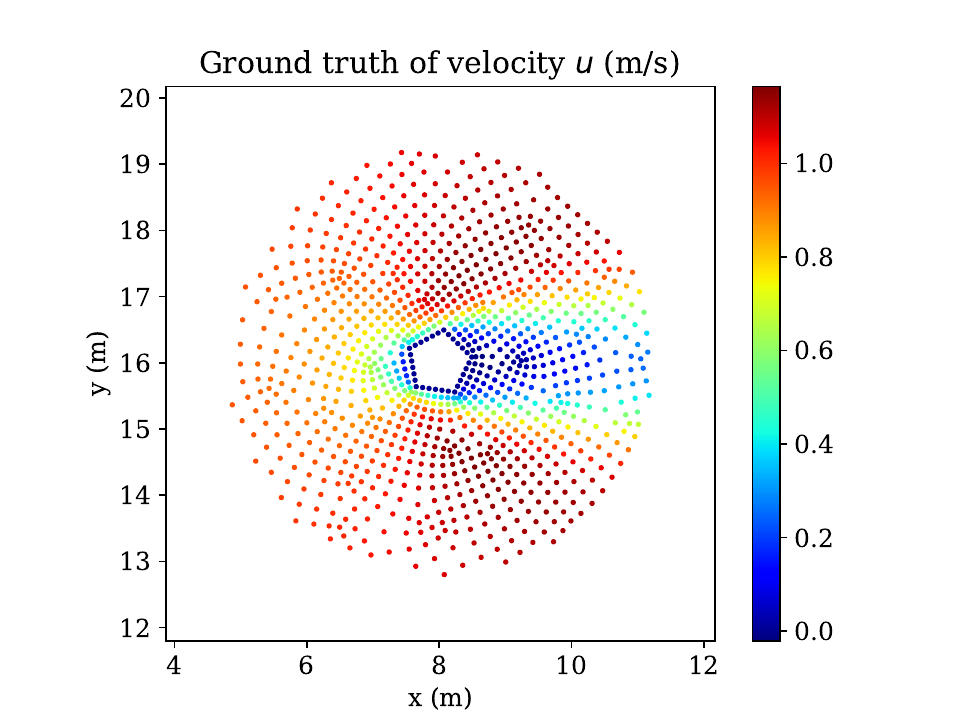}
    \end{subfigure}
    \begin{subfigure}[b]{0.32\textwidth}
        \centering
        \includegraphics[width=\textwidth]{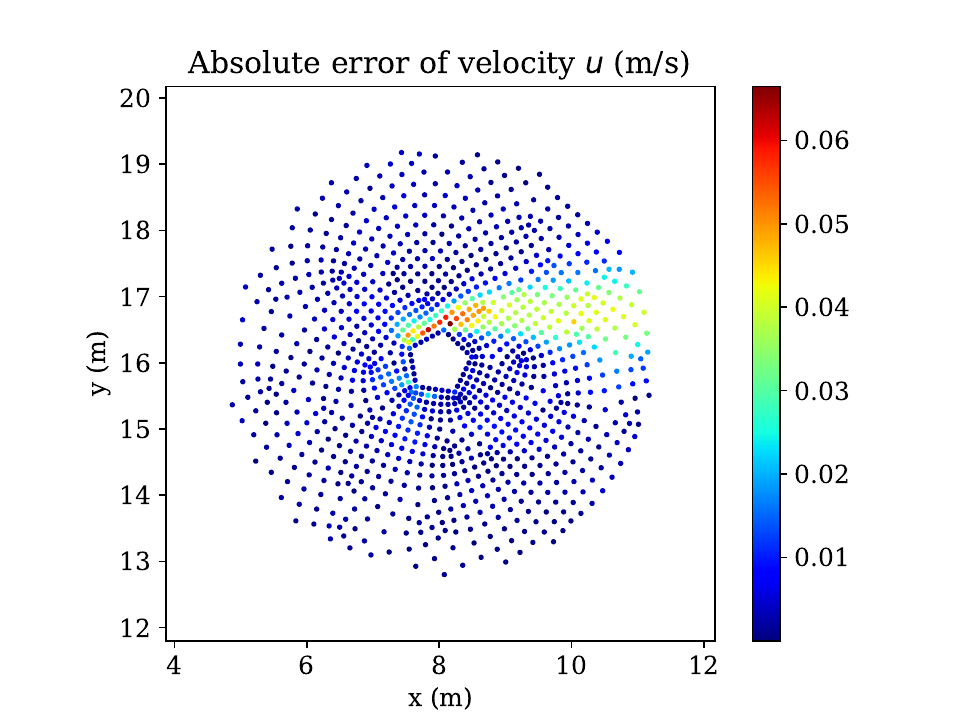}
    \end{subfigure}

    
    \begin{subfigure}[b]{0.32\textwidth}
        \centering
        \includegraphics[width=\textwidth]{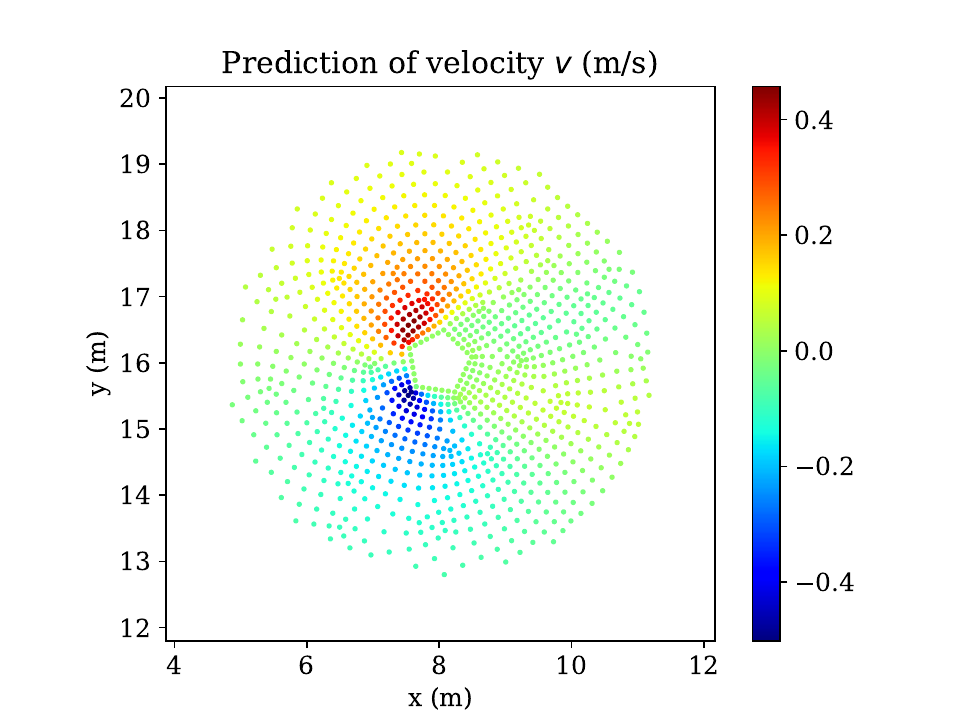}
    \end{subfigure}
    \begin{subfigure}[b]{0.32\textwidth}
        \centering
        \includegraphics[width=\textwidth]{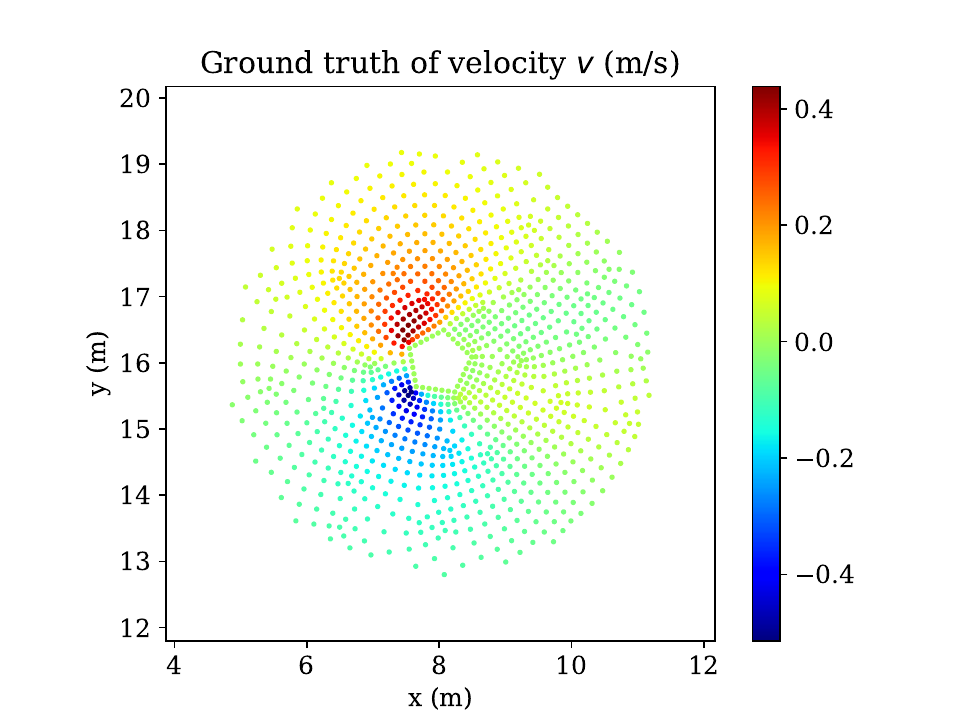}
    \end{subfigure}
    \begin{subfigure}[b]{0.32\textwidth}
        \centering
        \includegraphics[width=\textwidth]{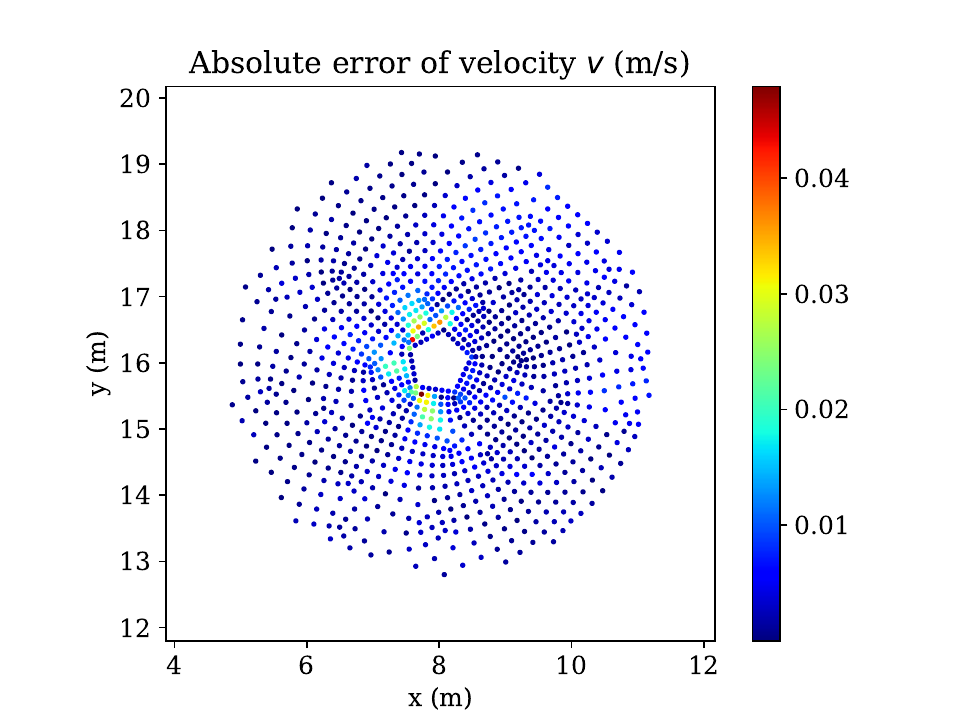}
    \end{subfigure}

    
    \begin{subfigure}[b]{0.32\textwidth}
        \centering
        \includegraphics[width=\textwidth]{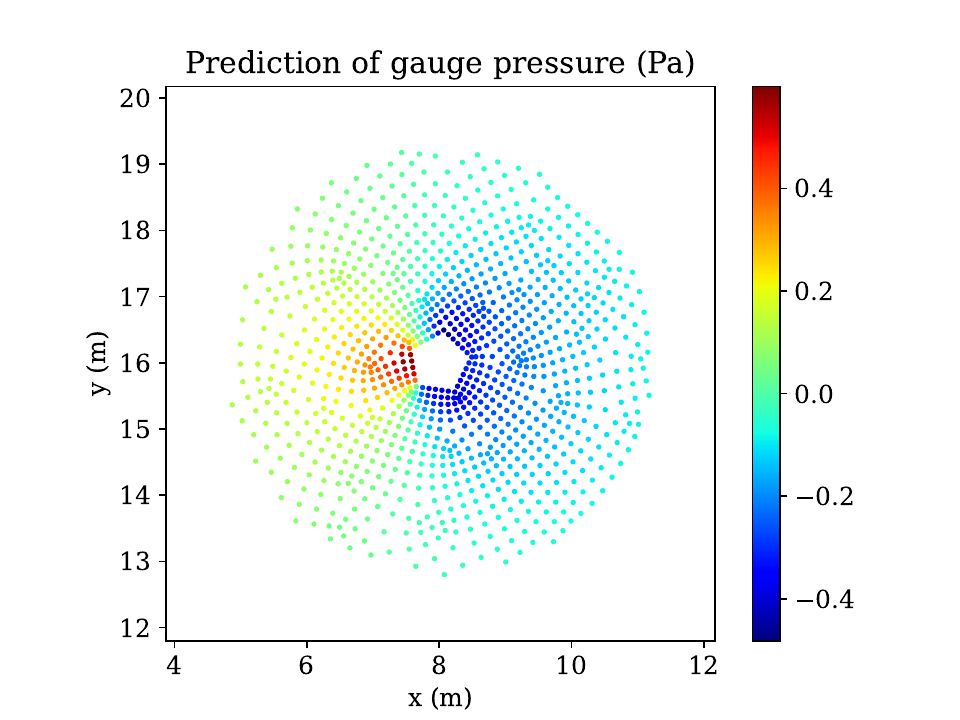}
    \end{subfigure}
    \begin{subfigure}[b]{0.32\textwidth}
        \centering
        \includegraphics[width=\textwidth]{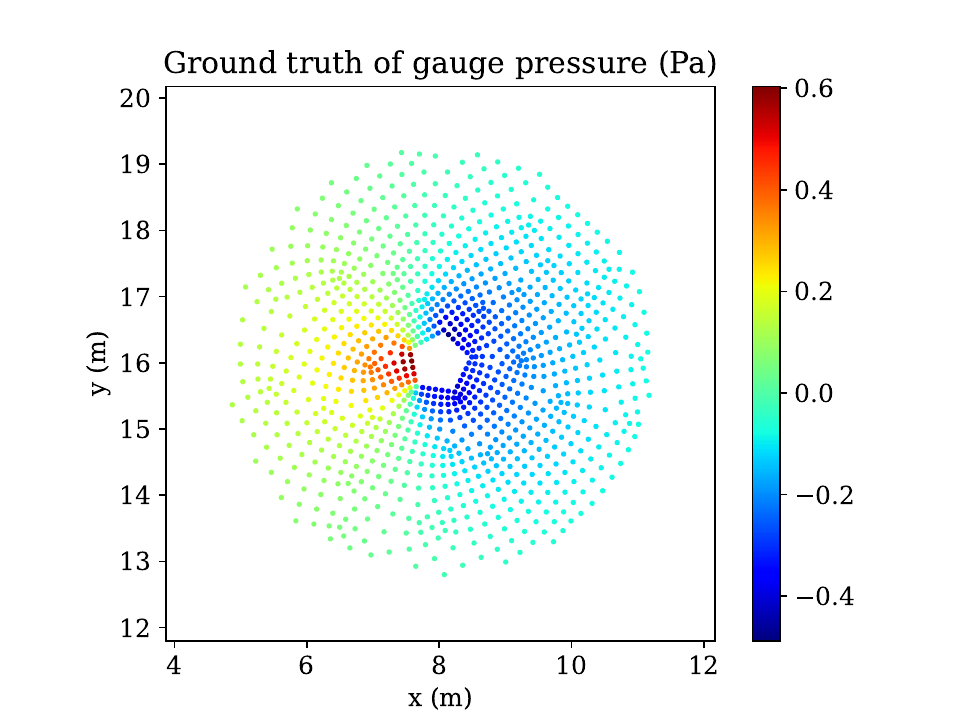}
    \end{subfigure}
    \begin{subfigure}[b]{0.32\textwidth}
        \centering
        \includegraphics[width=\textwidth]{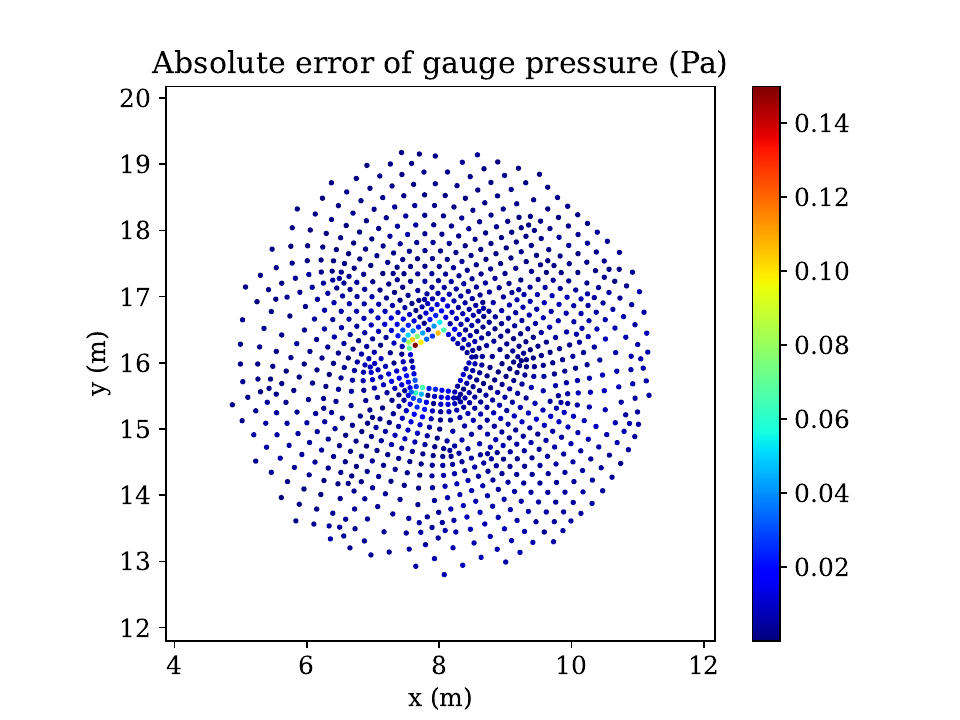}
    \end{subfigure}

  \caption{The first set of examples comparing the ground truth to the predictions of Kolmogorov-Arnold PointNet (i.e., KA-PointNet) for the velocity and pressure fields from the test set. The Jacobi polynomial used has a degree of 5, with $\alpha=\beta=1$. Here, $n_s=1$ is set.}
  \label{Fig5}
\end{figure}

\begin{figure}[!htbp]
  \centering 
      \begin{subfigure}[b]{0.32\textwidth}
        \centering
        \includegraphics[width=\textwidth]{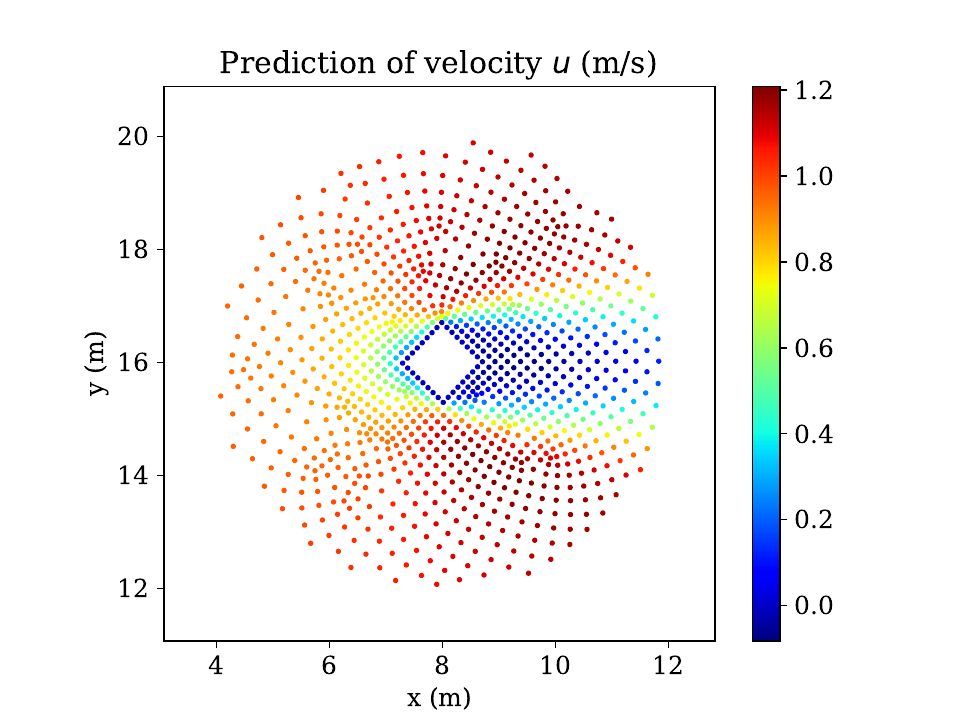}
    \end{subfigure}
    \begin{subfigure}[b]{0.32\textwidth}
        \centering
        \includegraphics[width=\textwidth]{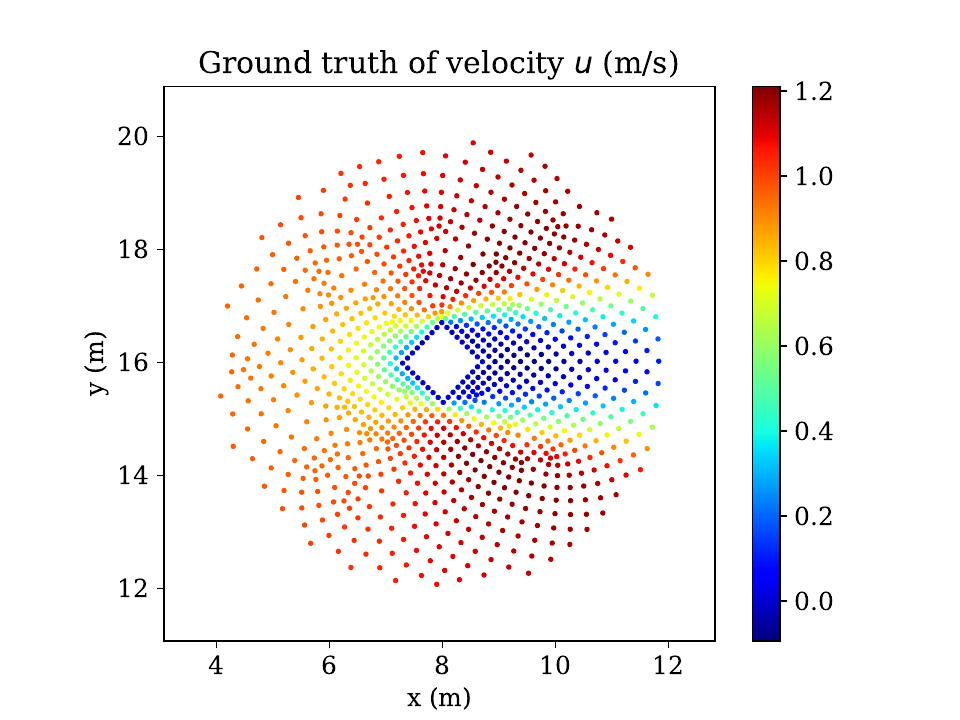}
    \end{subfigure}
    \begin{subfigure}[b]{0.32\textwidth}
        \centering
        \includegraphics[width=\textwidth]{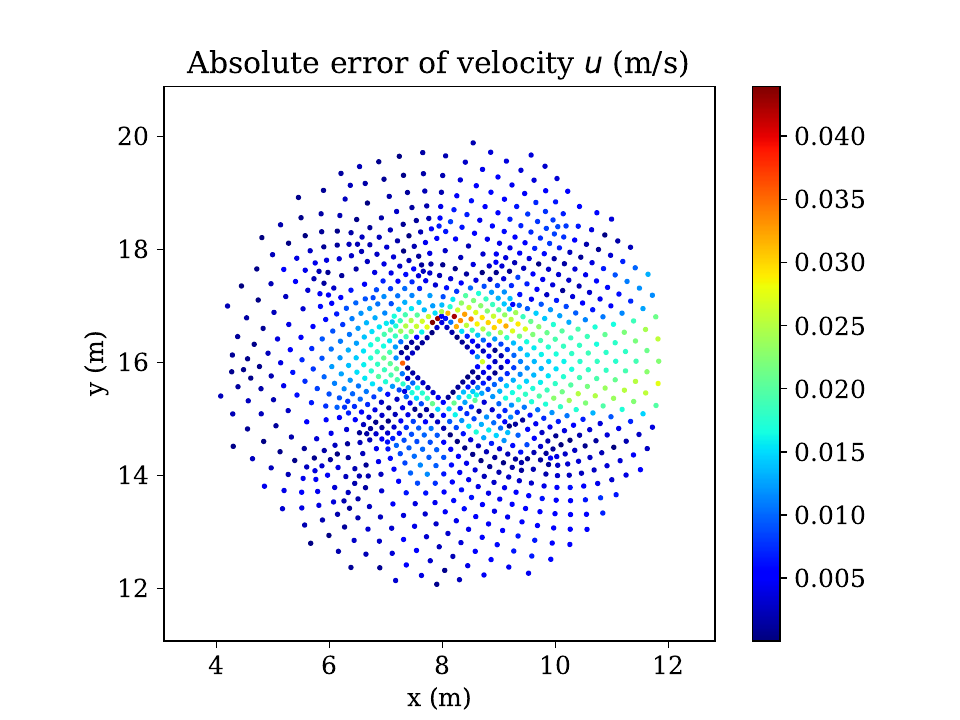}
    \end{subfigure}

    
    \begin{subfigure}[b]{0.32\textwidth}
        \centering
        \includegraphics[width=\textwidth]{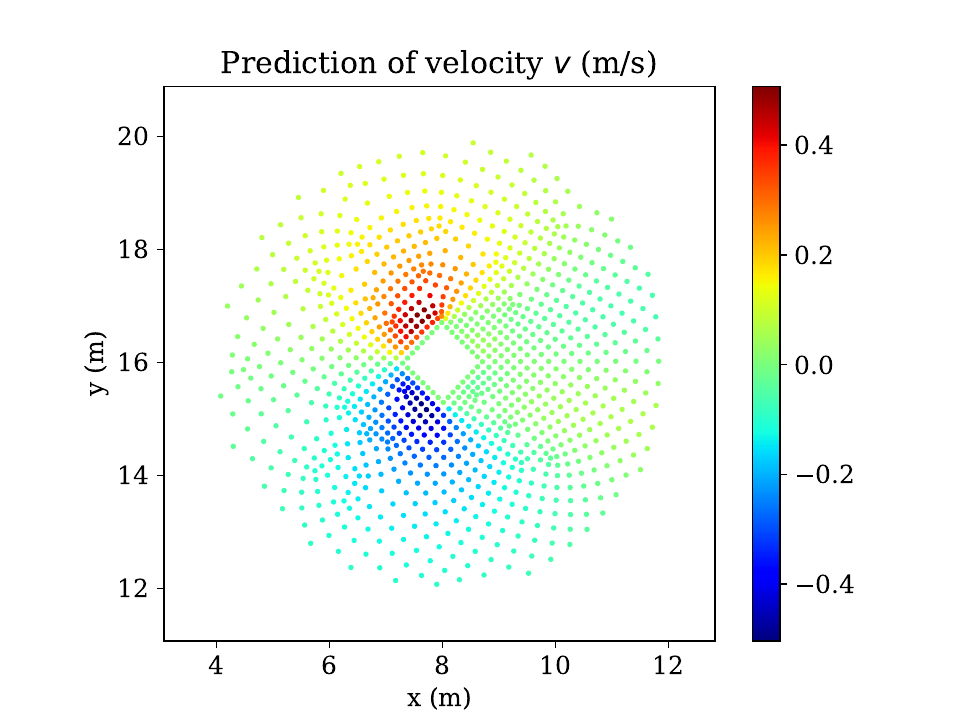}
    \end{subfigure}
    \begin{subfigure}[b]{0.32\textwidth}
        \centering
        \includegraphics[width=\textwidth]{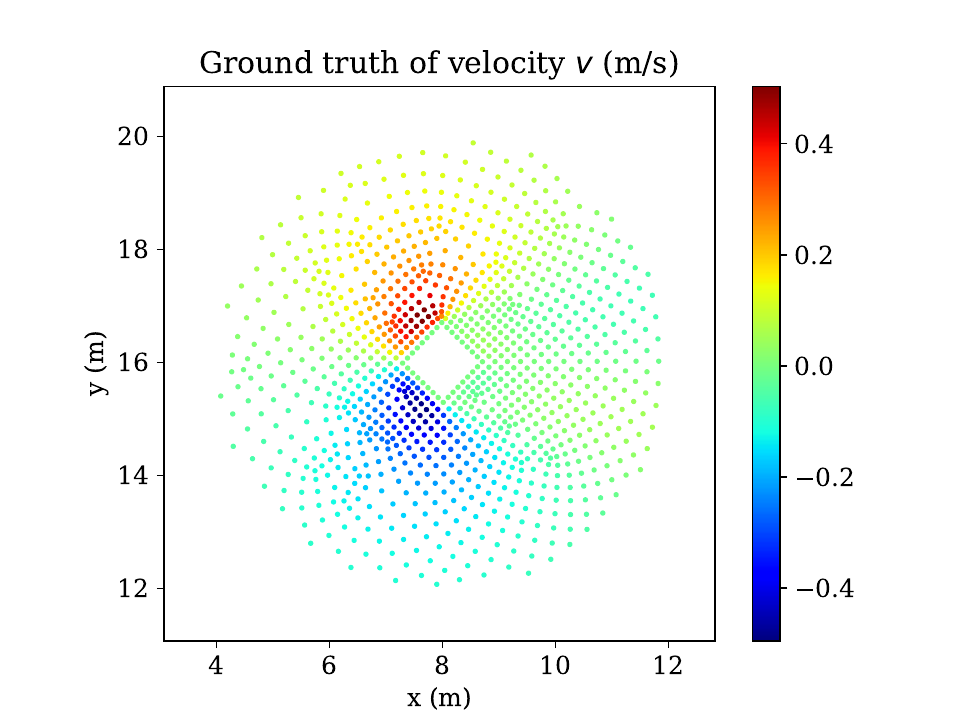}
    \end{subfigure}
    \begin{subfigure}[b]{0.32\textwidth}
        \centering
        \includegraphics[width=\textwidth]{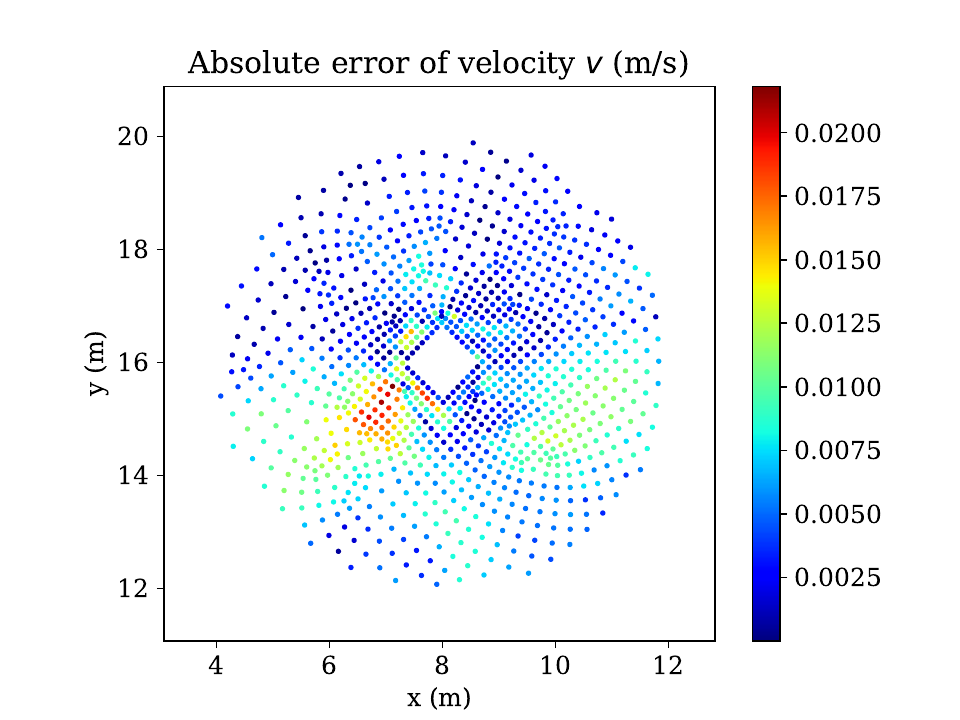}
    \end{subfigure}

    
    \begin{subfigure}[b]{0.32\textwidth}
        \centering
        \includegraphics[width=\textwidth]{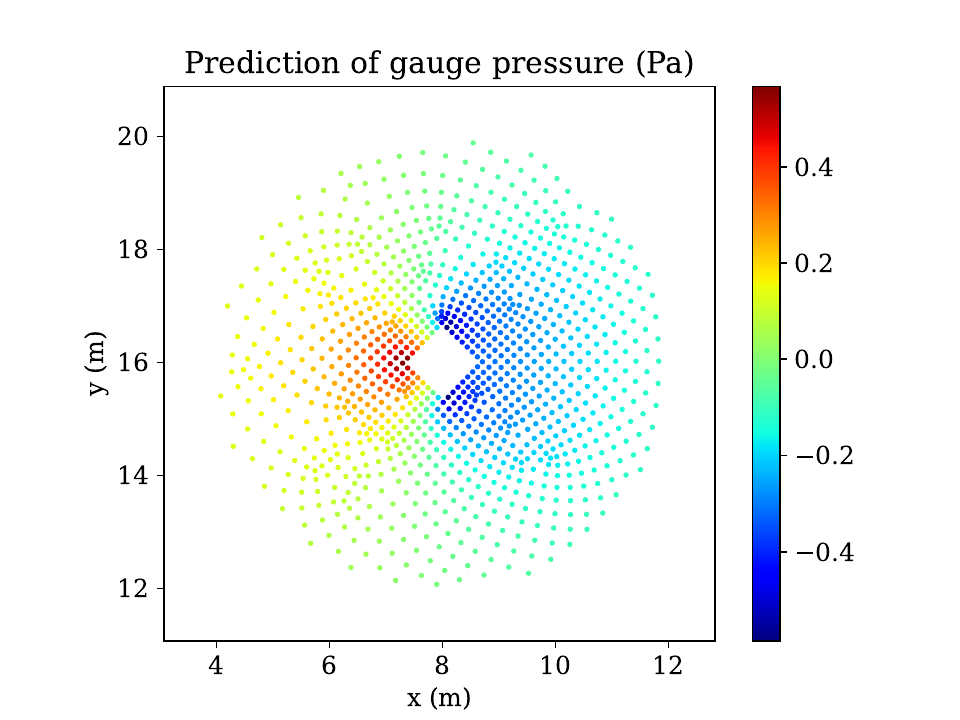}
    \end{subfigure}
    \begin{subfigure}[b]{0.32\textwidth}
        \centering
        \includegraphics[width=\textwidth]{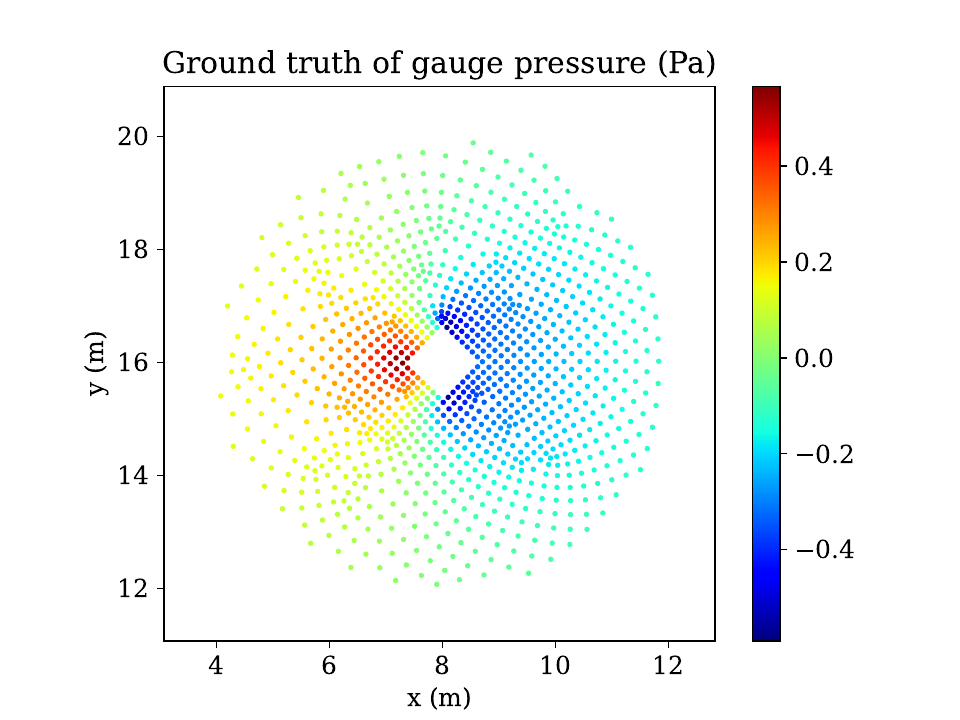}
    \end{subfigure}
    \begin{subfigure}[b]{0.32\textwidth}
        \centering
        \includegraphics[width=\textwidth]{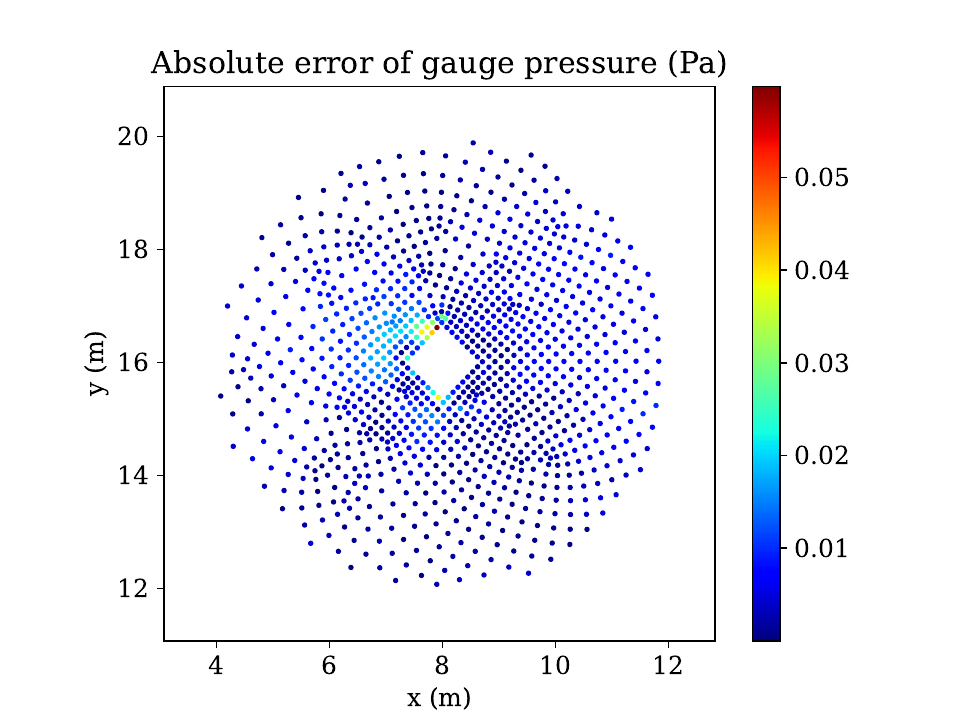}
    \end{subfigure}

  \caption{The second set of examples comparing the ground truth to the predictions of Kolmogorov-Arnold PointNet (i.e., KA-PointNet) for the velocity and pressure fields from the test set. The Jacobi polynomial used has a degree of 5, with $\alpha=\beta=1$. Here, $n_s=1$ is set.}
  \label{Fig6}
\end{figure}

\begin{figure}[!htbp]
  \centering 
      \begin{subfigure}[b]{0.32\textwidth}
        \centering
        \includegraphics[width=\textwidth]{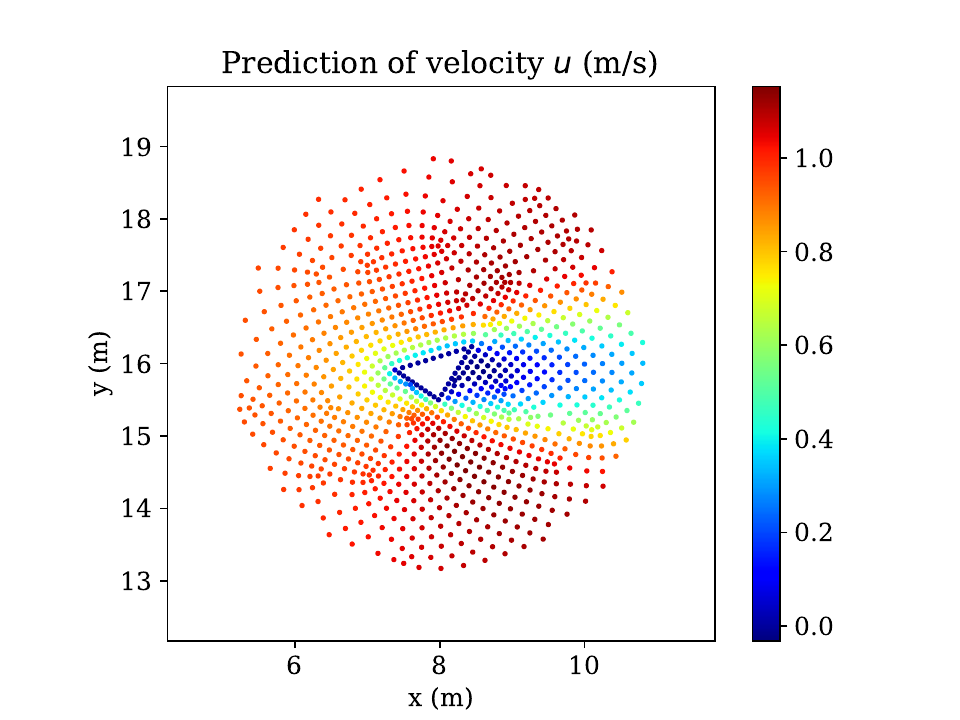}
      \end{subfigure}
    \begin{subfigure}[b]{0.32\textwidth}
        \centering
        \includegraphics[width=\textwidth]{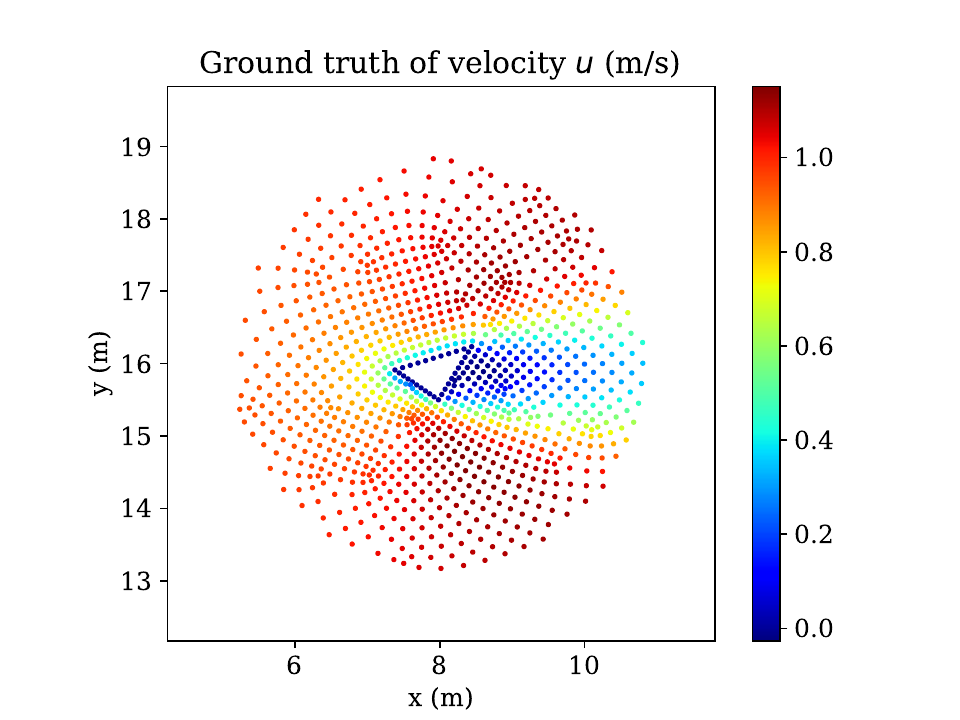}
      \end{subfigure}
    \begin{subfigure}[b]{0.32\textwidth}
        \centering
        \includegraphics[width=\textwidth]{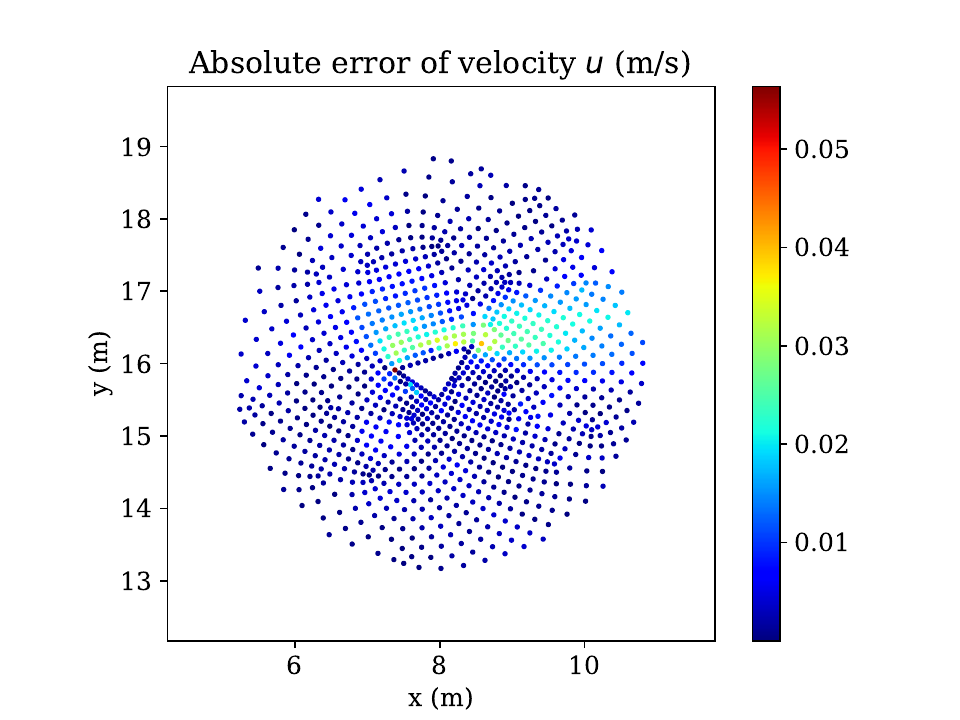}
      \end{subfigure}

    
    \begin{subfigure}[b]{0.32\textwidth}
        \centering
        \includegraphics[width=\textwidth]{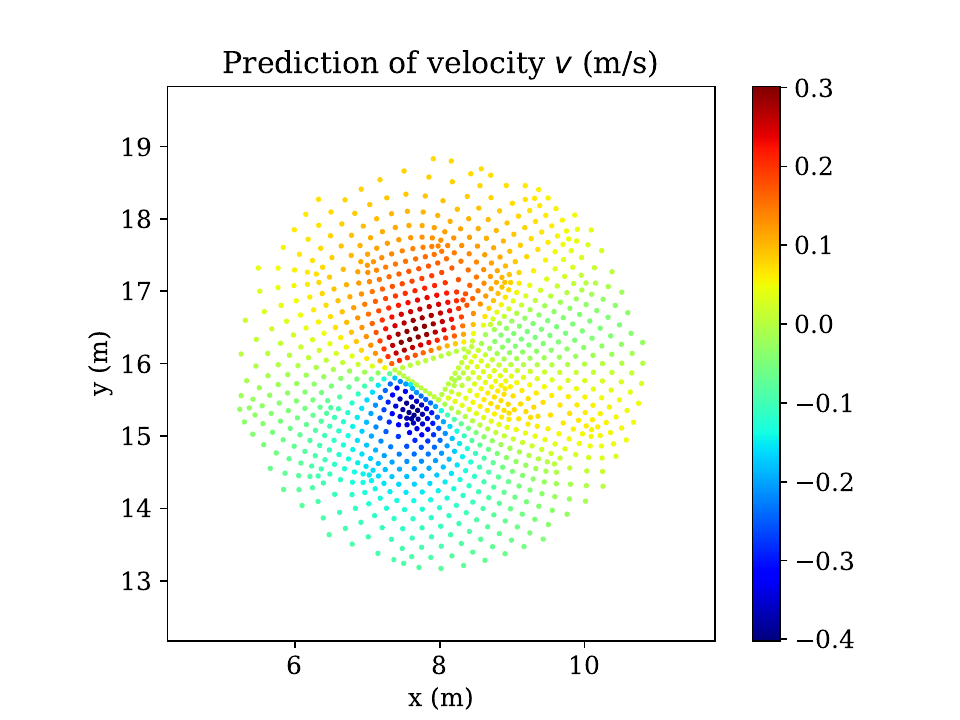}
      \end{subfigure}
    \begin{subfigure}[b]{0.32\textwidth}
        \centering
        \includegraphics[width=\textwidth]{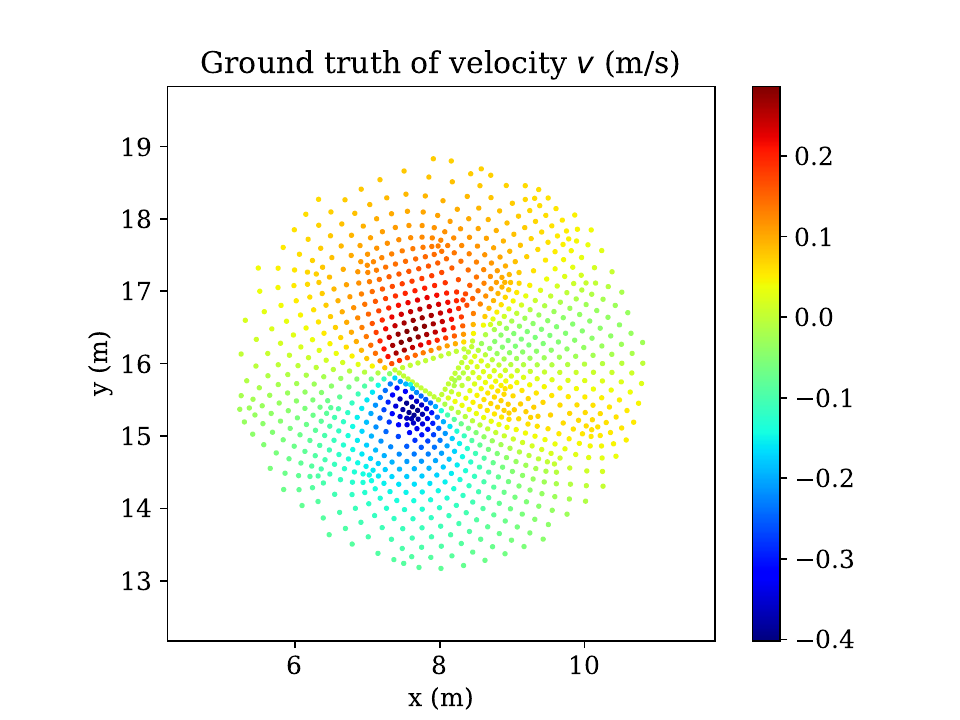}
      \end{subfigure}
    \begin{subfigure}[b]{0.32\textwidth}
        \centering
        \includegraphics[width=\textwidth]{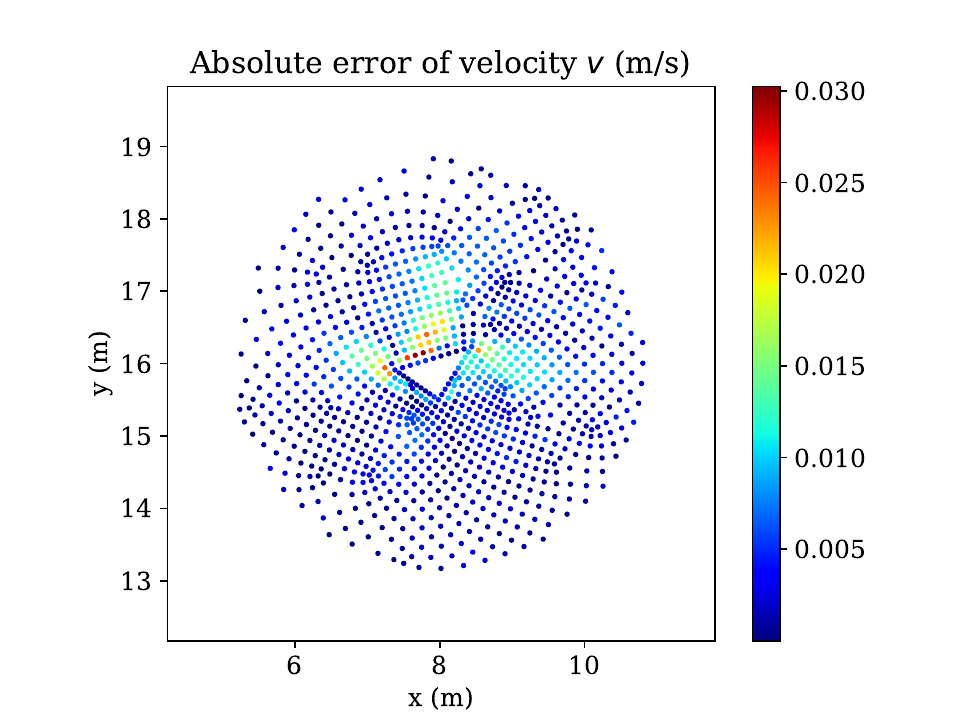}
      \end{subfigure}

    
    \begin{subfigure}[b]{0.32\textwidth}
        \centering
        \includegraphics[width=\textwidth]{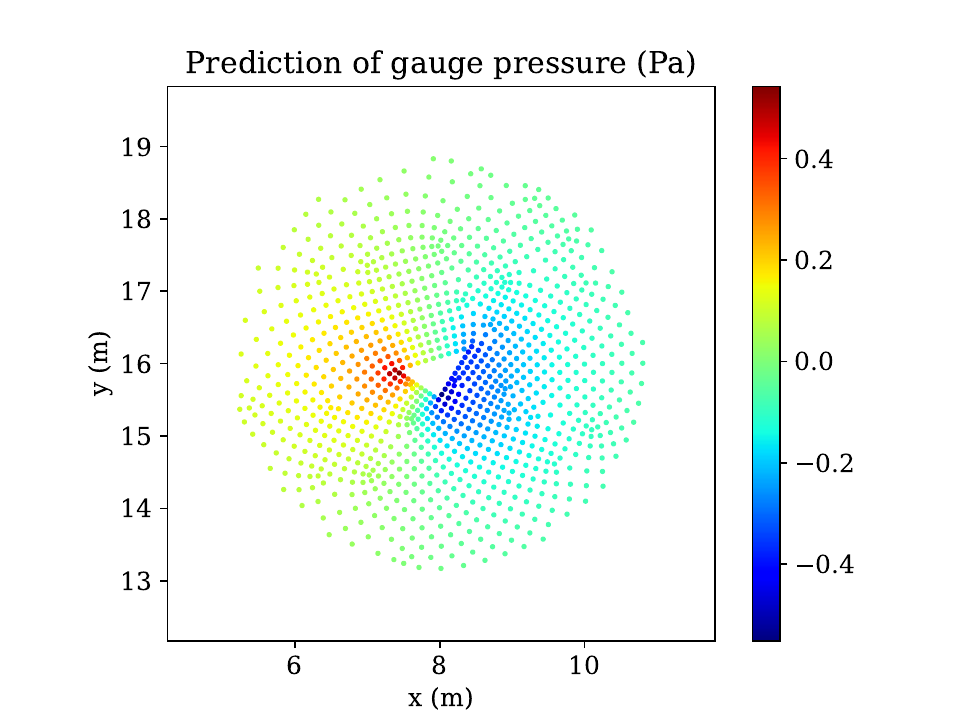}
      \end{subfigure}
    \begin{subfigure}[b]{0.32\textwidth}
        \centering
        \includegraphics[width=\textwidth]{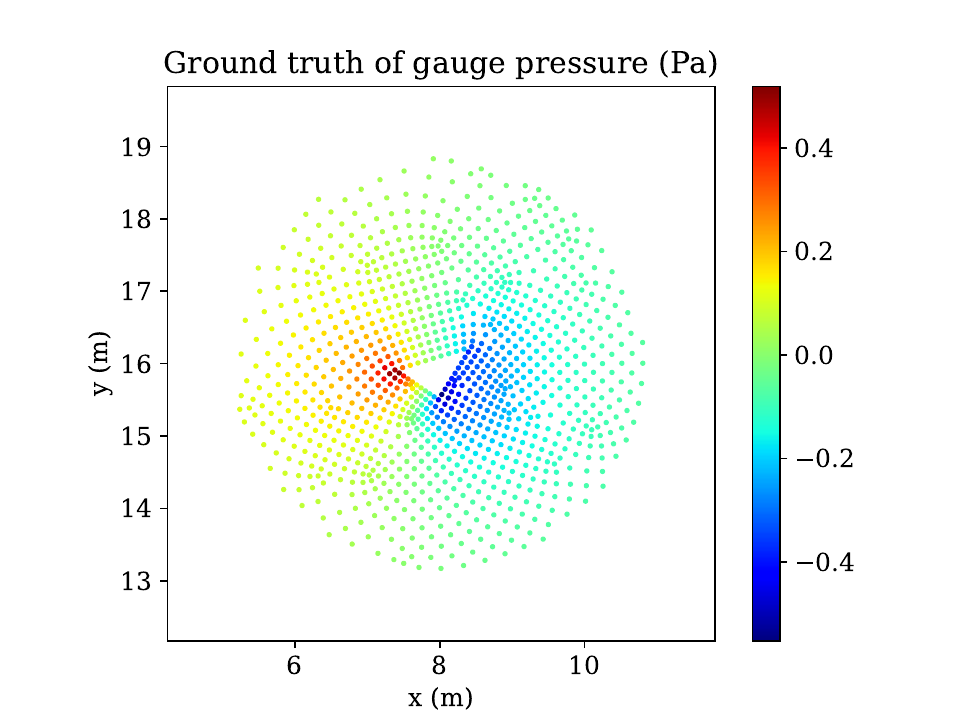}
      \end{subfigure}
    \begin{subfigure}[b]{0.32\textwidth}
        \centering
        \includegraphics[width=\textwidth]{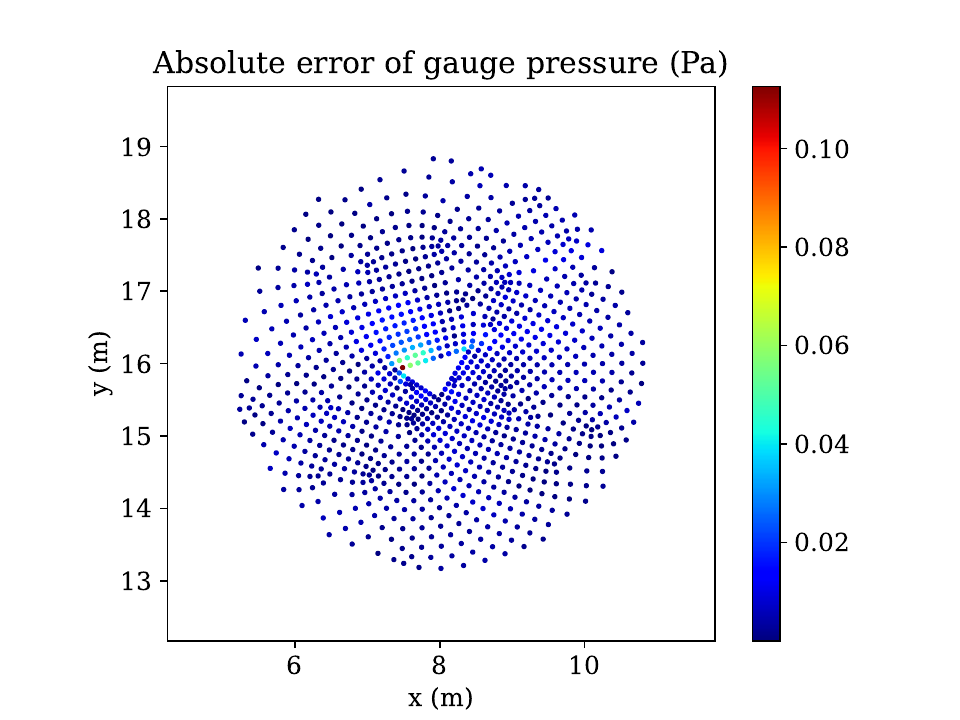}
      \end{subfigure}

  \caption{The third set of examples comparing the ground truth to predictions of the Kolmogorov-Arnold PointNet for velocity and pressure fields from the test set. The Jacobi polynomial used has a degree of 5, with $\alpha=\beta=1$. Here, $n_s=1$ is set.}
  \label{Fig7}
\end{figure}

\begin{figure}[!htbp]
  \centering 
      \begin{subfigure}[b]{0.32\textwidth}
        \centering
        \includegraphics[width=\textwidth]{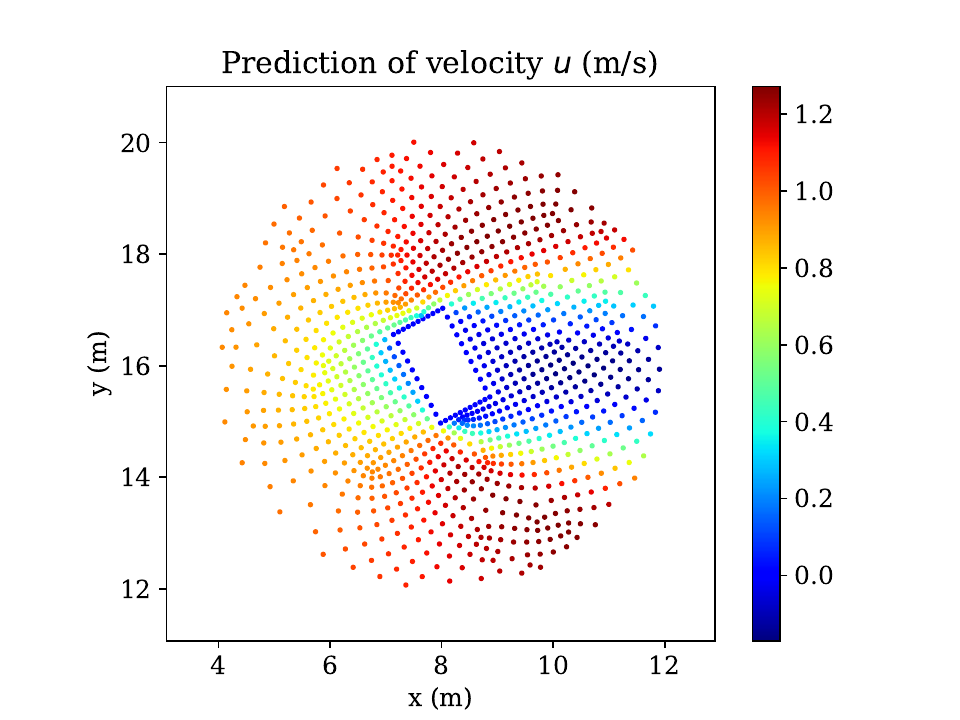}
      \end{subfigure}
    \begin{subfigure}[b]{0.32\textwidth}
        \centering
        \includegraphics[width=\textwidth]{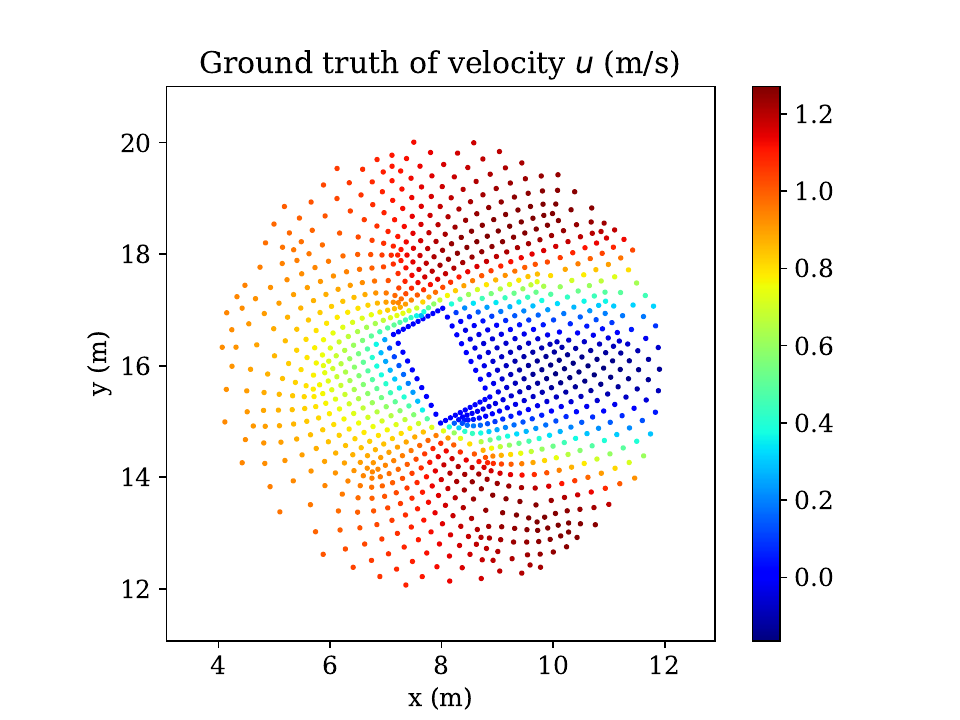}
      \end{subfigure}
    \begin{subfigure}[b]{0.32\textwidth}
        \centering
        \includegraphics[width=\textwidth]{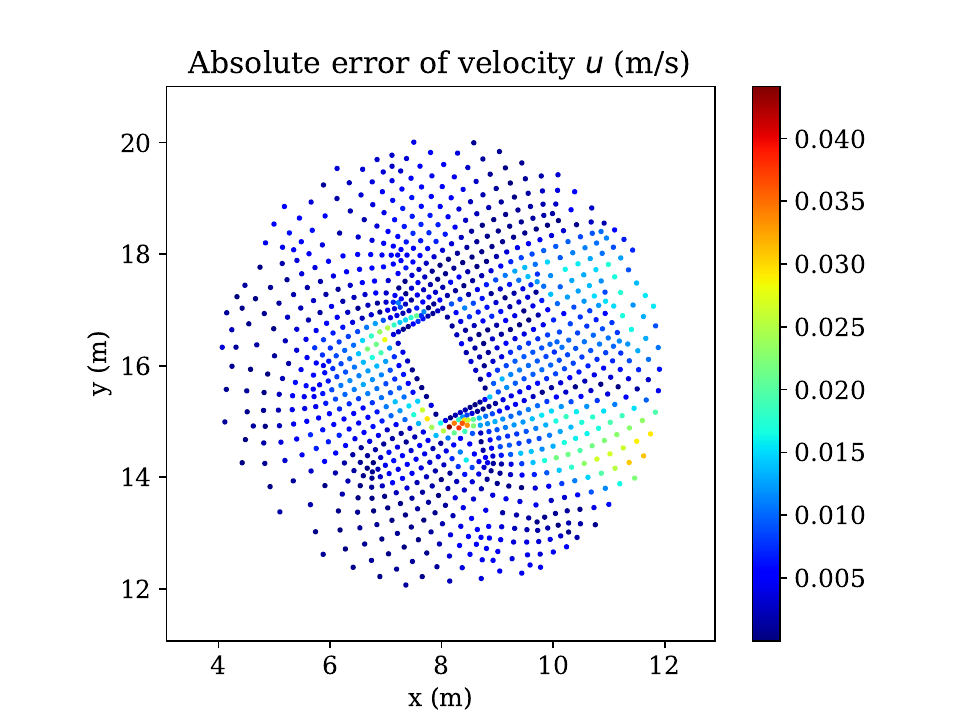}
      \end{subfigure}

    
    \begin{subfigure}[b]{0.32\textwidth}
        \centering
        \includegraphics[width=\textwidth]{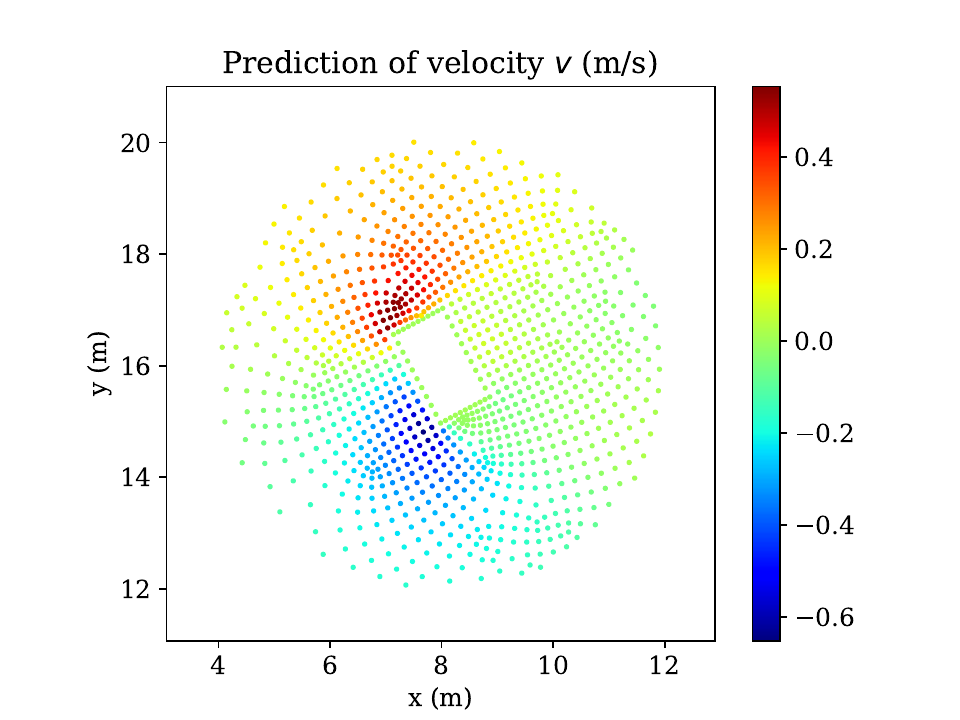}
      \end{subfigure}
    \begin{subfigure}[b]{0.32\textwidth}
        \centering
        \includegraphics[width=\textwidth]{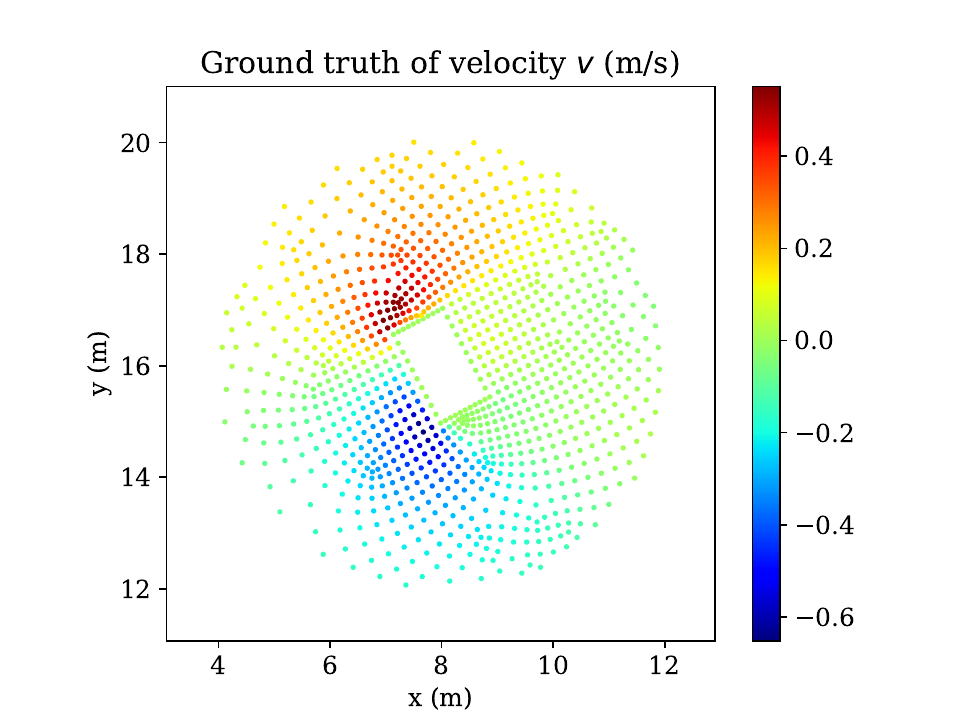}
      \end{subfigure}
    \begin{subfigure}[b]{0.32\textwidth}
        \centering
        \includegraphics[width=\textwidth]{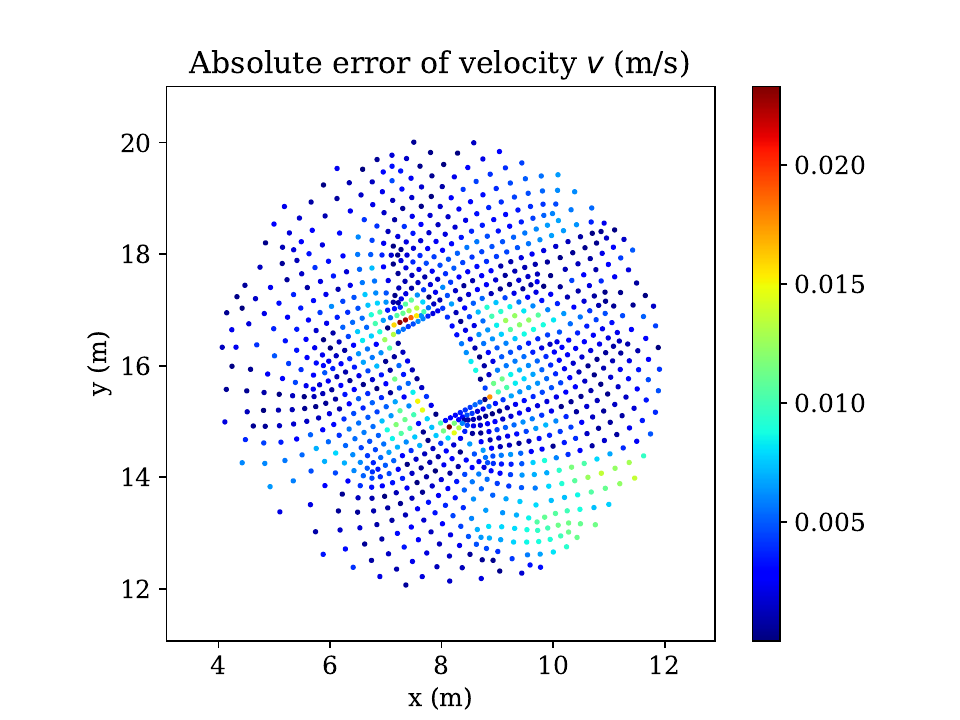}
      \end{subfigure}

    
    \begin{subfigure}[b]{0.32\textwidth}
        \centering
        \includegraphics[width=\textwidth]{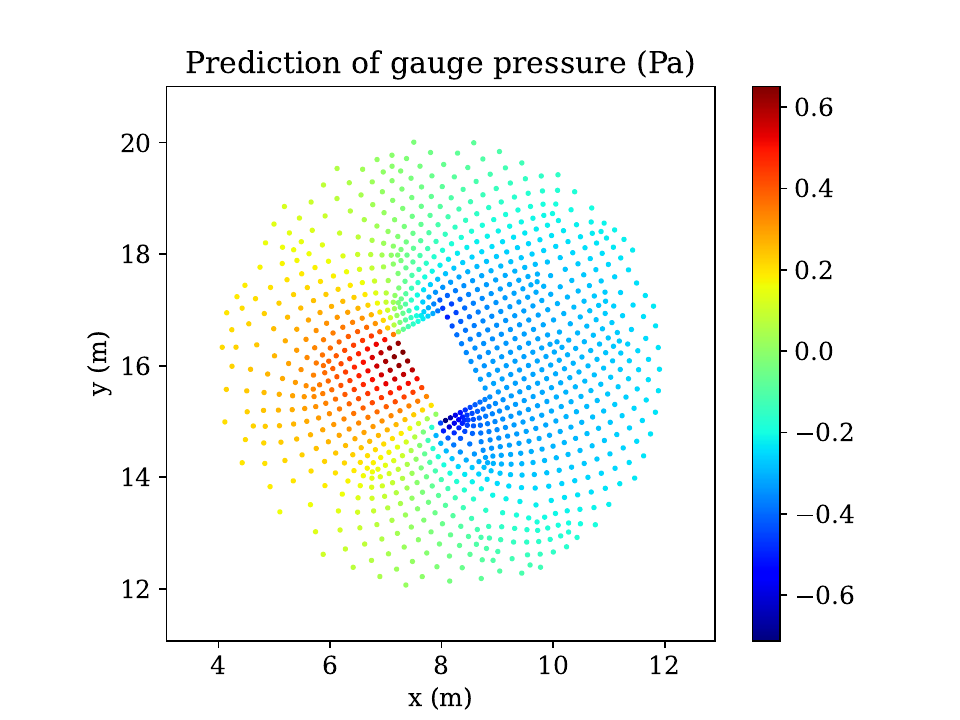}
      \end{subfigure}
    \begin{subfigure}[b]{0.32\textwidth}
        \centering
        \includegraphics[width=\textwidth]{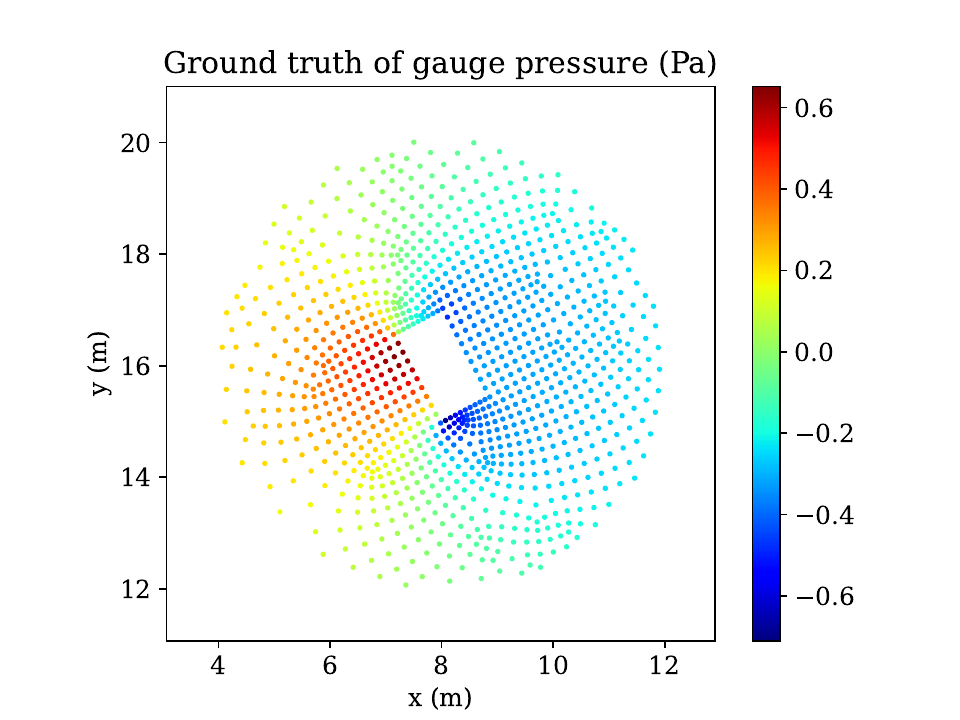}
      \end{subfigure}
    \begin{subfigure}[b]{0.32\textwidth}
        \centering
        \includegraphics[width=\textwidth]{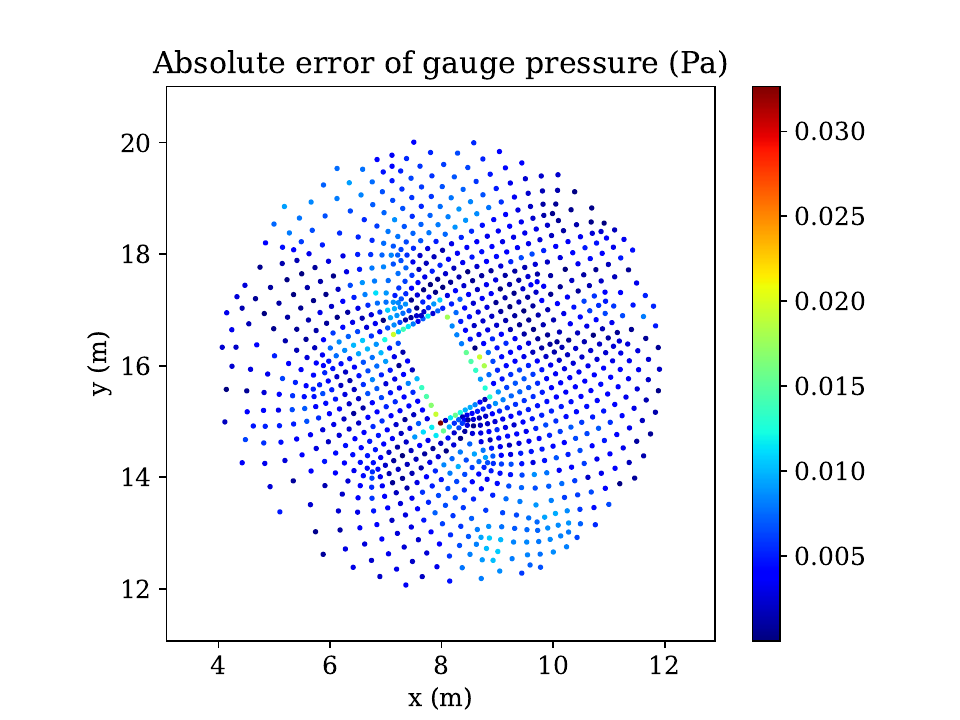}
      \end{subfigure}

  \caption{The fourth set of examples comparing the ground truth to the predictions of Kolmogorov-Arnold PointNet (i.e., KA-PointNet) for the velocity and pressure fields from the test set. The Jacobi polynomial used has a degree of 5, with $\alpha=\beta=1$. Here, $n_s=1$ is set.}
  \label{Fig8}
\end{figure}

\begin{figure}[!htbp]
  \centering 
      \begin{subfigure}[b]{0.32\textwidth}
        \centering
        \includegraphics[width=\textwidth]{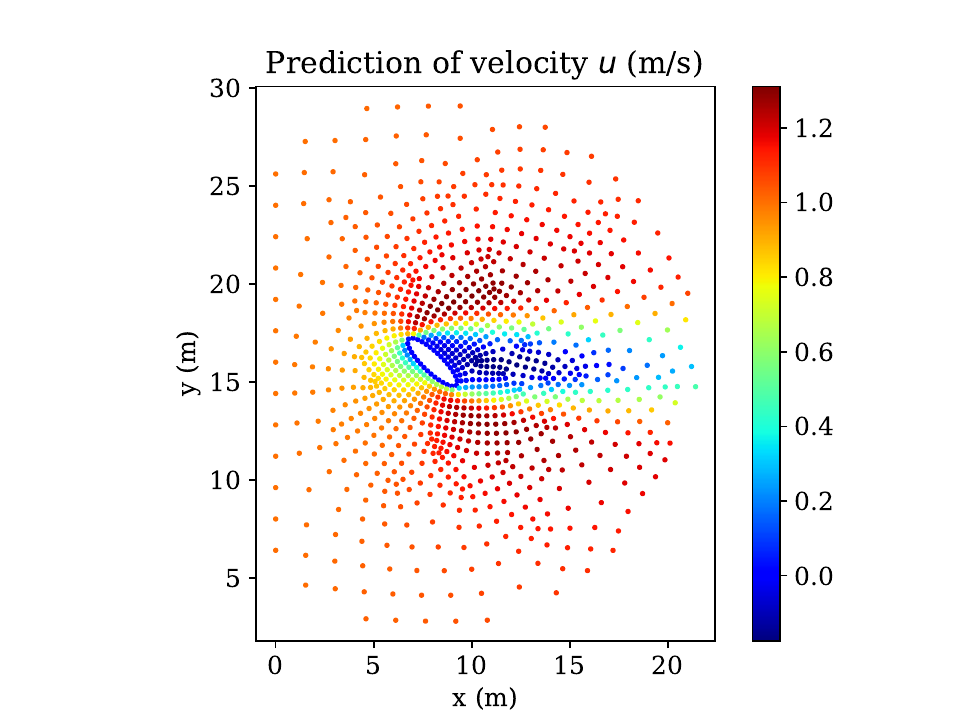}
      \end{subfigure}
    \begin{subfigure}[b]{0.32\textwidth}
        \centering
        \includegraphics[width=\textwidth]{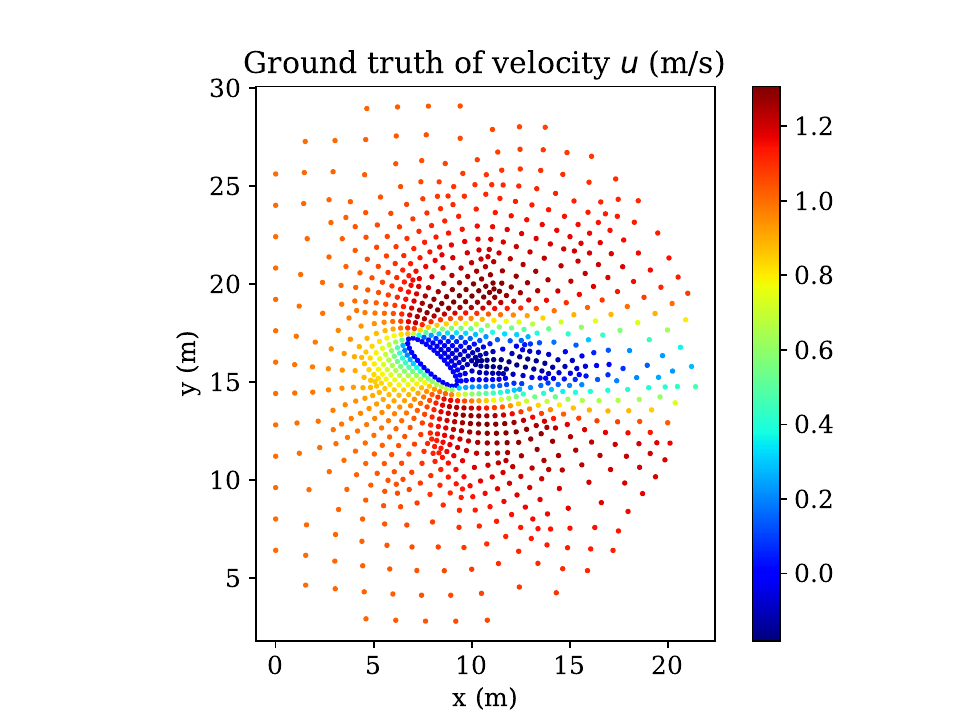}
      \end{subfigure}
    \begin{subfigure}[b]{0.32\textwidth}
        \centering
        \includegraphics[width=\textwidth]{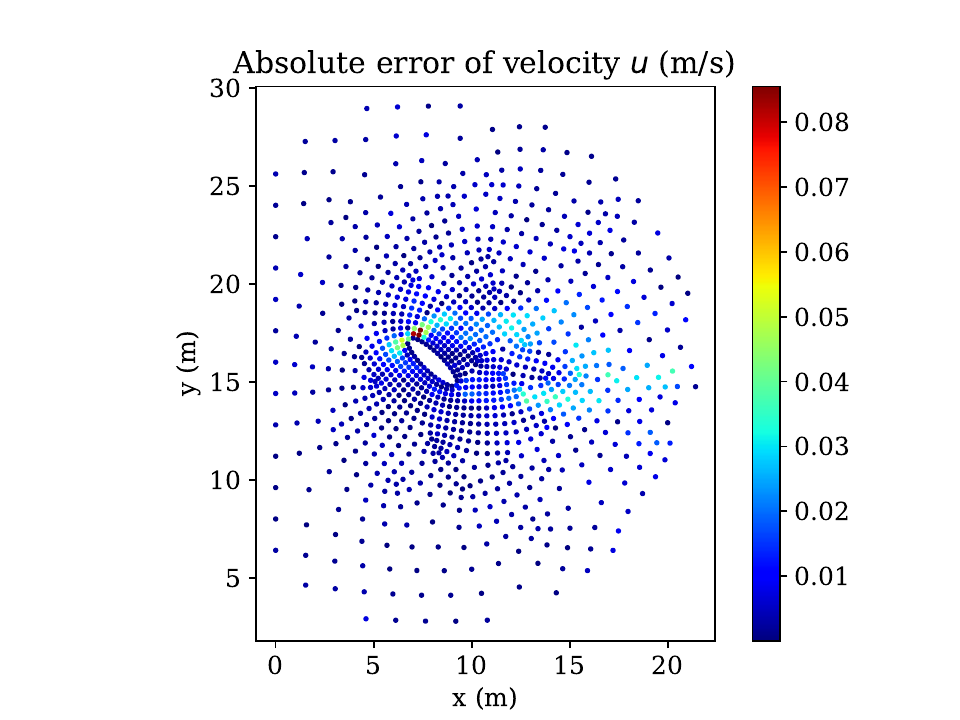}
      \end{subfigure}

    
    \begin{subfigure}[b]{0.32\textwidth}
        \centering
        \includegraphics[width=\textwidth]{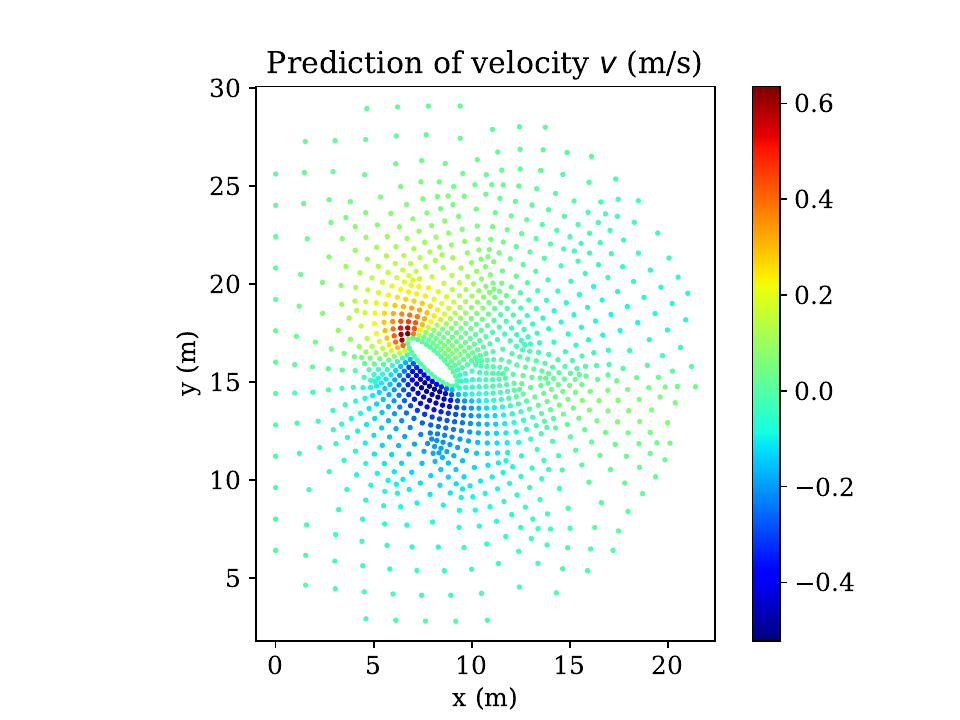}
      \end{subfigure}
    \begin{subfigure}[b]{0.32\textwidth}
        \centering
        \includegraphics[width=\textwidth]{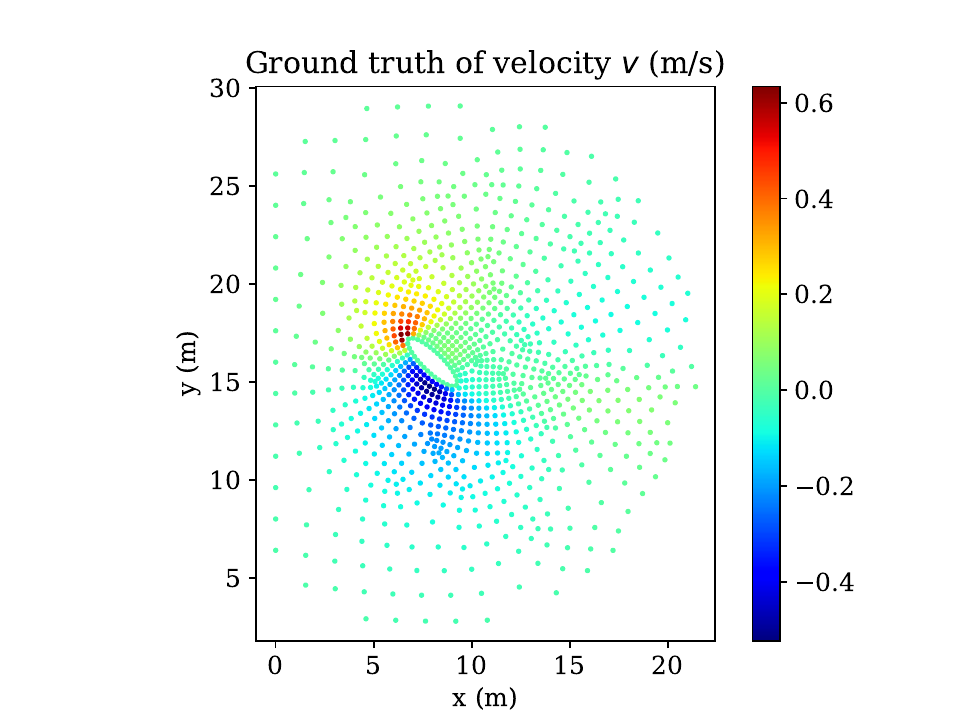}
      \end{subfigure}
    \begin{subfigure}[b]{0.32\textwidth}
        \centering
        \includegraphics[width=\textwidth]{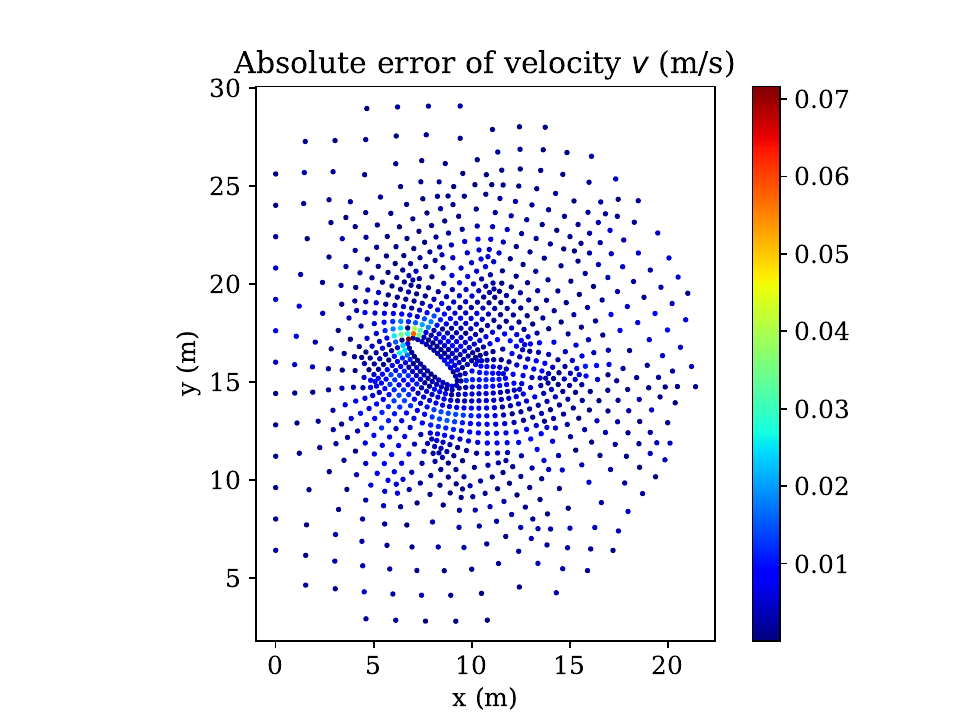}
      \end{subfigure}

    
    \begin{subfigure}[b]{0.32\textwidth}
        \centering
        \includegraphics[width=\textwidth]{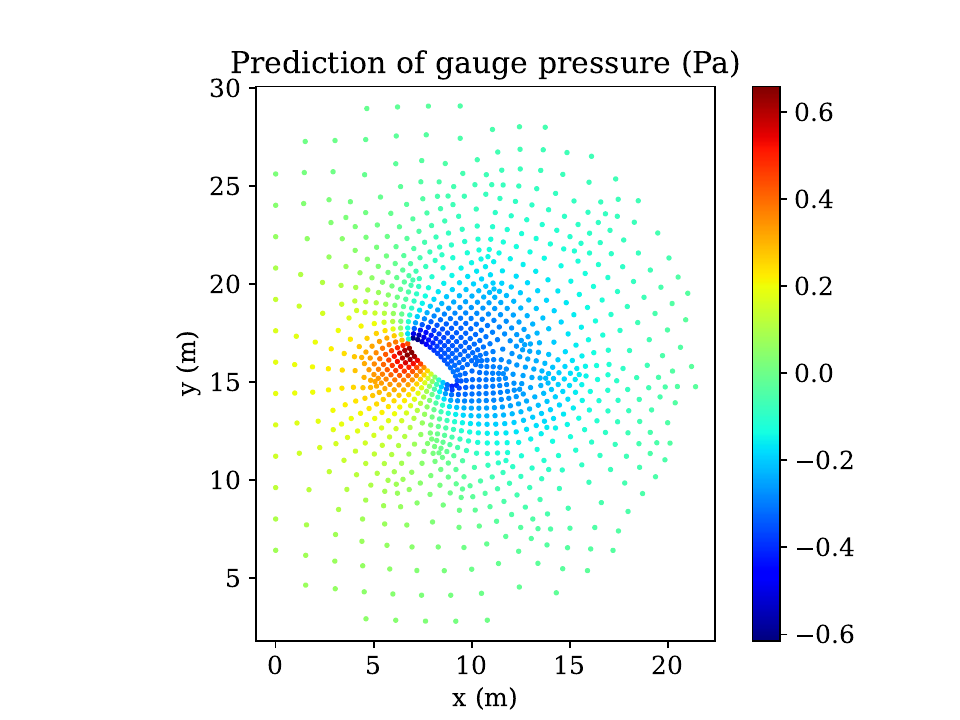}
      \end{subfigure}
    \begin{subfigure}[b]{0.32\textwidth}
        \centering
        \includegraphics[width=\textwidth]{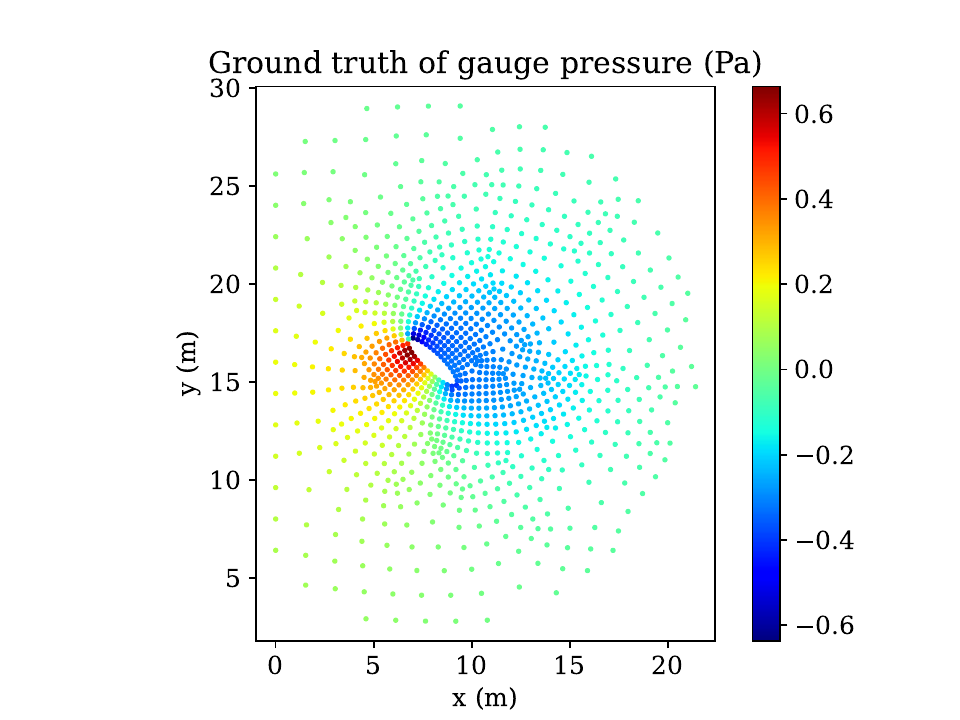}
      \end{subfigure}
    \begin{subfigure}[b]{0.32\textwidth}
        \centering
        \includegraphics[width=\textwidth]{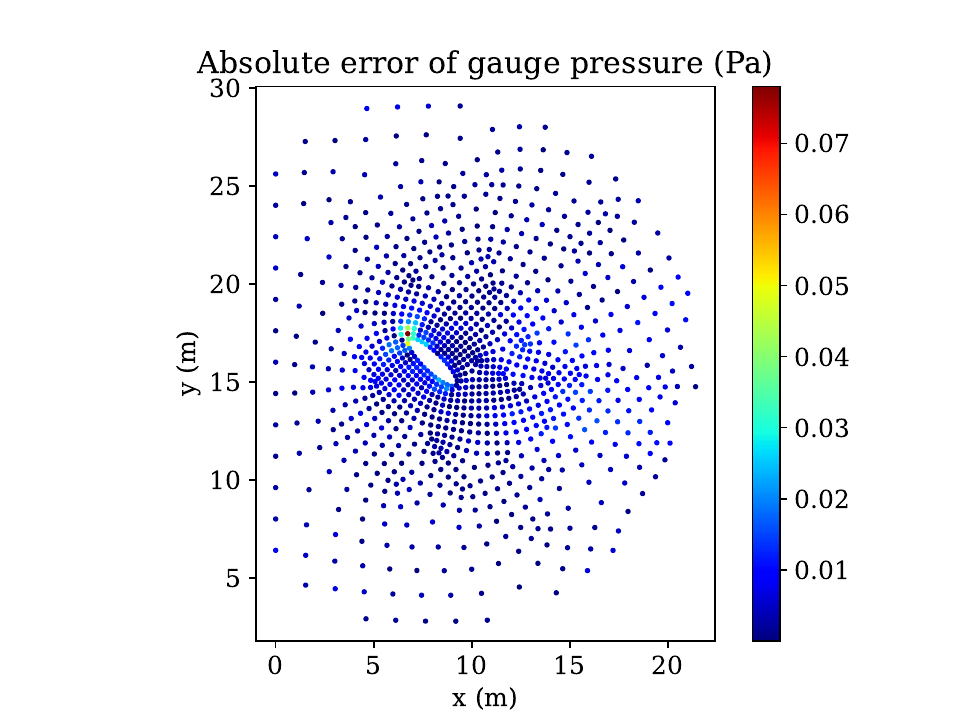}
      \end{subfigure}

  \caption{The fifth set of examples comparing the ground truth to the predictions of Kolmogorov-Arnold PointNet (i.e., KA-PointNet) for the velocity and pressure fields from the test set. The Jacobi polynomial used has a degree of 5, with $\alpha=\beta=1$. Here, $n_s=1$ is set.}
  \label{Fig9}
\end{figure}

\begin{figure}[!htbp]
  \centering 
      \begin{subfigure}[b]{0.32\textwidth}
        \centering
        \includegraphics[width=\textwidth]{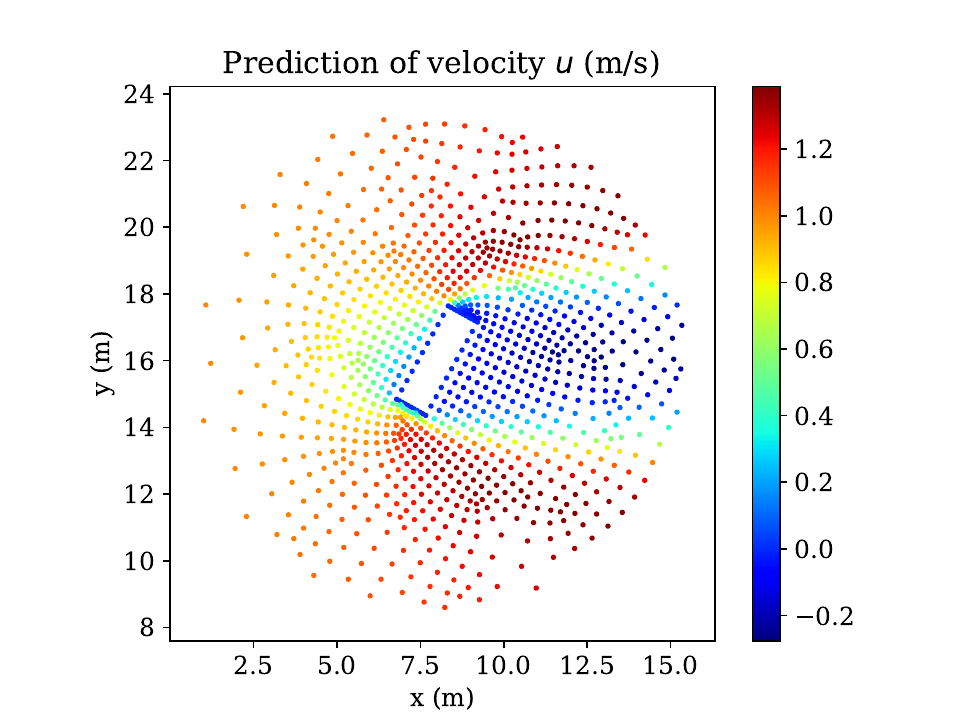}
      \end{subfigure}
    \begin{subfigure}[b]{0.32\textwidth}
        \centering
        \includegraphics[width=\textwidth]{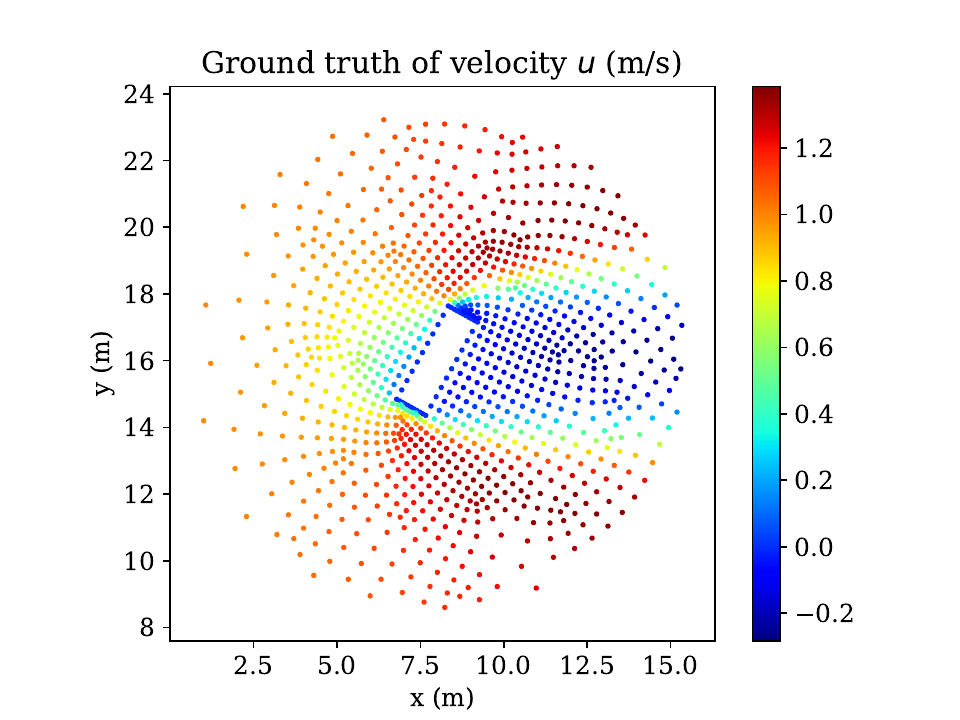}
      \end{subfigure}
    \begin{subfigure}[b]{0.32\textwidth}
        \centering
        \includegraphics[width=\textwidth]{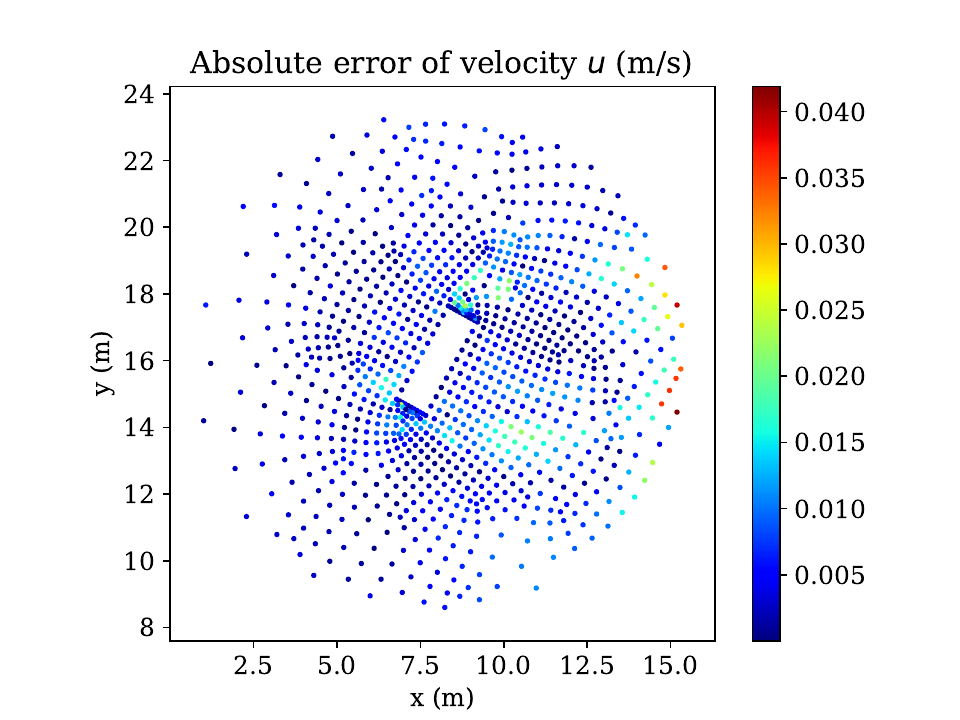}
      \end{subfigure}

    
    \begin{subfigure}[b]{0.32\textwidth}
        \centering
        \includegraphics[width=\textwidth]{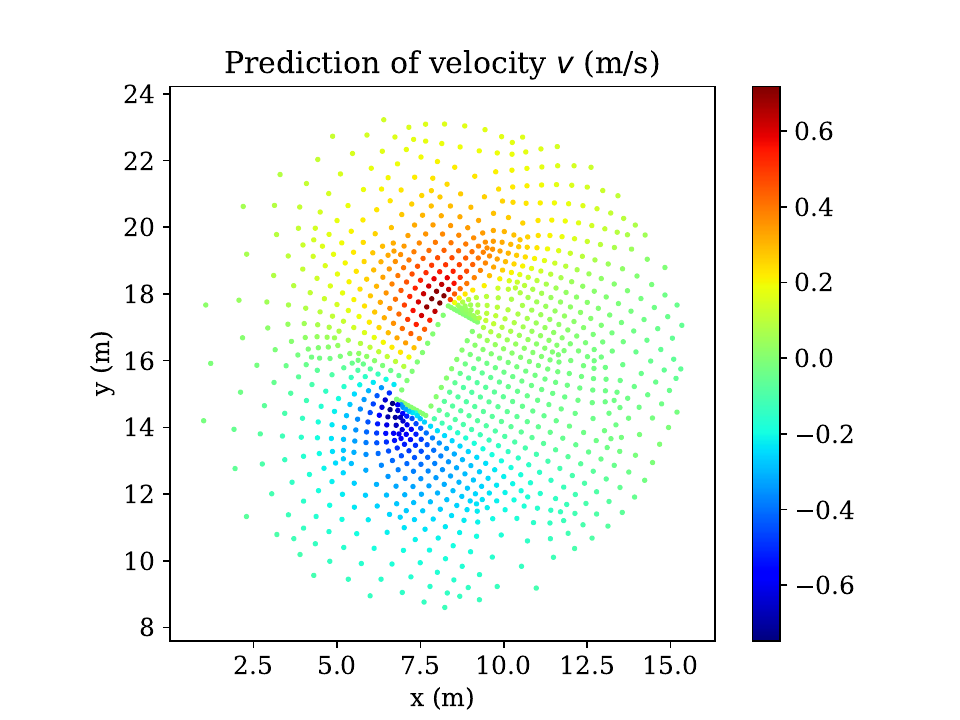}
      \end{subfigure}
    \begin{subfigure}[b]{0.32\textwidth}
        \centering
        \includegraphics[width=\textwidth]{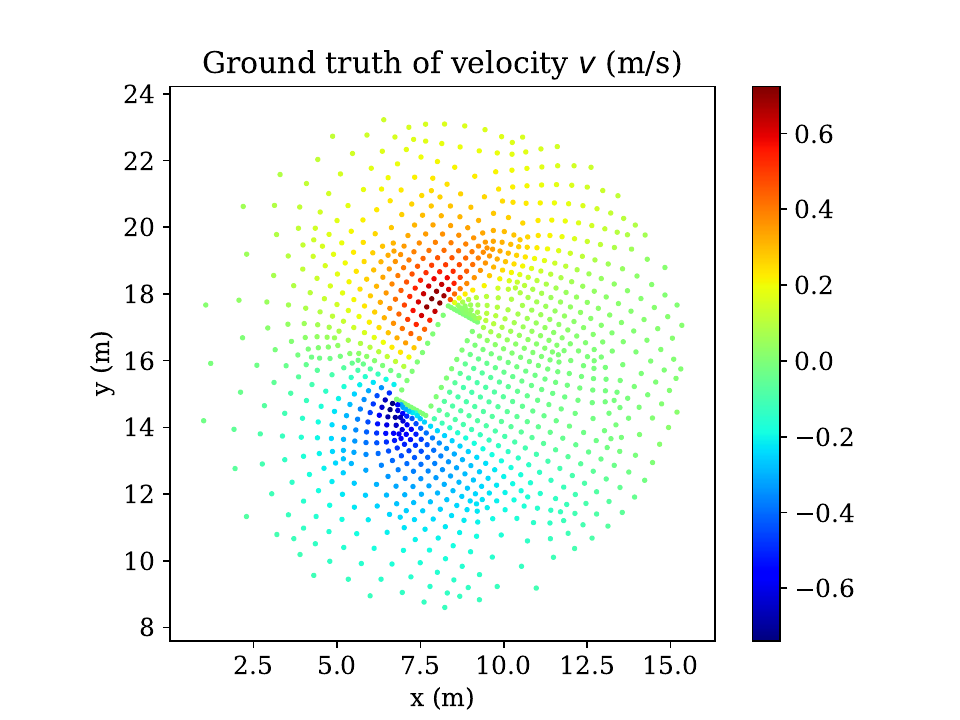}
      \end{subfigure}
    \begin{subfigure}[b]{0.32\textwidth}
        \centering
        \includegraphics[width=\textwidth]{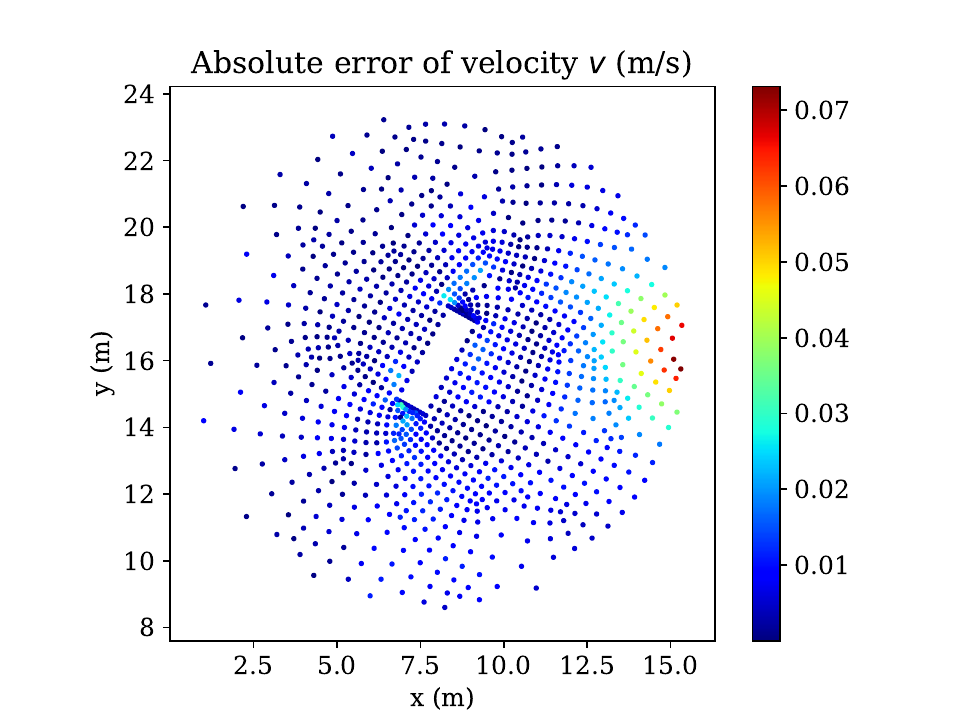}
      \end{subfigure}

    
    \begin{subfigure}[b]{0.32\textwidth}
        \centering
        \includegraphics[width=\textwidth]{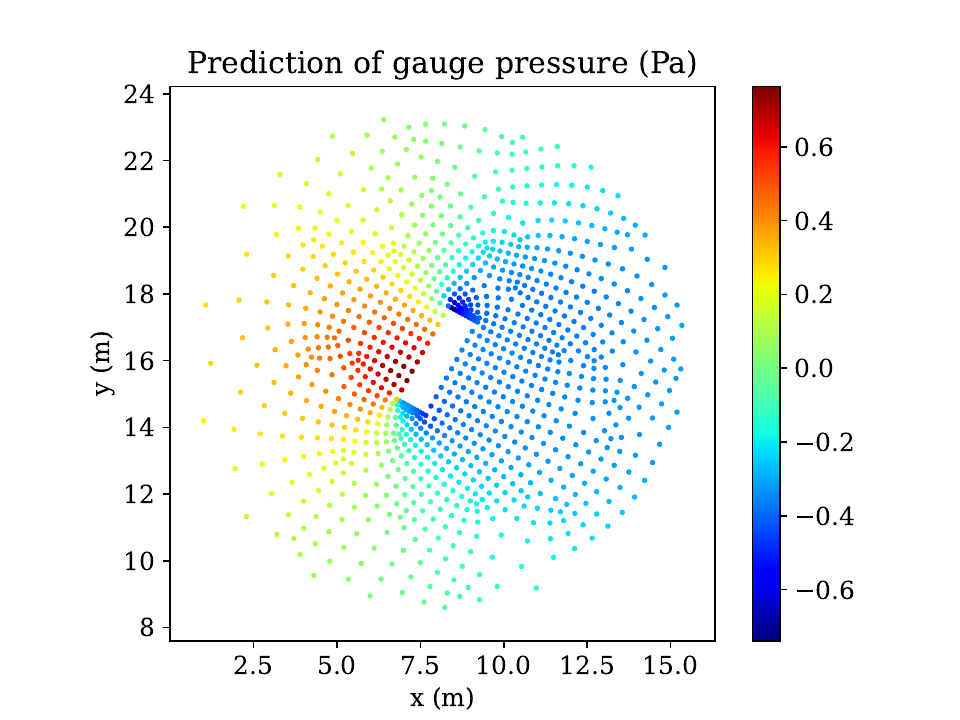}
      \end{subfigure}
    \begin{subfigure}[b]{0.32\textwidth}
        \centering
        \includegraphics[width=\textwidth]{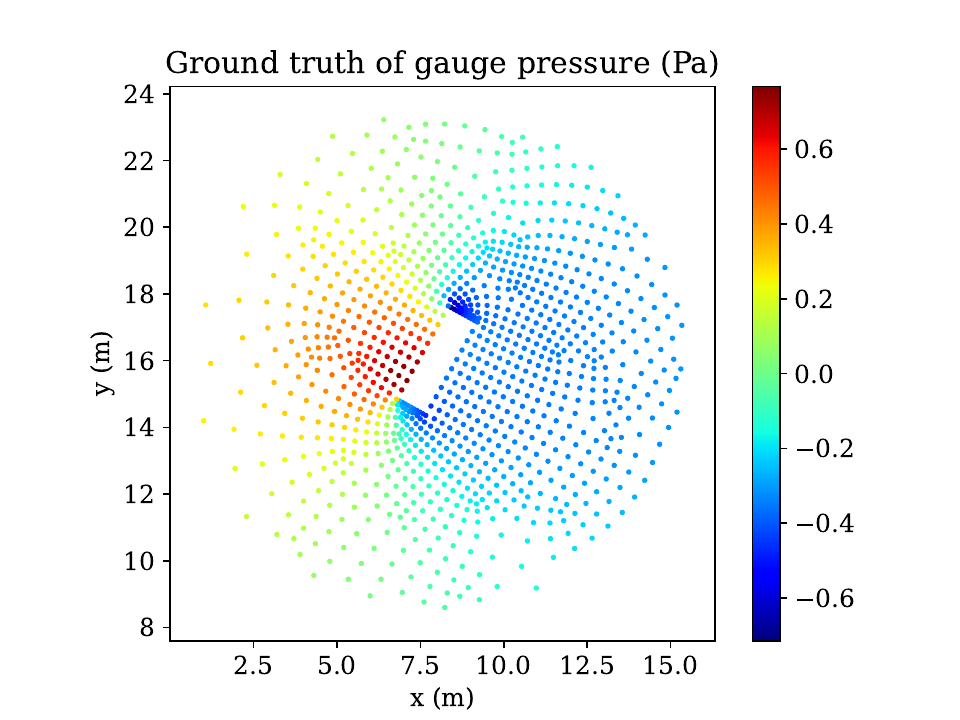}
      \end{subfigure}
    \begin{subfigure}[b]{0.32\textwidth}
        \centering
        \includegraphics[width=\textwidth]{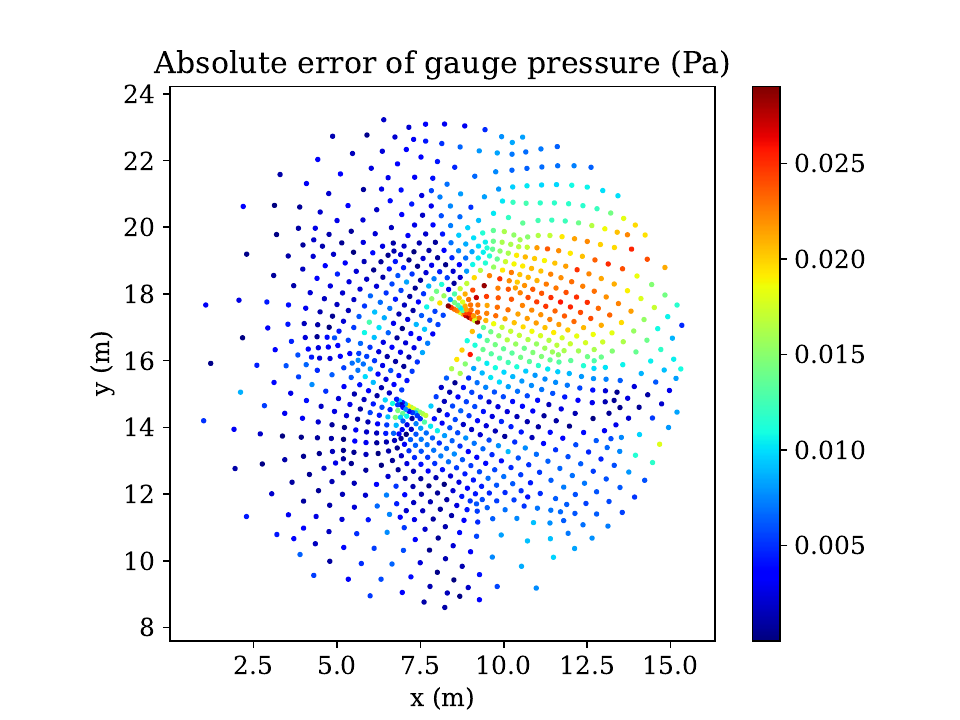}
      \end{subfigure}

  \caption{The sixth set of examples comparing the ground truth to the predictions of Kolmogorov-Arnold PointNet (i.e., KA-PointNet) for the velocity and pressure fields from the test set. The Jacobi polynomial used has a degree of 5, with $\alpha=\beta=1$. Here, $n_s=1$ is set.}
  \label{Fig10}
\end{figure}

\begin{figure}[!htbp]
  \centering 
      \begin{subfigure}[b]{0.32\textwidth}
        \centering
        \includegraphics[width=\textwidth]{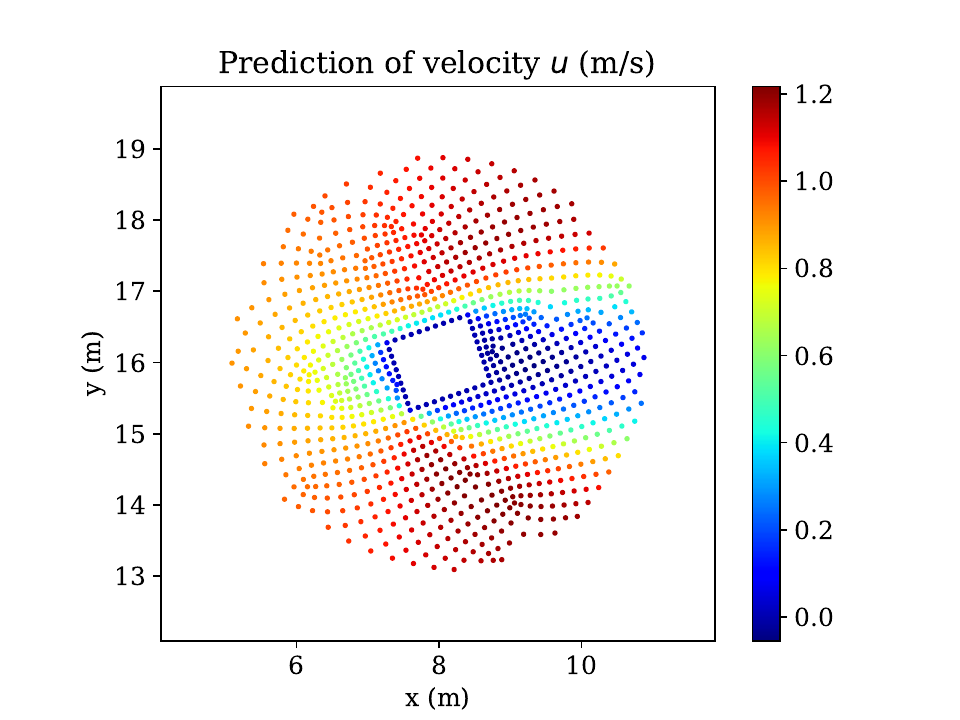}
      \end{subfigure}
    \begin{subfigure}[b]{0.32\textwidth}
        \centering
        \includegraphics[width=\textwidth]{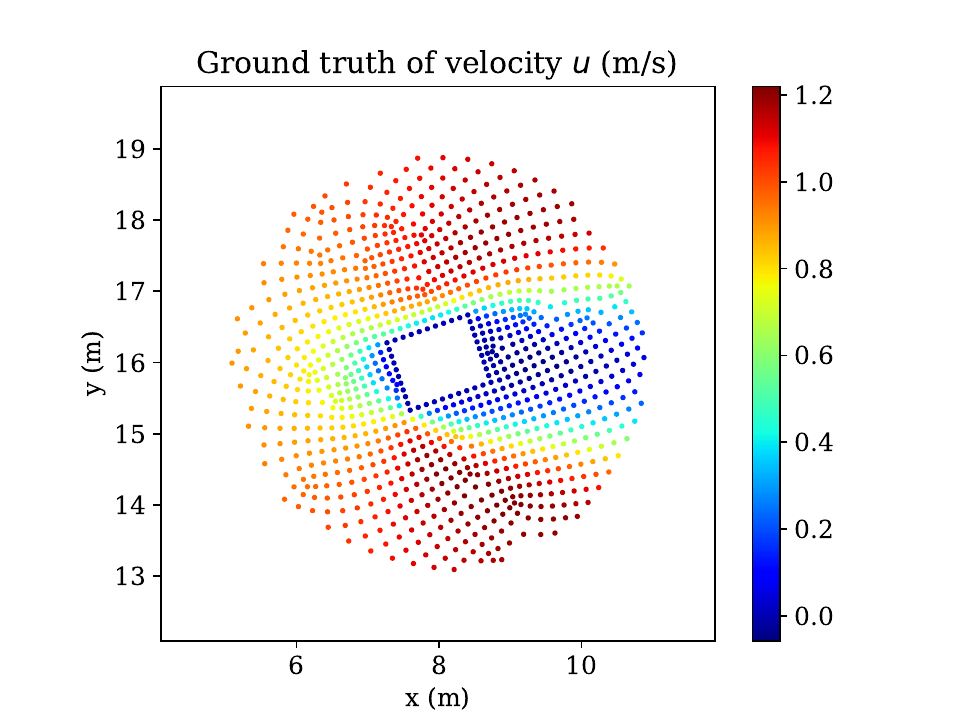}
      \end{subfigure}
    \begin{subfigure}[b]{0.32\textwidth}
        \centering
        \includegraphics[width=\textwidth]{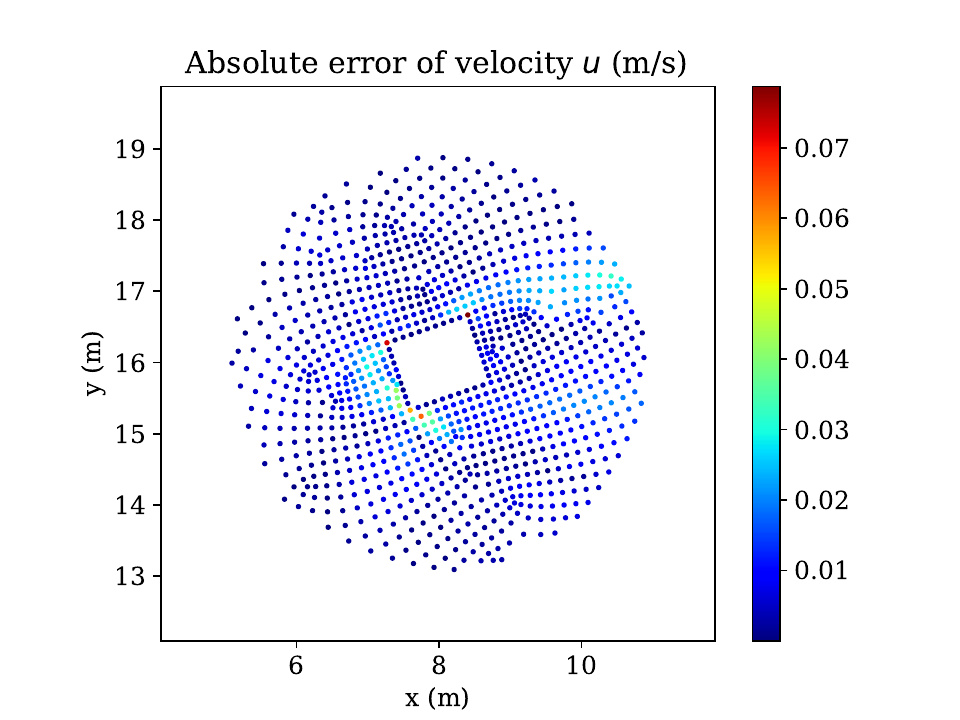}
      \end{subfigure}

    
    \begin{subfigure}[b]{0.32\textwidth}
        \centering
        \includegraphics[width=\textwidth]{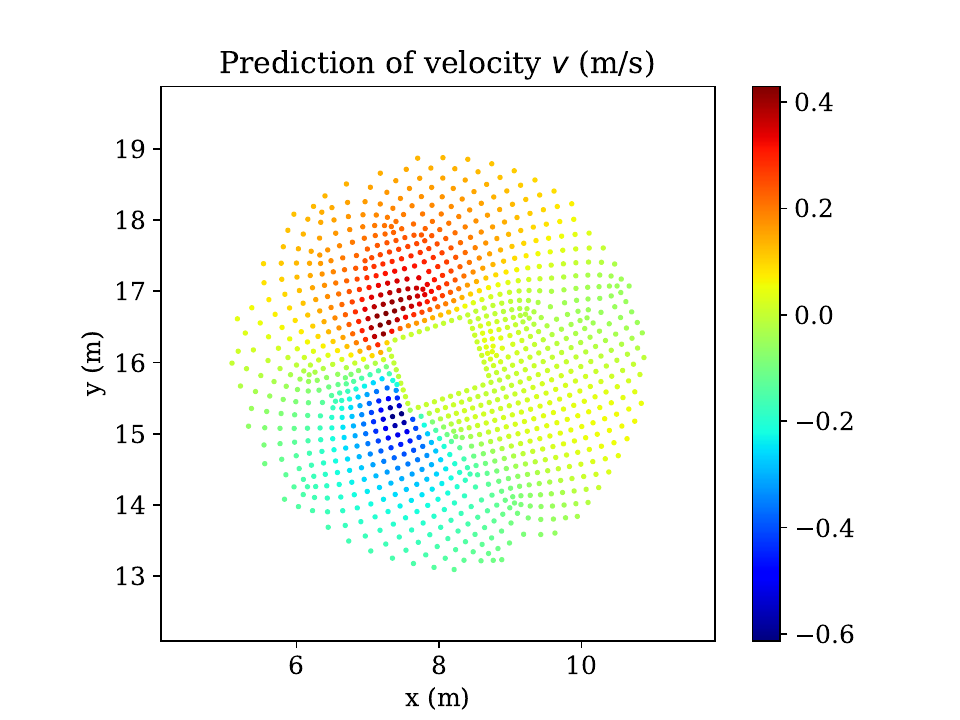}
      \end{subfigure}
    \begin{subfigure}[b]{0.32\textwidth}
        \centering
        \includegraphics[width=\textwidth]{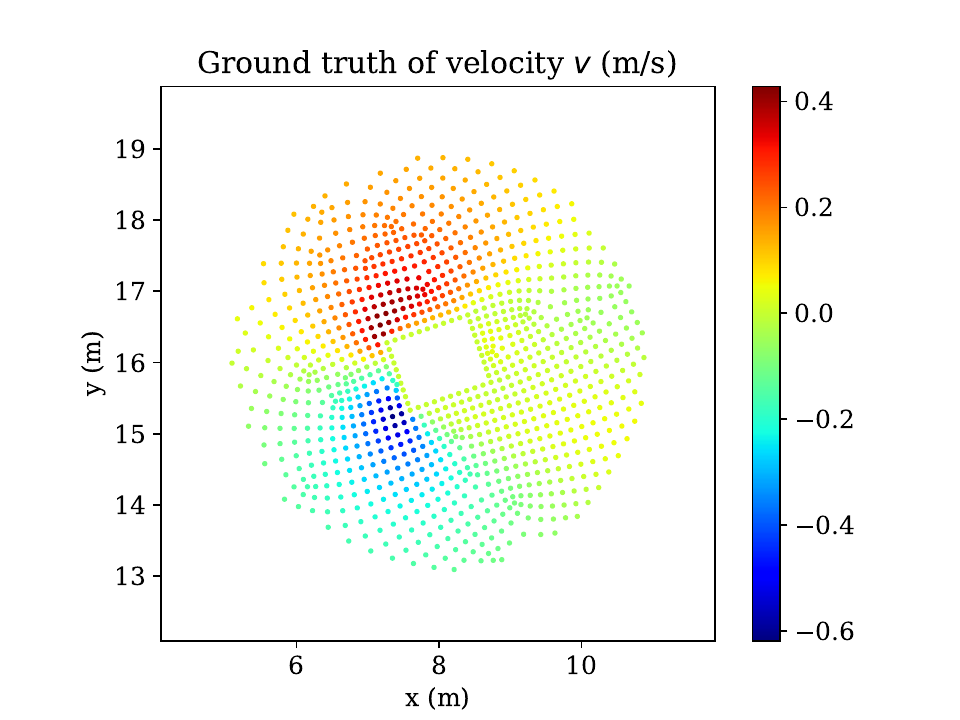}
      \end{subfigure}
    \begin{subfigure}[b]{0.32\textwidth}
        \centering
        \includegraphics[width=\textwidth]{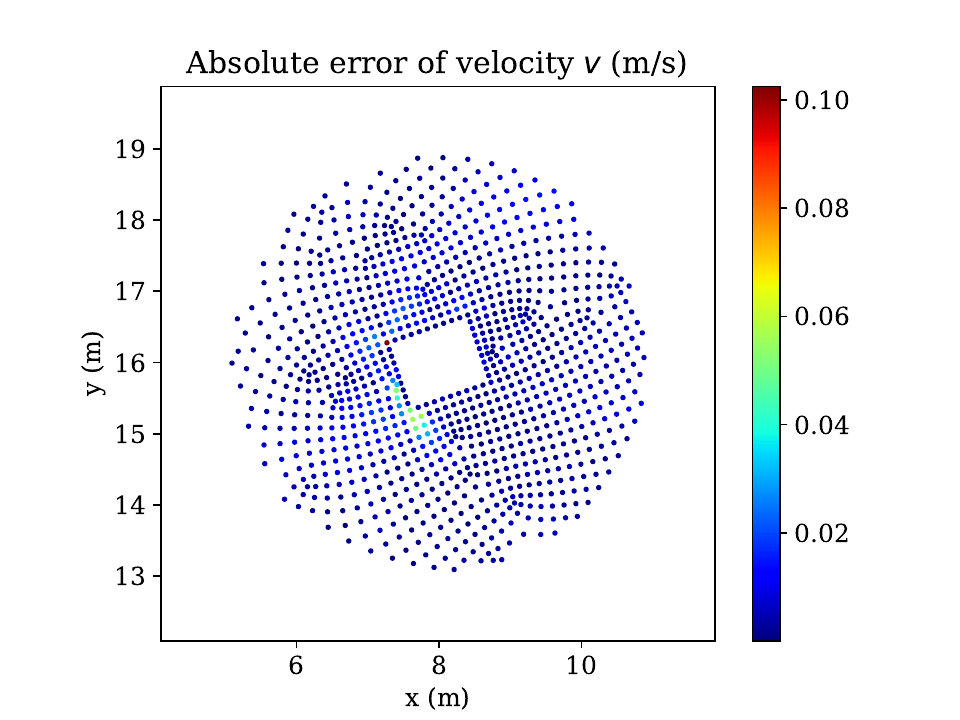}
      \end{subfigure}

    
    \begin{subfigure}[b]{0.32\textwidth}
        \centering
        \includegraphics[width=\textwidth]{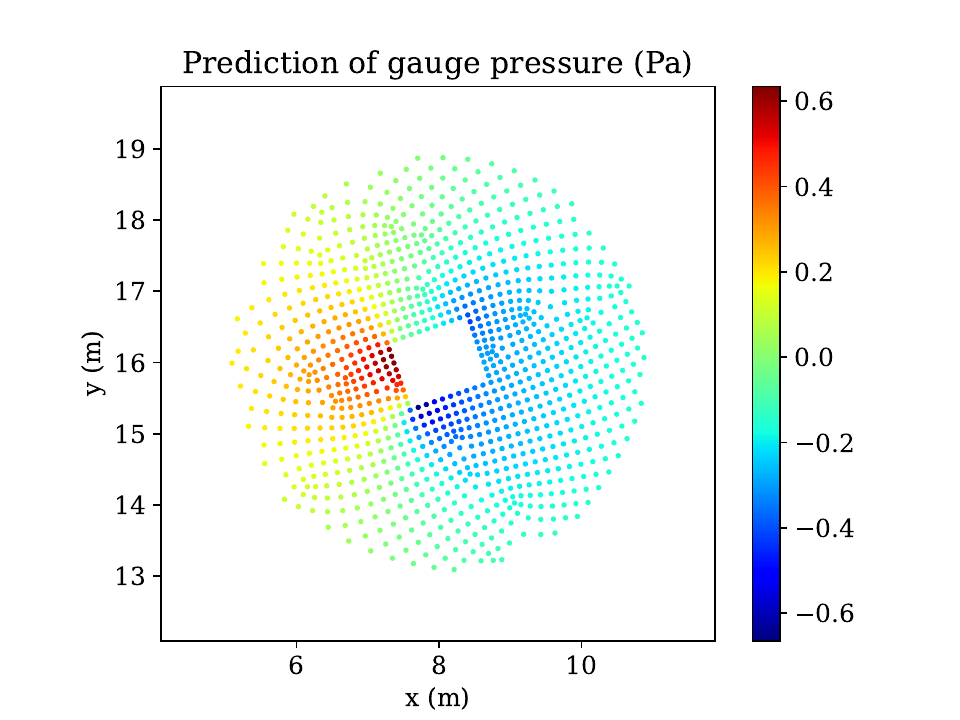}
      \end{subfigure}
    \begin{subfigure}[b]{0.32\textwidth}
        \centering
        \includegraphics[width=\textwidth]{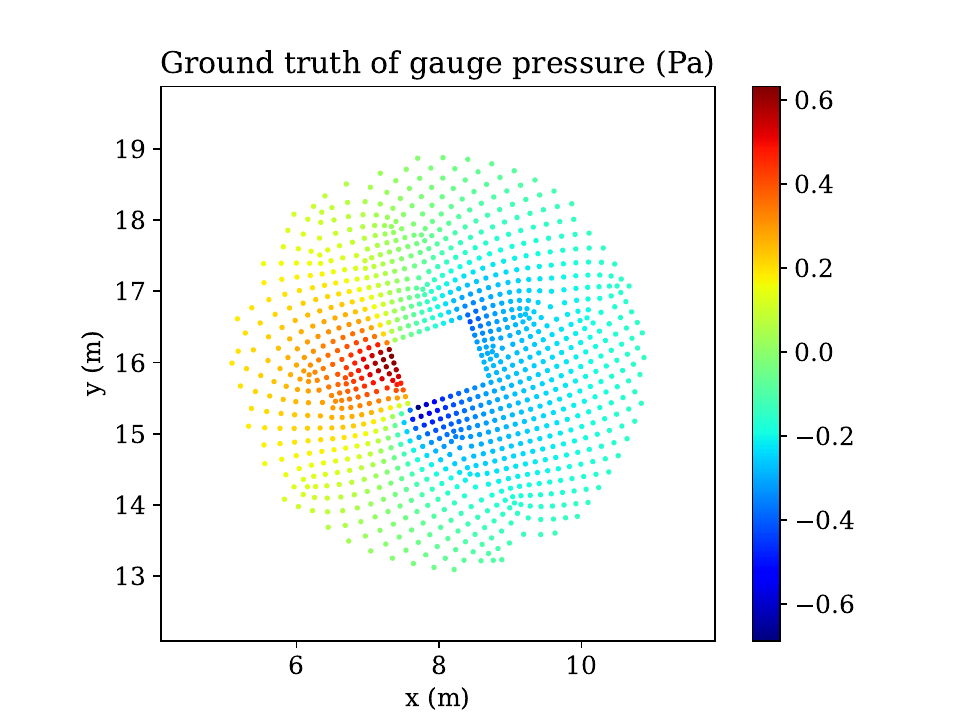}
      \end{subfigure}
    \begin{subfigure}[b]{0.32\textwidth}
        \centering
        \includegraphics[width=\textwidth]{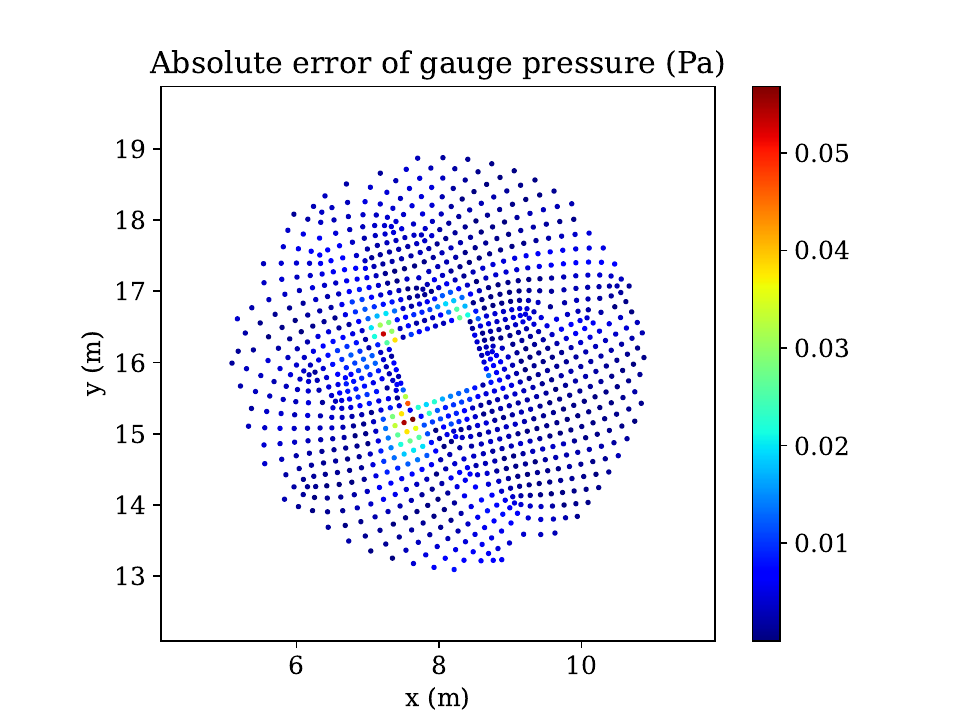}
      \end{subfigure}

    
  \caption{The seventh set of examples comparing the ground truth to the predictions of Kolmogorov-Arnold PointNet (i.e., KA-PointNet) for the velocity and pressure fields from the test set. The Jacobi polynomial used has a degree of 5, with $\alpha=\beta=1$. Here, $n_s=1$ is set.}
  \label{Fig11}
\end{figure}

\begin{figure}[!htbp]
  \centering 
      \begin{subfigure}[b]{0.32\textwidth}
        \centering
        \includegraphics[width=\textwidth]{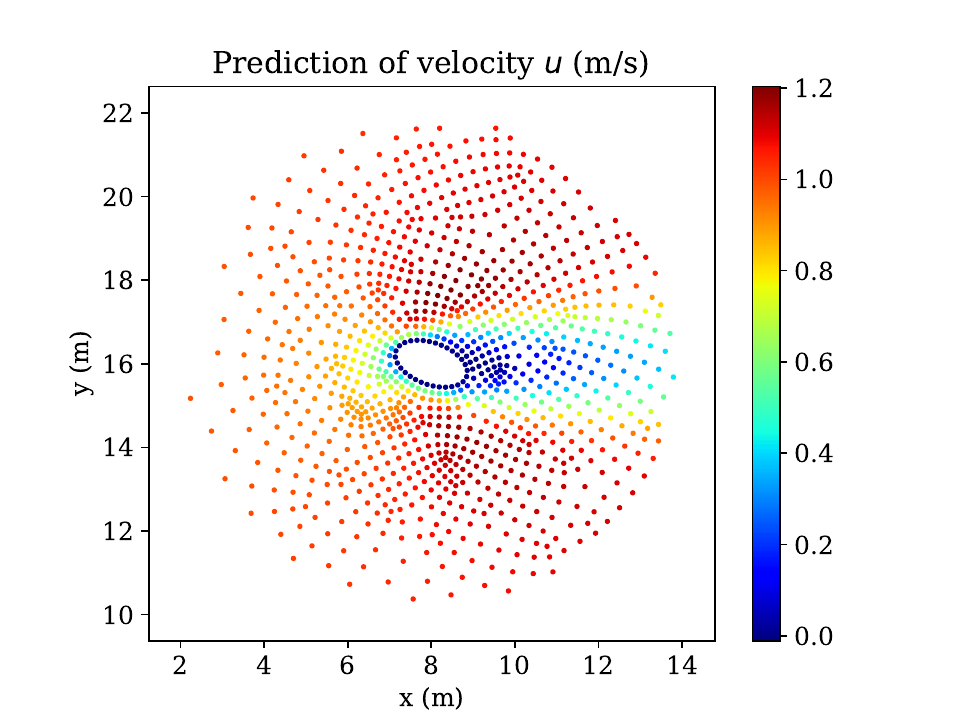}
      \end{subfigure}
    \begin{subfigure}[b]{0.32\textwidth}
        \centering
        \includegraphics[width=\textwidth]{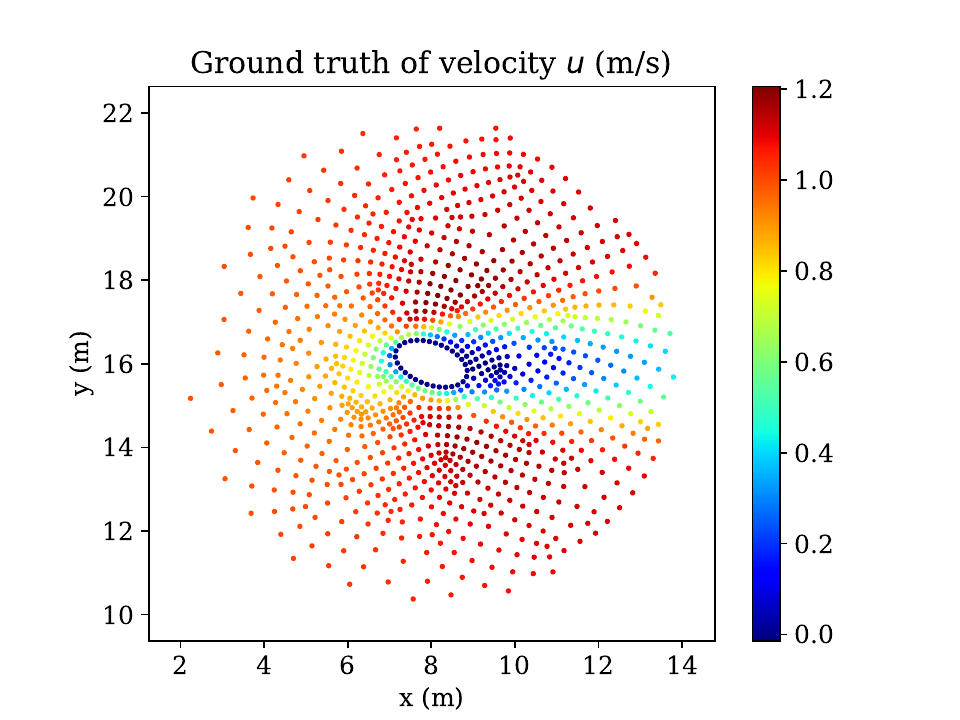}
      \end{subfigure}
    \begin{subfigure}[b]{0.32\textwidth}
        \centering
        \includegraphics[width=\textwidth]{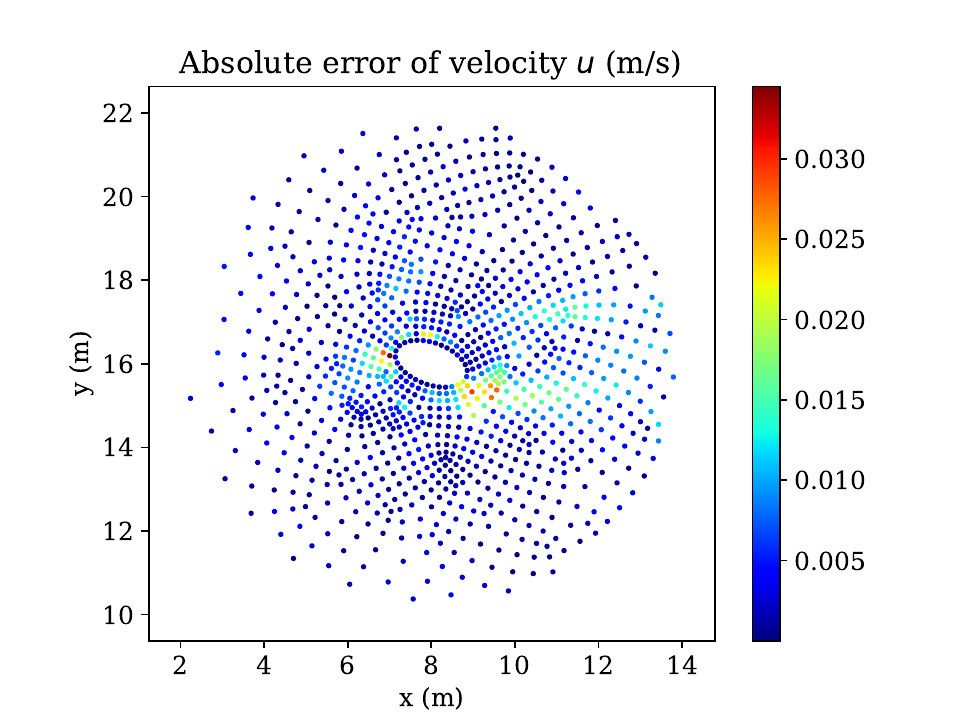}
      \end{subfigure}

    
    \begin{subfigure}[b]{0.32\textwidth}
        \centering
        \includegraphics[width=\textwidth]{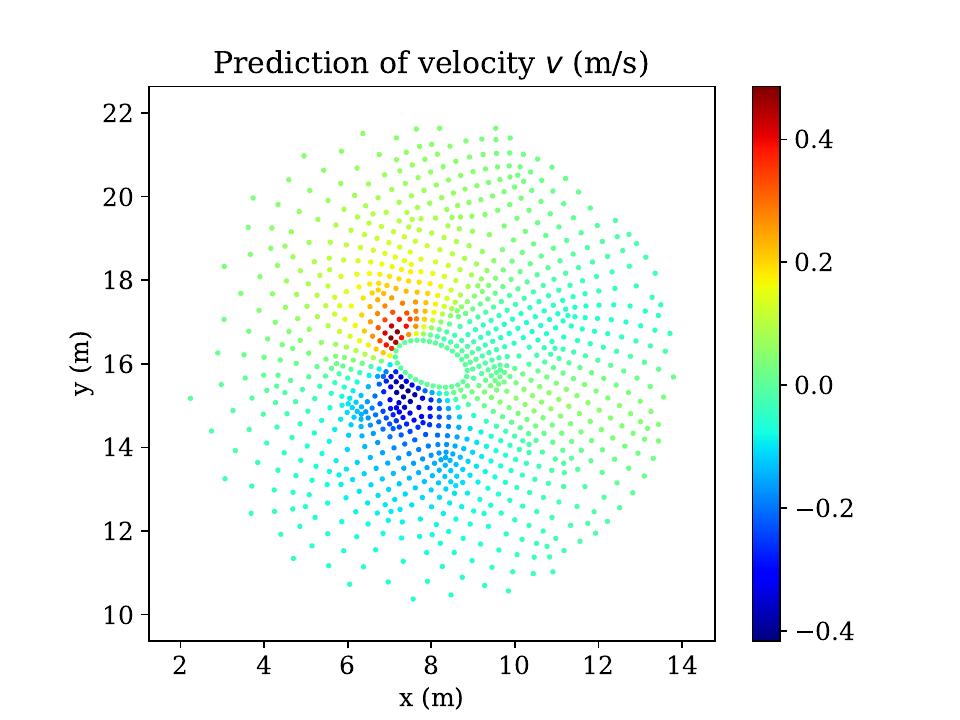}
      \end{subfigure}
    \begin{subfigure}[b]{0.32\textwidth}
        \centering
        \includegraphics[width=\textwidth]{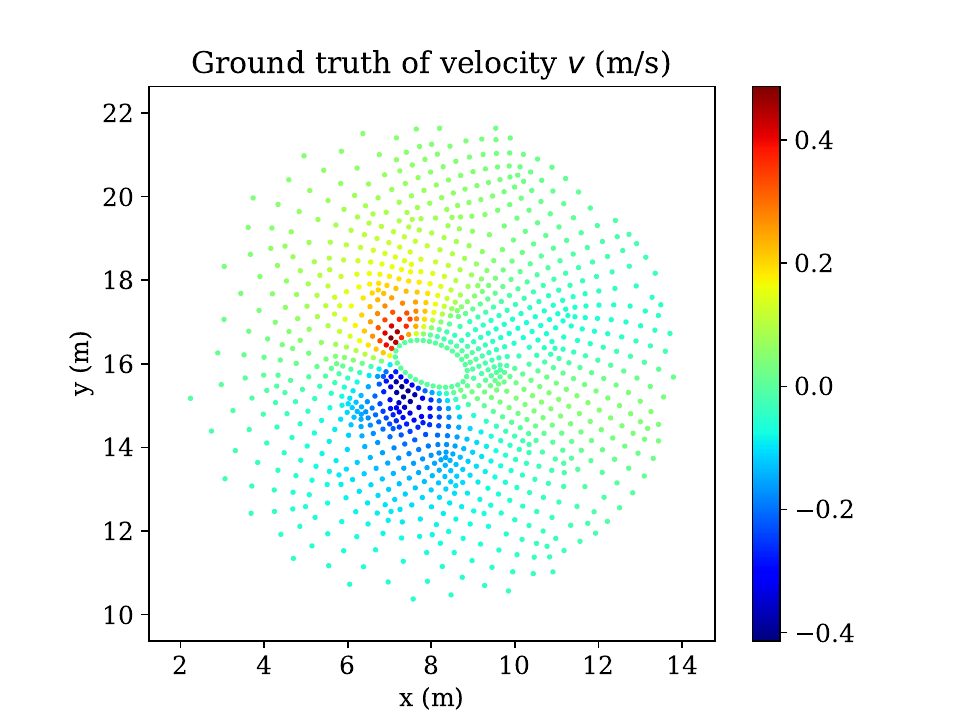}
      \end{subfigure}
    \begin{subfigure}[b]{0.32\textwidth}
        \centering
        \includegraphics[width=\textwidth]{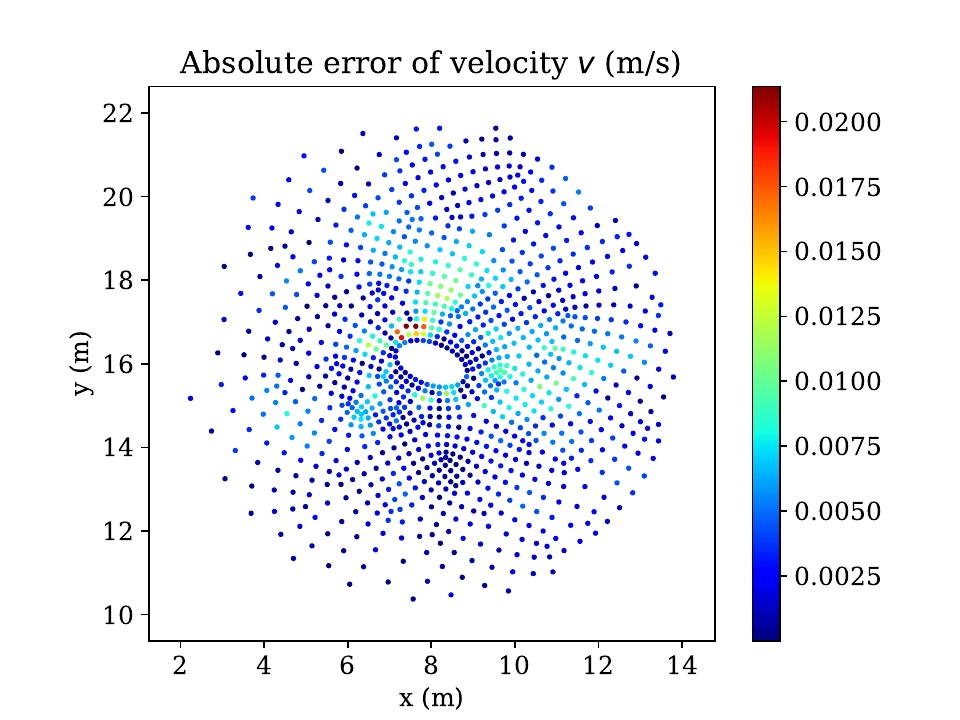}
      \end{subfigure}

    
    \begin{subfigure}[b]{0.32\textwidth}
        \centering
        \includegraphics[width=\textwidth]{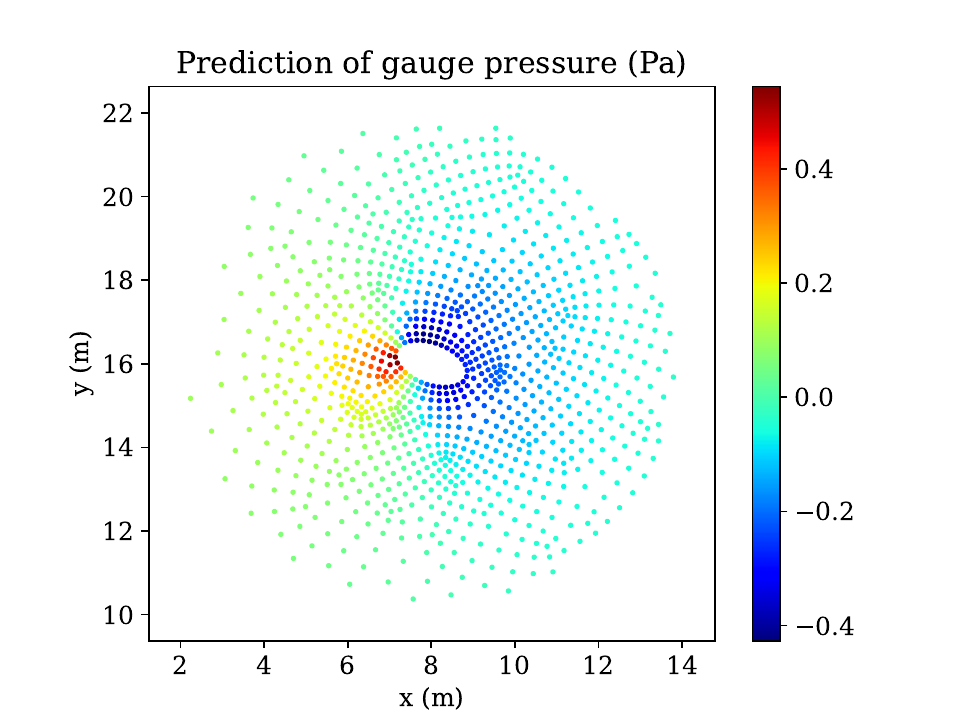}
      \end{subfigure}
    \begin{subfigure}[b]{0.32\textwidth}
        \centering
        \includegraphics[width=\textwidth]{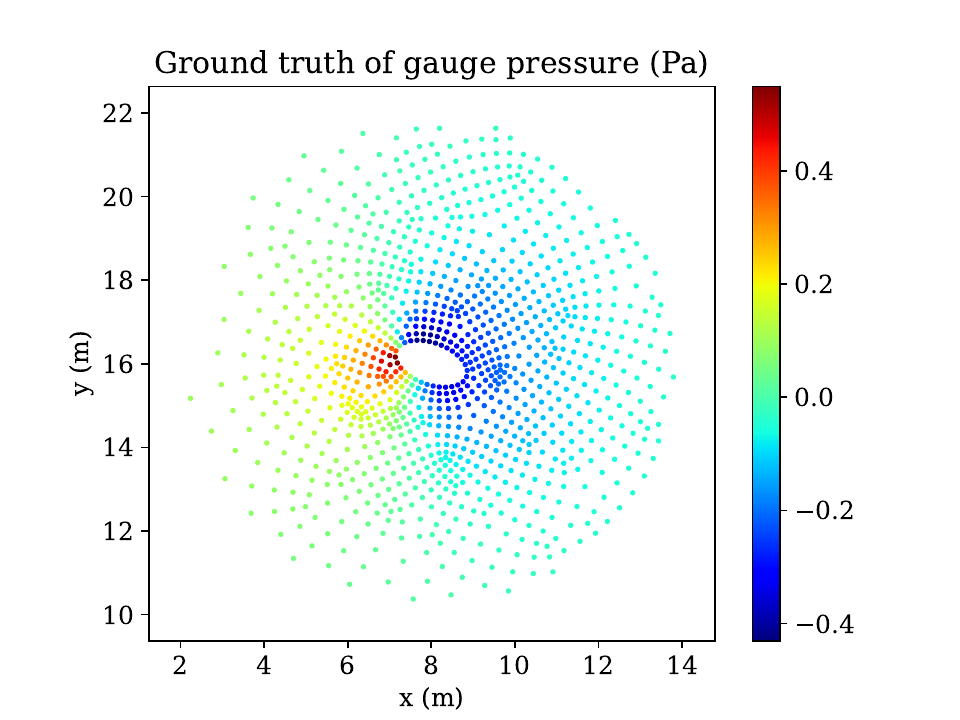}
      \end{subfigure}
    \begin{subfigure}[b]{0.32\textwidth}
        \centering
        \includegraphics[width=\textwidth]{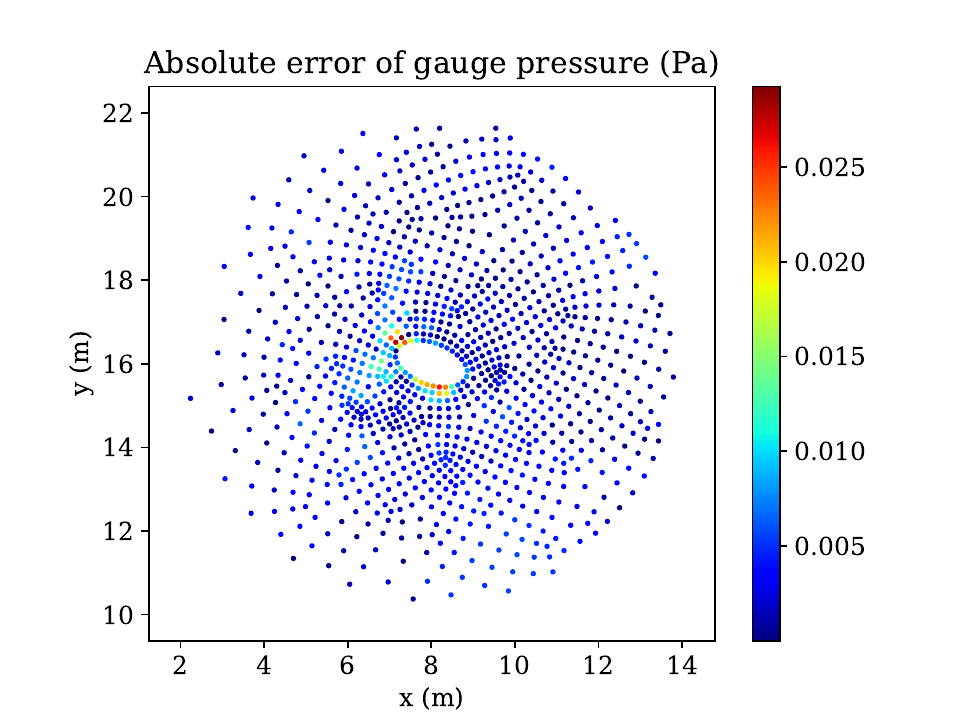}
      \end{subfigure}

  \caption{The eighth set of examples comparing the ground truth to the predictions of Kolmogorov-Arnold PointNet (i.e., KA-PointNet) for the velocity and pressure fields from the test set. The Jacobi polynomial used has a degree of 5, with $\alpha=\beta=1$. Here, $n_s=1$ is set.}
  \label{Fig12}
\end{figure}

\begin{figure}[!htbp]
  \centering 
      \begin{subfigure}[b]{0.32\textwidth}
        \centering
        \includegraphics[width=\textwidth]{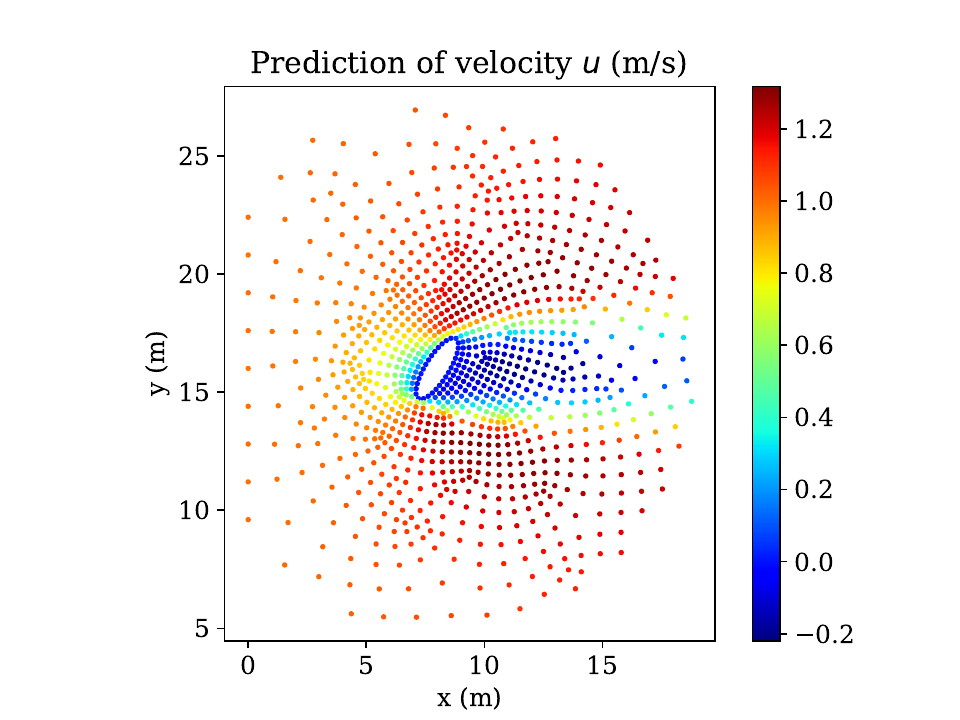}
      \end{subfigure}
    \begin{subfigure}[b]{0.32\textwidth}
        \centering
        \includegraphics[width=\textwidth]{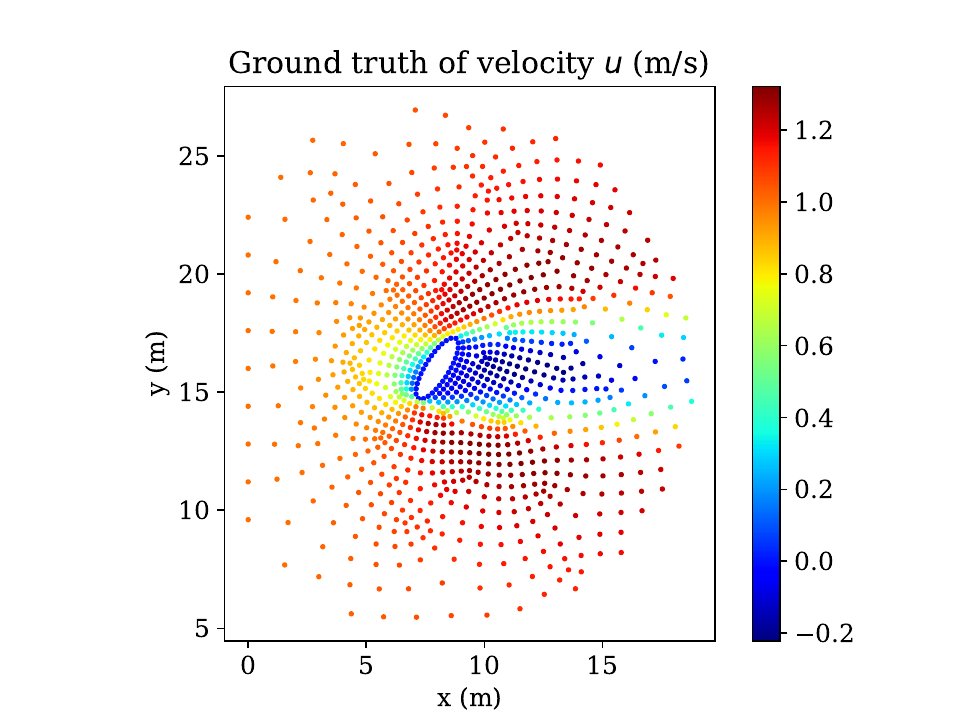}
      \end{subfigure}
    \begin{subfigure}[b]{0.32\textwidth}
        \centering
        \includegraphics[width=\textwidth]{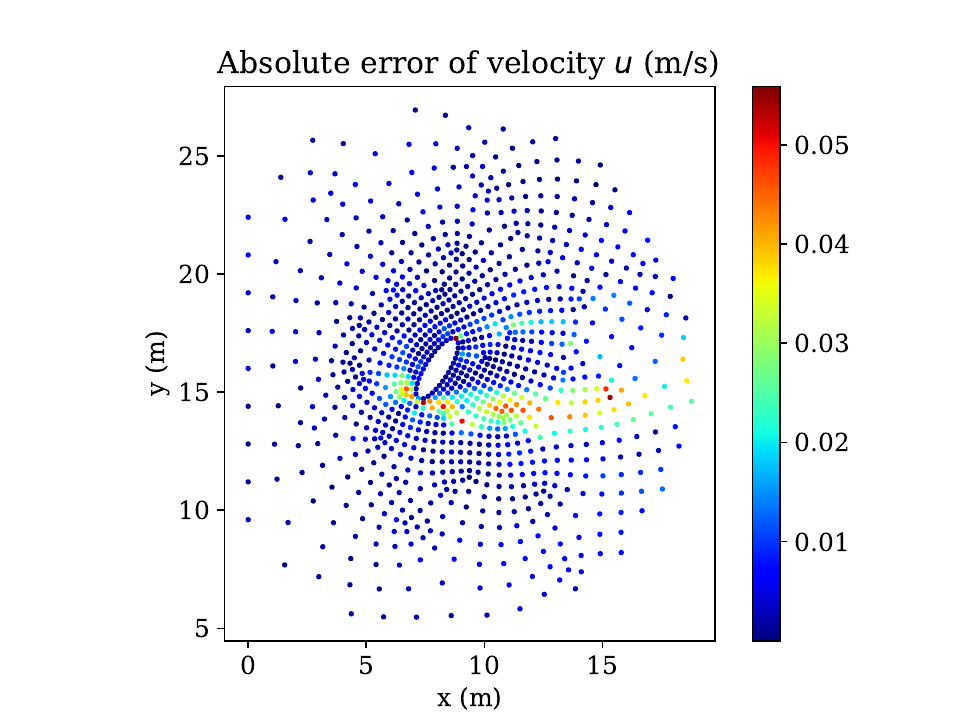}
      \end{subfigure}

    
    \begin{subfigure}[b]{0.32\textwidth}
        \centering
        \includegraphics[width=\textwidth]{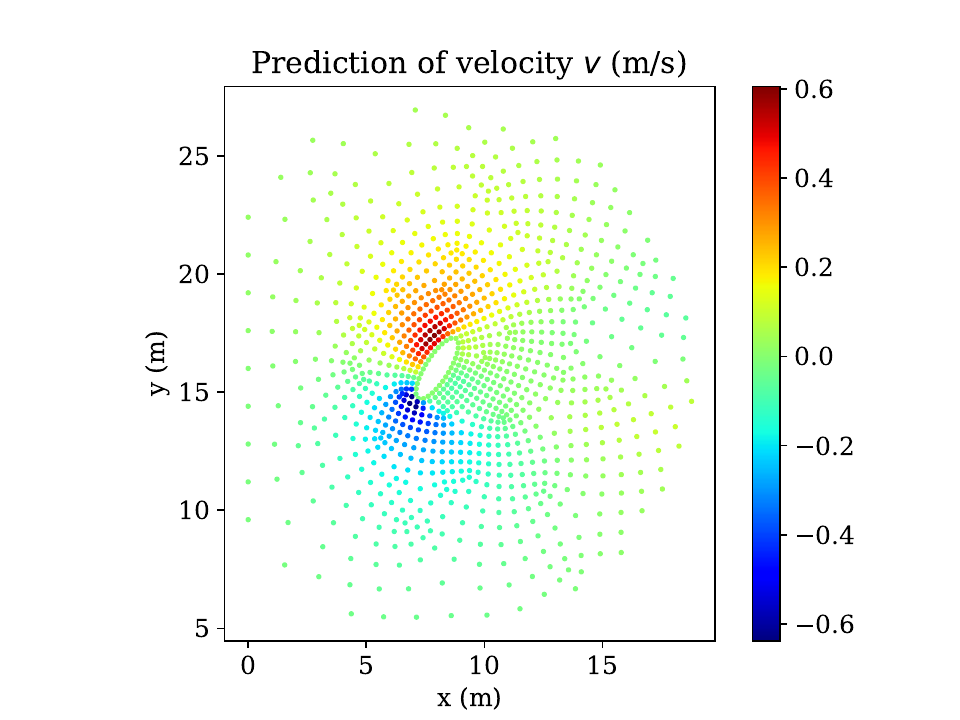}
      \end{subfigure}
    \begin{subfigure}[b]{0.32\textwidth}
        \centering
        \includegraphics[width=\textwidth]{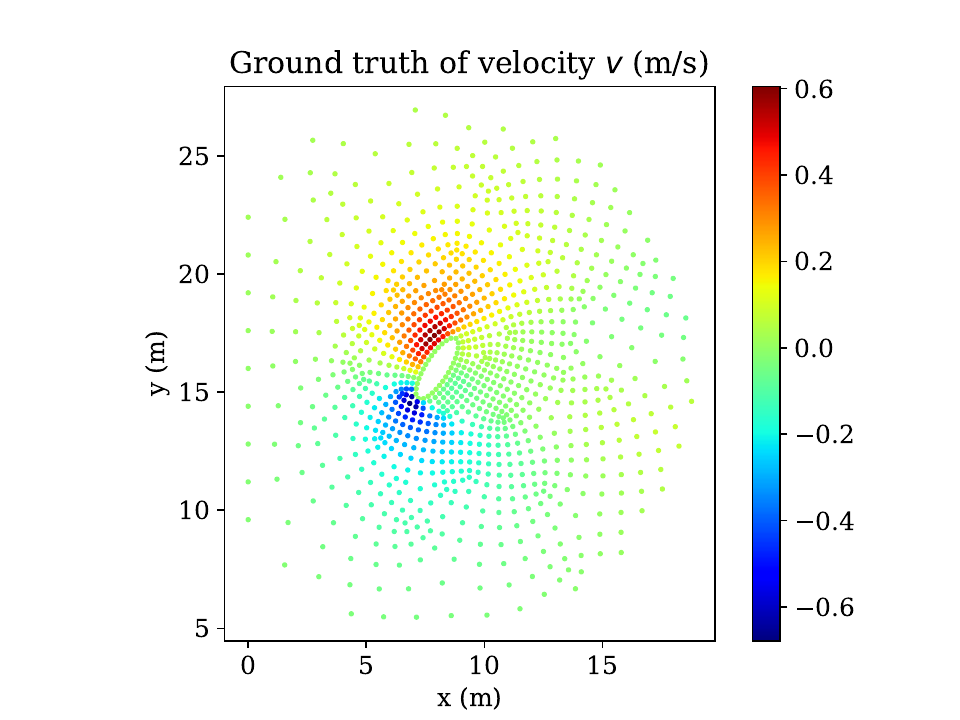}
      \end{subfigure}
    \begin{subfigure}[b]{0.32\textwidth}
        \centering
        \includegraphics[width=\textwidth]{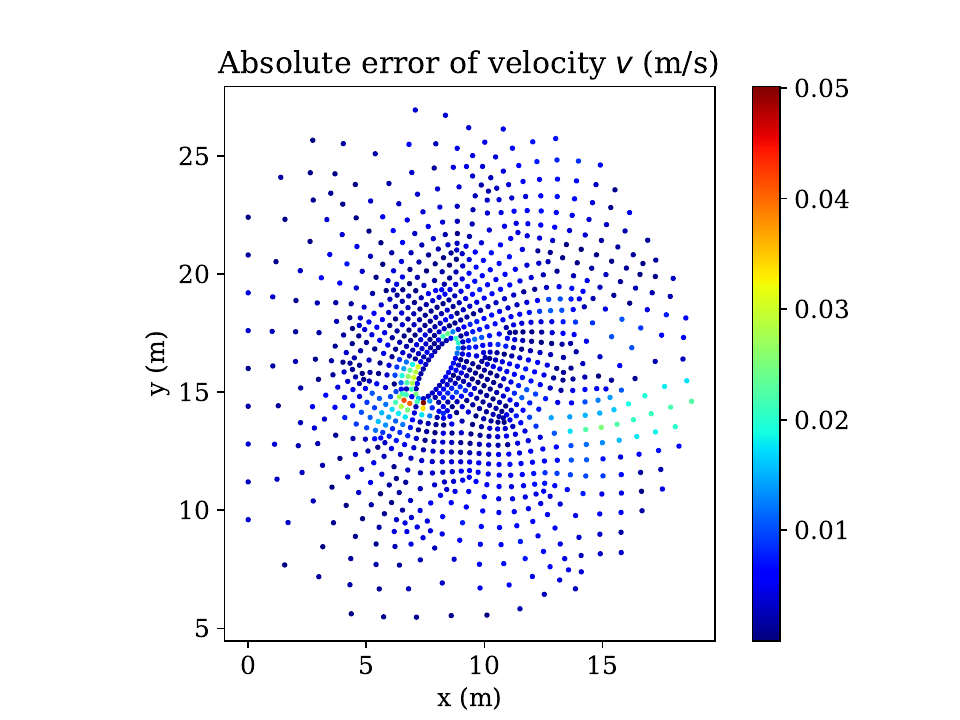}
      \end{subfigure}

    
    \begin{subfigure}[b]{0.32\textwidth}
        \centering
        \includegraphics[width=\textwidth]{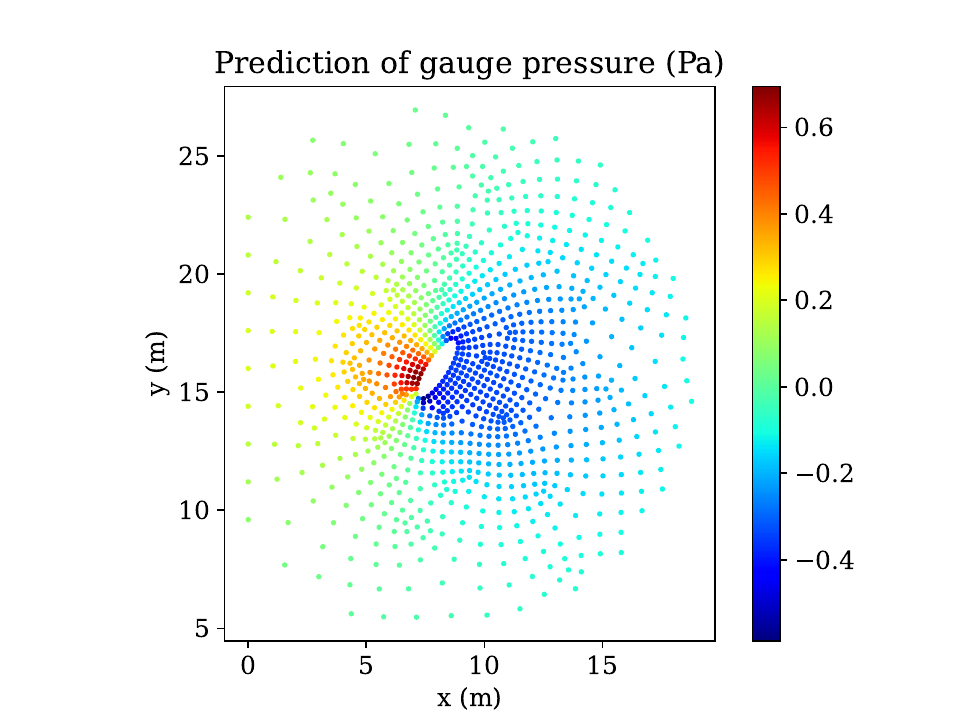}
      \end{subfigure}
    \begin{subfigure}[b]{0.32\textwidth}
        \centering
        \includegraphics[width=\textwidth]{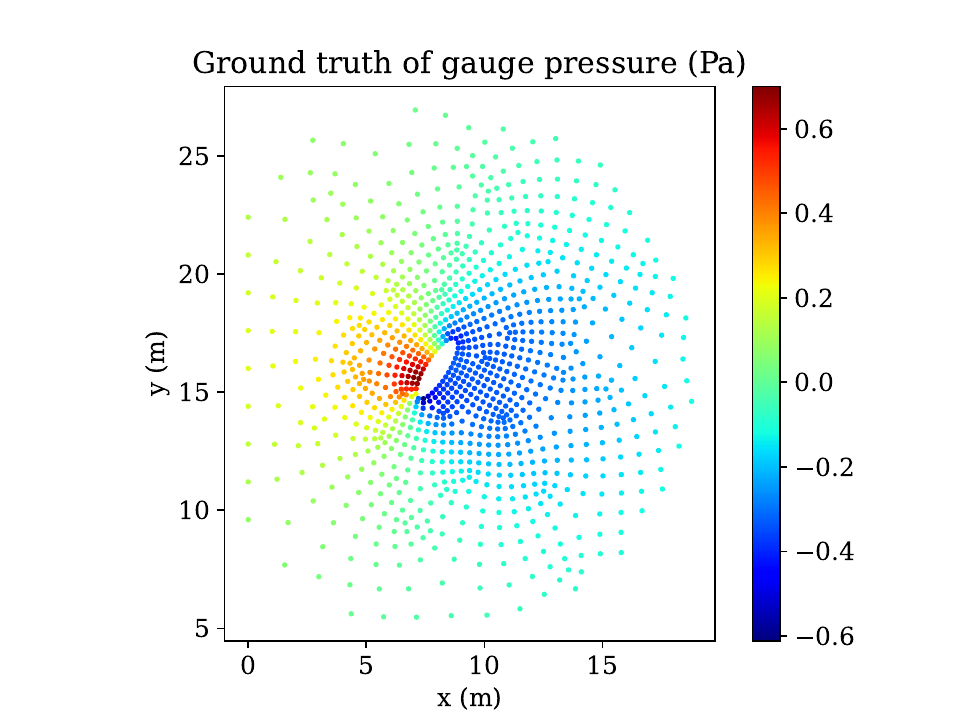}
      \end{subfigure}
    \begin{subfigure}[b]{0.32\textwidth}
        \centering
        \includegraphics[width=\textwidth]{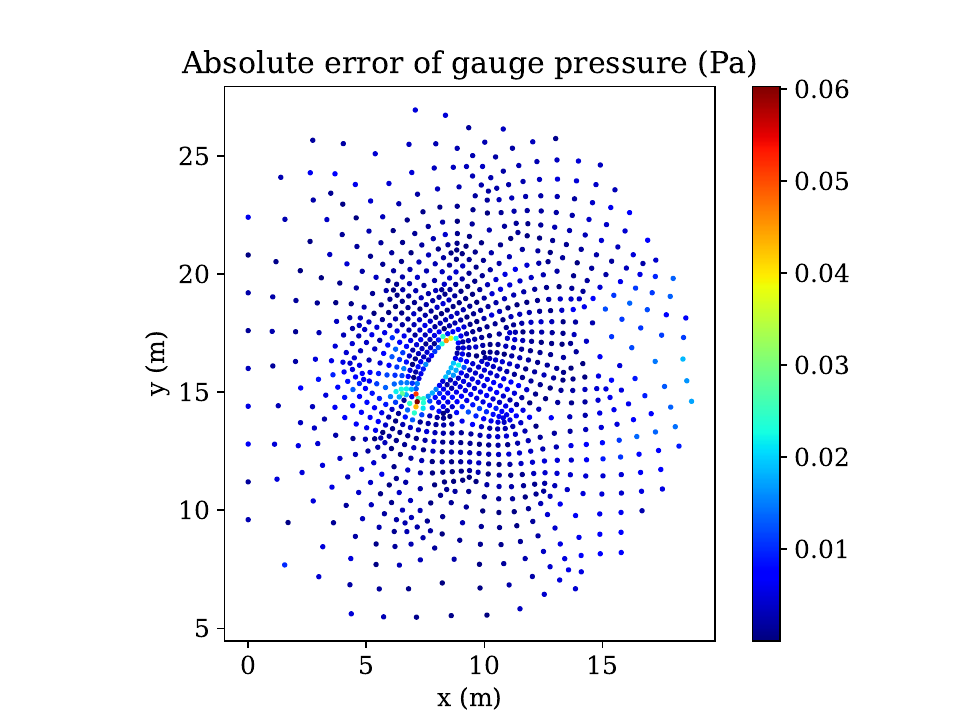}
      \end{subfigure}

  \caption{The ninth set of examples comparing the ground truth to the predictions of Kolmogorov-Arnold PointNet (i.e., KA-PointNet) for the velocity and pressure fields from the test set. The Jacobi polynomial used has a degree of 5, with $\alpha=\beta=1$. Here, $n_s=1$ is set.}
  \label{Fig13}
\end{figure}

\begin{figure}[!htbp]
  \centering 
      \begin{subfigure}[b]{0.32\textwidth}
        \centering
        \includegraphics[width=\textwidth]{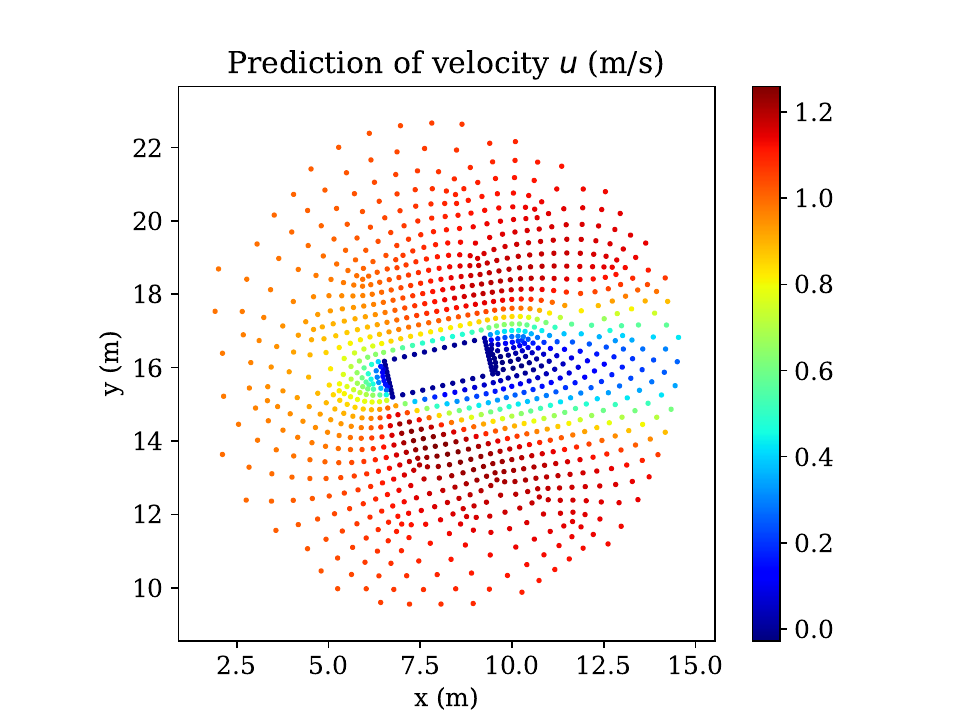}
      \end{subfigure}
    \begin{subfigure}[b]{0.32\textwidth}
        \centering
        \includegraphics[width=\textwidth]{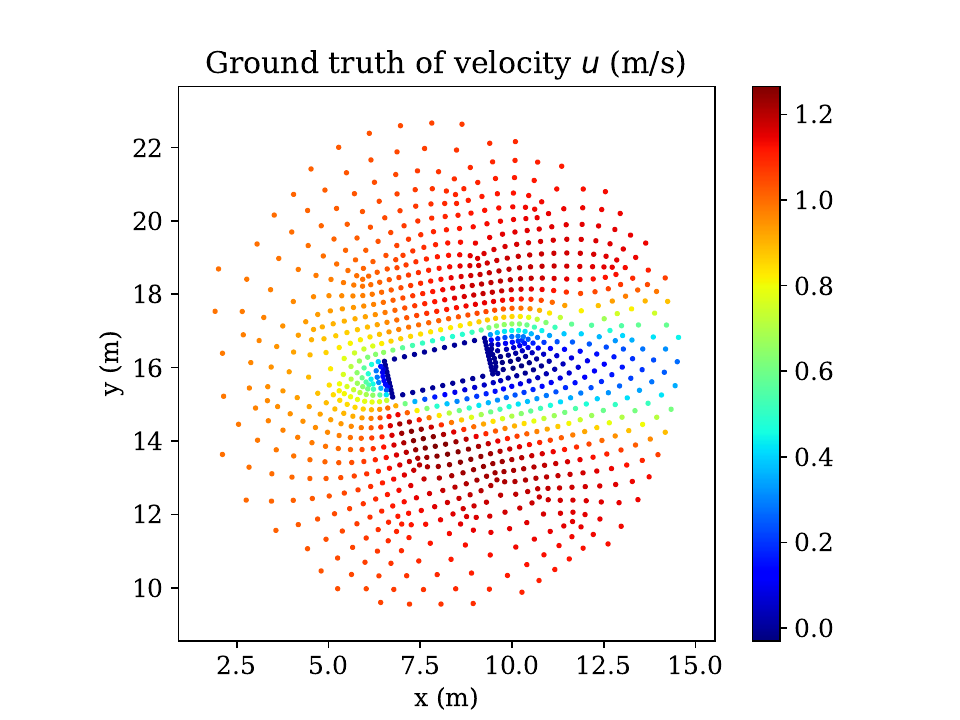}
      \end{subfigure}
    \begin{subfigure}[b]{0.32\textwidth}
        \centering
        \includegraphics[width=\textwidth]{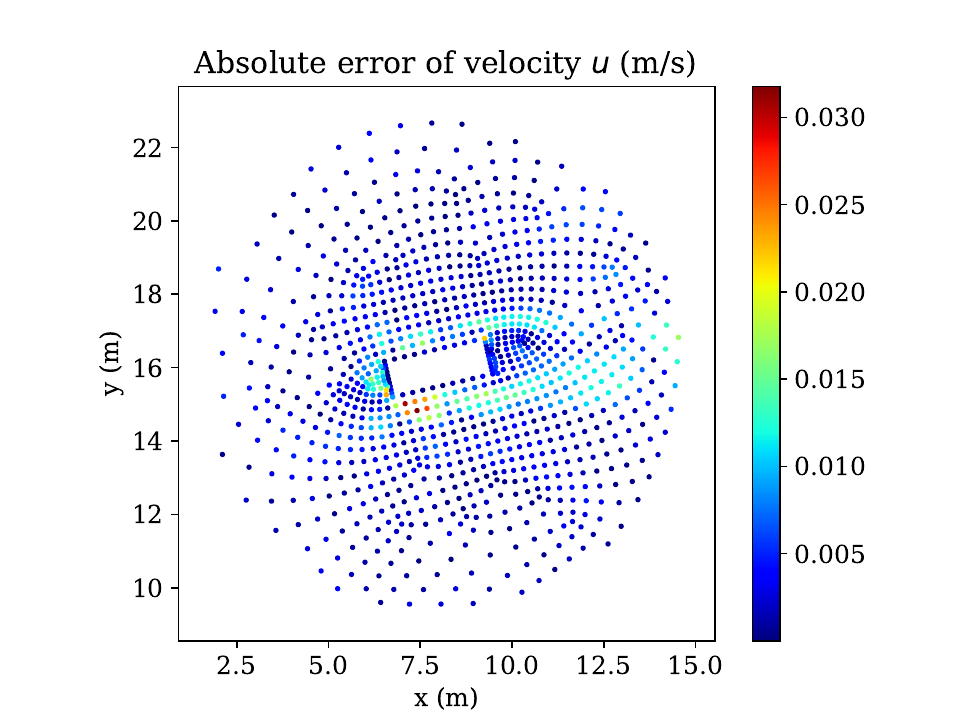}
      \end{subfigure}

    
    \begin{subfigure}[b]{0.32\textwidth}
        \centering
        \includegraphics[width=\textwidth]{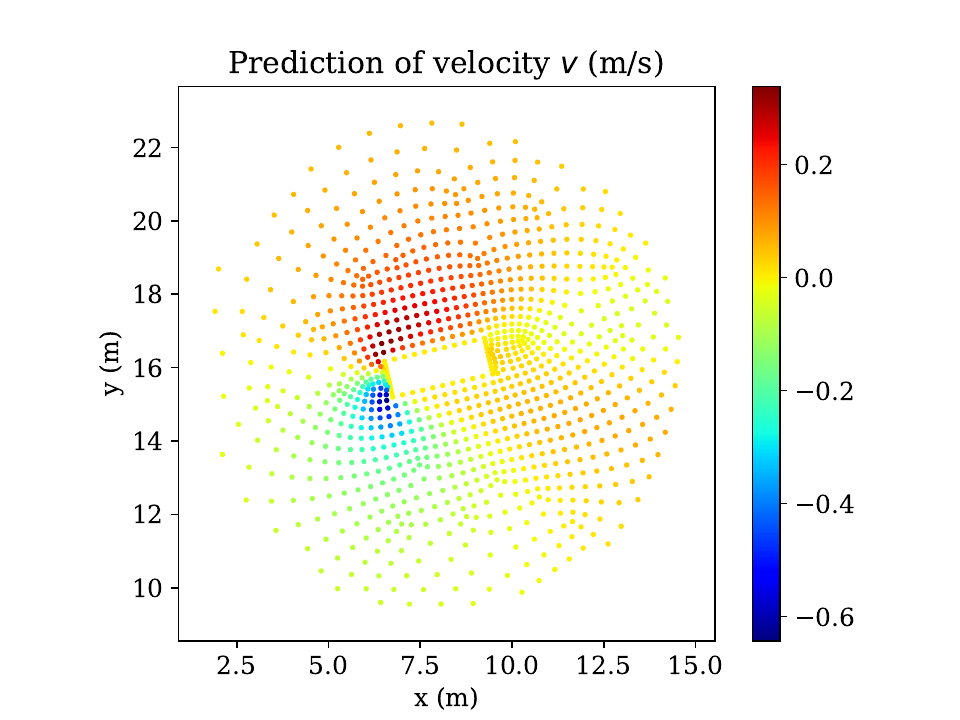}
      \end{subfigure}
    \begin{subfigure}[b]{0.32\textwidth}
        \centering
        \includegraphics[width=\textwidth]{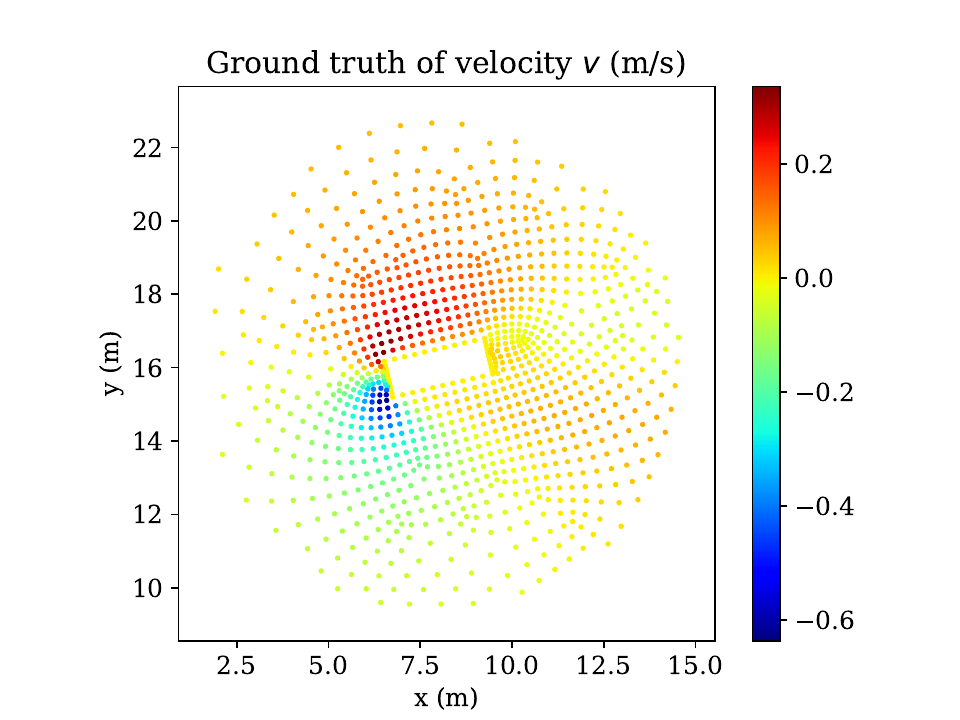}
      \end{subfigure}
    \begin{subfigure}[b]{0.32\textwidth}
        \centering
        \includegraphics[width=\textwidth]{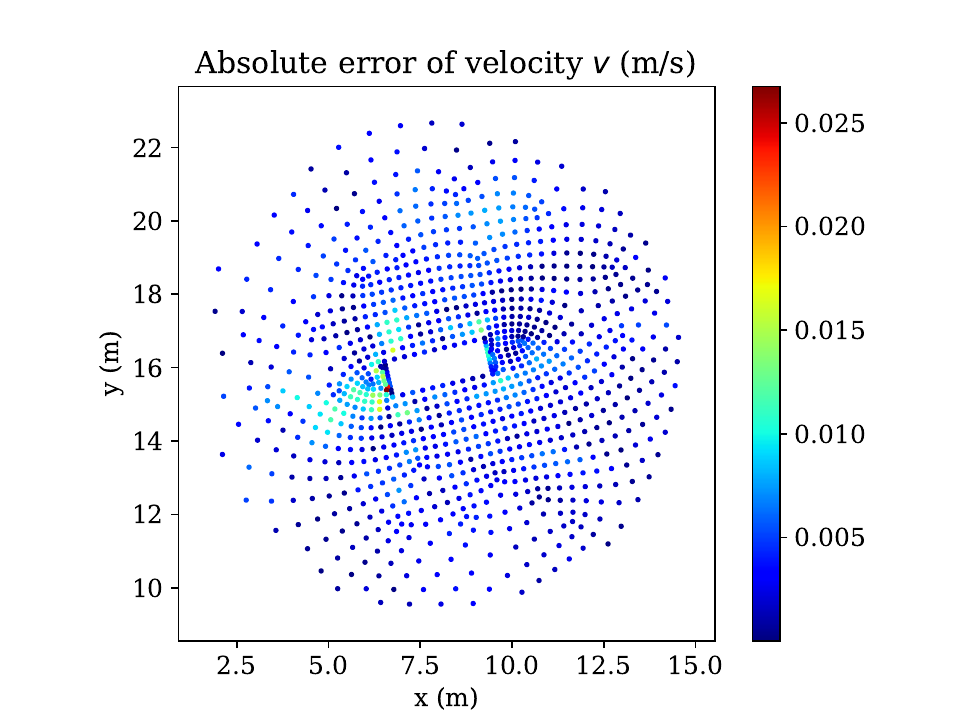}
      \end{subfigure}

    
    \begin{subfigure}[b]{0.32\textwidth}
        \centering
        \includegraphics[width=\textwidth]{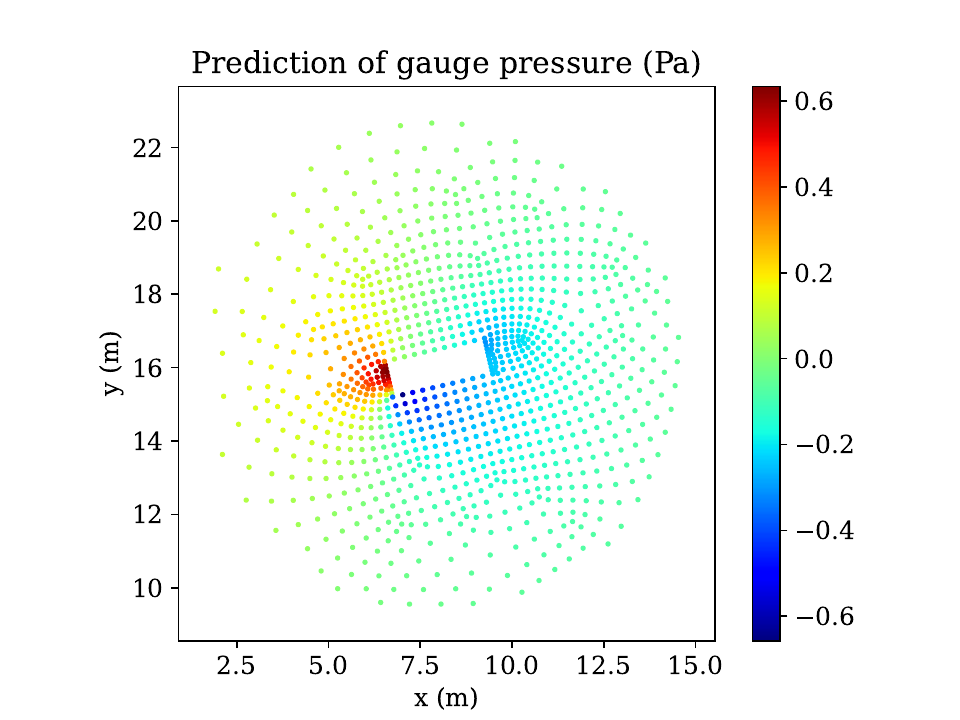}
      \end{subfigure}
    \begin{subfigure}[b]{0.32\textwidth}
        \centering
        \includegraphics[width=\textwidth]{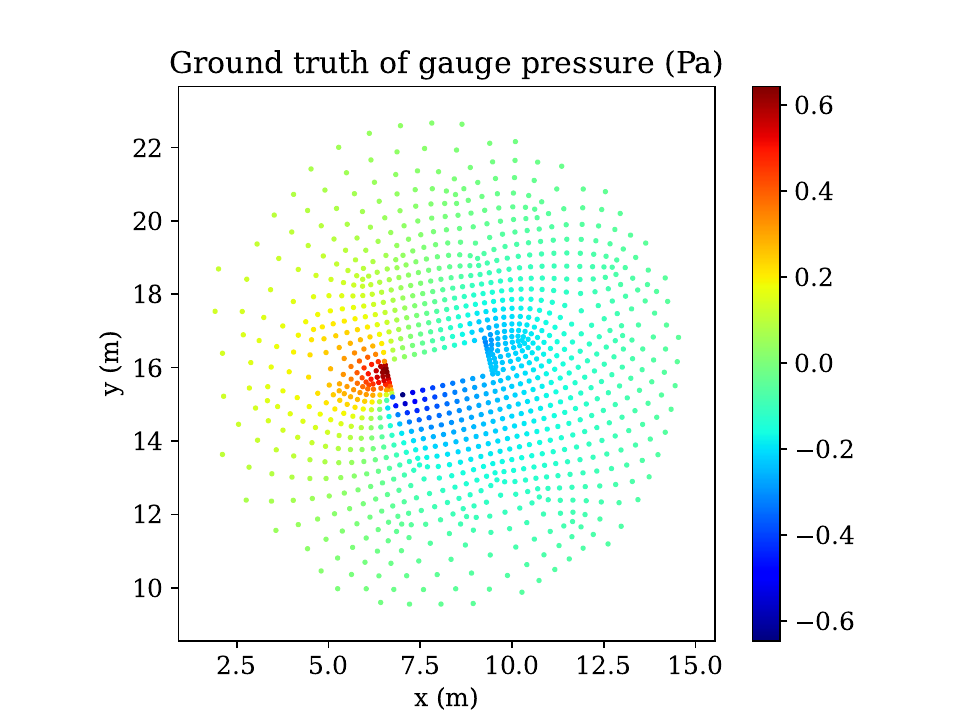}
      \end{subfigure}
    \begin{subfigure}[b]{0.32\textwidth}
        \centering
        \includegraphics[width=\textwidth]{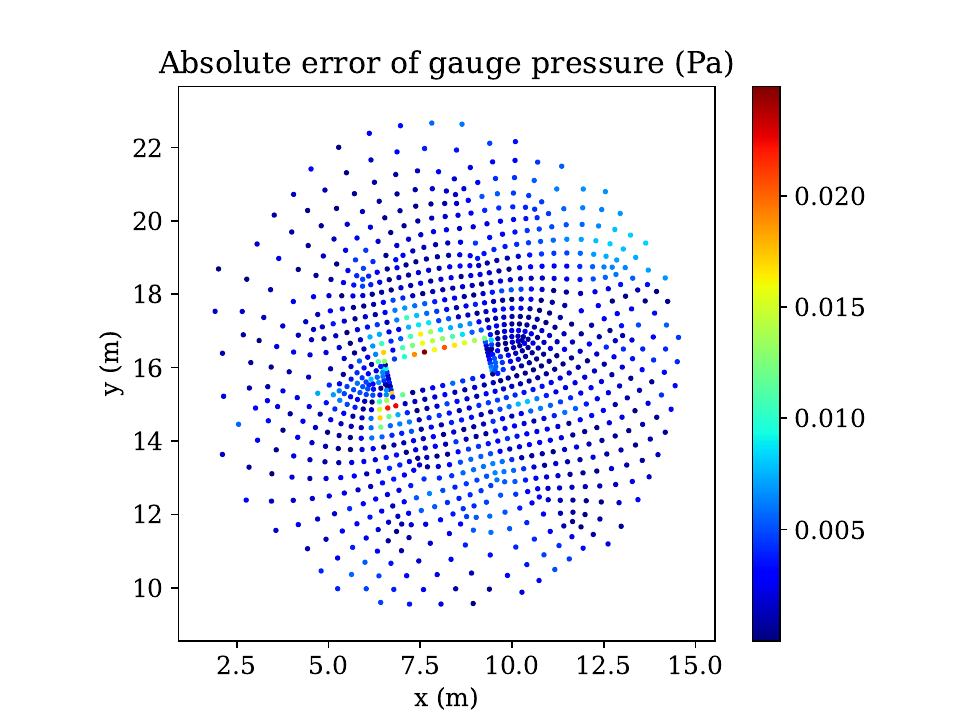}
      \end{subfigure}

  \caption{The tenth set of examples comparing the ground truth to the predictions of Kolmogorov-Arnold PointNet (i.e., KA-PointNet) for the velocity and pressure fields from the test set. The Jacobi polynomial used has a degree of 5, with $\alpha=\beta=1$. Here, $n_s=1$ is set.}
  \label{Fig14}
\end{figure}


\section{Results and discussion}
\label{Sect4}

\subsection{General analysis}
\label{Sect41}

\subsubsection{Visual comparison and overall outcomes}

To analyze the performance of KA-PointNet, we first focus on a specific case where the degree of the Jacobi polynomial (see Eqs. (\ref{Eq15})--(\ref{Eq16})) is set to 5, $\alpha = \beta = 1$ (leading to the Gegenbauer polynomial), and $n_s = 1$. All the information provided in this paragraph and the next three paragraphs is associated with this setup for KA-PointNet. Analysis of the relative pointwise error ($L^2$ norm) of the predicted velocity and pressure fields for 222 geometries of the test set is tabulated in Table \ref{Table2} (see the fifth column). Based on the information collected in Table \ref{Table2}, the average relative error over all these 222 geometries for $u$, $v$, and $p$ variables are approximately 1.18\%, 4.82\%, and 3.31\%, respectively, showing a strong performance of KA-PointNet. The histogram of the distribution of the relative errors ($L^2$ norm) is shown in Fig. \ref{Fig15}. In general, we observed that the $y$ component of the velocity vector ($v$) experiences a higher level of errors compared to the $x$ component of the velocity vector ($u$) and the pressure field ($p$).

Figure \ref{Fig4} visually compares the ground truth and the predictions of velocity and pressure fields by KA-PointNet (with the Jacobi polynomial of degree 5, $\alpha = \beta = 1$, and $n_s = 1$) after 10, 100, and 1000 epochs for one of the geometries of the test set. As shown in Fig.  \ref{Fig4}, KA-PointNet accurately predicts the general structure of flow fields after just 10 epochs. After 100 epochs, the prediction accuracy improves, particularly near the object surface, where the no-slip velocity boundary condition is better satisfied. Additionally, the scale of predicted values becomes more accurate. It is observed that after just 100 epochs, the prediction of the $u$ variable achieves a higher level of accuracy compared to the predictions of the $v$ and $p$ variables. This observation can explain why, in general, the relative error of the $u$ variable is lower than that of the $v$ and $p$ variables over the test set, as shown in the histogram in Fig. \ref{Fig15} as well as the data in Table \ref{Table2}. Comparing the predictions at 1000 epochs with those at 100 epochs, the velocity vector in the $y$ direction ($v$) and the pressure field ($p$) experience further modifications and increased accuracy, while the velocity vector in the $x$ direction ($u$) remains approximately unchanged.

Figures \ref{Fig5}--\ref{Fig14} present a comparison between the ground truth and the KA-PointNet predictions, along with the absolute pointwise error for the velocity and pressure fields, for ten cylinders with different cross-sectional geometries from the test set. As observed in Figs. \ref{Fig5}--\ref{Fig14}, there is excellent agreement between the ground truth and the predictions, despite the non-uniform distributions of the point clouds and varying spatial coordinate ranges within the domains of the test set. As seen in Fig. \ref{Fig10}, KA-PointNet successfully predicts the flow separation (i.e., when the $u$ variable becomes negative). As a general observation in Figs. \ref{Fig5}--\ref{Fig14}, the maximum absolute pointwise error typically occurs at or near the surface of the cylinders, where there are the highest variations in the geometry of the cylinders across different data in the test set.

\begin{figure}[!htbp]
  \centering 
      \begin{subfigure}[b]{0.32\textwidth}
        \centering
        \includegraphics[width=\textwidth]{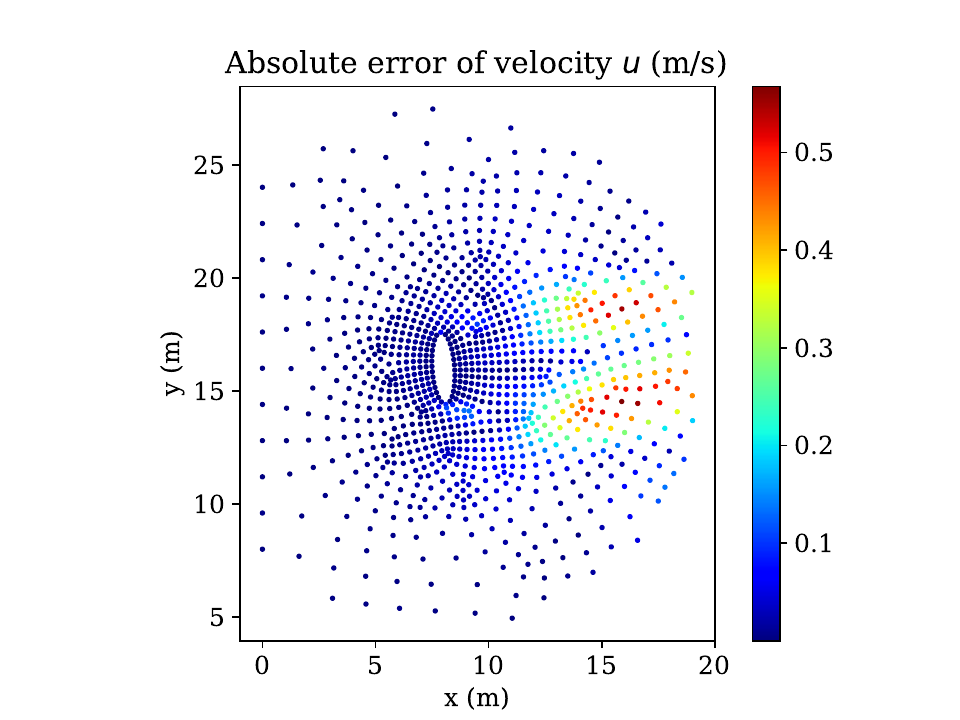}
      \end{subfigure}
    \begin{subfigure}[b]{0.32\textwidth}
        \centering
        \includegraphics[width=\textwidth]{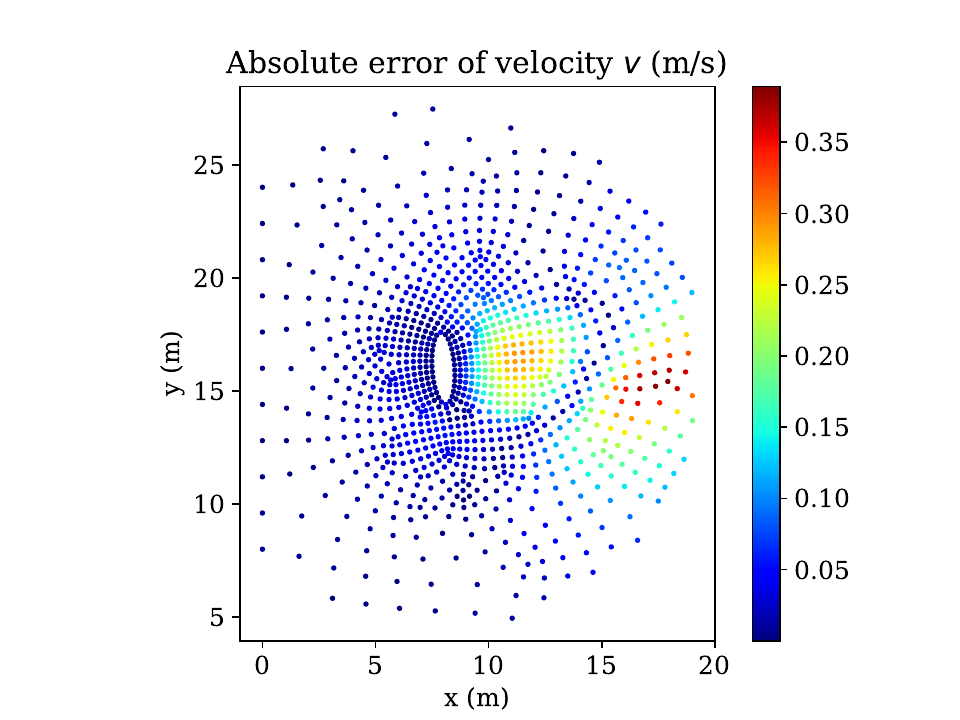}
      \end{subfigure}
    \begin{subfigure}[b]{0.32\textwidth}
        \centering
        \includegraphics[width=\textwidth]{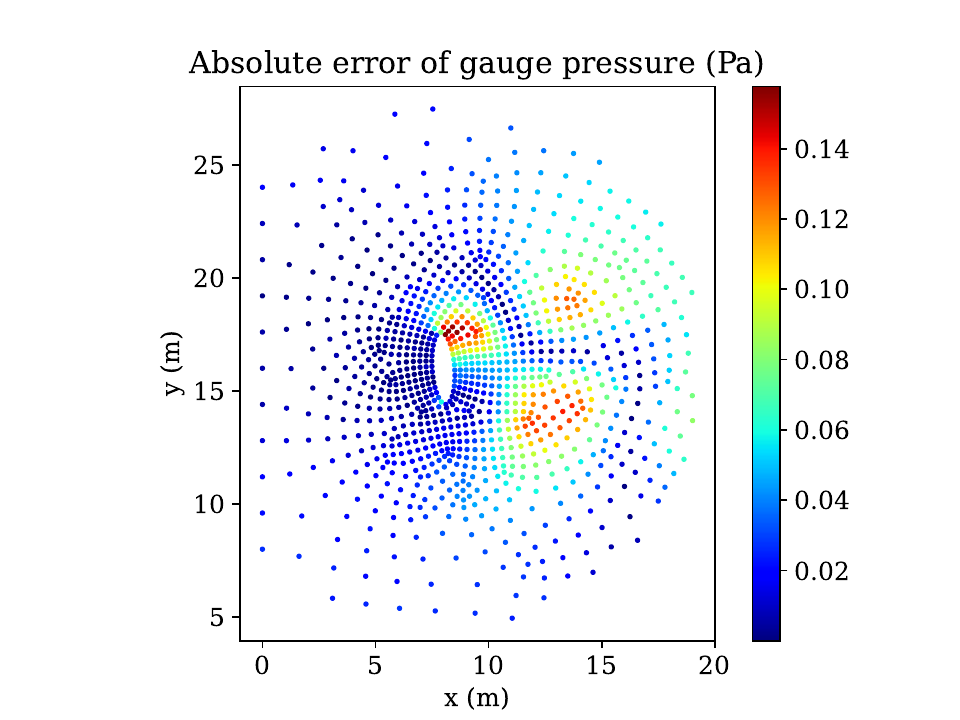}
      \end{subfigure}

    
    \begin{subfigure}[b]{0.32\textwidth}
        \centering
        \includegraphics[width=\textwidth]{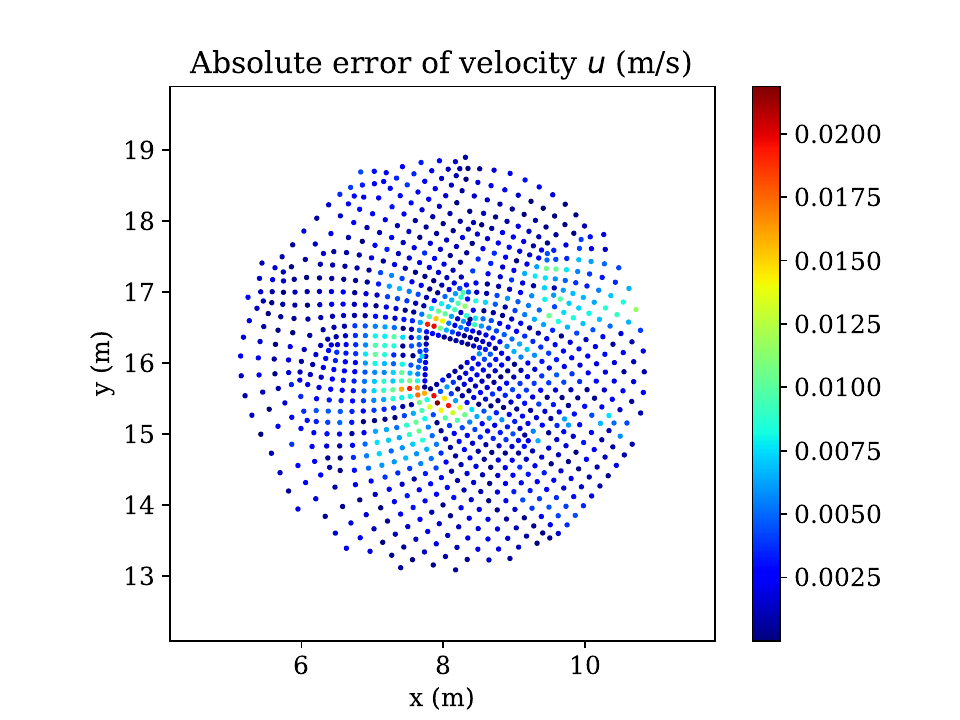}
      \end{subfigure}
    \begin{subfigure}[b]{0.32\textwidth}
        \centering
        \includegraphics[width=\textwidth]{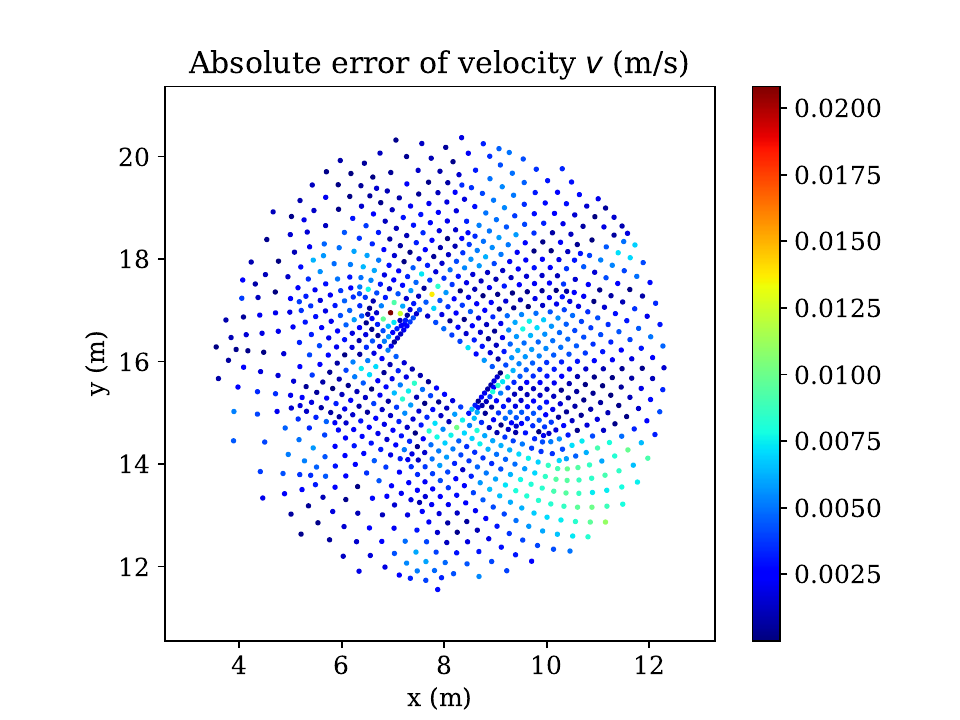}
      \end{subfigure}
    \begin{subfigure}[b]{0.32\textwidth}
        \centering
        \includegraphics[width=\textwidth]{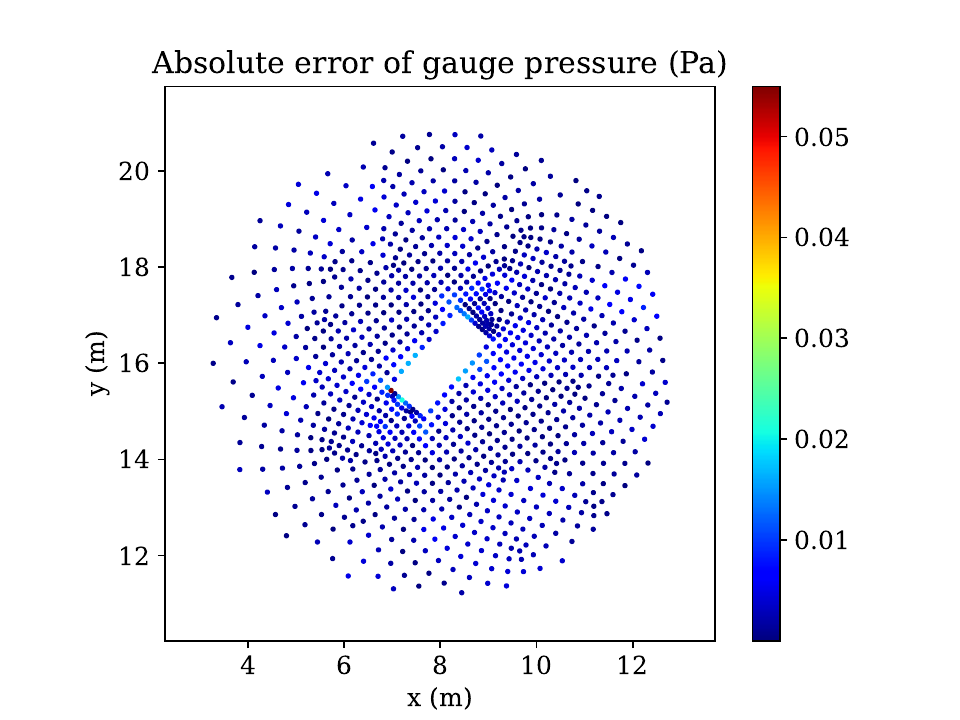}
      \end{subfigure}

  \caption{Distribution of absolute pointwise error for the prediction of the velocity and pressure fields by Kolmogorov-Arnold PointNet (i.e., KA-PointNet) for the velocity and pressure fields for the geometries from the test set when the relative pointwise error ($L^2$ norm) becomes maximum (first row) and minimum (second row). The Jacobi polynomial used has a degree of 5, with $\alpha=\beta=1$. Here, $n_s=1$ is set.}
  \label{Fig16}
\end{figure}

Distributions of absolute pointwise errors are shown in Fig. \ref{Fig16} for geometries from the test set where the relative pointwise error ($L^2$ norm) for the velocity vector components ($u$ and $v$) and the pressure field ($p$) becomes maximum (displayed in the first row of Fig. \ref{Fig16}) and minimum (displayed in the second row of Fig. \ref{Fig16}). The numerical values of these errors are listed in Table \ref{Table2}. From Fig. \ref{Fig16}, it is observed that the maximum relative pointwise error ($L^2$ norm) for the velocity and pressure fields occurs in a single domain with an elliptical cross-section for the cylinder, whereas the minimum relative pointwise error ($L^2$ norm) for the velocity and pressure fields occurs in three different domains, each with a distinct geometry for the cross-section of the cylinder. Comparing geometries with maximum and minimum relative pointwise error ($L^2$ norm), it is recognized that the maximum error occurs when the spatial coordinates of the domain (as a point cloud) reach their maximum possible value compared to other domains in the test set. Note that from the histogram shown in Fig. \ref{Fig15}, we see that the geometry with the maximum error is indeed a single outlier and falls outside the relative errors ($L^2$ norm) distribution of the histogram.

\subsubsection{Effect of Jacobi polynomial degree}

Returning to Table \ref{Table2}, with $n_s = 1$ and $\alpha = \beta = 1.0$, we analyze the performance of KA-PointNet as a function of the degree of the Jacobi polynomial. Based on the relative pointwise error ($L^2$ norm) tabulated in Table \ref{Table2}, the average error across the test set (222 unseen data) varies approximately from 1.07\% to 1.74\% for the $u$ variable, 4.82\% to 5.76\% for the $v$ variable, and 3.31\% to 4.76\% for the $p$ variable. Additionally, we observe that increasing the polynomial degree does not necessarily improve prediction accuracy. For instance, when comparing the average relative errors from a degree of 2 to a degree of 3, the prediction accuracy decreases by 0.30160\%, 0.47600\%, and 1.25832\% for the $u$, $v$, and $p$ variables, respectively. On the other hand, the Jacobi polynomials of degrees 5 and 6 yield the lowest errors, as indicated in Table \ref{Table2}. Regarding the computational cost of training, higher degrees of Jacobi polynomials increase the number of trainable parameters (see Eq. (\ref{Eq15})), thus extending the training time. Comparing the computational cost of using the higher-order Jacobi polynomials with the improvement in prediction accuracy, it appears that choosing even the polynomial of degree 2 is an optimized decision.

\subsubsection{Effect of Jacobi polynomial type}

Next, we investigate the effect of the choice of $\alpha$ and $\beta$ in the Jacobi polynomial (see Eqs. (\ref{Eq16})--(\ref{Eq19})) on the performance of KA-PointNet. To reach this goal, we set the degree of the Jacobi polynomial to 3 and $n_s = 1$. Specifically, we consider the following selections: $\alpha = \beta = 0$ (leading to the Legendre polynomial), $\alpha = \beta = -0.5$ (leading to the Chebyshev polynomial of the first kind), $\alpha = \beta = 0.5$ (leading to the Chebyshev polynomial of the second kind), $\alpha = \beta = 1$ (leading to the Gegenbauer polynomial), $2\alpha = \beta = 2$, and $\alpha = 2\beta = 2$. The outcomes of this investigation are presented in Table \ref{Table3}. According to Table \ref{Table3}, the Chebyshev polynomials (i.e., $\alpha = \beta = -0.5$ and $\alpha = \beta = 0.5$) lead to the highest accuracy among the options, considering the average relative pointwise error ($L^2$ norm) as the judgment criterion. More specifically, the Chebyshev polynomial of the first kind leads to more accurate predictions compared to the Chebyshev polynomial of the second kind. The superior performance of Chebyshev polynomials over other members of the Jacobi polynomial family has been noted in various applications of KANs \cite{shukla2024comprehensive,KANwithTANH,guo2024physicsKANFluids}. This can be relevant to the distribution of their roots within the interval, which minimizes the absolute error in function approximation and mitigates potential oscillatory behaviors. The lowest performance, in terms of prediction accuracy, occurs when $\alpha = 2\beta = 2$. Although the choice of $\alpha$ and $\beta$ affects the accuracy of predictions, we observe that all the choices result in errors within an acceptable range. Considering all the choices, the average relative pointwise error ($L^2$ norm) of the $u$, $v$, and $p$ variables over the test set containing 222 data with various geometries do not exceed 2.99\%, 9.48\%, and 6.40\%, respectively.

\subsubsection{Effect of network size}

To investigate the influence of the size of KA-PointNet, we tabulate the relative pointwise error ($L^2$ norm) of the predicted velocity and pressure fields as a function of the defined global scaling parameter ($n_s$) in Table \ref{Table4}. Using the Jacobi polynomial of degree 3 and setting $\alpha = \beta = 1$, we conduct machine learning experiments with different values of $n_s$. Based on the error analysis presented in Table \ref{Table4}, we observe that increasing $n_s$ and thereby enlarging KA-PointNet enhances prediction accuracy as $n_s$ is increased from 0.5 to 0.75 and from 0.75 to 1. This scenario changes when selecting $n_s$ values greater than 1 (i.e., $n_s = 1.25$, $n_s = 1.5$, and $n_s = 2$). As shown in Table \ref{Table4}, there is no clear trend in the relative error for a specific field. For example, increasing $n_s$ beyond 1 may lead to an increase in the error for the velocity prediction while reducing the error for the pressure prediction. Nonetheless, all errors for all fields are lower than those obtained with $n_s = 0.75$ and $n_s = 0.5$. Generally speaking, increasing the size of KA-PointNet while keeping the number of training data fixed raises the potential for overfitting. This explains why we observe a slight decrease in the accuracy for predicting some fields when $n_s$ is increased beyond a certain threshold (e.g., $n_s = 1$ in the current experiment). According to Table \ref{Table4}, it is noticeable that increasing $n_s$ leads to a rise in the number of trainable parameters and, consequently, the training time. For instance, the average training time per epoch for $n_s = 2$ is approximately 2.56 times greater than that for $n_s = 1$. Therefore, the design of KA-PointNet and the setting of hyperparameters such as $n_s$ are critical considerations from both accuracy and computational cost perspectives. Additionally, the machine learning experiments discussed here demonstrate that despite the complexity of the KA-PointNet architecture, a desirable level of accuracy can be achieved by fine-tuning the single parameter, $n_s$.

Based on the design and architecture of KA-PointNet, the degree of Jacobi polynomials ($n$) and the global scaling parameter ($n_s$) are two important factors for controlling the size of the network. Considering Table \ref{Table2} and Table \ref{Table4} simultaneously, for instance, the number of trainable parameters for KA-PointNet with the Jacobi polynomial degree of 5, $\alpha = \beta = 1$, and setting $n_s = 1$, is 5316224, with an average training time of 14.28542 seconds per epoch. On the other hand, consider KA-PointNet with the Jacobi polynomial degree of 3, $\alpha = \beta = 1$, and setting $n_s = 1.25$. This network has 5538080 trainable parameters, leading to an average training time of 8.85618 seconds per epoch. These two networks have approximately equal numbers of trainable parameters. Comparing the information in Table \ref{Table4} and Table \ref{Table2}, the prediction accuracy for all the velocity and pressure fields is higher in the first configuration (i.e., higher-order Jacobi polynomials but a smaller value of $n_s$). Although the number of trainable parameters is approximately equal in both cases, the training time of the first configuration is approximately 1.614 times greater than that of the second configuration. From a computer science perspective, this means that given limited RAM, if there is no restriction on training time, the first configuration produces more accurate outcomes.

\subsubsection{Comparison of layer normalization and batch normalization in KA-PointNet}

The last topic addressed in this subsection is the investigation of the performance of utilizing layer normalization in the architecture of KA-PointNet. One may refer to Ref. \citep{ba2016layer} for details of layer normalization and its differences with batch normalization from a computer science perspective. Note that all the results reported up to this point pertain to the architecture of KA-PointNet with batch normalization. As a machine learning experiment, we replace batch normalization with layer normalization in a KA-PointNet with a Jacobi polynomial of degree 3, setting $\alpha = \beta = 1$ and $n_s = 1$. The first consequence of using layer normalization is an increase in the number of trainable parameters from 3545728 to 8390656, requiring more RAM. Measuring the average relative pointwise error ($L^2$ norm) over the test set (222 data points) for the $u$, $v$, and $p$ variables, they are respectively 1.32751E$-$2, 4.77946E$-$2, and 4.20826E$-$2. The corresponding quantities for KA-PointNet with batch normalization are 1.73537E$-$2, 5.75906E$-$2, and 4.75586E$-$2. Accordingly, we observe an improvement in the prediction accuracy, albeit with a relatively high cost for RAM usage. Depending on the user's criteria and available computational resources, the decision to use layer normalization may vary.


\subsection{Comparison of PointNet with shared KANs and PointNet with shared MLPs}
\label{Sect42}

\subsubsection{A brief overview of PointNet with shared MLPs}
\label{Sect421}

\begin{figure}[!htbp]
  \centering 
        \includegraphics[width=\textwidth]{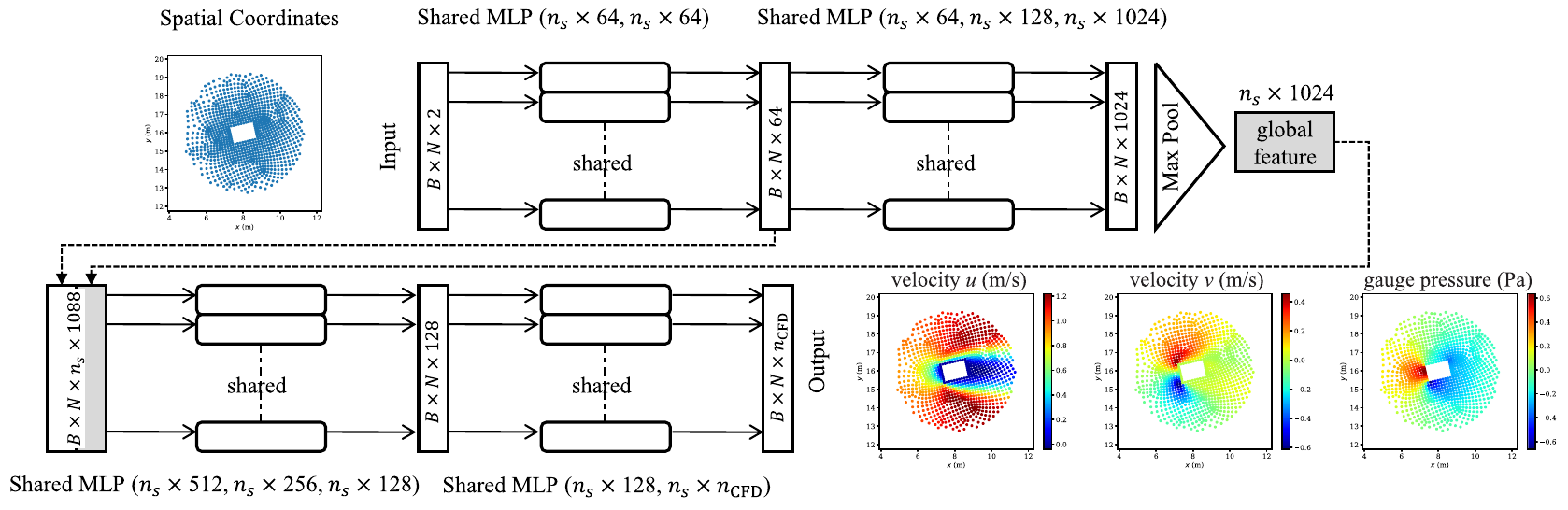}
  \caption{Architecture of PointNet with shared Multilayer Perceptron (MLP). Shared MLPs with the labels $(\mathcal{B}_1, \mathcal{B}_2)$ and $(\mathcal{B}_1, \mathcal{B}_2, \mathcal{B}_3)$ are explained in the text. $n_{\text{CFD}}$ denotes the number of CFD variables. $N$ is the number of points in the point clouds. $B$ represents the batch size. $n_s$ is the global scaling parameter used to control the network size. Note that the velocity and pressure fields shown are schematic.}
  \label{Fig17}
\end{figure}

Before performing a comparison between PointNet with shared KANs (i.e., KA-PointNet) and PointNet with shared MLPs, we briefly review the structure of the latter, shown in Fig. \ref{Fig17}. The goal is to make the differences clear so that the performance comparison becomes meaningful. As can be seen in Fig. \ref{Fig17}, similar to the notation used for shared KANs, we use ($\mathcal{B}_1$, $\mathcal{B}_2$) and ($\mathcal{B}_1$, $\mathcal{B}_2$, $\mathcal{B}_3$) for shared MLPs with two and three layers, respectively. The connection between the input vector $\mathbf{r}$ of size $d_\text{input}$ and the output vector $\mathbf{s}$ of size $d_\text{output}$ for one layer in an MLP can be formulated as follows:

\begin{equation}
    \mathbf{s}_{d_{\text{output}\times 1}} = \sigma \left(\textbf{W}_{d_{\text{output}}\times {d_\text{input}}}\mathbf{r}_{d_{\text{input}\times 1}} + \textbf{b}_{d_\text{output}\times 1} \right),
     \label{Eq27}
\end{equation}
where $\textbf{W}$ is the weight matrix and $\textbf{b}$ is the bias vector, containing trainable parameters. The nonlinear activation function is shown by $\sigma$, which operates elementwise. The number of trainable parameters in this layer is equal to $d_\text{input} \times d_\text{input} + d_\text{output}$. One may observe the difference in the structure of MLPs and KANs by comparing Eq. (\ref{Eq13}) and Eq. (\ref{Eq27}). The concept of shared MLPs is explained as follows. Let us apply the same example as we discussed in Sect. \ref{Sect33} for shared KANs. Taking $n_s=1$ and focusing on the first layer of the first MLP in the first branch of PointNet (see Fig. \ref{Fig17}), the resulting output vectors are:

\begin{equation}
\begin{aligned}
\mathbf{s}^{(1)}_{64 \times 1} &= \sigma \left( \mathbf{W}_{64 \times 2} \begin{bmatrix} x'_1 \\ \\ y'_1 \end{bmatrix} + \mathbf{b}_{64 \times 1} \right), \\
\mathbf{s}^{(2)}_{64 \times 1} &= \sigma \left( \mathbf{W}_{64 \times 2} \begin{bmatrix} x'_2 \\ \\ y'_2 \end{bmatrix} + \mathbf{b}_{64 \times 1} \right), \\
&\vdots \\
\mathbf{s}^{(N)}_{64 \times 1} &= \sigma \left( \mathbf{W}_{64 \times 2} \begin{bmatrix} x'_N \\ \\ y'_N \end{bmatrix} + \mathbf{b}_{64 \times 1} \right),
\end{aligned}
 \label{Eq33}
\end{equation}
where $\mathbf{W}$ and $\mathbf{b}$ are respectively the shared weight matrix and shared bias vector applied to all [$x_j$ $y_j$]$^\text{tr}$ for $1 \leq j \leq N$. One may compare Eqs. (\ref{Eq25}) and Eqs. (\ref{Eq33}). Further details about the implementation of shared MLP can be found in Ref. \citep{lin2014networknetwork}. In the case of using PointNet with shared MLPs, our machine learning experiments demonstrate that the accuracy of predictions is higher if we normalize the output of the network (i.e., $u^*$, $v^*$, and $p^*$) between [0, 1], instead of [$-$1, 1], using the following scaling formula:

\begin{equation}
    \left\{\phi'\right\} = \frac{\left\{\phi\right\}  - \min(\left\{\phi\right\} )}{\max(\left\{\phi\right\} ) - \min(\left\{\phi\right\} )}.
    \label{Eq28}
\end{equation}
Note that we still scale the spatial coordinates (i.e., $x'$ and $y'$) to the range of [$-$1, 1] using Eq. (\ref{Eq10}). The activation function of the Rectified Linear Unit (ReLU), defined as follows:

\begin{equation}
     \sigma(\gamma) = \max(0,\gamma),
      \label{Eq30}
\end{equation}
is implemented after all the layers except the last layer. Similar to KA-PointNet, each layer is followed by a batch normalization \citep{ioffe2015batch}, except the last layer. The activation function in the last layer is the sigmoid function defined as follows:

\begin{equation}
    \sigma(\gamma) = \frac{1}{1 + e^{-\gamma}},
     \label{Eq29}
\end{equation}
which covers the range of [0, 1]. The loss function, batch size, learning rate, and the optimizer model with its associated hyperparameters are all set the same as those for KA-PointNet (see Sect. \ref{Sect34}).


\begin{table}[width=.9\linewidth,cols=7,pos=!htbp]
\caption{Comparison between the performance of PointNet with shared Kolmogorov-Arnold Networks (KANs), i.e. KA-PointNet, and PointNet with shared Multilayer Perceptrons (MLPs), when the number of trainable parameters is approximately equal, for prediction of the velocity and pressure fields for the test set containing 222 unseen geometries. In the Jacobi polynomial of KA-PointNet, we set $\alpha=\beta=1$. $||\cdots||$ indicates the $L^2$ norm.}\label{Table5}
\begin{tabular*}{\tblwidth}{@{} LLLLLLL@{} }
\toprule
 &  KANs & MLPs & KANs & MLPs & KANs & MLPs\\
\midrule
$n_s$ & 0.5 & 1 & 1 & 2 & 2 & 4 \\
\midrule
Polynomial degree & 3 & - & 3 & - & 3 & - \\
\midrule
Number of trainable & 888128 & 892355 & 3545728 & 3554179 & 14169344 & 14186243\\
parameters &  &  &  &  &  & \\
\midrule
Training time & 2.83741 & 0.42597 & 6.52910 & 0.97970 & 16.69120 & 2.47669 \\
per epoch (s) &  &  &  &  &  & \\
\midrule
Average $||\Tilde{u}-u||/||u||$ & 2.63597E$-$2 & 5.01625E$-$2 & 1.73537E$-$2 & 3.83902E$-$2 & 1.69209E$-$2 & 3.80471E$-$2\\
Maximum $||\Tilde{u}-u||/||u||$ & 1.30750E$-$1 & 1.57523E$-$1 & 1.40088E$-$1 & 1.51597E$-$1 & 1.50143E$-$1 & 1.47946E$-$1\\
Minimum $||\Tilde{u}-u||/||u||$ & 1.30994E$-$2 & 2.03904E$-$2 & 7.22247E$-$3 & 1.24053E$-$2 & 1.08372E$-$2 & 1.40001E$-$2\\
\midrule
Average $||\Tilde{v}-v||/||v||$ & 1.00292E$-$1 & 1.61943E$-$1 & 5.75906E$-$2 & 1.39018E$-$1 & 4.48352E$-$2 & 1.47633E$-$1\\
Maximum $||\Tilde{v}-v||/||v||$ & 4.19706E$-$1 & 5.07943E$-$1 & 4.42245E$-$1 & 4.85987E$-$1 & 4.85161E$-$1 & 4.87821E$-$1\\
Minimum $||\Tilde{v}-v||/||v||$ & 5.58101E$-$2 & 6.56595E$-$2 & 2.05299E$-$2 & 5.60436E$-$2 & 2.49093E$-$2 & 5.42824E$-$2\\
\midrule
Average $||\Tilde{p}-p||/||p||$ & 9.86696E$-$2 & 1.54159E$-$1 & 4.75586E$-$2 & 1.05939E$-$1 & 4.25412E$-$2 & 7.22403E$-$2\\
Maximum $||\Tilde{p}-p||/||p||$ & 2.88919E$-$1 & 4.03240E$-$1 & 1.55694E$-$1 & 2.36421E$-$1 & 1.63125E$-$1 & 1.70085E$-$1 \\
Minimum $||\Tilde{p}-p||/||p||$ & 4.67306E$-$2 & 4.04501E$-$2 & 1.75673E$-$2 & 3.77024E$-$2 & 1.74415E$-$2 & 2.84674E$-$2 \\
\bottomrule
\end{tabular*}
\end{table}


\begin{table}[width=.9\linewidth,cols=7,pos=!htbp]
\caption{Comparison between the performance of PointNet with shared Kolmogorov-Arnold Networks (KANs), i.e., KA-PointNet, and PointNet with shared Multilayer Perceptrons (MLPs), when the training time per epoch is approximately equal, for the prediction of the velocity and pressure fields on the test set containing 222 unseen geometries. In the Jacobi polynomial of KA-PointNet, we set $\alpha = \beta = 1$. The notation $||\cdots||$ indicates the $L^2$-norm.}\label{Table6}
\begin{tabular*}{\tblwidth}{@{} LLLLLLL@{} }
\toprule
 &  KANs & MLPs & KANs & MLPs & KANs & MLPs\\
\midrule
$n_s$ & 0.5 & 4.25 & 1 & 7.25 & 2 & 11.75 \\
\midrule
Polynomial degree & 3 & - &  3 & - & 3 & - \\
\midrule
Number of trainable & 888128 & 16012915 & 3545728 & 46559155 & 14169344 & 122238355 \\
parameters &  &  &  &  &  & \\
\midrule
Training time & 2.83741 & 2.85774 & 6.52910 & 6.81708 & 16.69120 & 16.58025 \\
per epoch (s) &  &  &  &  &  & \\
\midrule
Average $||\Tilde{u}-u||/||u||$ & 2.63597E$-$2 & 3.33798E$-$2 & 1.73537E$-$2 & 7.98696E$-$2 & 1.69209E$-$2 & 1.10399E$-$2 \\
Maximum $||\Tilde{u}-u||/||u||$ & 1.30750E$-$1 & 1.32978E$-$1 & 1.40088E$-$1 & 1.79377E$-$1 & 1.50143E$-$1 & 1.72963E$-$1 \\
Minimum $||\Tilde{u}-u||/||u||$ & 1.30994E$-$2 & 1.39605E$-$2 & 7.22247E$-$3 & 2.38381E$-$2 & 1.08372E$-$2 & 3.36158E$-$3 \\
\midrule
Average $||\Tilde{v}-v||/||v||$ & 1.00292E$-$1 & 1.09335E$-$1 & 5.75906E$-$2 & 2.56850E$-$1 & 4.48352E$-$2 & 4.67447E$-$2 \\
Maximum $||\Tilde{v}-v||/||v||$ & 4.19706E$-$1 & 4.17927E$-$1 & 4.42245E$-$1 & 6.09449E$-$1 & 4.85161E$-$1 & 5.74375E$-$1 \\
Minimum $||\Tilde{v}-v||/||v||$ & 5.58101E$-$2 & 5.08140E$-$2 & 2.05299E$-$2 & 5.16830E$-$2 & 2.49093E$-$2 & 1.91595E$-$2 \\
\midrule
Average $||\Tilde{p}-p||/||p||$ & 9.86696E$-$2 & 9.81417E$-$2 & 4.75586E$-$2 & 1.90045E$-$1 & 4.25412E$-$2 & 3.79285E$-$2 \\
Maximum $||\Tilde{p}-p||/||p||$ & 2.88919E$-$1 & 2.84515E$-$1 & 1.55694E$-$1 & 4.68130E$-$1 & 1.63125E$-$1 & 2.04002E$-$1 \\
Minimum $||\Tilde{p}-p||/||p||$ & 4.67306E$-$2 & 3.48307E$-$2 & 1.75673E$-$2 & 4.69678E$-$2 & 1.74415E$-$2 & 1.48129E$-$2 \\
\bottomrule
\end{tabular*}
\end{table}


\begin{table}[width=.6\linewidth,cols=3,pos=!htbp]
\caption{Comparison between the performance of PointNet with shared Kolmogorov-Arnold Networks (KANs), i.e., KA-PointNet, and PointNet with shared Multilayer Perceptrons (MLPs), when the training data is polluted with 10\% Gaussian noise, for the prediction of the velocity and pressure fields on the test set containing 222 unseen geometries. The Jacobi polynomial used has a degree of 3 with $\alpha = \beta = 1$. The notation $||\cdots||$ shows the $L^2$ norm.}\label{Table7}
\begin{tabular*}{\tblwidth}{@{} LLL@{}}
\toprule
 &  KANs & MLPs \\
\midrule
$n_s$ & 1 & 2  \\
\midrule
Number of trainable & 3545728 & 3554179 \\
parameters &  &  \\
\midrule
Average $||\Tilde{u}-u||/||u||$ & 5.41813E$-$2 & 1.16279E$-$1 \\
Maximum $||\Tilde{u}-u||/||u||$ & 1.65142E$-$1 & 1.99973E$-$1  \\
Minimum $||\Tilde{u}-u||/||u||$ & 3.60475E$-$2 & 6.39549E$-$2 \\
\midrule
Average $||\Tilde{v}-v||/||v||$ & 1.43525E$-$1 & 3.19472E$-$1  \\
Maximum $||\Tilde{v}-v||/||v||$ & 5.05227E$-$1 & 5.04655E$-$1  \\
Minimum $||\Tilde{v}-v||/||v||$ & 7.06387E$-$2 & 1.93812E$-$1  \\
\midrule
Average $||\Tilde{p}-p||/||p||$ & 9.76559E$-$2 & 3.28236E$-$1 \\
Maximum $||\Tilde{p}-p||/||p||$ & 2.18898E$-$1 & 5.55212E$-$1 \\
Minimum $||\Tilde{p}-p||/||p||$ & 5.33931E$-$2 & 1.66042E$-$1 \\
\bottomrule
\end{tabular*}
\end{table}


\begin{table}[width=1.0\linewidth,cols=7,pos=!htbp]
\caption{Comparison between the performance of PointNet with shared Kolmogorov-Arnold Networks (KANs), i.e., KA-PointNet, and PointNet with shared Multilayer Perceptrons (MLPs) for predicting pressure drag and pressure lift on a test set containing 222 unseen geometries. Lift and drag are computed from the predicted pressure fields. For PointNet with shared MLPs, we set $n_s=2$. For KA-PointNet, we use a Jacobi polynomial with $\alpha = \beta = 1$ and set $n_s=1$.}\label{Table8}
\begin{tabular*}{\tblwidth}{@{} LLLLLLL@{} }
\toprule
 &  KANs & KANs & KANs & KANs & KANs & MLPs\\
\midrule
Polynomial degree & 2 & 3 & 4 & 5 & 6 & - \\
\midrule
Average absolute error of drag & 1.25407E$-$2 & 1.89780E$-$2 & 1.71568E$-$2 & 1.17999E$-$2 & 1.78657E$-$2 & 7.27047E$-$2 \\
Maximum absolute error of drag & 1.89992E$-$1 & 2.38099E$-$1 & 3.03330E$-$1 & 1.78330E$-$1 & 1.89298E$-$1 & 2.87231E$-$1 \\
Minimum absolute error of drag & 8.63075E$-$5 & 1.43647E$-$4 & 6.69360E$-$5 & 6.94990E$-$5 & 6.19888E$-$6 & 1.87975E$-$3 \\
\midrule
Average absolute error of lift & 1.28938E$-$2 & 2.15022E$-$2 & 1.61056E$-$2 & 1.19543E$-$2 & 1.56660E$-$2 & 6.41688E$-$2 \\
Maximum absolute error of lift & 1.26349E$-$1 & 1.22340E$-$1 & 1.38436E$-$1 & 1.11378E$-$1 & 1.07055E$-$1 & 2.92440E$-$1 \\
Minimum absolute error of lift & 6.61910E$-$5 & 2.04789E$-$5 & 1.08272E$-$4 & 1.59726E$-$4 & 4.33922E$-$5 & 5.25713E$-$4 \\
\bottomrule
\end{tabular*}
\end{table}


\begin{table}[width=1.0\linewidth,cols=7,pos=!htbp]
\caption{Robustness test of Kolmogorov-Arnold PointNet (i.e., KA-PointNet) by evaluating the error in predicted velocity and pressure when a certain percentage of points is randomly removed from point clouds in the test set, containing 222 unseen geometries. The Jacobi polynomial degree is set to 3, with $\alpha = \beta = 1$ and $n_s=1$. The $L^2$ norm is indicated by $||\cdots||$.}\label{Table9}
\begin{tabular*}{\tblwidth}{@{} LLLLLLL@{} }
\toprule
Percentage & 2\% & 5\% & 8\% & 10\% &  12\% & 15\% \\
\midrule
Average $||\Tilde{u}-u||/||u||$ & 1.70912E$-$2 & 1.73327E$-$2 & 1.97478E$-$2 & 2.25826E$-$2 & 2.70720E$-$2 & 2.97099E$-$2 \\
Average $||\Tilde{v}-v||/||v||$ & 5.77925E$-$2 & 5.86408E$-$2 & 6.43273E$-$2 & 7.96581E$-$2 & 9.12054E$-$2 & 1.01258E$-$1 \\
Average $||\Tilde{p}-p||/||p||$ & 4.95328E$-$2 & 5.16014E$-$2 & 5.73122E$-$2 & 6.73804E$-$2 & 7.70515E$-$2 & 8.37937E$-$2 \\
\bottomrule
\end{tabular*}
\end{table}


\begin{table}[width=0.9\linewidth,cols=4,pos=!htbp]
\caption{Error analysis of Kolmogorov-Arnold PointNet (KA-PointNet) for predicting velocity and pressure fields with different train, validation, and test set ratios. The Jacobi polynomial degree is set to 3, with $\alpha = \beta = 1$ and $n_s=1$. The $L^2$ norm is denoted by $||\cdots||$.} \label{Table10}
\begin{tabular*}{\tblwidth}{@{} LLLL@{} }
\toprule
Train-Validation-Test Split (\%) &  90\%, 5\%, 5\% & 80\%, 10\%, 10\% & 70\%, 15\%, 15\% \\
\midrule
Average $||\Tilde{u}-u||/||u||$ & 1.73140E$-$2 & 1.73537E$-$2 &  2.34218E$-$2 \\
Average $||\Tilde{v}-v||/||v||$ & 5.49319E$-$2 & 5.75906E$-$2 &  7.39440E$-$2  \\
Average $||\Tilde{p}-p||/||p||$ & 4.01143E$-$2 & 4.75586E$-$2 &  4.98218E$-$2 \\
\bottomrule
\end{tabular*}
\end{table}

\begin{figure}[!htbp]
  \centering 
      \begin{subfigure}[b]{0.49\textwidth}
      \caption{PointNet with shared KANs}
        \centering
        \includegraphics[width=\textwidth]{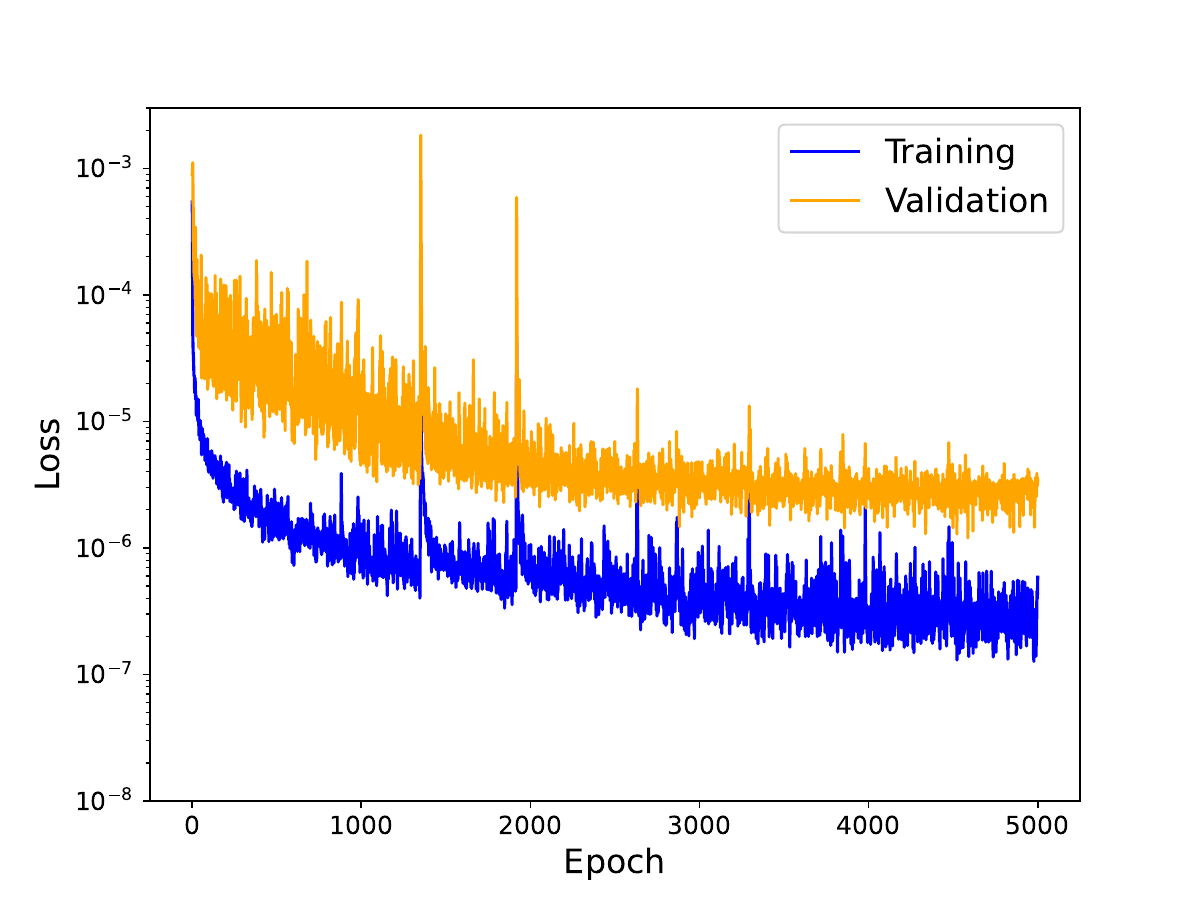}
    \end{subfigure}
    \begin{subfigure}[b]{0.49\textwidth}
     \caption{PointNet with shared MLPs}
        \centering
        \includegraphics[width=\textwidth]{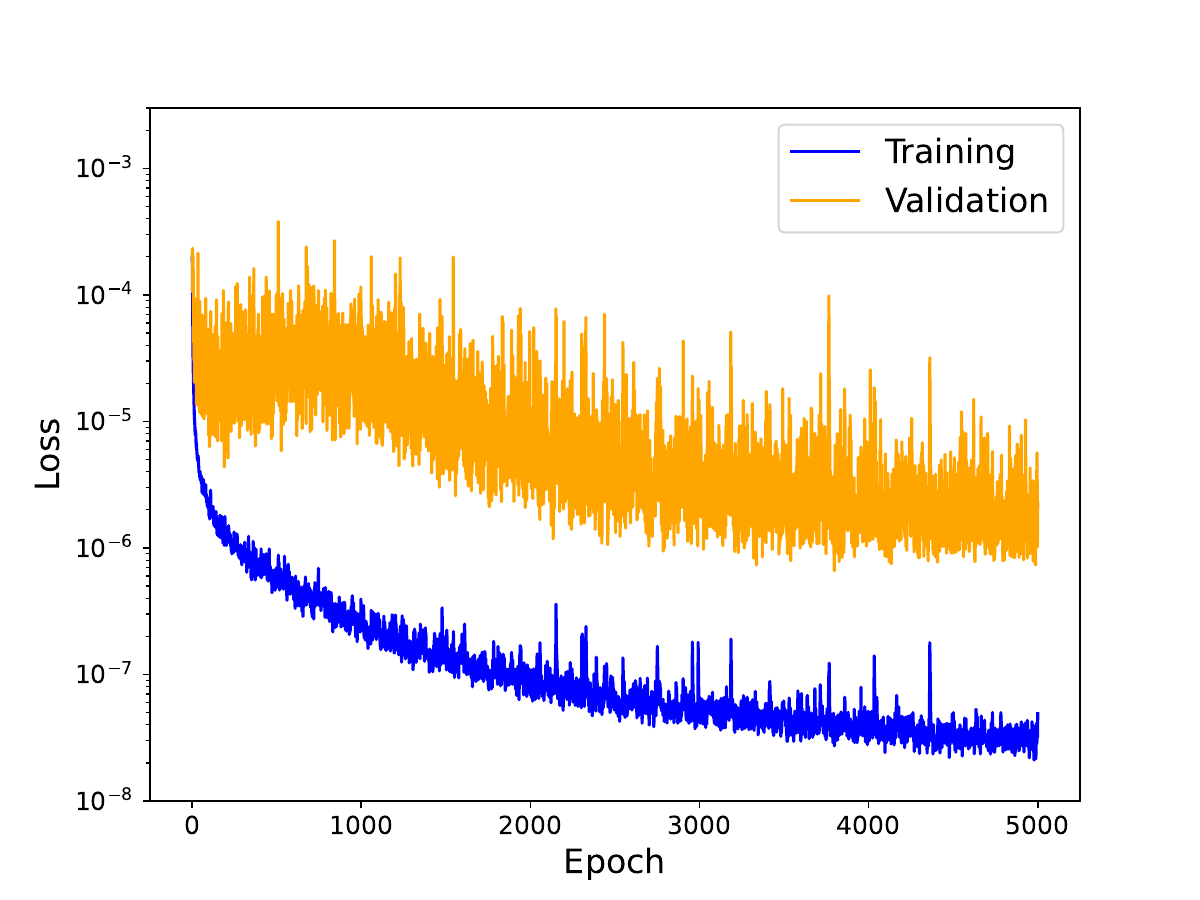}
    \end{subfigure}
    
\caption{Comparison of the loss evolution for the training and validation sets between PointNet with shared KANs, i.e., KA-PointNet, ($n_s=2$, Jacobi polynomial of degree 3, $\alpha=\beta=1$, and 14169344 trainable parameters) and PointNet with shared MLPs ($n_s=4$ and 14186243 trainable parameters)}
  \label{Fig18}
\end{figure}


\begin{figure}[!htbp]
  \centering 
      \begin{subfigure}[b]{0.32\textwidth}
      \caption{PointNet with shared KANs}
        \centering
        \includegraphics[width=\textwidth]{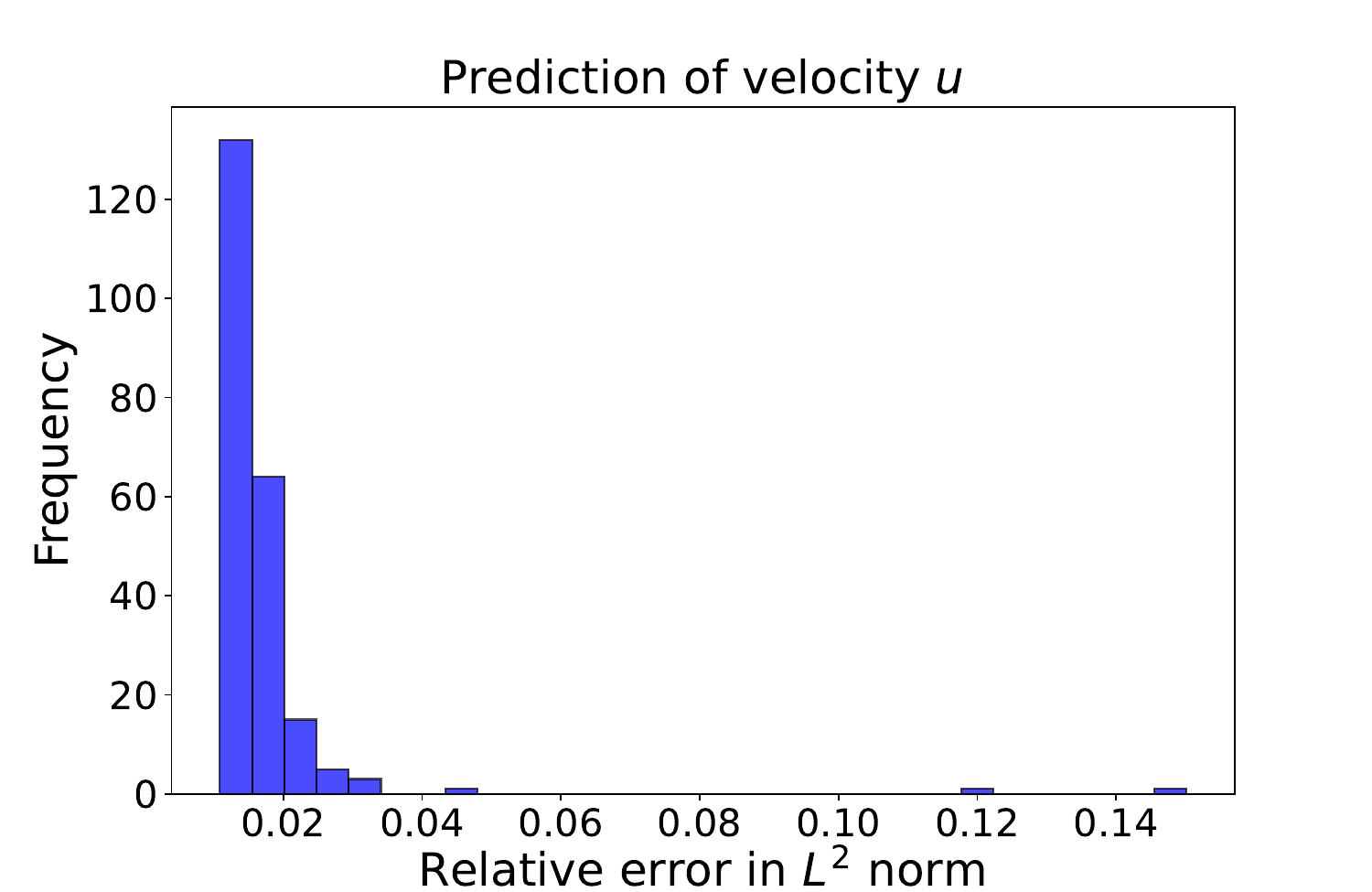}
    \end{subfigure}
    \begin{subfigure}[b]{0.32\textwidth}
     \caption{PointNet with shared KANs}
        \centering
        \includegraphics[width=\textwidth]{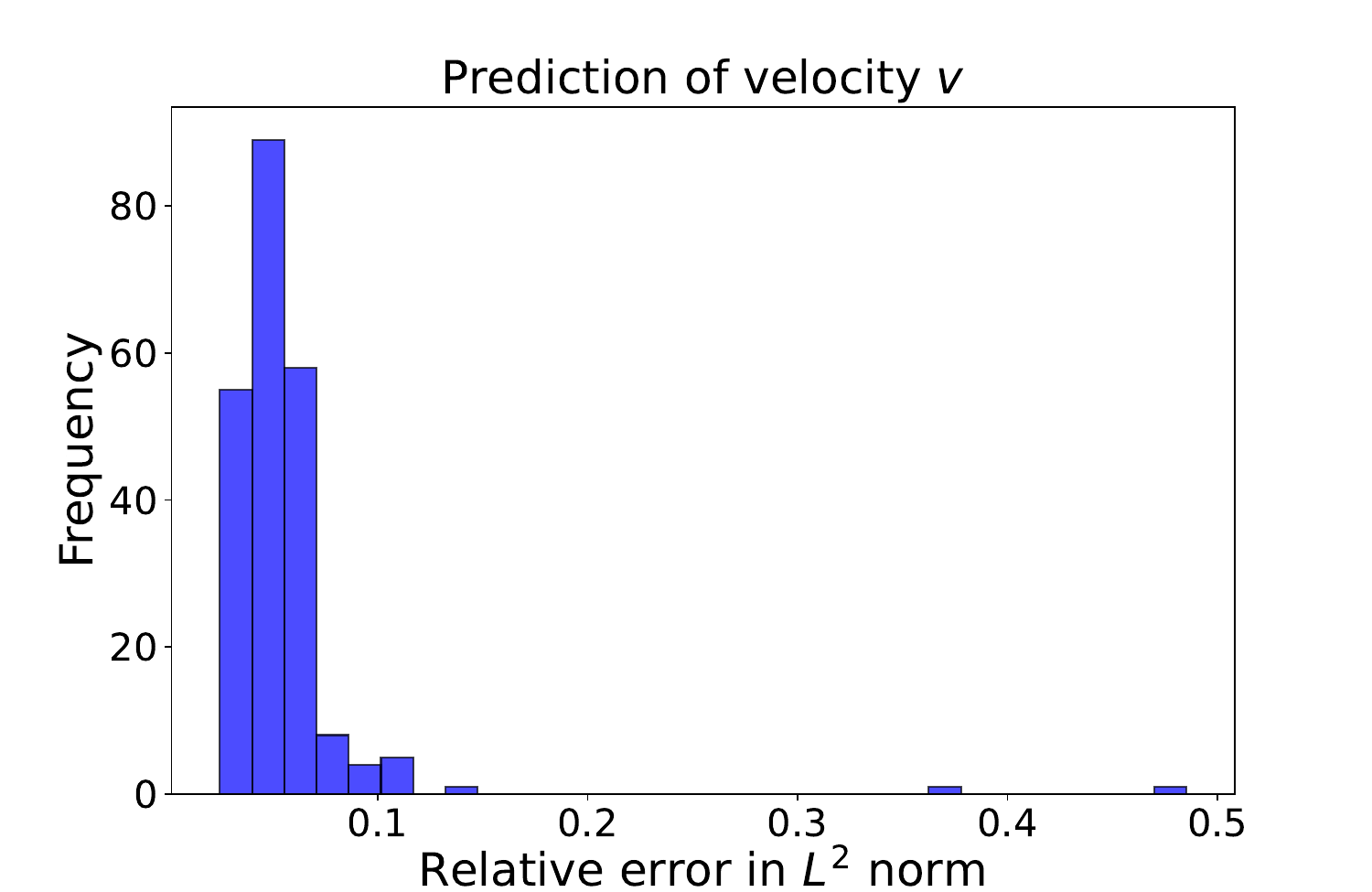}
    \end{subfigure}
    \begin{subfigure}[b]{0.32\textwidth}
    \caption{PointNet with shared KANs}
        \centering
        \includegraphics[width=\textwidth]{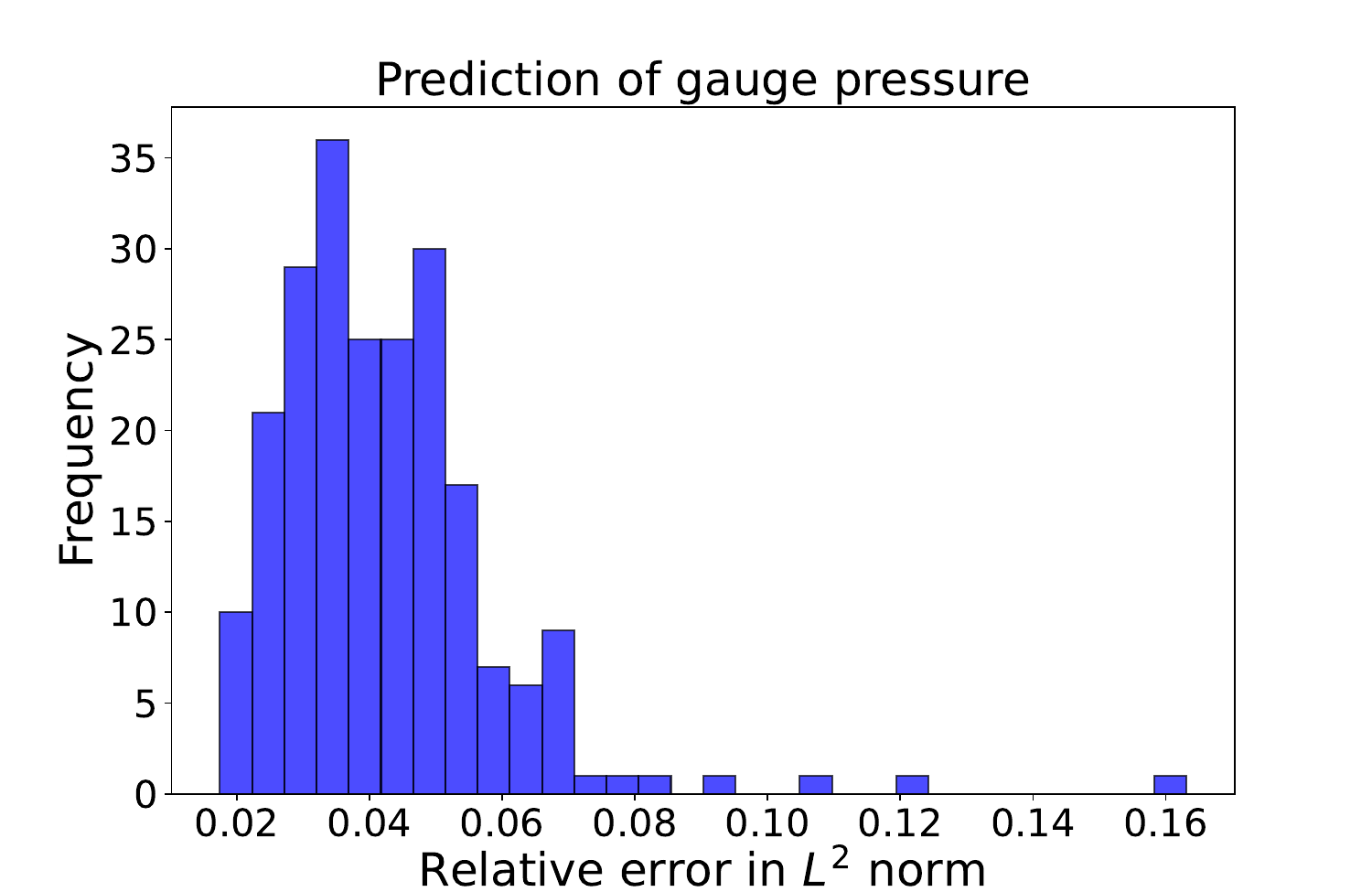}
    \end{subfigure}

    \begin{subfigure}[b]{0.32\textwidth}
    \caption{PointNet with shared MLPs}
        \centering
        \includegraphics[width=\textwidth]{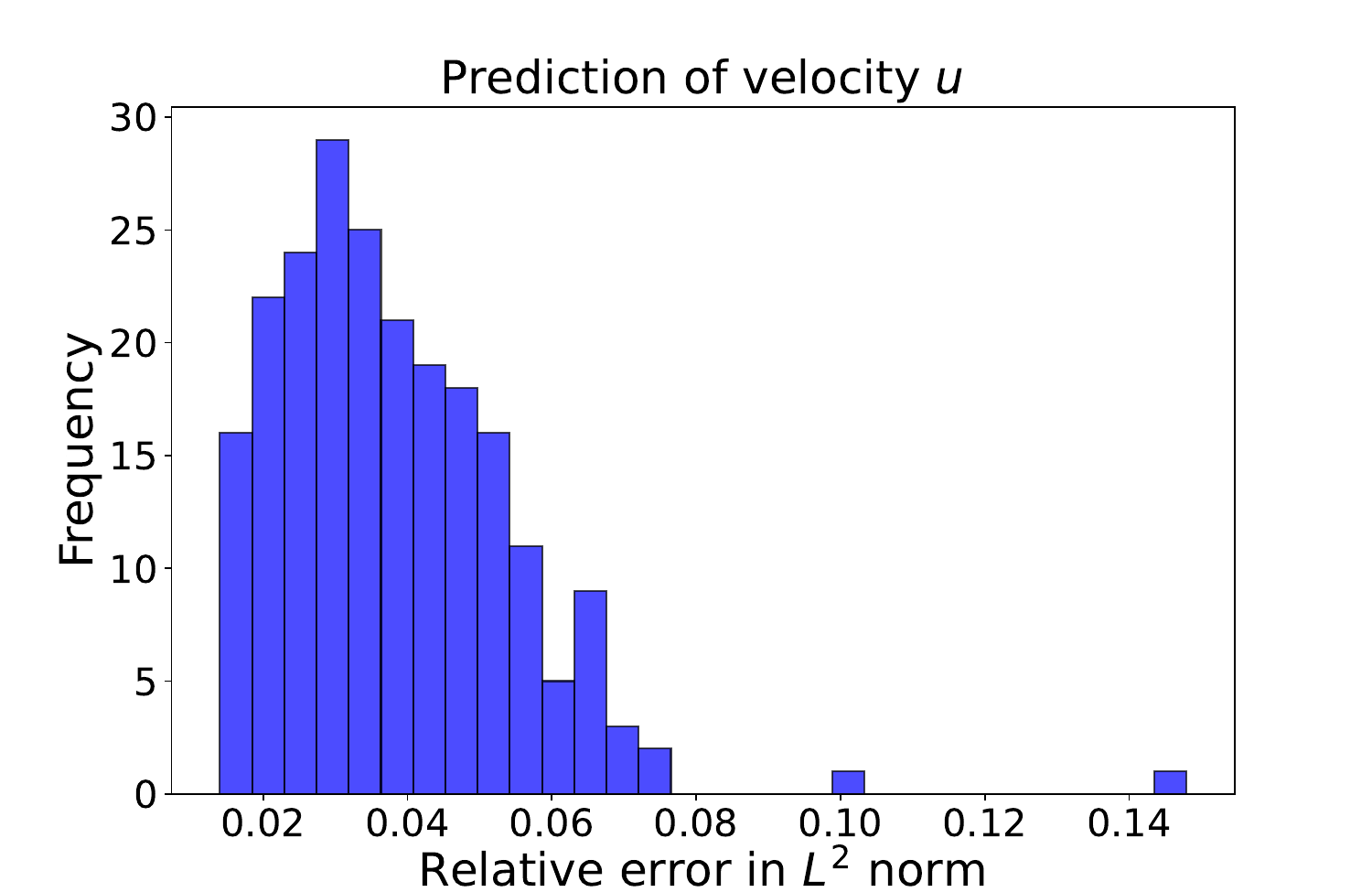}
    \end{subfigure}
    \begin{subfigure}[b]{0.32\textwidth}
    \caption{PointNet with shared MLPs}
        \centering
        \includegraphics[width=\textwidth]{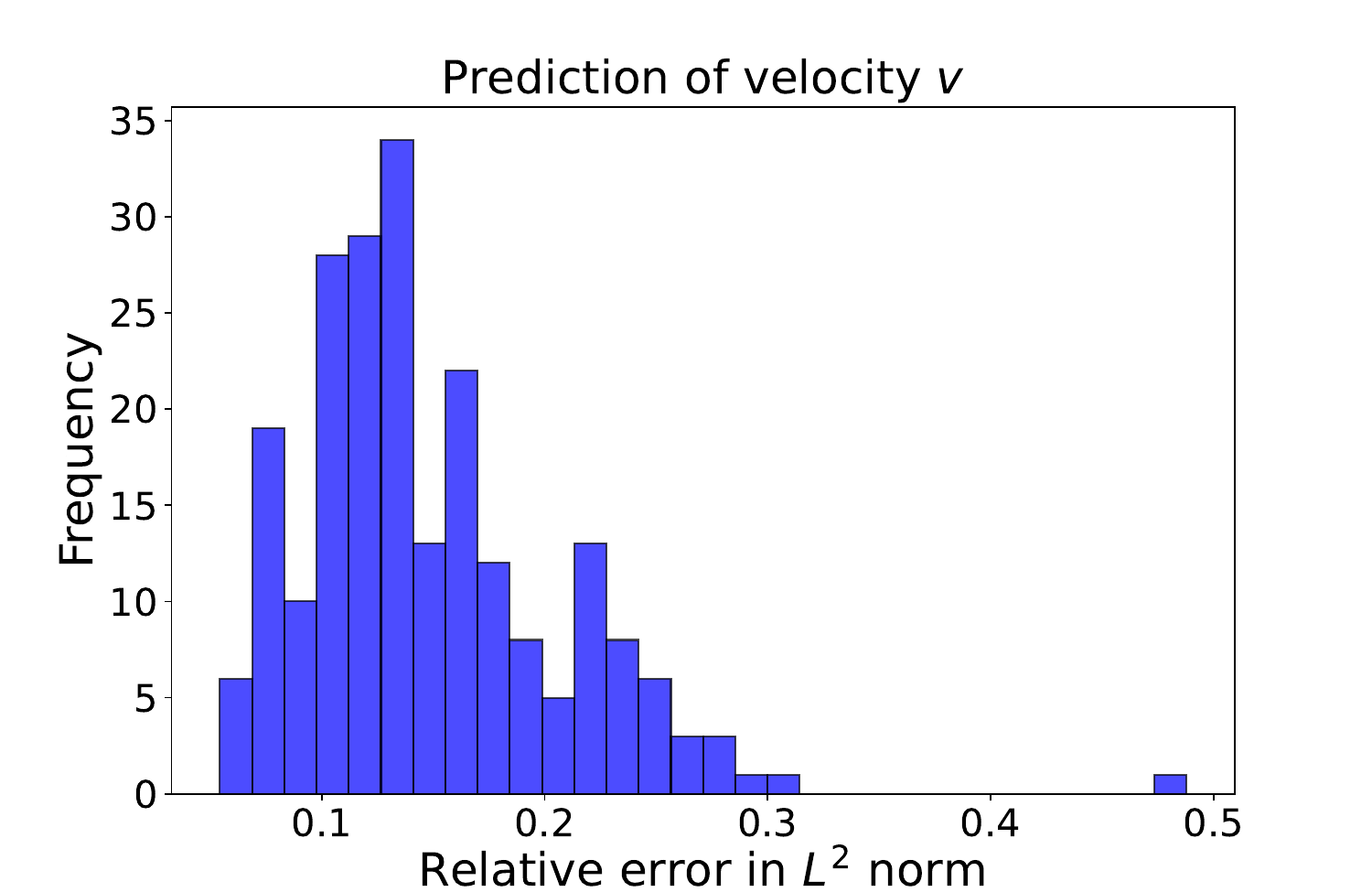}
    \end{subfigure}
    \begin{subfigure}[b]{0.32\textwidth}
    \caption{PointNet with shared MLPs}
        \centering
        \includegraphics[width=\textwidth]{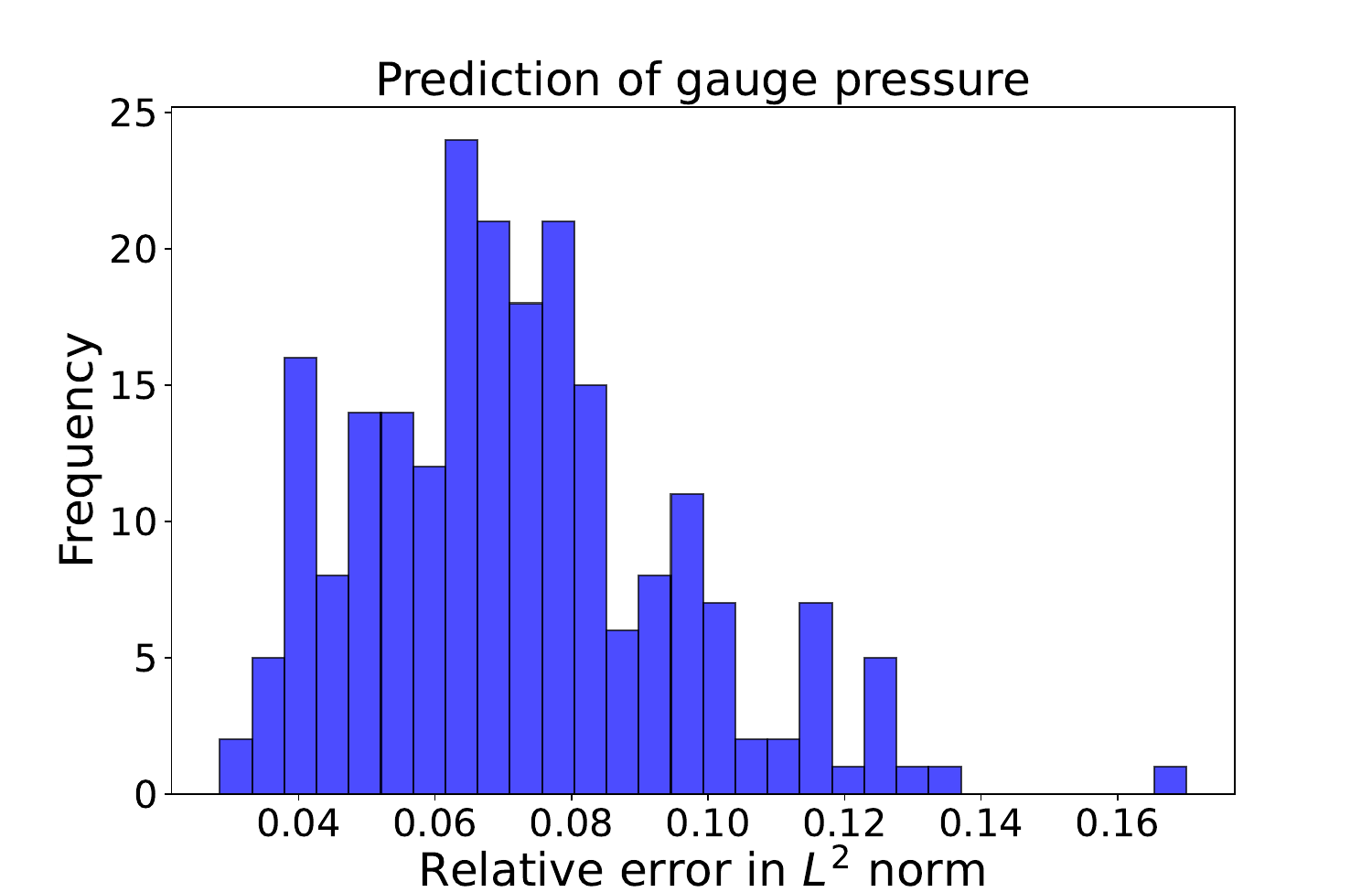}
    \end{subfigure}
    
  \caption{Comparison of histograms of the relative error in $L^2$ norm for the velocity and pressure fields predicted by PointNet with shared Kolmogorov-Arnold Networks (KANs), i.e., KA-PointNet, ($n_s=2$, Jacobi polynomial of degree 3, $\alpha=\beta=1$, and 14169344 trainable parameters) and PointNet with shared Multilayer Perceptrons (MLPs) ($n_s=4$ and 14186243 trainable parameters).}
  \label{Fig19}
\end{figure}


\begin{figure}[!htbp]
  \centering 
      \begin{subfigure}[b]{0.32\textwidth}
        \centering
        \includegraphics[width=\textwidth]{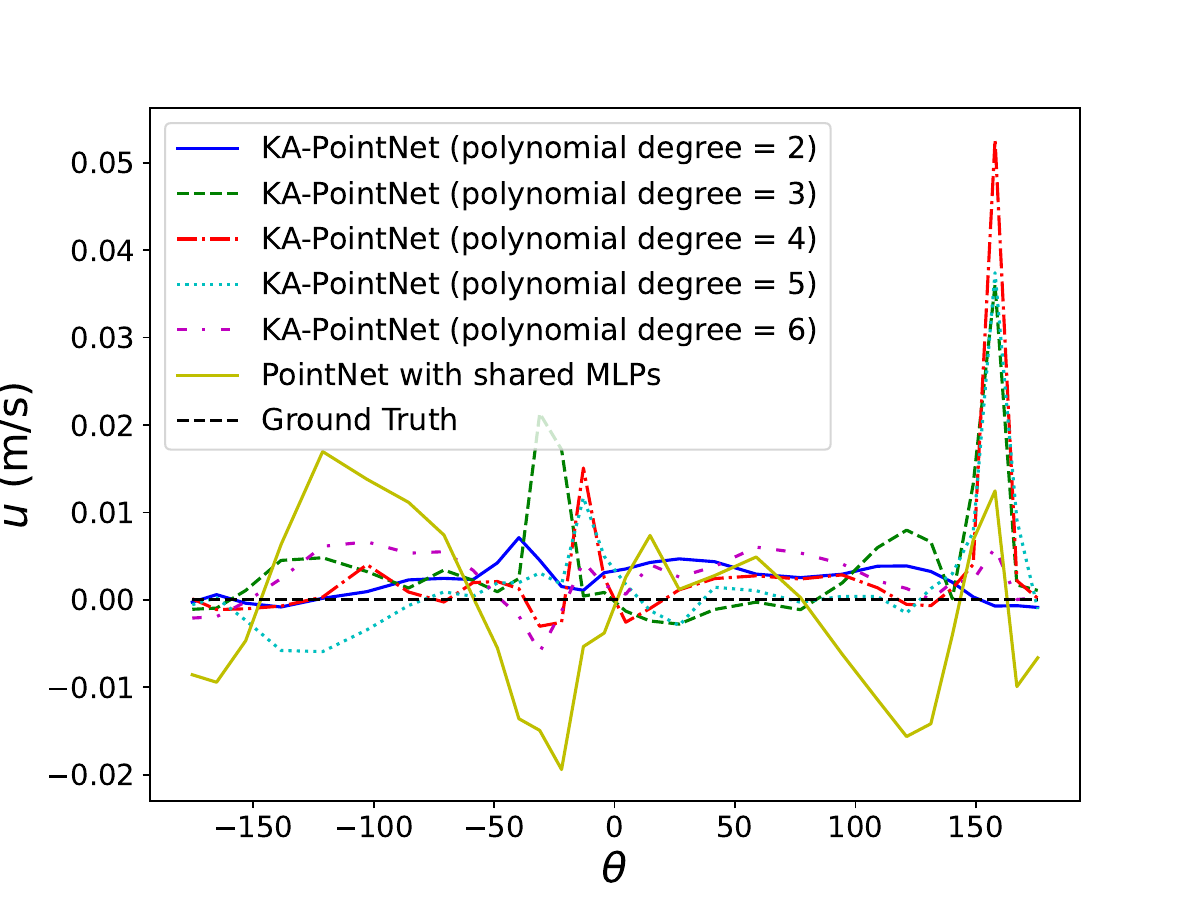}
    \end{subfigure}
    \begin{subfigure}[b]{0.32\textwidth}
        \centering
        \includegraphics[width=\textwidth]{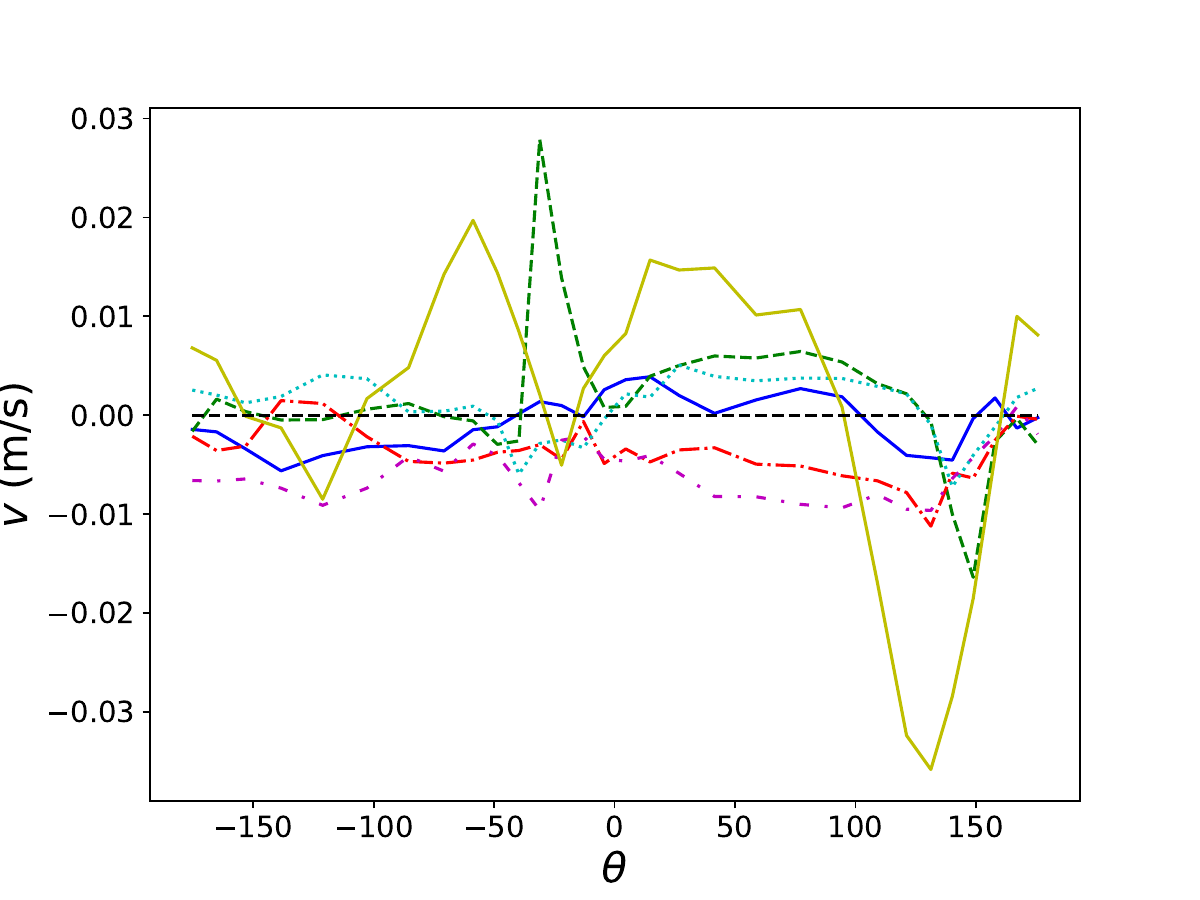}
    \end{subfigure}
    \begin{subfigure}[b]{0.32\textwidth}
        \centering
        \includegraphics[width=\textwidth]{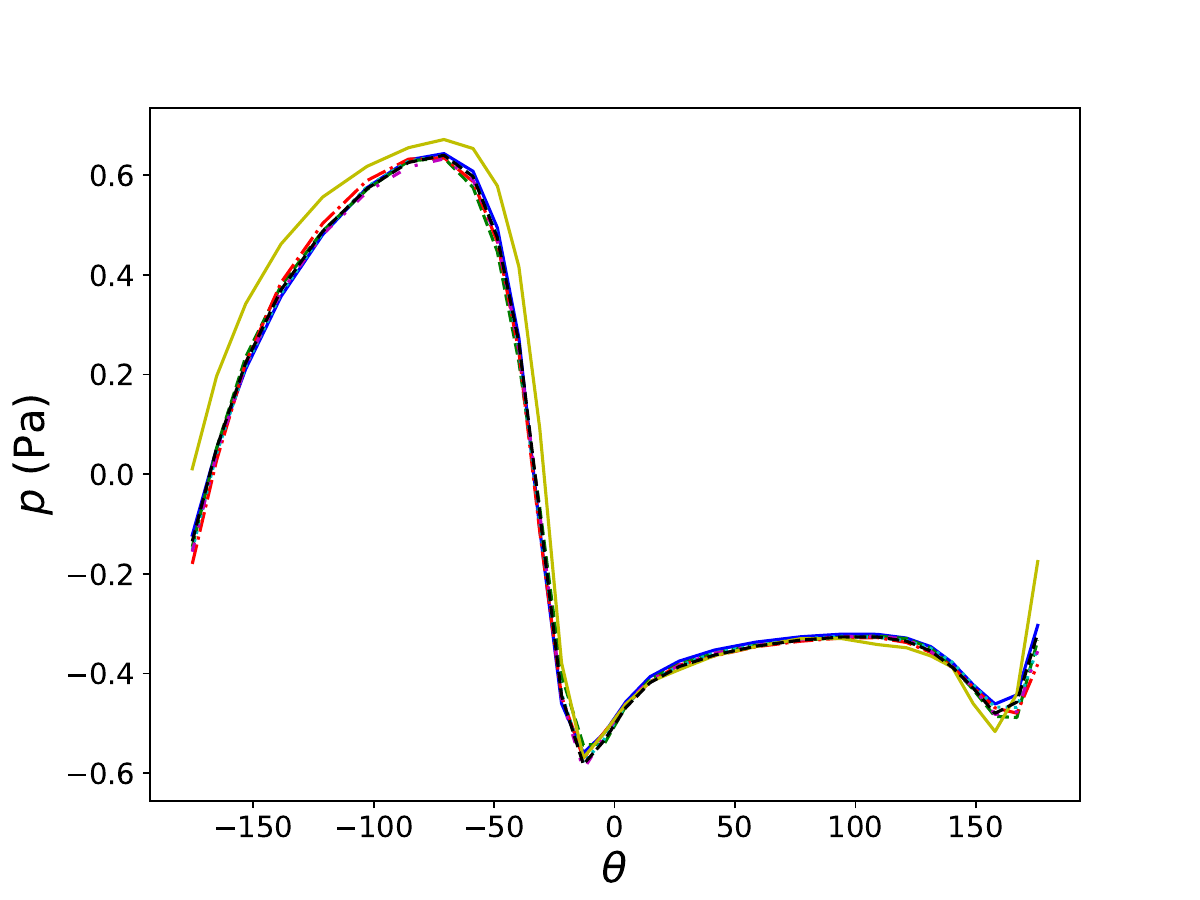}
    \end{subfigure}

  \caption{
  An example of velocity and pressure distributions predicted by KA-PointNet along the surface of a cylinder with an elliptical cross-section from the test set. The angle (\(\theta\)) is measured counterclockwise from the positive x-axis by connecting each surface point to the center of the ellipse at \((8, 16)\). KA-PointNet is used with $\alpha = \beta = 1$ and $n_s = 1$. For PointNet with shared MLPs, $n_s$ is set to 2.}
  \label{Fig20}
\end{figure}


\subsubsection{Performance analysis}
\label{Sect422}

To ensure a fair comparison between PointNet with shared KANs (KA-PointNet) and PointNet with shared MLPs, we select hyperparameters for each network to achieve an approximately equal number of trainable parameters. Table \ref{Table5} summarizes the results of this comparison, examining training time and error analysis. We conduct the comparison using three different setups. For PointNet with shared KANs, we set the degree of the Jacobi polynomial to 3 with $\alpha = \beta = 1$. Next, we choose $n_s$ for each of these two networks. For instance, the second and third columns of Table \ref{Table5} present the results for PointNet with shared KANs with $n_s = 0.5$ and 888128 trainable parameters, and PointNet with shared MLPs with $n_s = 1$ and 892355 trainable parameters, respectively. A similar scenario applies to the pairs in the fourth and fifth columns, as well as the pairs in the sixth and seventh columns of Table \ref{Table5}. In all three comparisons, we observe that the average relative pointwise error ($L^2$ norm) of the predicted velocity and pressure fields, over the test set, obtained by PointNet with shared KANs (i.e., KA-PointNet) is lower than those predicted by PointNet with shared MLPs. The pattern of errors is similar, with the $v$ variable experiencing the highest level of errors and the $u$ variable experiencing the lowest. This superior performance of PointNet with shared KANs (i.e., KA-PointNet) is particularly significant for the prediction of the $v$ variable. For instance, the average relative pointwise error ($L^2$ norm) over 222 unseen geometries in the test set for PointNet with shared KANs with $n_s = 1$ (see the fourth column of Table \ref{Table5}) is approximately 5.76\%, compared to 13.9\% for PointNet with shared MLPs with $n_s = 2$ (see the fifth column of Table \ref{Table5}). A notable difference in performance is also observed for the $p$ variable with this setup (4.76\% vs. 10.6\%). Interestingly, increasing the size of the PointNet with shared MLPs (i.e., choosing greater $n_s$) does not necessarily improve performance, while PointNet with shared KANs shows improvement. For example, comparing the error of the $v$ variable listed in the fifth column ($n_s = 2$) and the seventh column ($n_s = 4$), the relative error of the $v$ variable increases from 13.9\% to 14.8\%. In contrast, a similar comparison for PointNet with shared KANs shows that the relative error of the $v$ variable decreases from 5.75\% to 4.48\% by increasing $n_s$ from 1 to 2. Additionally, the average training time per epoch of KA-PointNet with shared KANs is approximately 6.7 times greater than that of PointNet with shared MLPs, as can be computed from the given training time in Table \ref{Table5}. The main reason for this is the recursive implementation of the Jacobi polynomial in PointNet with shared KANs. An improvement in this implementation from a software engineering perspective could potentially reduce the computational cost of shared KANs, making them more efficient compared to shared MLPs.

For a more in-depth comparison, let us focus on the last two columns of Table \ref{Table5}, where we have PointNet with shared KANs (using a Jacobi polynomial of degree 3, $\alpha = \beta = 1$, and $n_s = 2$), with PointNet with shared MLPs ($n_s = 4$). The loss evolution for the training set (1772 data) and the validation set (241 data) of these two networks is illustrated in Fig. \ref{Fig18}. For a few epochs, we observe a tendency for divergence in KA-PointNet; however, it eventually stabilizes during the training process, a tendency not present in PointNet with shared MLPs. Furthermore, PointNet with shared MLPs exhibits a higher degree of bias (i.e., a greater difference between the training loss and the validation loss) compared to PointNet with shared KANs. This suggests a tendency for PointNet with shared MLPs to overfit the training data, as evidenced by tracking the training loss values in Fig. \ref{Fig18}. Considering this case for comparison, the distribution and frequency of relative pointwise error ($L^2$ norm) of the velocity and pressure fields predicted by PointNet with shared KANs (i.e., KA-PointNet) and PointNet with shared MLPs are plotted in Fig. \ref{Fig19}. We observe that the frequency of predictions with lower errors is significantly higher for PointNet with shared KANs compared to PointNet with shared MLPs. In both architectures, there are a few predictions that appear as outliers, experiencing extremely high errors. Overall, PointNet with shared KANs (i.e., KA-PointNet) provides more accurate results with higher computational expenses for training compared to PointNet with shared MLPs, despite having approximately the same number of trainable parameters.

The comparison between PointNet with shared KANs (i.e., KA-PointNet) and PointNet with shared MLPs can be approached differently by considering approximately equal training times per epoch (i.e., approximately equal computational cost). The results of this machine learning experiment are presented in Table \ref{Table6}. For each pair, we set $n_s$ in PointNet with shared MLPs such that the training time is approximately equal to that of KA-PointNet, where the Jacobi polynomial has a degree of 3 with $\alpha = \beta = 1$. In the case of KA-PointNet with $n_s = 3$ and PointNet with shared MLPs with $n_s = 4.25$, we observe that KA-PointNet provides slightly more accurate predictions for the $u$-component of the velocity field. However, the accuracy of the predictions for the $v$-component of the velocity field and the pressure field remains approximately the same. This scenario changes for the other two pairs: KA-PointNet with $n_s = 1$ and PointNet with shared MLPs with $n_s = 7.25$, as well as KA-PointNet with $n_s = 2$ and PointNet with shared MLPs with $n_s = 11.75$. In these cases, PointNet with shared MLPs yields more accurate predictions for the fields of interest, primarily because the number of trainable parameters in PointNet with shared MLPs is significantly greater, approximately 10 times more than that in KA-PointNet. The main conclusion of this comparison is that the recursive implementation of KAN layers in KA-PointNet should be optimized from a software engineering perspective, particularly for large and deep networks such as PointNet, where a high number of hidden layers and operations between layers (such as max pooling and concatenation) are required.

Additionally, we compare the robustness of KA-PointNet and PointNet with shared MLPs by introducing 10\% random Gaussian noise into the training data. The results of this comparison are presented in Table \ref{Table7}. Specifically, we evaluate two models with approximately equal numbers of trainable parameters to ensure a fair comparison. As expected, the prediction errors for the velocity and pressure fields increase compared to the case with clean training data. However, KA-PointNet consistently achieves significantly lower error norms across all predicted variables and demonstrates greater robustness to noise. The data of Table \ref{Table7} can also be interpreted from another perspective by examining the increase in error compared to clean data, as detailed in Table \ref{Table5}. The average relative $L^2$ errors for the velocity components $u$ and $v$, as well as the pressure $p$, increase by factors of approximately 3.12, 2.49, and 2.05, respectively, for KA-PointNet. For PointNet with shared MLPs, these factors are 3.10, 2.29, and 3.09. This comparison indicates that while both models experience a similar impact from the introduction of noise, KA-PointNet maintains superior performance, as it initially outperforms PointNet with shared MLPs even in the clean data scenario.

Furthermore, we investigate the accuracy of velocity and pressure field predictions across the surface of the cylinders in the test set. In Fig. \ref{Fig20}, we visualize the velocity and pressure distribution on the surface of a cylinder with an elliptical cross-section, selected from the test set. As shown in Fig. \ref{Fig20}, the no-slip condition remains generally satisfied. However, increasing the polynomial degree from 2 to 3 or 4 introduces oscillatory behavior in the velocity predictions on the cylinder's surface, with the maximum local error reaching approximately 5\%. This oscillation is conjectured to be related to Runge's phenomenon. Interestingly, when the polynomial degree is further increased to 5 or 6, the oscillatory behavior disappears despite the use of higher-order polynomials. This can be attributed to the fact that increasing the polynomial degree also increases the number of trainable parameters, leading to more effective training and reducing fluctuations. Therefore, the choice of polynomial order must be carefully considered to balance accuracy and stability in KA-PointNet. Moreover, as discussed earlier and observed in Fig. \ref{Fig4}, training the network to enforce the no-slip condition on the cylinder surfaces is the most time-consuming part of the training (i.e., requires more epochs). As shown in Fig. \ref{Fig20}, in contrast to velocity predictions, pressure predictions along the surface exhibit smooth behavior across all polynomial orders, given that pressure is not a constant zero along the surface and that the pressure gradient is a linear component in the Navier-Stokes equations. Similarly, Fig. \ref{Fig4} demonstrates that pressure prediction is less challenging for the network than velocity prediction, as discussed earlier. Comparing the performance of KA-PointNet and PointNet with shared MLPs, we observe that KA-PointNet consistently outperforms the MLP-based model in all cases. This can be explained by the architectural differences: KA-PointNet incorporates learnable polynomials in its final layer, whereas PointNet with shared MLPs relies on a fixed sigmoid activation function. This additional flexibility enables KA-PointNet to produce more accurate and physically realistic predictions. Furthermore, our findings in Fig. \ref{Fig20} align with the results presented in Table \ref{Table8}, which quantifies the accuracy of pressure lift and drag computations derived from pressure predictions across the test set. Since KA-PointNet produces more accurate pressure predictions (as explained above), it consequently leads to more precise estimations of lift and drag. As mentioned in the previous paragraphs, KA-PointNet with $n_s = 1$ and a Jacobi polynomial degree of 3, and PointNet with shared MLPs with $n_s = 2$, result in approximately the same number of trainable parameters. Therefore, for a more direct and fair comparison, one may focus on these two models, although the discussion provided in this section holds true for other pairs of KAN- and MLP-based models.


\subsection{Ablation studies}

In the previous subsections, we examined the impact of various parameters, including the Jacobi polynomial type, polynomial degree, network size, and the influence of noisy training data on test case performance. These analyses can be considered formal ablation studies. In this subsection, we focus on two additional ablation studies, which are common practices in computer science: the effects of dropping points from the point clouds representing the geometry of the test set, and the impact of altering the ratio of training to test data.

\begin{figure}[!htbp]
  \centering 
      \begin{subfigure}[b]{0.32\textwidth}
        \centering
        \includegraphics[width=\textwidth]{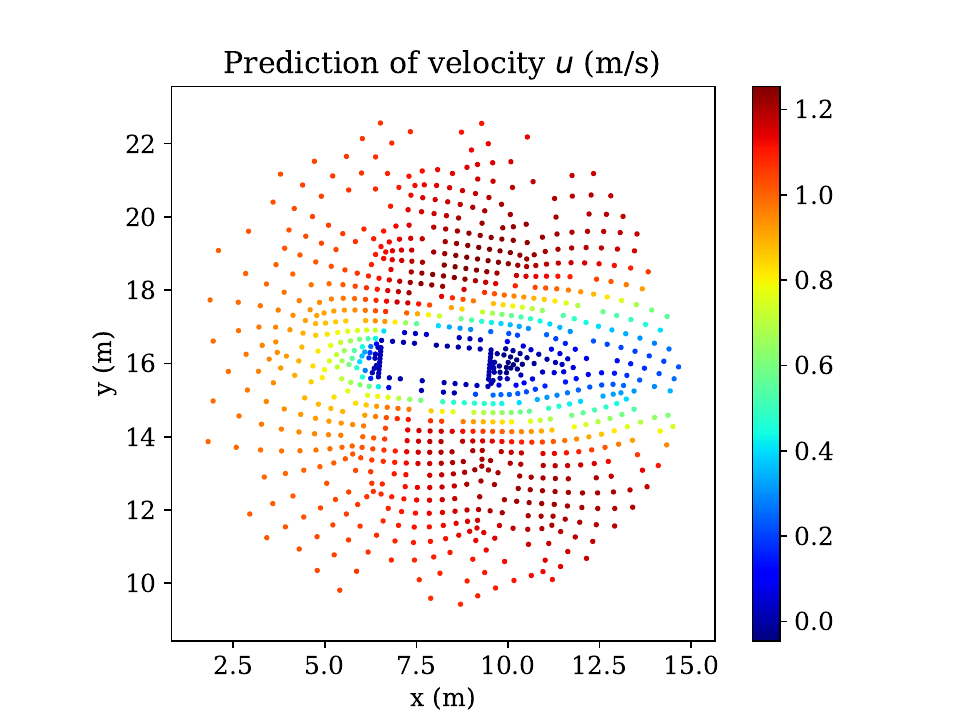}
    \end{subfigure}
    \begin{subfigure}[b]{0.32\textwidth}
        \centering
        \includegraphics[width=\textwidth]{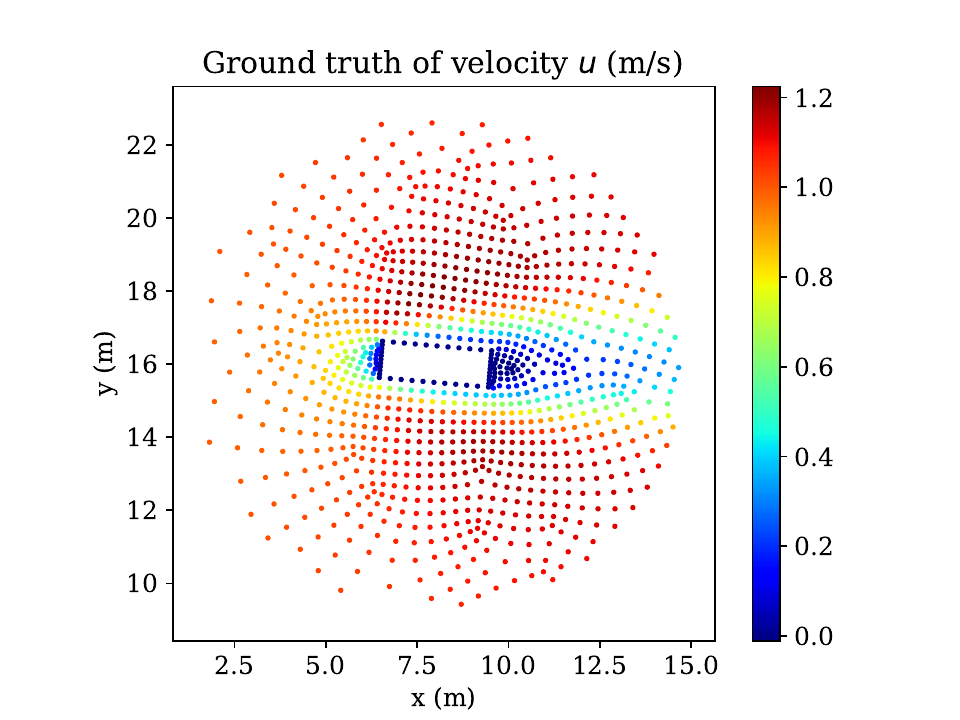}
    \end{subfigure}
    \begin{subfigure}[b]{0.32\textwidth}
        \centering
        \includegraphics[width=\textwidth]{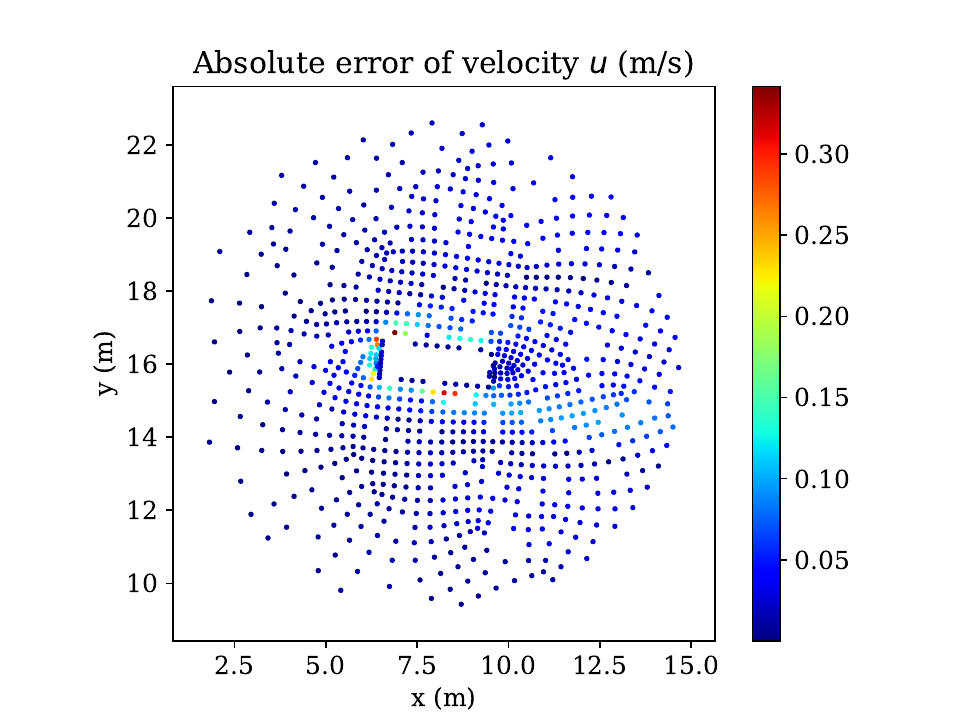}
    \end{subfigure}

    
    \begin{subfigure}[b]{0.32\textwidth}
        \centering
        \includegraphics[width=\textwidth]{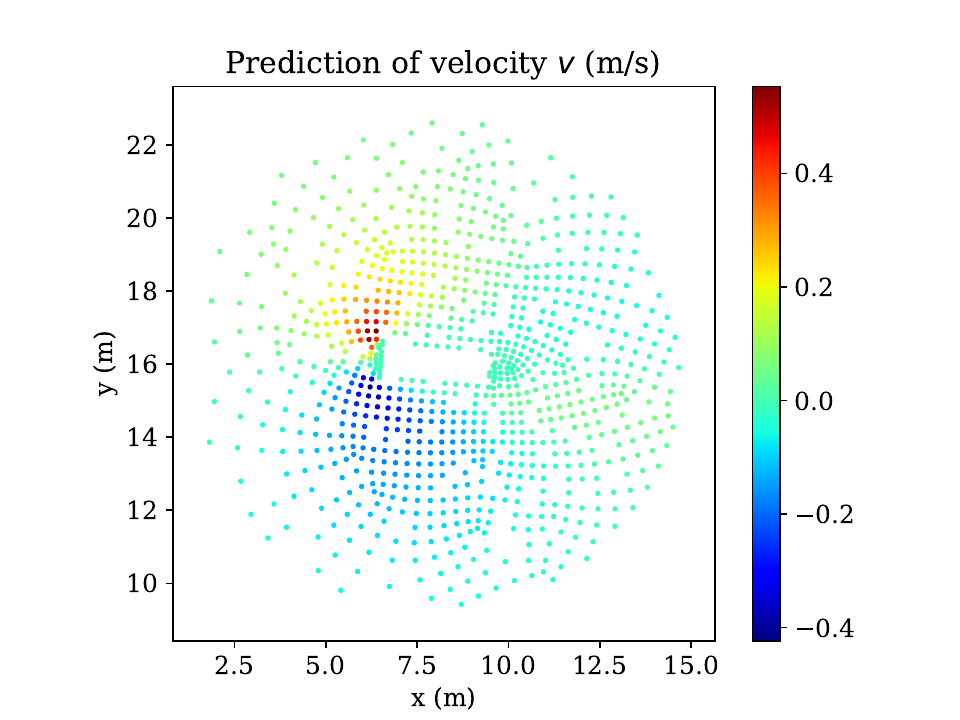}
    \end{subfigure}
    \begin{subfigure}[b]{0.32\textwidth}
        \centering
        \includegraphics[width=\textwidth]{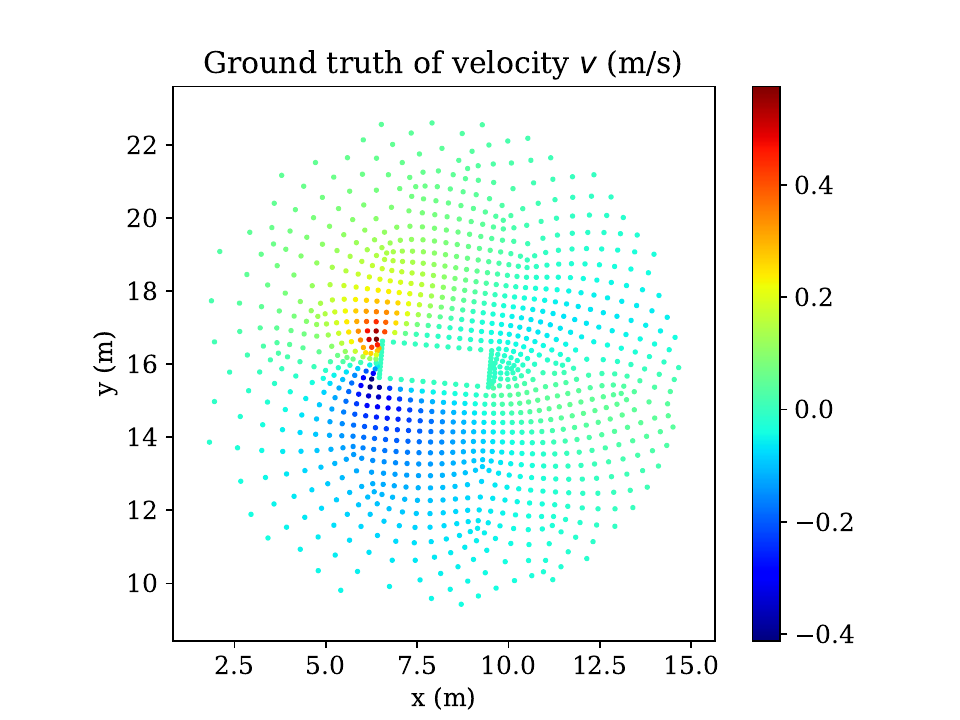}
    \end{subfigure}
    \begin{subfigure}[b]{0.32\textwidth}
        \centering
        \includegraphics[width=\textwidth]{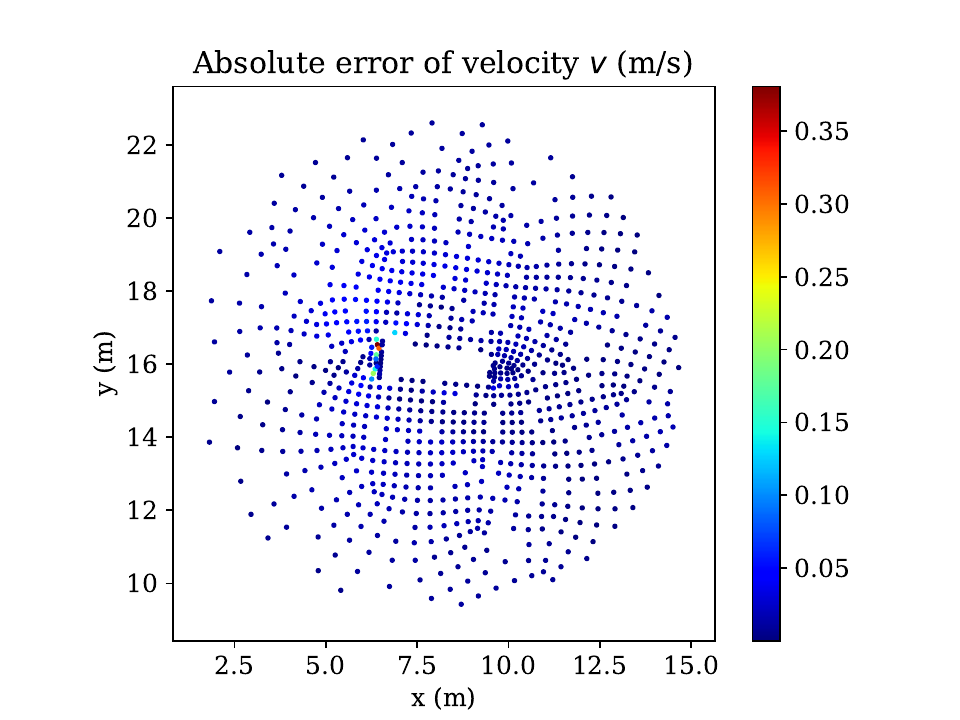}
    \end{subfigure}

    
    \begin{subfigure}[b]{0.32\textwidth}
        \centering
        \includegraphics[width=\textwidth]{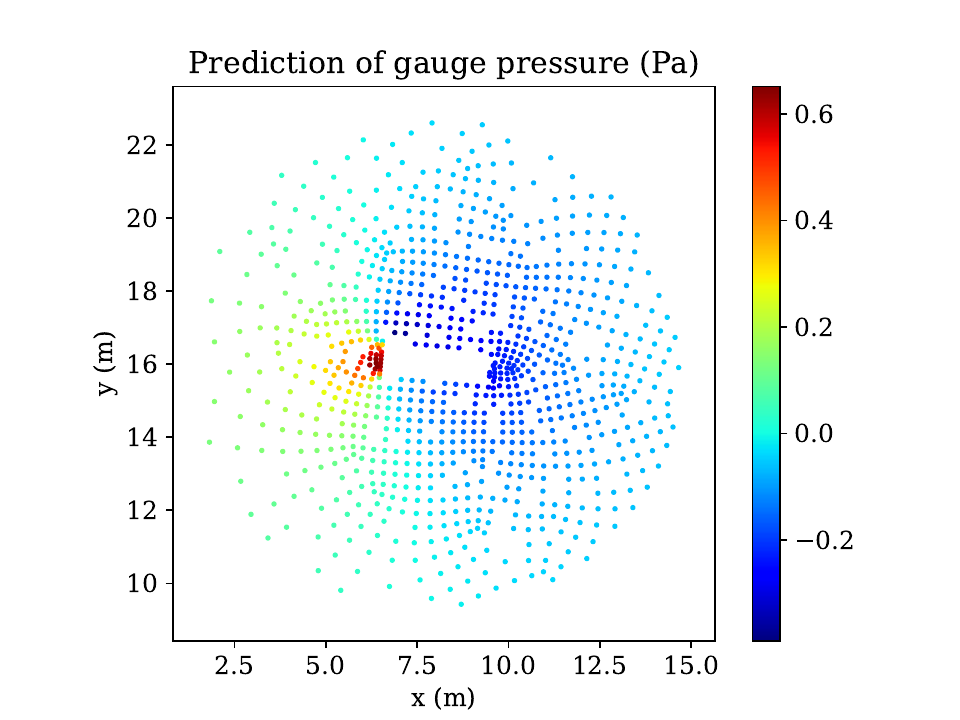}
    \end{subfigure}
    \begin{subfigure}[b]{0.32\textwidth}
        \centering
        \includegraphics[width=\textwidth]{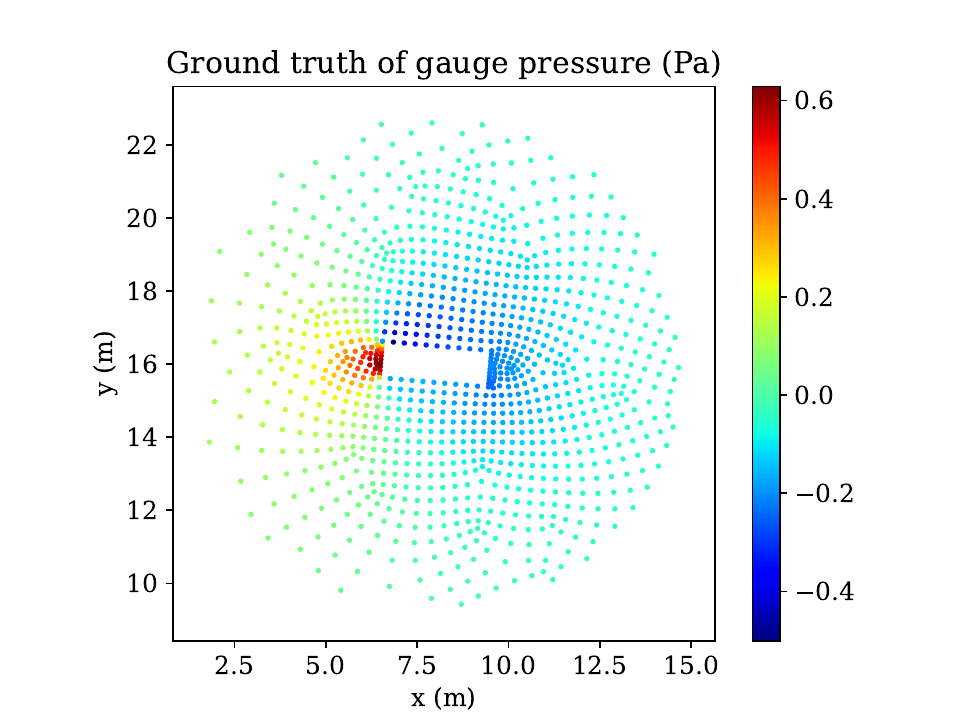}
    \end{subfigure}
    \begin{subfigure}[b]{0.32\textwidth}
        \centering
        \includegraphics[width=\textwidth]{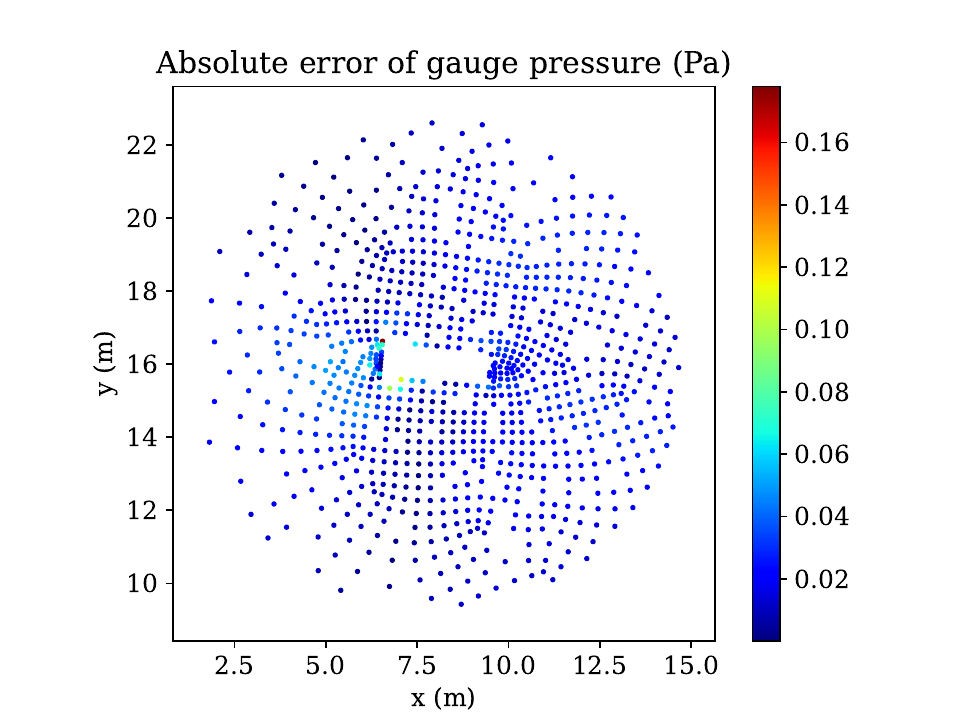}
    \end{subfigure}
    
    \caption{
    An example comparing the ground truth to the predictions of Kolmogorov-Arnold PointNet (i.e., KA-PointNet) for the velocity and pressure fields from the test set, where 15\% of points are randomly removed from the point clouds, and an imperfect geometry of the test set is fed into the network. Note that the ground truth is shown with all 1024 points. The Jacobi polynomial used has a degree of 3, with $\alpha=\beta=1$. Here, $n_s=1$ is set.}
  \label{Fig21}
\end{figure}

Table \ref{Table9} presents a robustness analysis of KA-PointNet by evaluating the error in predicted velocity and pressure fields when a certain percentage of points is randomly removed from the point clouds in the test set. The results indicate that KA-PointNet maintains relatively stable performance even when up to 15\% of the points are missing. Specifically, for the velocity components $u$ and $v$, the relative pointwise error in the $L^2$ norm increases gradually from 1.70912E$-$2 and 5.77925E$-$2 at 2\% missing points to 2.97099E$-$2 and 1.01258E$-$1 at 15\% missing points, respectively. Similarly, for pressure predictions, the relative pointwise error in the $L^2$ norm increases from 4.95328E$-$2 to 8.37937E$-$2 as the percentage of missing points rises. This trend further demonstrates that KA-PointNet exhibits strong resilience to incomplete input data, with errors increasing smoothly rather than abruptly. Connecting to the information tabulated in Table. \ref{Table9}, Figure \ref{Fig21} illustrates an example where 15\% of the point cloud is missing for one of the geometries in the test set. As shown in Fig. \ref{Fig21}, there is a good agreement between the predicted and ground truth fields. Remarkably, as can be seen from Fig. \ref{Fig21}, we observe that some of the missing points form part of the boundary of the cylinder surface. Since the solution of the governing equation is a function of the geometry of the cylinder, one might expect a significant degradation in performance. However, KA-PointNet still produces accurate predictions. It highlights the KA-PointNet robustness in handling such scenarios that might occur in industrial applications.

Table \ref{Table10} presents the average relative pointwise errors ($L^2$ norm) for the velocity and pressure fields of the test set across three different train-validation-test split ratios. Comparing the 90\%, 5\%, 5\% split with the 80\%, 10\%, 10\% split, the errors for the velocity and pressure fields increase slightly. However, the errors become more pronounced when the split changes from 80\%, 10\%, 10\% to 70\%, 15\%, 15\%, particularly for the velocity fields. This trend can be attributed to the Navier-Stokes equations being nonlinear in velocity and linear in pressure. Nevertheless, the average relative errors for all fields remain below 10\%. While a larger training set improves KA-PointNet's accuracy in predicting the test set, the model demonstrates remarkable robustness even with relatively smaller training sets.

\section{Summary and future studies}
\label{Sect5}

In this work, we presented Kolmogorov-Arnold PointNet (KA-PointNet), a novel supervised deep learning framework for predicting fluid flow fields in irregular geometries. Instead of using regular shared Multilayer Perceptrons (MLPs), this framework incorporated shared Kolmogorov-Arnold Networks (KANs) within the segmentation branch of PointNet \citep{qi2017pointnet}. Jacobi polynomials were utilized to construct the shared KANs, and batch normalization \citep{ioffe2015batch} was applied after each shared KAN layer. A two-dimensional steady-state incompressible flow over a cylinder, with varying cross-sectional geometries, was considered as a test case. The Adam optimizer \citep{kingma2014adam} was efficiently used to train KA-PointNet on the training set. Quantitative and visual error analyses were conducted to examine the capability of KA-PointNet. Specifically, the effects of the degree of Jacobi polynomials, specific types of Jacobi polynomials (e.g., Legendre polynomials, Chebyshev polynomials of the first and second kinds, and Gegenbauer polynomials), and the global size of KA-PointNet on the accuracy of the predicted velocity and pressure fields of the test set were assessed. According to one of our machine learning experiments, the average relative pointwise error ($L^2$ norm) of the $x$ component of the velocity vector, the $y$ component of the velocity vector, and the pressure field of the test set predicted by KA-PointNet were approximately 1.2\%, 4.9\%, and 3.4\%, respectively, demonstrating the strong performance of KA-PointNet. Jacobi polynomials of degree 5 and degree 6 generally led to more accurate results but required higher computational training time compared to lower-order Jacobi polynomials. Among the specific cases of Jacobi polynomials, KA-PointNet with Chebyshev polynomials of the first and second kinds showed the lowest error in predicting the velocity and pressure fields of the test set. The single global scaling parameter defined in this study to control the size of KA-PointNet was shown to be an effective parameter for fine-tuning KA-PointNet, despite its complex architecture. A comparison between KA-PointNet with batch normalization and layer normalization demonstrated that, although the prediction accuracy with layer normalization is higher, it requires a significant increase in RAM, which may or may not be available depending on the user's resources. A comprehensive comparison was conducted between PointNet with shared KANs (i.e., KA-PointNet) and PointNet with shared MLPs. PointNet with shared KANs (i.e., KA-PointNet) provided more accurate results than PointNet with shared MLPs, despite having approximately the same number of trainable parameters. However, this increased accuracy came at the expense of higher computational costs for training.

Because KANs are a relatively new deep learning paradigm, there are still many unexplored areas that can be investigated. A few possible directions include extending KA-PointNet to three-dimensional computational mechanics problems, investigating the capacity of KA-PointNet in solid mechanics, such as the prediction of stress and displacement fields in nonlinear elasticity and plasticity problems, and developing physics-informed KA-PointNet for solving inverse problems in scientific computing. Another direction is combining KANs with advanced versions of PointNet \citep{qi2017pointnet}, such as PointNet++ \citep{qi2017pointnet++} and KPConv \citep{thomas2019kpconv}, to improve the accuracy of predictions. Integrating KANs into Fourier neural operators \citep{li2020fourier,anandkumar2020neural,wen2022u,bonev2023spherical,kashefi2024novelFNO} and large language models \citep{lewkowycz2022solving,imani2023mathprompter,frieder2024mathematical,Kashefi2024misleading,kashefi2023chatgpt} could be potentially beneficial to the fields of computational physics and computational mathematics.

In this study, we focused on Jacobi polynomials for constructing shared KAN layers. As part of our future research, we plan to compare the performance of PointNet combined with alternative basis functions in shared KANs, such as B-splines \cite{liu2024kan}, wavelet functions \cite{bozorgasl2024wav}, and radial basis functions \cite{li2024KANradial}. A comparative analysis in terms of computational time, memory usage, the number of trainable parameters, prediction accuracy of physical fields, and critical design parameters such as drag and lift could provide valuable insights.

Another direction for future research is extending the current framework to physics-informed KA-PointNet for solving inverse problems in scientific computing, particularly when only sparse data is available. We previously introduced PIPN \citep{kashefi2022physics}, which integrates PointNet with automatic differentiation to enforce governing equations of a problem within the loss function. Building on this approach, we plan to replace fixed hyperbolic tangent activation functions with learnable Jacobi polynomials. A comparative analysis between PIPN and physics-informed KA-PointNet could offer valuable insights. Furthermore, we aim to develop a physics-informed PointNet architecture that incorporates both MLPs and KANs, such as using shared MLPs in the encoder and shared KANs in the decoder in order to leverage the strengths of both architectures within a unified PointNet.

As another research project, we aim to extend KA-PointNet to unsteady flow problems, particularly transient fluid dynamics involving moving objects or interfaces within the computational domain. Such geometric variations require the point cloud representation of the domain to evolve throughout the simulation. A classical benchmark problem in this context is the motion of objects falling into a free stream \citep{lohner1989adaptive}. To enable KA-PointNet to handle such unsteady problems, we propose integrating it with recently introduced temporal KAN models \citep{genet2024tkan,han2024kan4tsf,yang2024kolmogorov,jiang2025incremental}.

Finally, we plan to extend the application of KA-PointNet to compressible flows, where shocks and discontinuities are present. Previous studies have shown that KANs are more robust to spectral bias compared to MLPs \cite{wang2024expressiveness,meshir2025study}. Therefore, investigating KA-PointNet in such contexts is crucial. One potential test case involves supersonic flows around airfoils with various geometries, where the objective is to predict the locations of shocks, as well as the flow variables within the domain and on the surfaces of unseen airfoils in the test set \cite{chen2023towards,zhang2024deep}.

\section*{CRediT authorship contribution statement}
\textbf{Ali Kashefi:} Conceptualization, Methodology, Software, Visualization, Writing – original draft, Writing – review \& editing.

\section*{Declaration of competing interest}
The author declares that he has no known competing financial interests or personal relationships that could have appeared to influence the work reported in this paper.

\section*{Acknowledgement}
The author wishes to thank the reviewers for their valuable comments and suggestions, which have helped improve the quality of this research article.

\section*{Data availability}
The Python codes and data are available on the following GitHub repository: \url{https://github.com/Ali-Stanford/KAN_PointNet_CFD}


\bibliographystyle{model1-num-names}

\bibliography{cas-refs}


\end{document}